\documentclass[master,oneside,en,table,xcdraw]{resources/dyplompwr}

\usepackage[utf8]{luainputenc}
\usepackage{hyperref}
\usepackage{babel}
\usepackage[T1]{fontenc}
\usepackage{url}
\usepackage{float}
\usepackage{subcaption}
\usepackage{amssymb}
\usepackage{booktabs}
\usepackage{longtable}
\usepackage{topiclongtable}
\usepackage{multirow}
\usepackage{emptypage}
\usepackage[numbers]{natbib}

\usepackage{blindtext}

\author{Kamil Raczycki}
\title{Predicting the location of bicycle-sharing stations using OpenStreetMap data}
\titlen{}
\supervisor{dr inż. Piotr Szymański}
\faculty{Faculty of Computer Science and Management}
\city{Wrocław}
\abstract{This thesis presents the application of spatial data embedding methods to the task of predicting the position of bicycle-sharing stations.}

\begin{document}
\maketitle

\addtocontents{toc}{\protect\setcounter{tocdepth}{0}}
\section*{Abstract}

Planning the layout of bicycle-sharing stations is a complex process, especially in cities where bicycle sharing systems are just being implemented. Urban planners often have to make a lot of estimates based on both publicly available data and privately provided data from the administration and then use the Location-Allocation model popular in the field. Many municipalities in smaller cities may have difficulty hiring specialists to carry out such planning. This thesis proposes a new solution to streamline and facilitate the process of such planning by using spatial embedding methods. Based only on publicly available data from OpenStreetMap, and station layouts from 34 cities in Europe, a method has been developed to divide cities into micro-regions using the Uber H3 discrete global grid system and to indicate regions where it is worth placing a station based on existing systems in different cities using transfer learning. The result of the work is a mechanism to support planners in their decision making when planning a station layout with a choice of reference cities.

\begingroup
\renewcommand{\cleardoublepage}{}
\renewcommand{\clearpage}{}
\begin{otherlanguage}{polish}
\section*{Streszczenie}

Planowanie rozmieszczenia stacji rowerów publicznych jest złożonym procesem,\linebreak szczególnie w miastach, w których systemy rowerów publicznych są dopiero wdrażane. Urbaniści często muszą dokonywać wielu szacunków na podstawie zarówno publicznie dostępnych danych, jak i prywatnych danych dostarczanych przez administrację, a następnie stosować popularny w tej dziedzinie model Lokalizacji-Alokacji. Wiele gmin w mniejszych miastach może mieć trudności z zatrudnieniem specjalistów do przepro\-wadzenia takiego planowania. Niniejsza praca dyplomowa proponuje nowe rozwiązanie usprawniające i ułatwiające proces takiego planowania poprzez wykorzystanie metod osadzania przestrzennego. Bazując jedynie na ogólnodostępnych danych z OpenStreetMap oraz układach stacji z 34 miast w Europie, opracowano metodę dzielenia miast na mikroregiony z wykorzystaniem systemu Uber H3 generującego dyskretne siatki globalne oraz wskazywania regionów, w których warto umieścić stację na podstawie istniejących systemów w różnych miastach z wykorzystaniem uczenia transferowego. Efektem pracy jest mechanizm wspomagający planistów w podejmowaniu decyzji przy planowaniu układu stacji przy wyborze miast referencyjnych.
\end{otherlanguage}
\endgroup
\addtocontents{toc}{\protect\setcounter{tocdepth}{2}}
\cleardoublepage

\tableofcontents
\cleardoublepage

\chapter{Introduction}
\label{ch:intro}

In recent years, in the era of increasingly thriving smart cities, the need to apply machine learning to geographic information system (GIS) tasks has emerged. A substantial amount of data collected using sensors spread throughout cities, remote sensing, or social media can be used to support the work of urban planners. Initial attempts to use machine learning in urban data tasks were based on the analysis of photos and verbal descriptions of point-of-interests (POIs) drawing from the developed fields of image analysis and natural language processing. Currently, the focus shifts towards graph embedding methods and recurrent neural networks to embed carefully selected spatio-temporal data. These methods produce promising results, but unfortunately, they have to be designed for a specific task, which means that they cannot be used more widely.

To use machine learning, one will need data. Often urban data is collected by the authorities but this data is not publicly available or is limited. The data is also stored in different formats, which results in the need to process it or adapt existing methods and models. Such tasks generate costs and often take a long time.

One of the tasks related to urban data is planning the layout of bicycle-sharing stations. The currently available methods focus on manual selection and processing of traffic and station neighbourhood features using a limited number of POI types. 

The region embedding method proposed in this thesis is intended to allow machine learning capabilities to be used to a greater extent than currently available methods. It will use publicly available data from OpenStreetMap, thanks to which it will be possible to apply the method in every city which has spatial data entered into this system. The method will focus on embedding a city region arbitrarily divided into regular polygons and predict whether a station should be located in a particular region or not. In the context of the task of planning the position of city bicycle sharing stations, it is supposed to propose an initial layout of stations in a city without any special preparation of data for a specific city, which will allow planners later to elaborate the plan in more detail. 

\section{Topic analysis}

Within the topic of this thesis, 4 main concepts are worth discussing because they are important for understanding the content of the thesis. Those four concepts are: \textbf{spatial data}, \textbf{spatial indexes}, \textbf{bicycle-sharing systems} and \textbf{data embedding}. 

\paragraph*{Spatial data} carries geographic information related to the Earth and its features. They consist of two main types of data: raster and vector. Examples of raster data are satellite images or scanned photographs, while vector data consists of points, lines, and polygons whose vertices are geographic coordinates \cite{Janipella2019}. These data may be associated with numerical or textual attributes that carry additional information beyond the position and shape of the stored geographic object itself. 

Spatial data analysis refers to a set of methods that aim to find patterns, detect anomalies, and confirm or refute hypotheses and theories. An analysis can be considered spatial if the location information is meaningful, meaning that the results change when the analysed objects are relocated \cite{EncyclopediaOfGIS}.  

Today's spatial data analysis is mostly carried out in GIS (Geographic Information System) software. They allow easy collection, management, analysis, and visualisation of spatial data. They are often used as part of a decision-making system that is based on mathematical models that allow predictions to be made about the future so that administrative planners can test their decisions before implementing them \cite{Maliene2011}.

\begin{figure}[h]
    \centering
    \includegraphics[width=0.5\textwidth]{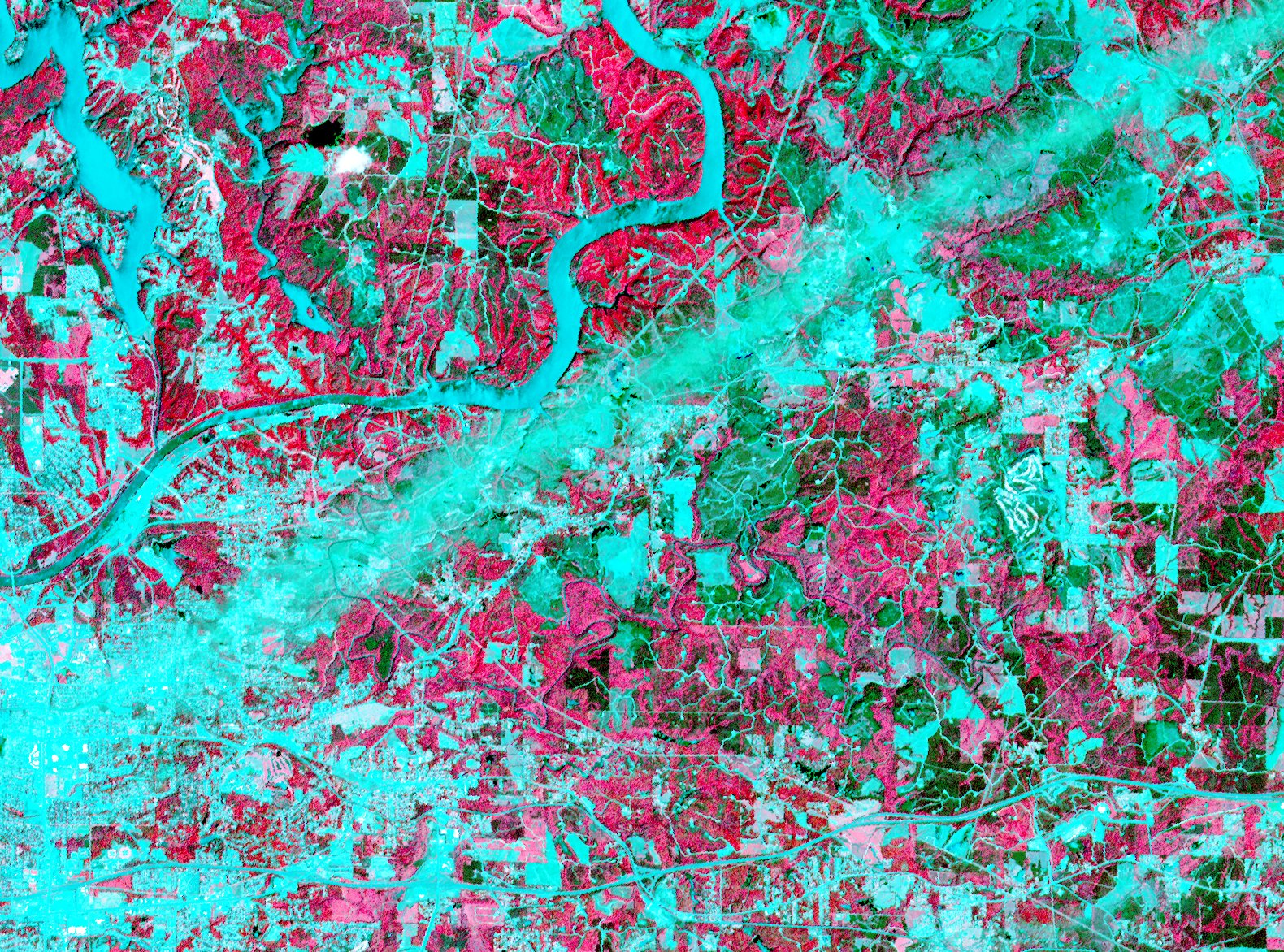}
    \caption[Example of remotely sensed satellite imagery]%
    {Example of remotely sensed satellite imagery. The image shows destruction left by a storm in Alabama in the 2011 year. An instrument aboard NASA's Terra satellite, called Advanced Spaceborne Thermal Emission and Reflection Radiometer or ASTER, captured the images show the scars from the outbreak.\par \small "Satellite View of Alabama Tornadoes (NASA, 05/06/11)" by NASA's Marshall Space Flight Center is licensed under CC BY-NC 2.0.}
    \label{fig:nasa_satellite}
\end{figure}

\begin{figure}[h]
    \centering
    \includegraphics[width=0.5\textwidth]{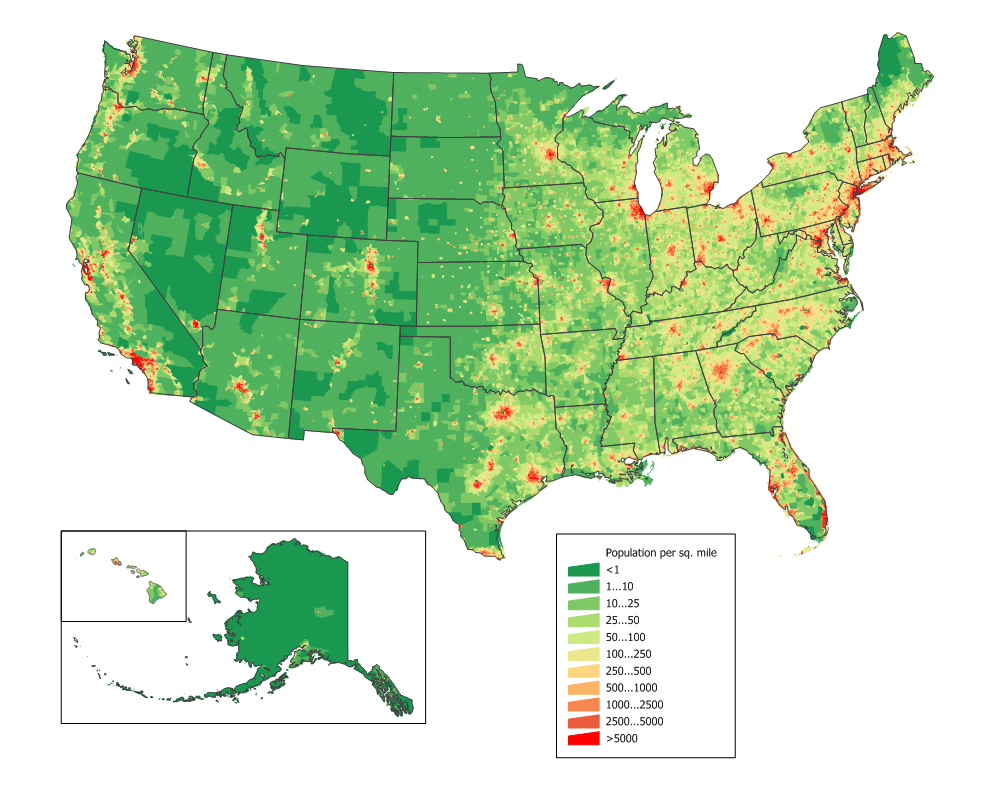}
    \caption[Example of demographic data associated with administrative regions]%
    {Example of demographic data associated with administrative regions. The figure shows population density in the USA based on Census 2010 data.\par \small "US population map" by JimIrwin is licensed under CC BY-SA 3.0, via Wikimedia Commons.}
    \label{fig:demographic_example}
\end{figure}

\begin{figure}[h]
    \centering
    \includegraphics[width=0.5\textwidth]{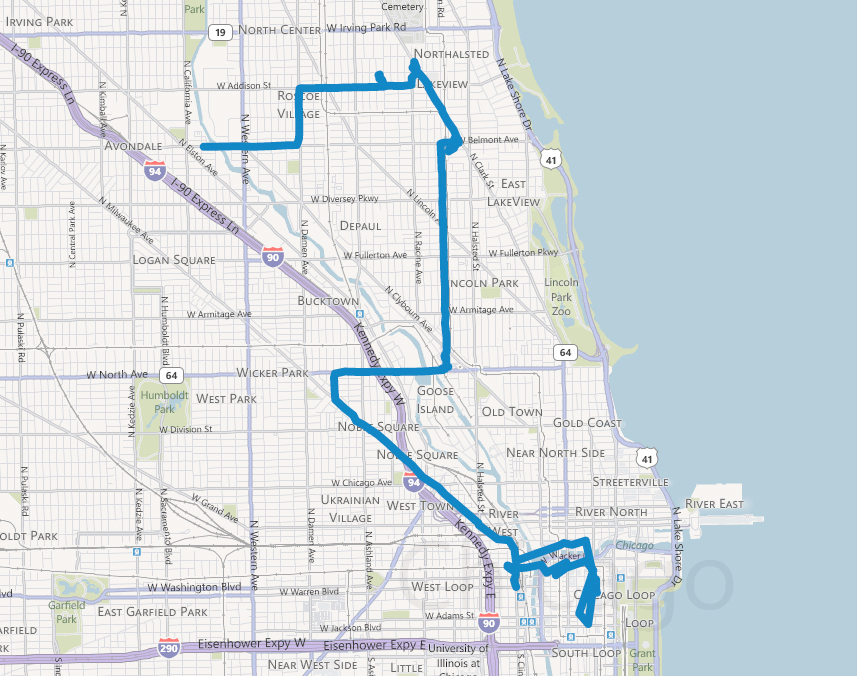}
    \caption[Example of spatio-temporal mobility data]%
    {Example of spatio-temporal mobility data. The figure shows a GPS route in Chicago.\par \small "Map of GPS route" by Steven Vance is licensed under CC BY-NC-SA 2.0.}
    \label{fig:mobility_example}
\end{figure}

\paragraph*{Spatial indexes} are structures that allow you to quickly find an object. They are used to efficiently search large data sets that may be included in spatial databases. They allow finding a specific object without the need for sequential database searches, thus reducing processing time \cite{Zhang2017}. The most popular spatial indexes currently used in GIS databases are Geohash and R-Tree, both of which allow for quick retrieval of a specific object in space.

To divide the study area into smaller regions, Discrete Global Grids (DGG) are used. They divide the Earth's surface into so-called cells. Each cell has a unique identifier so that it can be used as a spatial index \cite{Bondaruk2020}. DGGs can be hierarchical or nonhierarchical and they can divide the Earth into different regular or irregular shapes and can have different granulations. Some of the most popular grids include the ISEA DGG \cite{Sahr_2003}, S2 \cite{googles2}, HEALPix \cite{Calabretta_2007} and H3 \cite{uberh3}.

\begin{figure}[H]
    \centering
    \begin{minipage}{.5\textwidth}
        \centering
        \subfloat[Uber's H3 hexagonal tiling.]{\label{fig:subim1}\includegraphics[width=0.95\linewidth, keepaspectratio]{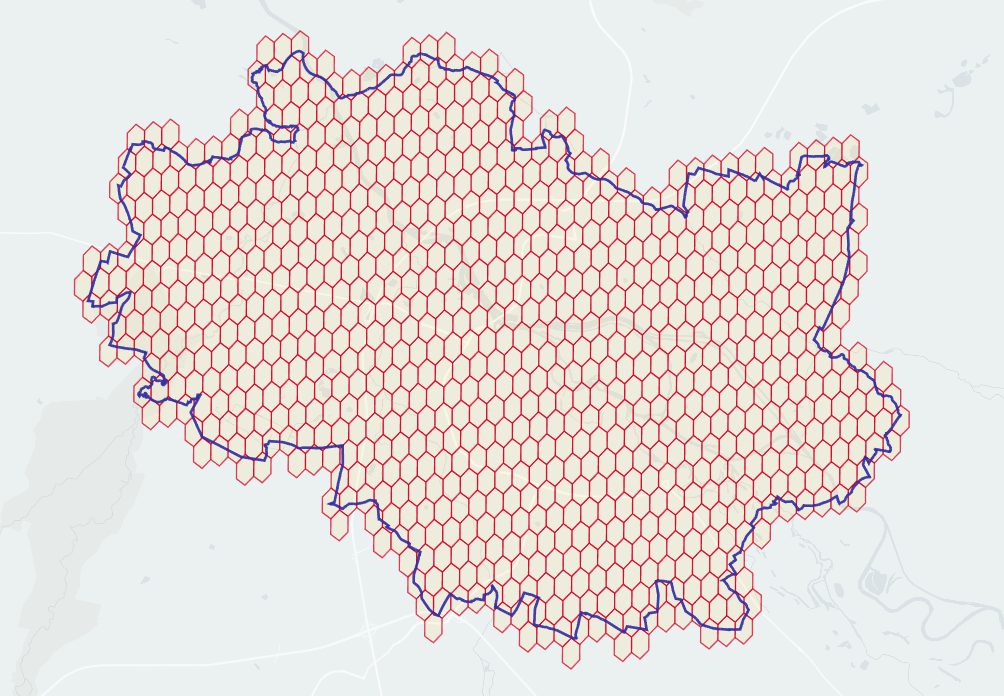}}
    \end{minipage}%
    \begin{minipage}{.5\textwidth}
        \centering
        \subfloat[Google's S2 quadrilateral tiling.]{\label{fig:subim2}\includegraphics[width=0.95\linewidth, keepaspectratio]{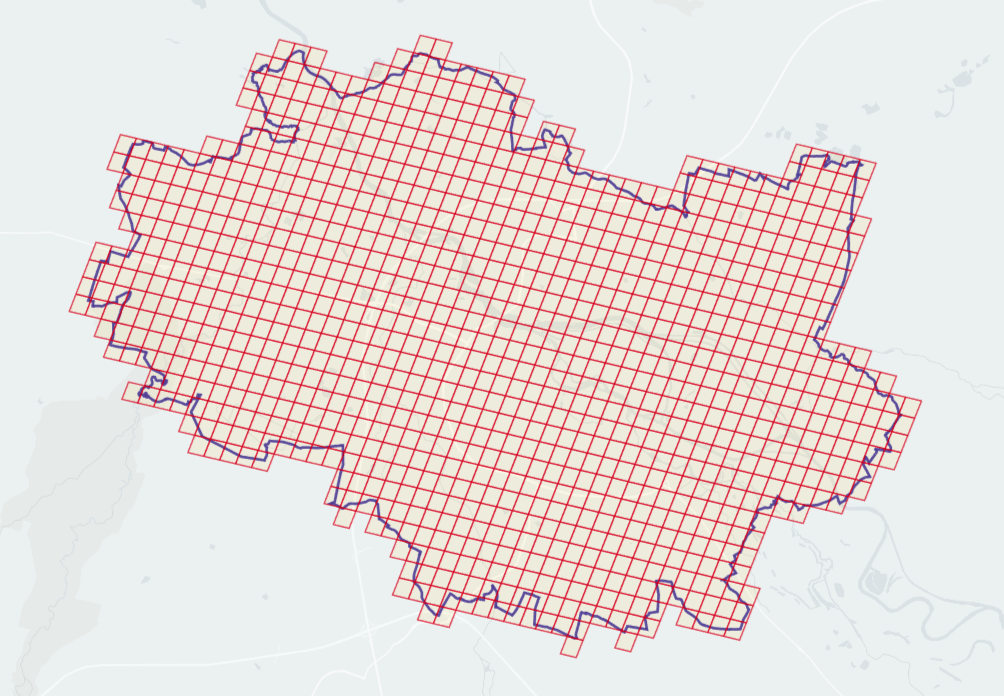}}
    \end{minipage}
    \caption[Comparison between H3 and S2 Discrete Global Grids]%
    {Comparison between H3 and S2 Discrete Global Grids on the example of the city of Wroclaw, Poland. \par \small Personal work. Rendered using kepler.gl library.}
\label{fig:h3_vs_s2}
\end{figure}

\begin{figure}[h]
    \centering
    \includegraphics[width=0.5\textwidth]{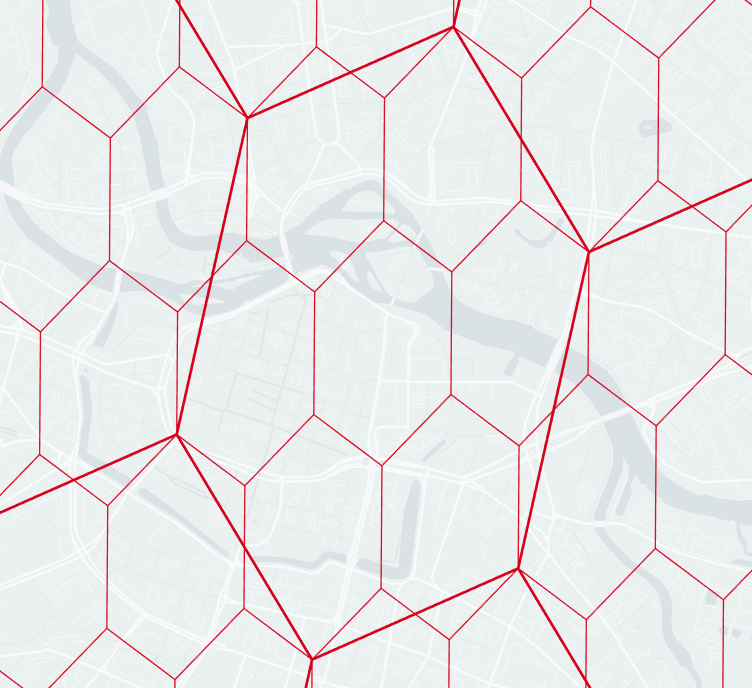}
    \caption[Example of two resolutions overlapping in Uber's H3 DGG]%
    {Example of two resolutions overlapping in Uber's H3 DGG. The thicker line shows hexagon of resolution 7 and smaller hexagons have resolution 8. \par \small Personal work. Rendered using kepler.gl library.}
    \label{fig:uber_overlap_hexagons}
\end{figure}

This thesis will use one spatial indexing system during the experiments, the H3 library provided by Uber \cite{uberh3}. This system uses a grid of hexagons that roughly divides the entire globe of the earth at different resolutions, allowing for a hierarchical subdivision of the region under study. This system was chosen because it is the latest (2018), is easy to use thanks to a library written in Python, among other things, and is well documented. As the system uses hexagons to subdivide the Earth's surface, it has the least distortion, the minimum overlap ratio within a single resolution, and each of its neighbour is the same distance away. Compared to the Google's S2 system (2015) \cite{googles2} which uses quadrilaterals, the distortion of the surveyed shapes is variable depending on the position on Earth due to the Mercator projection (Figure \ref{fig:h3_vs_s2}). The surveyed hexagons do not perfectly overlap between resolution changes (Figure \ref{fig:uber_overlap_hexagons}), but this problem is not relevant in the context of the task solved by this thesis. Unfortunately, this grid contains 12 pentagons which result from the construction of a grid on the basis of the truncated icosahedron, but are mainly distributed over the oceans and are therefore also not important in the context of this thesis.

\paragraph*{Bicycle-sharing system} is a shared transport service in which bikes are made available for use by individuals for short periods of time for a fee or for free \cite{Vogel2011}. They are used for inner-city transport, increasing the flexibility and comfort of movement around the city, reducing the need for cars in city centres. The most frequent systems are based on an extensive infrastructure of docking stations from which bikes are rented and returned. The strategic positioning of these stations is crucial for the efficient operation of the system and its maintenance. The methods used so far to position those stations are based on the manual development of features for analysis in GIS by planners. This thesis aims to develop a simple model to roughly distribute the stations on the city grid, which can help planners in the final detailed positioning of individual stations without the need for manual preparation of features for analysis.

\begin{figure}[h]
    \centering
    \includegraphics[width=0.5\textwidth]{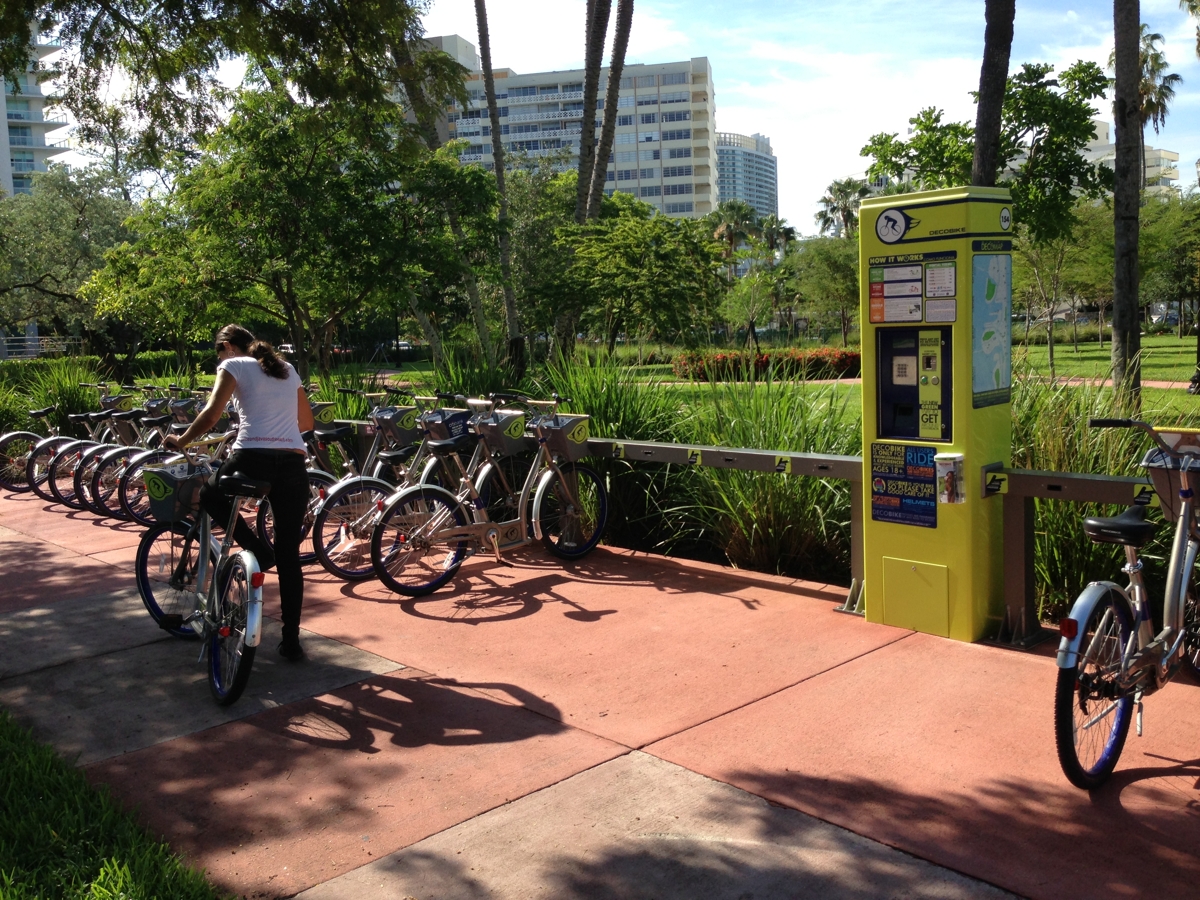}
    \caption[Example of a bicycle-sharing system station]%
    {Example of a bicycle-sharing system station. \par \small "Miami Beach DecoBike Bicycle Sharing Station" image by Wayan Vota is licensed under CC BY-NC-SA 2.0.}
    \label{fig:bike_station}
\end{figure}

\begin{figure}
    \centering
    \includegraphics[width=0.8\textwidth]{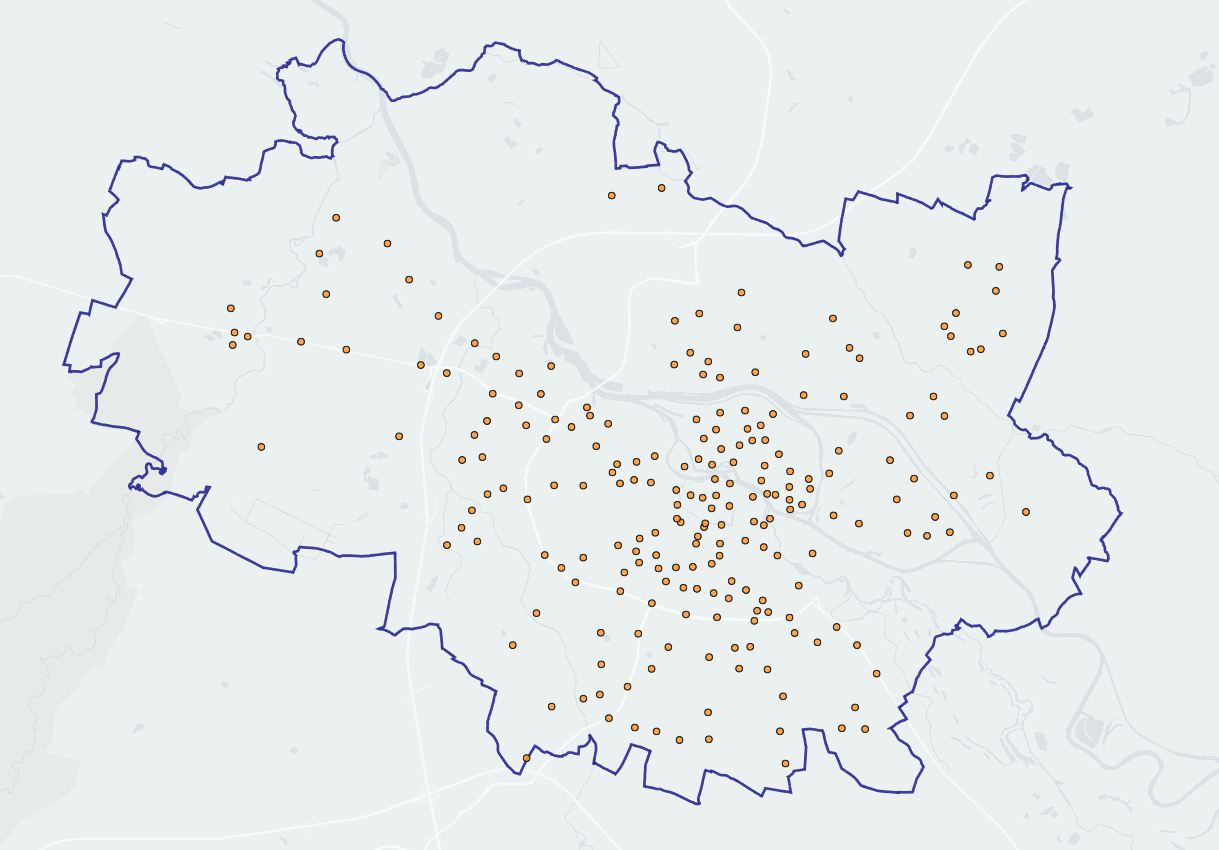}
    \caption[Map of bicycle-sharing system stations in Wrocław, Poland]%
    {Map of bicycle-sharing system stations in Wrocław, Poland. \par \small Personal work. Rendered using kepler.gl library and OpenStreetMap data.}
    \label{fig:bike_map_wroclaw}
\end{figure}

\paragraph*{Data embedding} is a representation learning method focused on reducing the dimensionality of the processed data. The goal of this method is to obtain a low-dimensional representation of the data by building its own feature vector in a given space in an unsupervised manner. The embedding space must be significantly smaller than the original data space \cite{nikhilbuduma2017}, because the smaller dimension of the vectors translates into lower computational complexity associated with their processing and thus reduced learning and evaluation costs for the machine learning model.

\begin{figure}[h]
    \centering
    \includegraphics[width=0.7\textwidth]{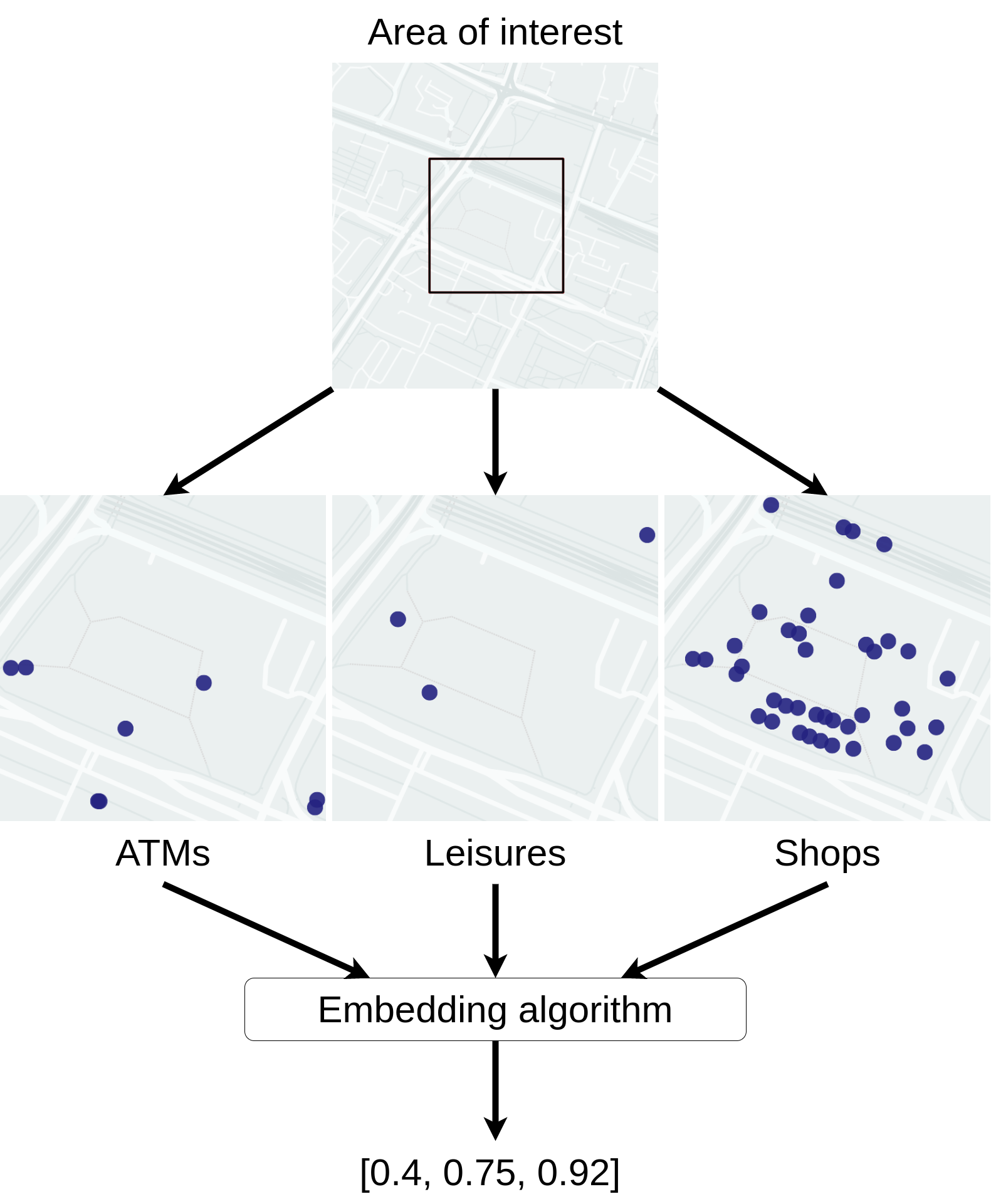}
    \caption[The idea of spatial embedding algorithm for an area of interest]%
    {The idea of spatial embedding algorithm for an area of interest. \par \small Personal work. Rendered using kepler.gl library and OpenStreetMap data.}
    \label{fig:map_embedding_diagmram}
\end{figure}

Embedding methods can be applied to different types of data, such as words, images, or graphs, but in this thesis, only spatial data will be considered. The idea of spatial data embedding is presented in Figure \ref{fig:map_embedding_diagmram}.

\section{Problem statement}

\paragraph*{Define a function that returns the probability of occurrence of a bicycle-sharing station in a specific city-region based on OpenStreetMap data.} \ \\*

The above problem can be broken down into smaller functions, which have been described below.

Having a region $R$, let's define a function $f(R) \rightarrow \left \{ r_{1}, r_{2}, ..., r_{i} \right \}$ that subdivides given region into smaller subregions. This function will be performed by the global discrete grid system and will not be implemented as part of this thesis.

Once the subregions are obtained, the next step is to assign each sub-region a set of objects from the OpenStreetMap data $O = \left \{ o_{1}, o_{2}, o_{3}, ..., o_{j} \right \}$ that it contains. Let us formally define this function as $f(r_{n}) \rightarrow \left \{ o_{1}, o_{2}, ..., o_{j} \right \}$. This step can be performed using spatial queries.

To allow the collected data to be used in machine learning, one more function needs to be used that will map the set of objects assigned to a specific subregion to a vector of real numbers $V$. 

\[
	f(\left \{  o_{1}, o_{2}, ..., o_{j} \right \}) \rightarrow V, V \in \mathbb{R}^{N}
\]

Once the vector form is achieved, the obtained data will be used in machine learning models to obtain the probability of occurrence of a bicycle-sharing station $P(S)$ in a given subregion. Formally, the function takes the following form. 

\[
	f(V) \rightarrow P(S), P(S) \in [0, 1]
\]

The last two functions defined above are the main objective of this thesis. Summarising the whole processing pipeline, the ultimate goal of the work is to propose sub-regions where it would be worthwhile to set up stations that could be important for the bicycle-sharing system network based only on static data from OpenStreetMap.

\section{Thesis objectives}

To respond to the objective defined above, the following steps need to be performed and documented:
\begin{enumerate}
    \item Prepare a method to retrieve OpenStreetMap data from a selected administrative area.
    \item Conduct an exploratory analysis of collected data and identify regions for use based on the topology of aggregated bicycle-sharing systems stations.
    \item Propose a new method for embedding city regions using previously extracted data for use in machine learning methods.
    \item Evaluation of the proposed method by predicting the occurrence of bicycle-sharing stations in specific regions. Examine the hyperparameter space.
\end{enumerate}

\noindent
To achieve the objectives outlined above, the following tasks will be carried out:
\begin{itemize}
    \item conduct research in the area of spatial data embedding and bicycle-sharing stations layout  optimisation,
    \item define comparison criteria and evaluate the found method against them,
    \item identify potential problems and weaknesses of those methods,
    \item propose a new region embedding method for spatial data using only publicly available OpenStreetMap data, that can be trained and applied to machine learning tasks,
    \item collect and prepare data for research,
    \item carry out an exploratory analysis of the data collected with a focus on selecting appropriate regions for research,
    \item implement proposed method,
    \item plan the experiments,
    \item evaluate the new solutions in the defined experiments,
    \item interpret the results,
    \item write the thesis.
\end{itemize}

\section{Thesis outline}

The remainder of this thesis project is outlined as follows. Section 2 reviews relevant previous literature. Section 3 describes the general experimental setup and elaborates on the selection, cleaning and transformation of the dataset as well as a description of used classifiers, general methods, metrics, and software. Furthermore, a list of important concepts is included. Section 4 contains the specific methods and results for the exploratory data analysis and the four research questions. Since each research question comprises a specific approach, the method and results are separated and listed per question. Finally, in Section 5 the research questions and problem statement are discussed, as well as limitations, recommendations and directions for future research.

\chapter{Literature review}
\label{ch:literature_review}

The problem addressed in this thesis brings together two main areas, which separately have already been explored quite thoroughly by researchers. These two fields are the embedding of spatial regions and the optimisation of the layout of bicycle-sharing system networks. For this reason, the review will be divided into two parts.

In the first part, existing methods for embedding spatial data will be surveyed. This will provide the knowledge needed to propose a method for embedding such a complex structure as regions and their associated spatial data. Unfortunately, there is no published literature review in the field of spatial embedding that would allow a simple search for state-of-the-art models. This may be since the field is quite young and increasingly thriving. The reviewed methods are sorted chronologically and start with the Loc2Vec \cite{spruyt2018loc2vec} method from 2018 and end with Region2Vec \cite{Xiang2020} published in 2020. 

The second part will summarise the studies that addressed the problem of optimising the layout of bicycle stations. This topic is quite well researched and is mainly based on optimising the existing layout in a region. However, the topic is still quite popular and newer techniques are being used to solve it. The earliest paper mentioned addresses the problem by researchers in 2012 in Madrid and the latest in 2021 addresses the optimisation problem in Istanbul. The subsections in this section will be named after the city from which the data for the study was used and the year in which the study was published.

\section{Spatial data embedding methods}

A natural first step in trying to solve a problem in a new field is to try to draw on solutions from other fields for which state-of-the-art models have already emerged that are widely used in science and industry. It is no different when trying to solve the problem of embedding spatial data. The authors of the methods described below have taken inspiration from the fields of word embedding using the very popular word2vec \cite{DBLP:journals/corr/abs-1301-3781} model, image embeddings from convolutional neural networks and graph embeddings using models such as DeepWalk \cite{10.1145/2623330.2623732} or Node2vec \cite{10.1145/2939672.2939754}. These models are either the inspiration for building a new model or are part of it.

\subsection{Loc2Vec (2018)}
One of the first methods to emerge in this area is \textbf{Loc2Vec} \cite{spruyt2018loc2vec}, which draws on the well-studied field of image embedding and convolutional neural networks. The method learns a region representation based on a raster square map slice.

Loc2Vec takes as input the coordinates of a location, which it then rasterises using data from OpenStreetMap. The authors were aware that the data in the OSM is much richer and can get individual objects, their shapes, and types, but decided to discard it and just use the image itself which displays all this information in a known and human-understandable format. However, to retain some of the information, instead of rasterising the data to just one RGB photo, they generate a 12-channel greyscale tensor in which each layer contains different object types.

The task of the model is to build an embedding space in which representations of two similar locations are close together. To do this, they used a self-supervised convolutional neural network with a triplet loss function to avoid having to label any data.

When learning the network, the authors used dropout between layers and batch normalisation and Leaky Relu activation functions. In addition, the authors added dropouts to the input, meaning they randomly skipped all one of the twelve channels.

The authors praise the successful development of an embedding space with the ability to move seamlessly between two points in space, the ability to add and subtract embeddings between each other to find another point in space and that the space is easy for humans to understand and analyse.

\subsection{Tile2Vec (2018)}
The second method discussed, \textbf{Tile2Vec} \cite{DBLP:journals/corr/abs-1805-02855}
, is also based on developments in computer vision but uses satellite images rather than map sections. 

The authors of this paper also taught their model using a triplet loss function and added the constraint that images of areas that are close together should also be close together in the learned embedding space. 
The architecture of the convolutional neural network model is based on the ResNet-18 model learned on the CIFAR-10 dataset. The authors tuned the model using their dataset of 300,000 examples. Before the layers of the ResNet model, there is an autoencoder reducing the input size of the images and finally, Principal Component Analysis / Independent Component Analysis and K-means clustering are used for dividing the data into 10 clusters and returned as a 10-dimensional vector of distances to each cluster centroid.

The embedding space learned on urban data allowed the authors to use it successfully in a poverty prediction task and a country health index.

The authors note that the model can return different results and generate different embedding spaces depending on the time of year the data was collected. It is also noted that the model tends to overfit. However, they do not address in the paper the computational effort that must have been invested in learning and using the model.

\subsection{Zone2Vec (2018)}
Another method, \textbf{Zone2Vec} \cite{Du2018}, allows the embedding of spatial regions based on mobility data, i.e. traffic trajectories and so-called check-ins obtained from social networks. On the basis of these data, connections between specific regions are generated, as well as a relationship matrix between zones that provides information on the frequency of visits. In addition, semantic data related to social networks are included, which are processed by the doc2vec \cite{DBLP:journals/corr/LeM14} method and allow the construction of a document-topic semantic matrix.

The model performs optimisation by maximising the average log probability of visiting a zone based on other zones visited within a single trajectory using the skip-gram method. Then, based on the Low-rank Matrix Factorization assumption, the model optimizes a function that reduces the distance between the Forbenius norms of the matrix.

The authors evaluated the model in the Beijing multi-label zone classification task and the city zone similarity function discovery task. The model allowed better results compared to models based on LDA (Latent Dirichlet Allocation), DMR (Dirichlet Multinomial Regression) and TF-IDF (Term Frequency-Inverse Document Frequency) where zones were treated as documents. However, no information on the computational effort of the method is available. 

\subsection{RegionEncoder (2019)}
The next paper combines data from different sources and uses both image embedding and graph embedding methods. \textbf{RegionEncoder} \cite{Jenkins2019} builds the region embedding space based on 4 sources of information: spatial distance between regions, mobility data in the form of trajectories, POI data and satellite images.

The architecture of the model consists of 3 main components: a denoising convolutional autoencoder which processes satellite images, a graph convolutional network (GCN) for learning network representations built from distributed POI data and mobility data, and a discriminator which combines the representations coming out of the two previous components into one common space.

The model is learned in an unsupervised manner and minimises the error combining the loss functions from the three components: reconstructing the autoencoder images, distinguishing the graph representation from the noise distribution together with the reconstruction of the trajectory spanned over the graph, and the binary cross-entropy function from the discriminator.

The authors used the model to solve two tasks: prediction of region popularity and prediction of a flat price. They compared their model with DeepWalk, Node2Vec and the previously mentioned Tile2Vec, and the proposed model produced better results. In addition, the authors performed an analysis of the impact of each component of the architecture on the results. This is one of the first works combining the construction of multimodal representations that are based on different types of data. However, there is no information about the computational complexity of the model and the time needed to train it.

\subsection{Urban2Vec (2020)}
At the beginning of 2020, a paper was published describing another method addressing the topic of spatial embedding of regions. \textbf{Urban2Vec} \cite{Wang2020} combines two types of data: texts about POIs and images, but not satellite images as in the previously mentioned works but images from the Google Street View API.

The model also learns in unsupervised mode, combines part of a convolutional neural network with a triplet loss function, using additionally the assumption proposed by the creators of Tile2Vec that items that are close to each other in the physical space should be close to each other in the embedding space. The image embedding part is based on the trained ImageNet architecture. The bag-of-words construction and the GloVe \cite{pennington-etal-2014-glove} model for vector extraction are used to embed textual data associated with POIs. 

The usability of the obtained embedding space is validated in a prediction task of demographic and socio-economic features. The model is compared with one previous work embedding only street photos and individual components of its model to show that the combined elements and the proposed methods perform better as a whole. However, there is no comparison to other works that have approached region embedding differently.

\subsection{Region2Vec (2020)}
The latest work found, \textbf{Region2Vec} \cite{Xiang2020}, focuses on combining POI data with mobility data. The authors of the paper point out that previous work in the field that relied only on text models, such as word2vec or GloVe, focused mainly on the frequencies and statistics attached to text POIs, neglecting the spatial aspect. However, it was highlighted that existing models combining multiple data types are not suitable for high-level feature extraction and are quite complex.

The authors of this paper used an existing GloVe model to generate embeddings and also learned their own LDA model on documents containing information about all POIs in the study area. Additionally, the model builds zone embeddings based on mobility data extracted from mobile phones by aggregating the number of people in a zone at each hour of the week (168-dimensional normalised vector).

After obtaining 3 vectors from each component, similarity matrices are built using Pearson's coefficient, which is then summed with different weights. Based on the final matrix using the K-means model, the embeddings are aggregated into 5 clusters that could be easily interpreted by humans as regions with a specific utility function.

The authors test the usefulness of their model in the task of land use classification of regions by comparing only with the base models and each of the 3 components used in the model. However, there is no comparison with the models of other researchers.

\section{Bicycle-sharing system network layout optimisation methods}

In the history of the last 50 years, bicycle-sharing systems have evolved and adapted to growing cities. Researchers on the subject categorise existing systems into five generations \cite{CHEN_2018}:
\begin{enumerate}
    \item free bikes available to the public (Amsterdam, Netherlands, 1965),
    \item bikes available for a cash deposit (Copenhagen, Denmark, 1991),
    \item bikes with locking stations unlocked by magnetic card (Portsmouth University, UK, 1996),
    \item bikes rented using a mobile application linked to an ITS system and providing real-time data,
    \item dockless bikes that can be picked up and returned anywhere in a service area. 
\end{enumerate}

The bicycle-sharing systems discussed in the context of this study are mainly of the 4th generation, as most of the older 3rd generation systems have been already upgraded and the 5th generation systems do not require the stations that are the subject of the optimization discussed in this thesis.

Although the problem of optimising the layout of bicycle-sharing stations was discussed by researchers even before 2010, the work presented below was published after 2010, because only then did methods using GIS and machine learning models start to emerge. 

\subsection{Madrid, Spain (2012)}
The first work discussed concerns the preparation of optimal station positions during the implementation of a bicycle-sharing system in Madrid. The authors, \citet*{GarcaPalomares2012} describe a GIS-based method that consists of four steps: estimating potential user demand, finding station positions based on demand, collecting characteristics of proposed stations and finally analysing these stations in terms of accessibility to potential destinations.

For the implementation of the method, the authors used the following data: the city street network with slopes and available speeds, the buildings in the city with the number of inhabitants and workplaces, the transport zones from mobility studies defining the traffic volume and the positions of all metro stations and public transport stops. Statistical and spatial analysis was carried out in ArcGIS-ArcINFO 10.

Based on the population density in the individual buildings in the city and the traffic intensity in the individual city zones, the authors calculated the number of routes made daily from each building and, based on the number of workplaces, the number of routes to each building. After adding the two obtained values, kernel density maps indicating the estimated spatial distribution of bicycle demand were generated. 

The Location-Allocation model \cite{iiasa1781} was then used to determine discrete values at the points of highest demand as well as to determine the locations of potential stations. The model optimised the positions in the modes of coverage maximisation (MCLP) and impedance minimisation (p-median). The authors analysed 5 scenarios of generating stations in a city based on other available systems in the world (station density per population): 100 stations, 200, 300, 400, and 500. The model also allowed the number of docks per station to be determined. 

Using GIS, the authors divided the proposed stations into 4 groups based on estimated demand: generator, mixed, normal attractor, and high attractor. Station accessibility was calculated based on the sum of the number of routes shorter than 5 km that would terminate at that station divided by the time required to cycle those routes squared. The higher the value, the more important the station is for the proposed layout.

According to the authors, a better coverage maximisation method gave better results and they pointed out that a higher number of stations does not linearly translate into better demand coverage and may unnecessarily generate higher implementation and maintenance costs. In addition, various limitations are pointed out: the analysis is only based on data from working days and there is no coverage for points such as parks or other places that can attract large numbers of residents and tourists.

\subsection{New York City, USA (2015)}

Another paper proposes a much more complex process for selecting suitable stations using neural networks and genetic algorithms. Using real mobility data from the CitiBike System operating in New York, the authors, \citet*{Liu_2015} developed a sophisticated method to optimise the existing station layout. In addition to bicycle mobility data, publicly available taxi traffic data and category information for more than 27,000 POIs were used.

Based on the existing station network, regions in the city were generated using Voronoi tessellation. For each station, the average demand per hour was calculated as well as the unavailability due to lack of bikes. The distance to the nearest car park, metro station, taxi stop and the number of fast bike routes were also determined. The number of docks is also added to the characteristics of a particular station, and a preference factor for cycling over taxi use in the region is calculated using historical data.

With the data described above, for each station, the authors trained a neural network to predict the demand and balance of bikes at a particular time of day. The network was learned on data for 320 stations in Manhattan and Brooklyn. A genetic algorithm was then used to find a layout of 252 stations from 1,720 candidates in Manhattan and 68 stations from 967 candidates in Brooklyn using the learned predictor.

The authors compare the results of their network with standard algorithms from the Scikit-Learn library \cite{scikit-learn}: K-Nearest Neighbor, Logistic Regression, SVR with RBF kernel, CART and Adaboost Decision Tree Regression. Comparing the coefficient of determination (R2 score) values, the neural network proposed by the authors performed best with a score of 0.88168, achieving a value more than 0.1 percentage points higher than the next method. The optimisation of the genetic algorithm converged after 109 generations achieving a much better demand score than the then-current station layout. In addition, the new layout reduced the number of unbalanced stations from 86 to 56.

\subsection{Seoul, South Korea (2017)}

An example of optimisation of the station network in Seoul focuses on the reduction of short routes, which are quite often covered by cars. The authors, \citet*{Park_2017} of the study focused on the city's administrative district of Gangnam-gu, which had no bicycle-sharing stations and was congested with car traffic.

The research framework consisted of 3 parts. The first part consisted of determining potential bike station positions, using the trajectories of taxis passing through the study district. The start and end positions of the journeys were selected as potential locations, along with the frequency of occurrence in the travel patterns.

The second step was to determine demand points from a selected set of points: metro stations, shopping malls, parks, and residences. Using mobility data from South Korean mobile operator SK Telecom, the researchers were able to determine the average number of people in the areas around the study points with hourly accuracy. By choosing different radii for different categories of points, they determined the cells that were supposed to reflect travel demand during the day.

The final step was to solve two modes of the Location-Allocation model \cite{iiasa1781}: the minimum impedance (p-median) and the maximum location coverage problem (MCLP). Both of these were discussed in the paper on Madrid, but the former showed better results there. In the case of Seoul, the authors were not able to identify one better model. The impedance minimisation method allowed an even distribution of stations across the district and the coverage maximisation method focused more on the density of station positions in the centre to better meet the estimated demand.

The authors investigated different numbers of stations in the district and found 80 to be the optimum value, emphasising that planners can use both models as support in determining the final station layout. However, the limitations of the study were pointed out. Firstly, taxi routes may not accurately reflect the mobility behaviour of residents. Secondly, both models optimise station positions using Euclidean distance rather than the exact walking distance resulting from the urban street layout. The third point raised was the lack of distinction between generative and attractor positions, as in the Madrid-related analysis.

\subsection{Malaga, Spain (2020)}
Paper by authors  \citet*{Cintrano_2020} proposes the use of metaheuristics to optimise the layout of stations in a city, also attempting to minimise impedance (i.e. distances between residents and bike stations) in a p-median problem. The following methods were investigated: genetic algorithm, iterated local search, particle swarm optimisation, variable neighbourhood search and simulated annealing.

Various publicly available data for the city of Malaga were used for the optimisation: neighbourhood centres, positions of current stations, use of the legacy system, and road layout in the city. These data were used to determine 363 settlement locations, 33,350 potential station positions and the demand for bike stations using the information on population density in the settlements and bike rental positions.

The evaluation of the layout was based on different modes: for population, the influence of an equal and weighted distribution based on the number of inhabitants in the settlement was studied and for distance, both Euclidean distance and distance based on the existing road network were studied.

After examining the hyperparameter space of the 5 metaheuristics studied, the researchers obtained the best results using the genetic algorithm and obtained a more than 50\% reduction in the distance that users have to walk to the station.

In further research, the authors would like to add information about the type of roads in the city, specific POIs, and also take into account the number of docks in the station. It is also proposed to use the presented methods to deploy electric vehicle charging stations in cities.

\subsection{Wuhan, China (2020)}

Another paper focuses on the Chinese city of Wuhan. Authors \citet*{Yang_2020}, propose to use the aspect of temporal demand variability with spatial information when planning station layout. Based on very accurate GPS data on bicycle use in agglomeration, the work builds a spatial-temporal bicycle demand cube. 

A demand map describing where the bikes are at each hour of the day was built from the GPS positions. These data were then superimposed on a grid into which the examined city region was divided. The result is a cube with time on one axis and spatial coordinates X and Y (or latitude and longitude) on the other two. On the constructed cube, with the use of genetic algorithms, the layout of bicycle stations is proposed. A predefined set of possible station locations is given for analysis and an evaluation function maximises demand coverage and minimises the average distance needed to reach the station.

The authors compare their results with the model used in the 2012 Madrid study. Based on the results, they claim that it performs better than the baseline Location-Allocation model due to the use of the temporality aspect. In addition, a new station layout in the study region is proposed, which increases the demand coverage and reduces the distance needed to reach the station. It is also proposed to use the model in other tasks such as optimising the location of petrol stations based on taxi trajectories. However, the method uses a lot of highly detailed GPS data, so it cannot always be used easily. If access to such data is provided, the authors claim that the model is worth using to obtain better results than baseline methods.

\subsection{Istanbul, Turkey (2021)}
The last paper discussed concerns Turkey's largest city, Istanbul. Authors \citet*{Guler_2021}, propose the use of the best worst-case method (BWM) based on several variables related to the demand for station presence. By dividing the study area into small regions, they calculated the distances to public parks, shopping centres, cycling infrastructure, educational facilities, public transport stations (excluding buses), and bus stops. In addition, the population density and the slope of the city were taken into account. 

Parameters were normalised to values of 0-1 by minimisation and some by maximisation (depending on whether the parameter was to be maximised or minimised). A value of 1 determined a high score for the region and 0 a low score. Different weights were then given to the different parameters and, using these, the positions where stations would be worthwhile were calculated for the study region. The values were then discretised into 7 classes from most fitting to most unsuitable.

Additionally, it was checked how these classes change when the weights are modified (from -20 to +20\%) - in this way the authors wanted to check the sensitivity of the proposed system to changes and how the ratio of the 7 classes mentioned above changes. The authors emphasise that their method is easily applicable in other regions and by combining semantic and spatial data, they allow to obtain more effective and realistic solutions. Although the method uses mainly data that is usually publicly available, developing the weights that allow the model to perform well requires the expertise of an experienced domain expert. Another disadvantage is the limited top-down selection of categories to be considered, but this can be changed and adjusted in another implementation.

\section{Summary}

This section reviews the studies that are most relevant to this thesis.
A summary of methods for the spatial embedding of regions will summarise current developments in the field and use the experience as a foundation for proposing a new method. 
An overview of the researchers' different approaches to optimising the layout of stations in a bicycle-sharing network will allow for a better understanding of the needs and to point out possible gaps in the research.

\subsection{Spatial data embedding}

Based on the information obtained from the reviewed works, the methods can be grouped according to the type of data used for embedding as well as the sources of these data. A comparison of these methods can be found in Tables \ref{tab:lit-rev-1} and \ref{tab:lit-rev-2}.

\TopicSetWidth{*}
\TopicSetVPos{t}
\TopicSetContinuationCode{\ (cont.)}
\begin{topiclongtable}{Fl|TcTcTcTc}

\toprule
    Method & Graph & Image & Numerical & Text
\\* \midrule
\endhead
\hline\endfoot
\endlastfoot

\rowcolor{gray!10}
Loc2Vec         & & \checkmark & & \\
Tile2Vec        & & \checkmark & \checkmark & \\
\rowcolor{gray!10}
Zone2Vec        & & & \checkmark & \checkmark \\
RegionEncoder   & \checkmark & \checkmark & \checkmark & \checkmark \\
\rowcolor{gray!10}
Urban2Vec       & & \checkmark & \checkmark & \checkmark \\
Region2Vec      & & & \checkmark & \checkmark \\
\bottomrule

\caption[Type of data used in embedding in different reviewed methods]%
    {Type of data used in embedding in different reviewed methods.}
    \label{tab:lit-rev-1}\\
\end{topiclongtable}

\TopicSetWidth{*}
\TopicSetVPos{t}
\TopicSetContinuationCode{\ (cont.)}
\begin{topiclongtable}{Fl|TcTcTcTcTcTcTc}

\toprule
    Method & CI & Map & MT & POI & SD & SI & SV
\\* \midrule
\endhead
\hline\endfoot
\endlastfoot

\rowcolor{gray!10}
Loc2Vec       &   & \checkmark &   &   &   &   &   \\
Tile2Vec      &   &   &   &   & \checkmark & \checkmark &   \\
\rowcolor{gray!10}
Zone2Vec      & \checkmark &   & \checkmark & \checkmark &   &   &   \\
RegionEncoder &   &   & \checkmark & \checkmark & \checkmark & \checkmark &   \\
\rowcolor{gray!10}
Urban2Vec     &   &   &   & \checkmark & \checkmark &   & \checkmark \\
Region2Vec    &   &   & \checkmark & \checkmark &   &   &   \\*
\bottomrule

\caption[Origin of the data used in embedding in different reviewed methods]%
    {Origin of the data used in embedding in different reviewed methods. \par The columns represent from left to right: \textbf{CI} (\textbf{C}heck-\textbf{I}ns), \textbf{Map}, \textbf{MT} (\textbf{M}obility \textbf{T}rajectory), \textbf{POI} (\textbf{P}oint-\textbf{O}f-\textbf{I}nterest), \textbf{SD} (\textbf{S}patial \textbf{D}istance), \textbf{SI} (\textbf{S}atellite \textbf{I}magery) and \textbf{SV} (\textbf{S}treet \textbf{V}iew).}
    \label{tab:lit-rev-2}\\
\end{topiclongtable}

Unfortunately, these methods cannot be compared due to computational complexity, as the authors did not include such information in their research. It is also not possible to easily compare the quality of performance of these methods, as most of them were validated on different datasets (including non-public data) and used for different types of tasks. Unfortunately, only implementations of the Loc2Vec and Tile2Vec models are publicly available. Additionally, since one of the assumptions of this work is to use only OpenStreetMap data, it will not be possible to use the solutions presented above and compare them with the method proposed later in this paper.

\subsection{Bicycle-sharing system network layout optimisation}

The topic of planning the layout of bicycle-sharing stations is important and is discussed quite widely in the literature. Most works focus on optimising the existing layout based on usage data, but works on planning the layout from scratch are also available. Unfortunately, these methods often use nonpublic data that are made available by the city for specialists. Another problem with these methods is that they are highly complex and complicated, or the knowledge of a domain expert is needed. Not all municipalities can afford such analysis or they simply do not collect the necessary data, so the methods described in the review cannot be applied.

Due to the lack of access to mobility data or population density, it will not be possible in this work to calculate metrics such as station accessibility or distance to stations that can later be compared with the new layouts proposed by the model. Also, for this reason, it will not be possible to repeat the experiments from the work described earlier and compare them with the results obtained in this work.

\chapter{General experimental setup}
\label{ch:experimental_setup}

This section presents the general experimental setup of this thesis. A definition of the source data is given at the beginning. Then, the data selection and cleaning procedure is described, which provides the basis for the experiments. The data transformation steps are then discussed. In the next section, the 3 embedding methods used in the experiments are presented. The next two sections present the proposed classifiers and metrics that will be used to evaluate the method. Additionally, the problem of class imbalance will be addressed. Finally, the technology stack used in the experiments is mentioned.

\section{Datasets}

\subsection{Bicycle-sharing stations positions}
In the 4th generation, bicycle-sharing systems information about the position of the bike stations as well as the bikes themselves is available in real-time and access to it is often public. Using the API provided by one of the main network operators Nextbike\footnote{\url{https://www.nextbike.net}} and the Bike Share Map\footnote{Oliver O'Brien's Bike Share Map website - \url{https://oobrien.com/bikesharemap/}} website, the positions of 11,826 bicycle-sharing stations were downloaded from 61 cities in 24 countries across Europe. The station positions were then entered into a MongoDB database.

\subsection{OpenStreetMap}

OpenStreetMap is an open-source project to collaboratively collect and share geographic data from around the world for free. In addition to a web interface for viewing maps, API services are available to retrieve data in various formats. The Nominatim\footnote{\url{https://nominatim.openstreetmap.org}} search engine is used to search for an administrative area based on a verbal description of the region, for example, "Greater London, UK", and returns information about it such as the region ID in the OSM database, region boundaries in geojson format as well as other metadata.
The second service used was the Overpass API\footnote{\url{http://www.overpass-api.de} with frontend available at \url{https://overpass-turbo.eu}} which allows for selective searches of individual objects from the OSM database.

For data retrieval, a library was developed containing the prepared queries and retrieving in bulk the objects containing the searched administrative region and saving them either to geojson files or entering data into the MongoDB database. The second method of data saving was used in the research.

For 61 cities, for which bicycle-sharing station positions were obtained beforehand, data for 10,941,423 objects in the OSM database were downloaded.

\section{Data selection and cleaning}

\begin{figure}[H]
    \centering
    \includegraphics[width=\textwidth]{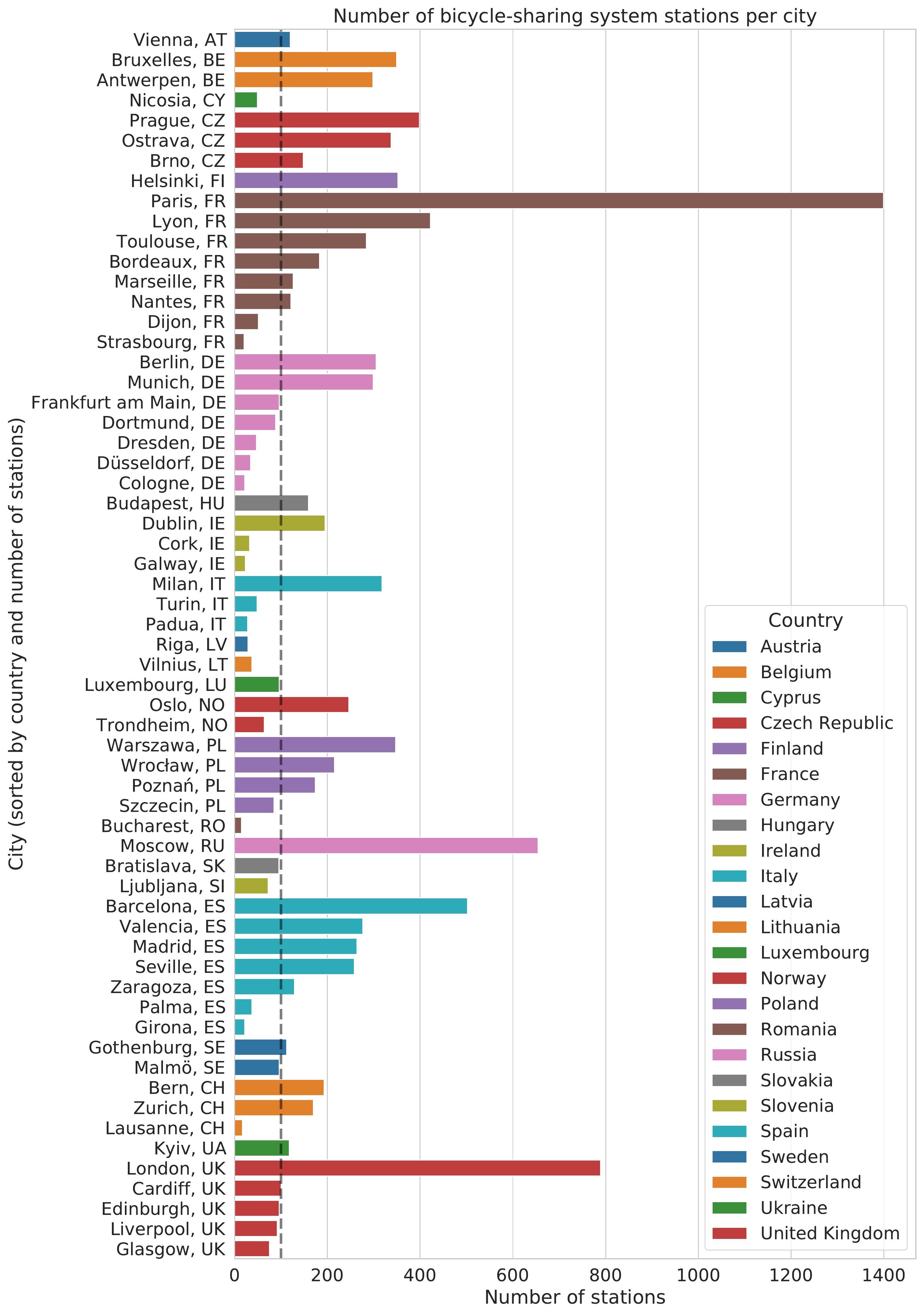}
    \caption[Number of bicycle-sharing system stations per city]%
    {Number of bicycle-sharing system stations per city. The vertical dashed line indicates the cut-off point equal to the number of 100 stations. \par \small Personal work.}
    \label{chart:stations_per_city_all}
\end{figure}

After downloading the data, the number of bicycle-sharing stations was first compared between cities. After analysing the number of stations per city, a minimum number of 100 stations was arbitrarily selected to limit the set of cities for further study. This step was taken due to the concern of too-small learning set for a particular city. The full distribution from all cities along with the cut-off point is included in Figure \ref{chart:stations_per_city_all}. After filtering out the cities due to the minimum number of stations, 34 cities remained in the set.

The next step was to select the studied resolutions from the Uber H3 index. It was decided to study 3 consecutive resolutions:

\begin{longtable}{c|cc}
    \toprule
        Resolution &
        \begin{tabular}[c]{@{}c@{}}Average hexagon\\ edge length (m)\end{tabular} &
        \begin{tabular}[c]{@{}c@{}}Average hexagon\\ area (ha)\end{tabular} \\* \midrule
    \endfirsthead
    \endhead
    \bottomrule
    \endfoot
    \endlastfoot
    9 & 174.38 & 10.53 \\
    10 & 65.91 & 1.50 \\
    11 & 24.91 & 0.21 \\* \bottomrule
    \caption[Selected Uber H3 resolutions]%
    {Selected Uber H3 resolutions and its properties.}
    \label{tab:uber-resolutions}\\
\end{longtable}

The choice was dictated by concern for prediction accuracy as well as computational complexity. Resolution 8 divides the region into hexagons of about 460 metres, so that there may be several residential blocks inside. As stations in cities are often less than half a kilometre apart, it was decided to discard this value from the study. On the other hand, at a resolution of 12, the hexagons are so small and there are so many of them that there could be tens of millions of microregions in the interior of one city, resulting in a large computational effort. This value may be considered in the future if the method proves valuable for resolution 11.

\newpage

\begin{figure}[H]
    \centering
    \subfloat[]{\label{fig:london_h3_example:a}\includegraphics[width=0.95\linewidth, keepaspectratio]{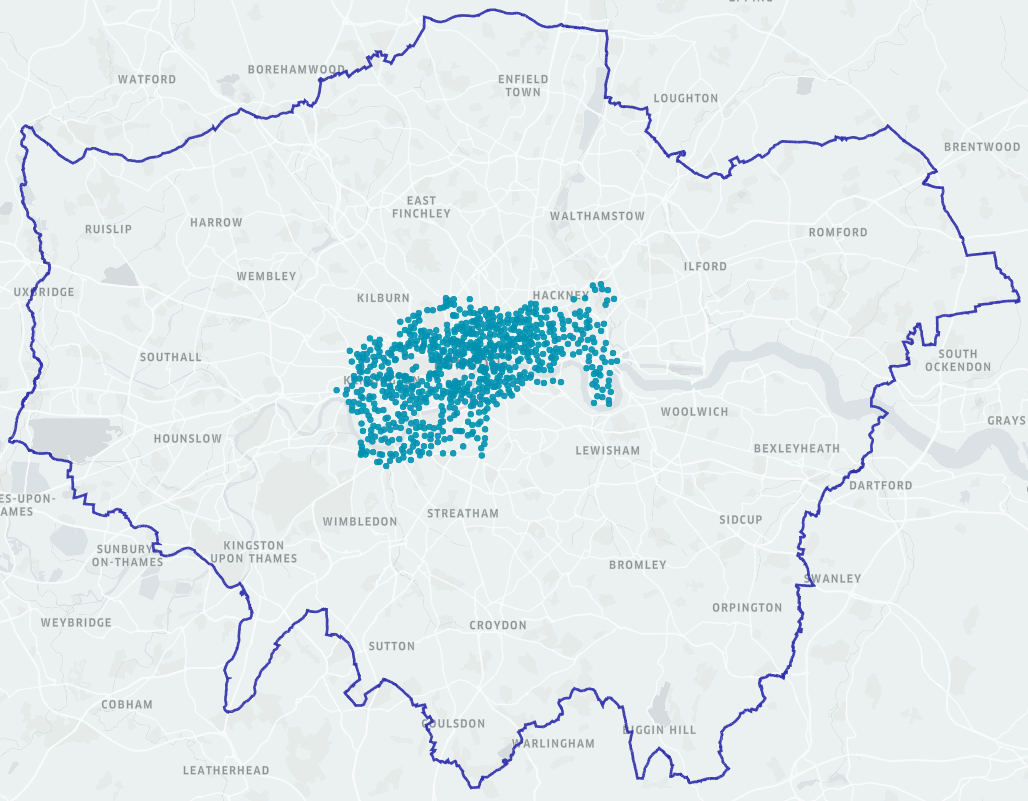}}
    \par\medskip
    \begin{minipage}{.5\textwidth}
        \centering
        \subfloat[]{\label{fig:london_h3_example:b}\includegraphics[width=0.9\linewidth, keepaspectratio]{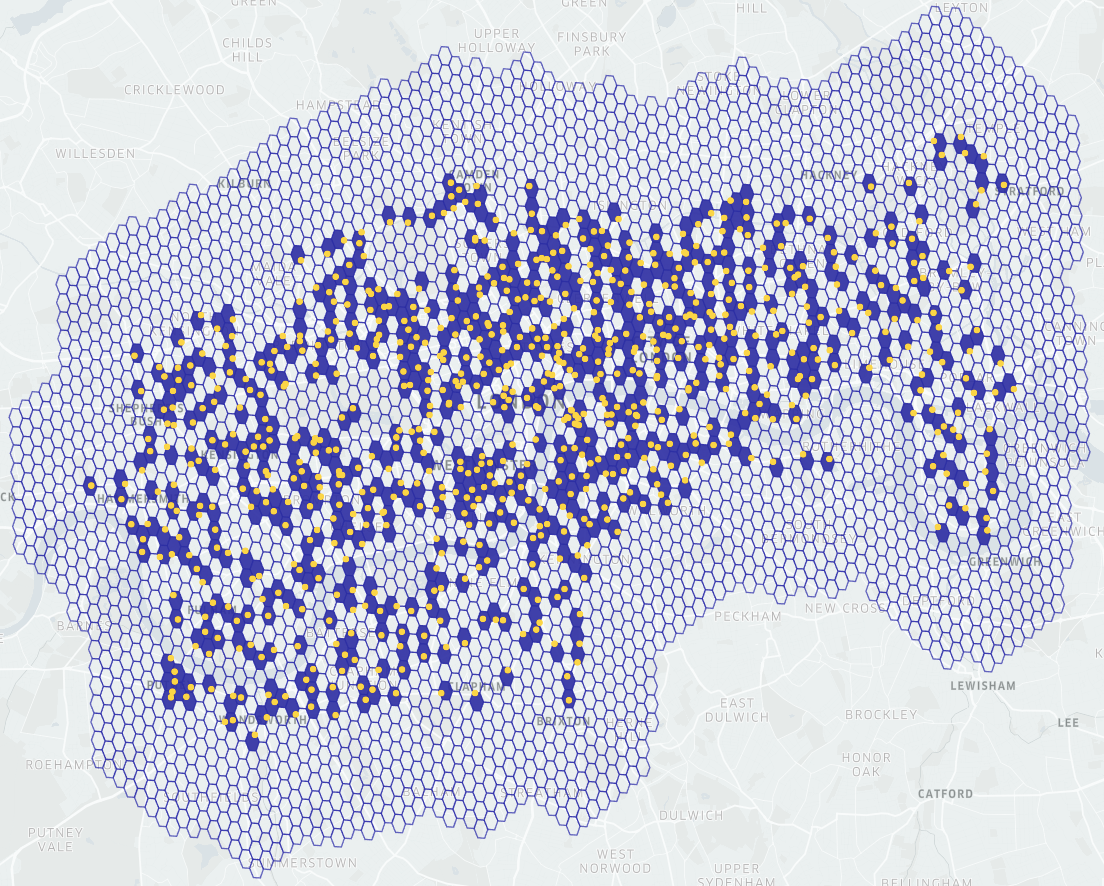}}
    \end{minipage}%
    \begin{minipage}{.5\textwidth}
        \centering
        \subfloat[]{\label{fig:london_h3_example:c}\includegraphics[width=0.9\linewidth, keepaspectratio]{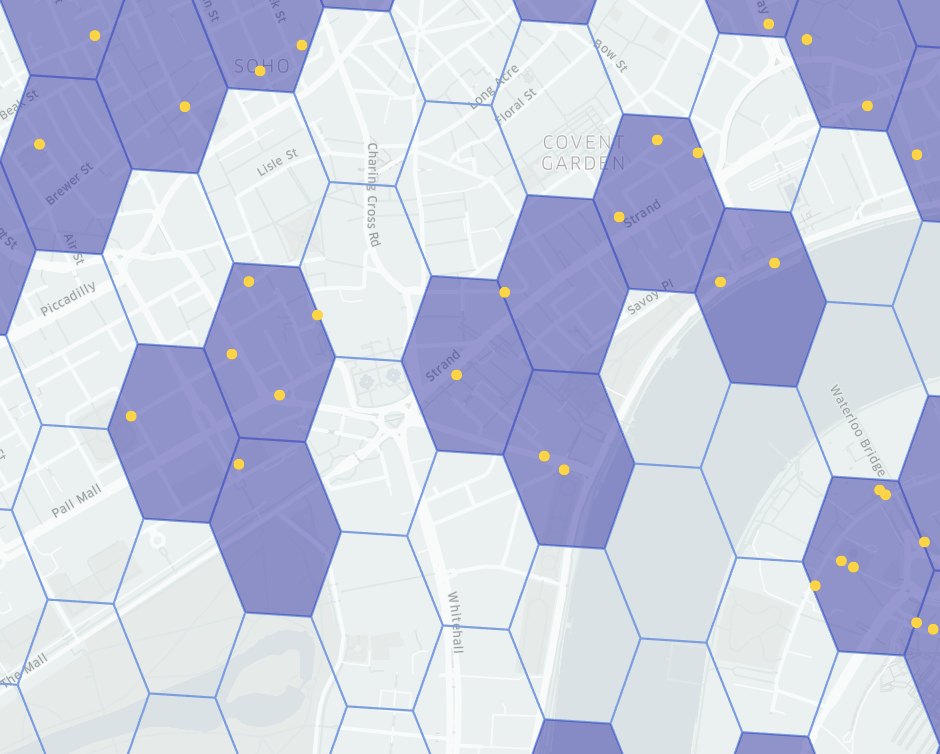}}
    \end{minipage}
    
    \caption[Data reduction and regional division using the H3 index on the example of Greater London]%
    {Data reduction and regional division using the H3 index on the example of Greater London administrative area. \par
    The figure \ref{fig:london_h3_example:a} shows the administrative area border with stations represented as teal dots. \par
    The figure \ref{fig:london_h3_example:b} shows all generated H3 hexagons of resolution 9 that are within 2 kilometres from any station. \par
    The figure \ref{fig:london_h3_example:c} is a close-up view of central London with generated hexagons. \par
    In the second and third figure, the hexagons that contain stations are shaded and the yellow dots indicate the stations. \par \small Personal work. Rendered using kepler.gl library.}
    \label{fig:london_h3_example}
\end{figure}

To download data from OpenStreetMap, administrative regions were used, which often included the city with its outskirts. As bicycle-sharing stations are often placed near city centres, the next step of data selection is to limit the number of relations downloaded. For each station in the city, microregions were generated at 3 resolutions within 2 km of the station. Then all geographical objects that intersected with any of the regions were filtered out. An example of this filtering is shown for the City of London in Figure \ref{fig:london_h3_example}.

After all operations, data from 34 cities from 17 European countries remained for the study. The total number of bike stations in these cities is 10,360 and the number of filtered objects from OpenStreetMap is 2,787,408. 103,878 microregions with resolution 9 (of which 9,304 contain stations), 638,319 microregions with resolution 10 (of which 10,218 contain stations) and 3,420,758 microregions with resolution 11 (of which 10,259 contain stations) were generated for the study.

\section{Used embedding methods}

To use spatial data in machine learning, as with words in natural language processing, it has to be transformed into numbers that can then be processed by machine learning algorithms. In this thesis, 4 different methods for building embedding vectors have been proposed and are described below.

\subsection{Category counting}

The first and simplest method. Vectorisation by counting is one of the basic methods used in natural language processing, where documents are converted into vectors containing information on how many times a particular word has occurred in the document. In this case, the occurrences of objects of a given category will be counted. These categories were developed on the basis of existing sub-groups available in the OSM documentation. In the library downloading the data, 20 main categories were determined: 
\begin{enumerate}
    \item \texttt{aerialway} - air transport elements like gondolas and cable cars;
    \item \texttt{airports} - air transport infrastructure;
    \item \texttt{buildings} - any buildings not included in other categories (like offices);
    \item \texttt{culture\_and\_entertainment} - cultural and entertainment facilities;
    \item \texttt{education} - education facilities from nurseries to university campuses;
    \item \texttt{emergency} - emergency facilities such as EDs, defibrillators and medical helicopter landing pads;
    \item \texttt{finances} - banks, exchange offices and ATMs;
    \item \texttt{healthcare} - all medical buildings and pharmacies;
    \item \texttt{historic} - historical sites such as ruins and historical monuments;
    \item \texttt{leisure} - leisure facilities;
    \item \texttt{other} - public buildings: orphanages, cemeteries, embassies, post offices, prisons, places of worship, police and fire stations, shopping centres, courts of law;
    \item \texttt{roads\_bike} - roads where bicycles can ride;
    \item \texttt{roads\_drive} - roads where motorised vehicles can ride;
    \item \texttt{roads\_walk} - roads where pedestrians can walk;
    \item \texttt{shops} - any shops;
    \item \texttt{sport} - sporting venues;
    \item \texttt{sustenance} - restaurants, cafes, bars, pubs;
    \item \texttt{tourism} - tourist facilities: hotels, lodging places, tourist attractions, museums, zoos, amusement parks;
    \item \texttt{transportation} - car parks, public transport stops, railway stations, also bicycle sharing stations (they were removed from the dataset so that the model would not learn on them);
    \item \texttt{water} - bodies of water, seas, beaches, rivers, canals.
\end{enumerate}

The final dimension of the vector is 20.

\subsection{Shape analysis per category}

To extend the first method and to take into account the shape of the surveyed objects, a method is proposed which analyses the polygon areas and the lengths of the road lines inside the region. 4 categories are treated specially: in case of \texttt{water} only the areas occupied by water inside the region are counted and in case of \texttt{roads\_bike}, \texttt{roads\_drive} and \texttt{roads\_walk} the sum of lengths of roads inside the region is counted. For the other 16 categories, the areas of objects inside the region are counted separately if they are marked as area and the number of occurrences if the object is marked as a point. The final dimension of the vector is 36 (16 times 2 plus 4 times 1).

\subsection{Shape analysis per all tags with dimensionality reduction}

\begin{figure}[H]
    \centering
    \includegraphics[width=\textwidth]{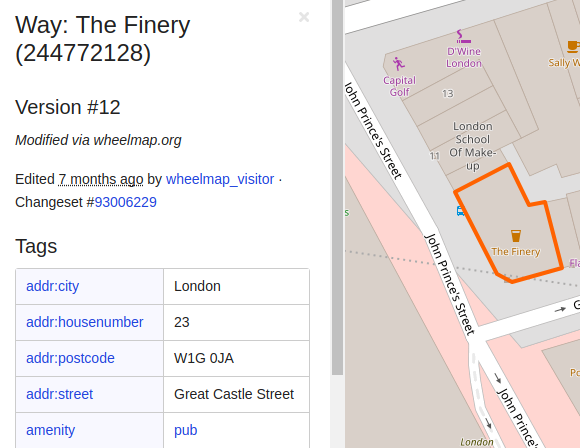}
    \caption[Example of a sustenance category POI]%
    {Example of a sustenance category POI. \par \small Source: OpenStreetMap.}
    \label{fig:the_finery_london}
\end{figure}

Objects in OpenStreetMap are mostly tagged according to rules that can be found in the guideline. However, nothing prevents the community to add their own new tags or new values for popular tags. For most of the 16 categories (except water and roads), it has been decided to choose a tag that uniquely identifies the second hierarchy of importance of the object. For example, the Figure \ref{fig:the_finery_london} shows a pub in central London that belongs to the \texttt{sustenance} category. Based on the amenity tag, it is possible to see in more detail which subcategory the object belongs to. Based on this second hierarchy, a vector of dimension 5702 is generated. However, based on different regions, this number can vary because it depends on all different values that users can add. This results in a very sparse vector. In order not to take this vector directly into the learning, an autoencoder was used to reduce its dimensionality to 300.

\subsection{Shape analysis per selected tags with dimensionality reduction}

The method works similarly to the previous one, however, to be independent of various tags added by the community, it was decided to limit the set of values that are taken into account while building the vector. Thanks to this, a vector of dimension 888 is built. Objects that do not fit into any of the most popular categories are rejected from the analysis. The full list of used tags can be found in the Table \ref{tab:app-osm-tags} in Appendix \ref{osm_tags}. For this method, it was also decided to use an autoencoder for dimensionality reduction and the following dimensions are examined: 20, 32, 50, 64, 100, 128, 200, 256, 300 and 500.

\section{Used neighbourhood embedding methods}

\begin{figure}[H]
    \centering
    \includegraphics[width=\textwidth]{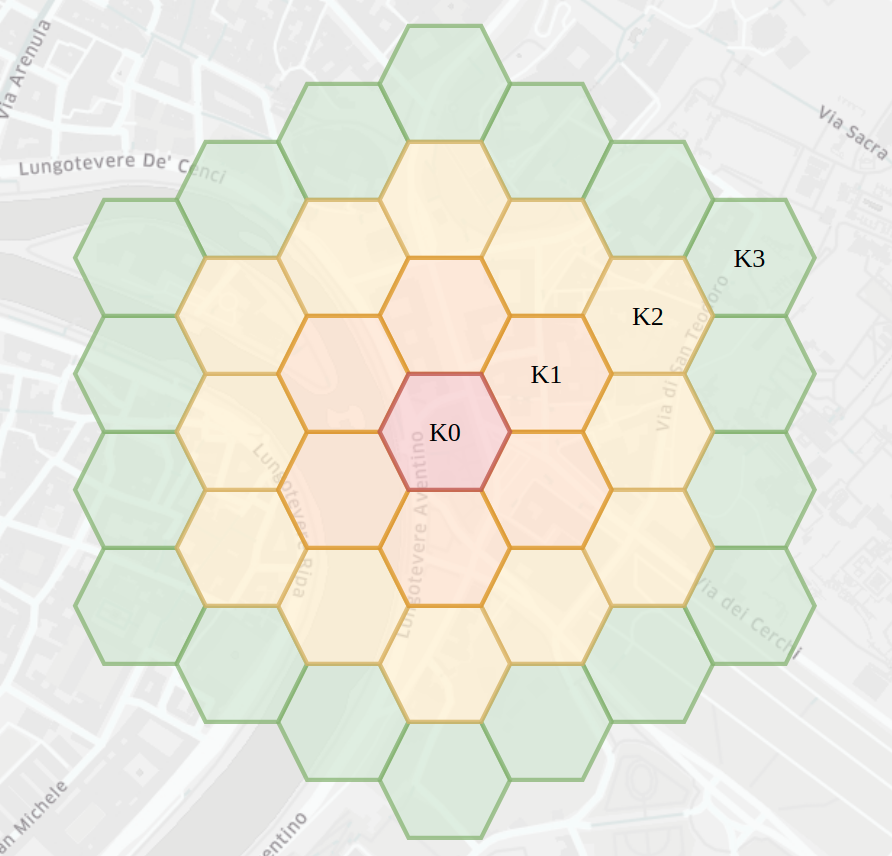}
    \caption[Example of different sizes of regions neighbourhood]%
    {Example of different sizes of regions neighbourhood. Each neighbourhood ring has been coloured and labelled correspondingly. For each ring, vectors obtained for all regions are averaged. \par \small Personal work. Rendered using kepler.gl library.}
    \label{fig:neighbourhood_example}
\end{figure}

To provide the classifier with more context for making predictions, it was decided that the embedding vector of a particular region would also contain information about the neighbourhood of that region. The neighbourhood of a region includes hexagons adjacent to the hexagon under study. In this paper, the neighbourhood size parameter will be investigated, which translates to the number of rings of consecutive hexagons surrounding the region under study. An illustrative diagram showing the neighbourhoods can be found in Figure \ref{fig:neighbourhood_example}. To facilitate the method's operation, it was decided that the embedding vectors generated for all regions included in one neighbourhood ring will be averaged and only the vectors of particular neighbourhood rings will be treated separately. Four methods of combining embedding vectors have been proposed and are described below.

\subsection{Concatenation}

A basic method that involves combining vectors of consecutive neighbourhoods with each other. On the one hand, this preserves all information, but on the other hand, it increases the computational complexity as the size of the neighbourhood considered increases. The final vector is of dimension $N\cdot (K+1)$ where $N$ is the dimension of the single region vector and $K$ is the size of the neighbourhood.

\subsection{Averaging}

The second base method takes all vectors for consecutive neighbourhoods and averages them together with the vector generated for the region under study. This ensures that the vector has a constant dimension regardless of the size of the neighbourhood, but with more rings of neighbours taken into account, the noise may increase as all vectors are averaged equally.

\subsection{Diminishing averaging}

Extending the previous method by adding a weighting aspect. The vector for the study region is assigned a weight of 1 and each subsequent neighbourhood is assigned a weight of $1 / (k+1)$ where $k$ is the distance of the ring of neighbours from the centre. Finally, the vectors are averaged taking into account the assigned weights, so that further neighbourhoods have less influence on the change in values.

\subsection{Diminishing averaging squared}

Another variant of weighted averaging, but this time the assigned weight for successive neighbourhoods equals $1 / (k+1)^2$. This was proposed to test which version of weight assignment might have a better impact on the predictions made by the proposed method.

There are also other possible options for assigning weights for averaging such as $1 / (k+1)^3$ (to see if it would work better than squaring) or $1 / 6\cdot (k+1)$ (which takes into account that there are 6 more regions in each neighbourhood ring than in the previous one), but these were discarded to limit the set of hyperparameters studied and may be explored in the future.

\section{Used classification methods}
\subsection{Base classifiers}

For the analysis of the base classifiers it was decided to choose 4 algorithms: k-nearest neighbours, Radial Basis Function (RBF) kernel SVM, random forest and AdaBoost metaclassifier. The performance of each of these classifiers in the station occurrence prediction task will be tested with attention to the mean and variance of the results.

\subsection{Neural Network}

In addition to the base classifiers, the performance of a neural network-based classifier will be checked to see if it performs better than the base classifiers mentioned above. Additionally, the embedding vectors coming out of the autoencoder will be checked only on the neural network without introducing their base classifiers.

\section{Used classification performance metrics}

Most of the experiments will use 3 metrics that will test the quality of the predictions made by the method. These three metrics include accuracy, the F1 score measure, and a custom metric designed to not penalize the model for predicting too many regions that would contain stations. The method assigns a value of 1 for TP (true positive) and TN (true negative) predictions and a value of 0 for FN (false negative), but for FP (false positive) predictions, it assigns a value of $1 / (k + 1)$ where $k$ is the distance in hex to the nearest region with a station. That is, if the model predicts that the immediate neighbour of the region containing the station should also contain it, it does not get a score of 0 but $1 / (1 + 1)$ i.e. 0.5.

In the final experiments, 3 further metrics will also be investigated which will give a better idea of the quality of the method's performance. These include precision, recall and balanced accuracy.

\section{The class imbalance problem}

The resulting dataset contains only (depending on the resolution tested) between $2.99\cdot 10^{-3}$ and $8.95\cdot 10^{-2}$ regions that contain stations. If the model was taught on the whole set, it could turn out that there are big problems in the prediction of station occurrence, because more than 99\% of regions are examples without stations. For this purpose, limiting the set of negative examples will be used during learning and the effect of the ratio between positive and negative classes on the quality of prediction and discrimination of regions will be investigated. When building the learning set for a given city, all regions containing stations will be used and an appropriate number of regions that do not contain stations will be drawn at random. Values between 1 and 5 will be examined and compared to see how they affect the method results and the meaningfulness of the results obtained as a heat map.

\section{Software}

The retrieved data from OpenStreetMap was stored in a MongoDB \cite{mongodb} database with spatial indexes to speed up the queries. Custom scripts written in Python were used for the experiments. From the most important libraries used, these can be mentioned: Scikit-learn \cite{scikit-learn} - most of the classifiers and data processing methods used for the experiments came from this library, NumPy \cite{harris2020array} - a library allowing to process numerical data in the form of vectors and matrices, TensorFlow \cite{tensorflow2015-whitepaper} - allowed to implement simple neural networks and an autoencoder used to build embeddings, Shapely \cite{shapely} - allowed to perform intersection operations and analysis of object shapes, pandas \cite{reback2020pandas} and GeoPandas \cite{kelsey_jordahl_2020_3946761} - libraries allowing to work with datasets, also those based on geographical data, h3-py \cite{uberh3} - library allowing to generate regions from H3 system and do operations connected with neighbourhoods, PyMongo \cite{pymongo} - library allowing to connect to MongoDB databases, matplotlib \cite{Hunter:2007} and seaborn \cite{Waskom2021} - libraries for generating graphs of results, Kepler.gl \cite{keplergl} - tool allowing to display geographical data on maps.

\chapter{Specific methods and results}
\label{ch:results}

This chapter will summarise the work done to analyse the dataset studied and the results of all the experiments carried out in the form of 7 research questions. At the end of the chapter the results of the method for 4 European cities of different sizes, which do not have a bicycle sharing system, are presented. 

\section{Exploratory data analysis}

An exploratory analysis of the data will allow the characteristics of the datasets from the studied cities, their similarities and differences to be recognised. By analysing station layouts in individual cities, comparisons can be made between systems and
the analysis of station surroundings can uncover their influence on station position planning in cities and possibly generalise these premises.

\subsection{EA1: Average density of bicycle-sharing stations per city}

\begin{figure}[H]
    \centering
    \includegraphics[width=\textwidth]{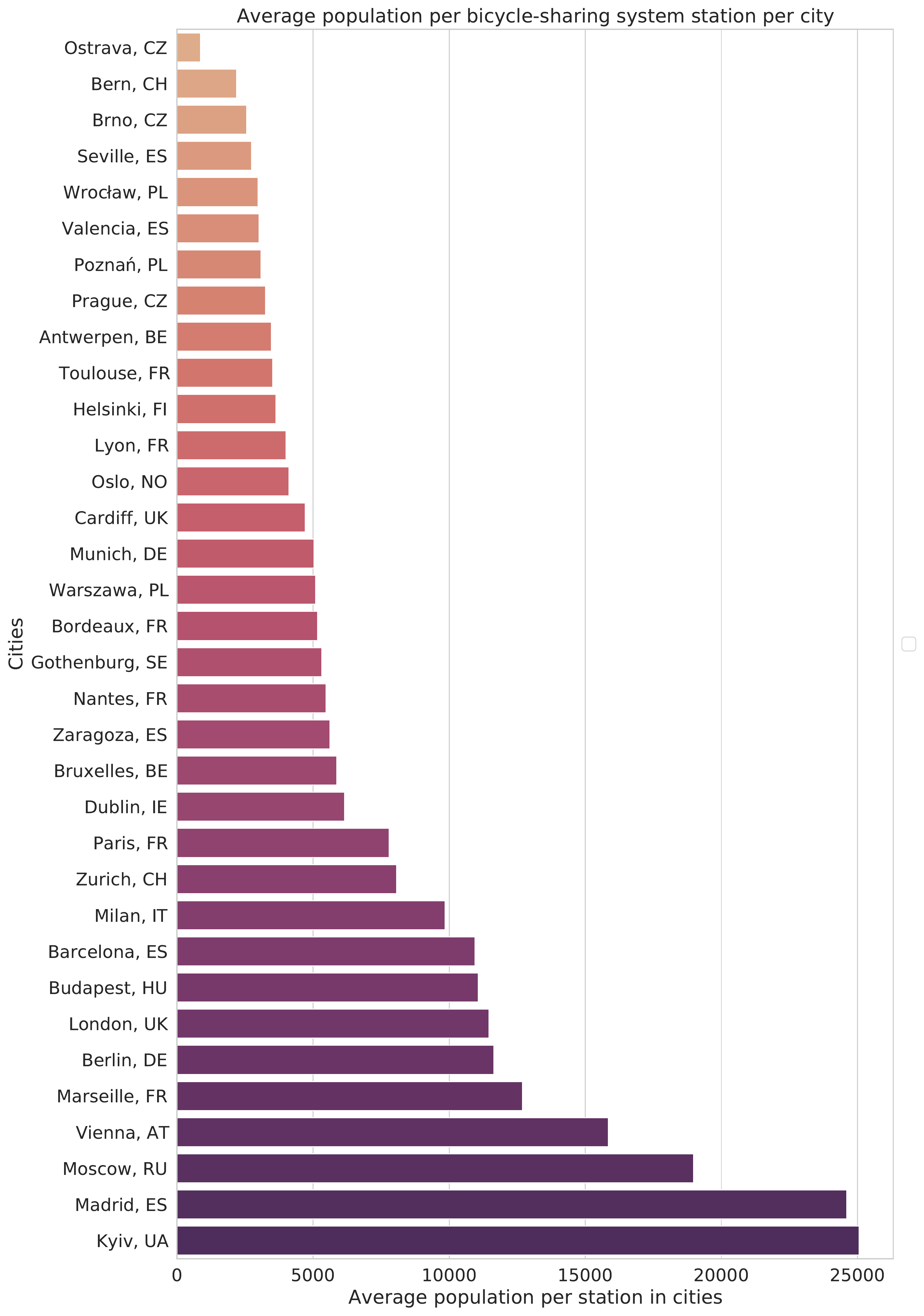}
    \caption[Average population per bicycle-sharing system stations per city]%
    {Average population per bicycle-sharing system stations per city. The smaller the value, the more stations there are per capita. \par \small Personal work. Demographic data: United Nations, Department of Economic and Social Affairs, Population Division (2018). World Urbanization Prospects: The 2018 Revision, Online Edition.}
    \label{chart:stations_density_per_city_accepted}
\end{figure}

\begin{figure}[H]
    \centering
    \includegraphics[width=0.8\textwidth]{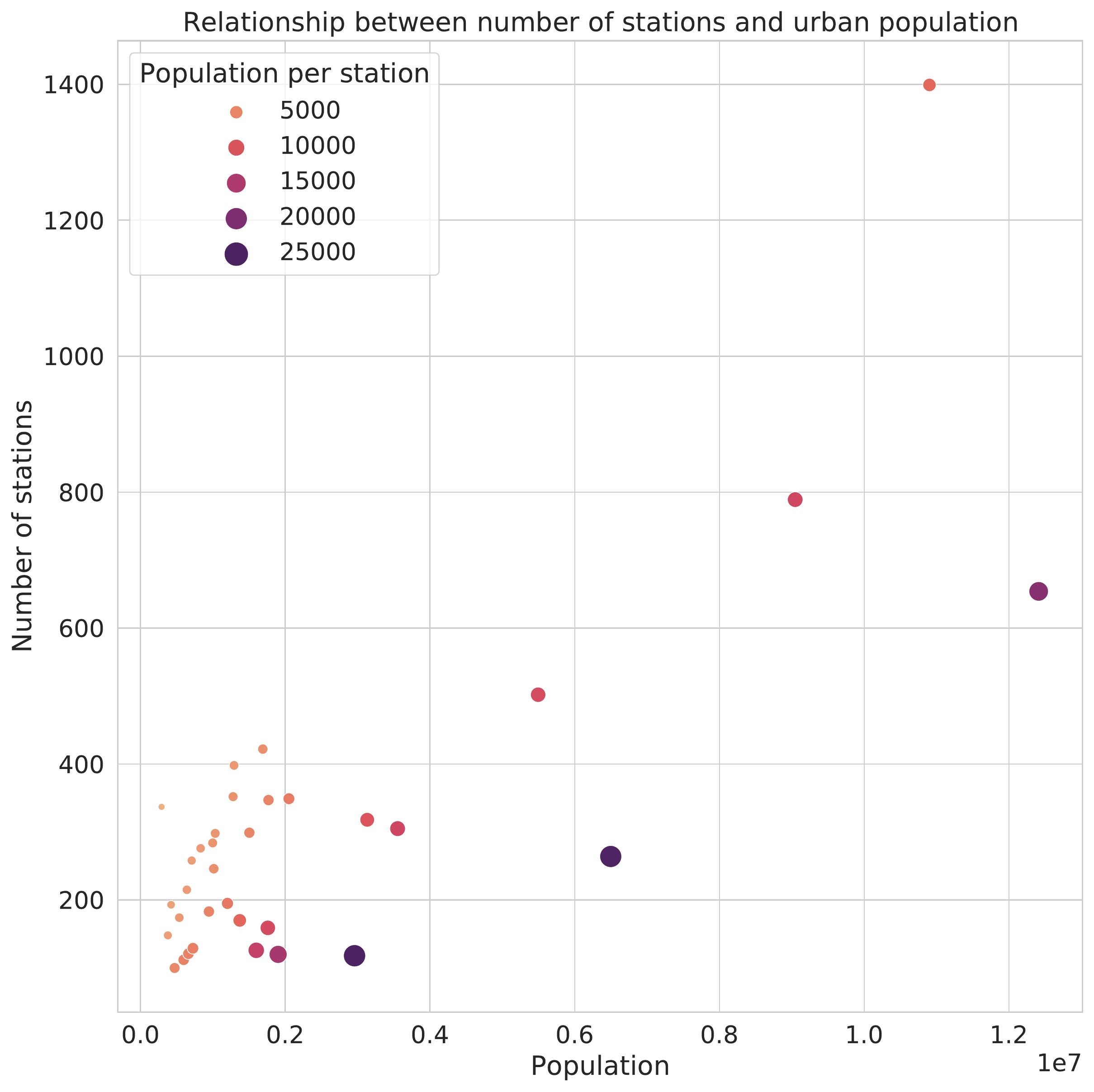}
    \caption[Relationship between the number of stations and urban population]%
    {Relationship between the number of stations and urban population. The population scale represents 10 million. \par \small Personal work.}
    \label{chart:stations_population_scatterplot}
\end{figure}

Determining the relationship between the size of the city (in terms of population) and the number of stations will allow recognition of whether there are similarities between cities of similar size and whether there is a correlation between population and the number of stations located in the city. For this purpose, a simple measure was determined for each city - the average population per station. Graphs visualising this measure were then generated and are presented in Figures \ref{chart:stations_density_per_city_accepted} and \ref{chart:stations_population_scatterplot}.

Based on this measure a clear division into 3 groups can be observed: cities with less than approx. 5000 inhabitants per station (Ostrava, Bern, Brno, Seville, Wrocław, Valencia, Poznań, Prague, Antwerp, Toulouse, Helsinki, Lyon, Oslo, Cardiff, Munich, Warsaw, Bordeaux, Gothenburg, Nantes, Zaragoza, Brussels, Dublin), cities with approx. 6 to 12 thousand inhabitants per station (Paris, Zurich, Milan, Barcelona, Budapest, London, Berlin, Marseille) and cities with more than 15 thousand inhabitants per station (Vienna, Moscow, Madrid, Kyiv). Additionally, the city of Ostrava stands out as it has a very large number of stations (337) for its small population of about 285,000 inhabitants.

The scatter plot in the Figure \ref{chart:stations_population_scatterplot} shows that most of the surveyed cities have less than 2 million inhabitants and less than 400 stations. The cluster of cities with more than 15,000 inhabitants per station can be clearly seen at the bottom of the graph as a slowly increasing line to the point representing Moscow for 654 stations and 12.5 million inhabitants. Another cluster consists of a diagonal line from the main cluster up to a point representing Paris for 1399 stations and almost 11 million inhabitants.

\subsection{EA2: Distribution of POIs categories in different cities}

To know the obtained dataset from OpenStreetMap, it is useful to determine the distribution of POI categories. A list of those categories has been discussed in a previous chapter.

\begin{figure}[H]
    \centering
    \includegraphics[width=\textwidth]{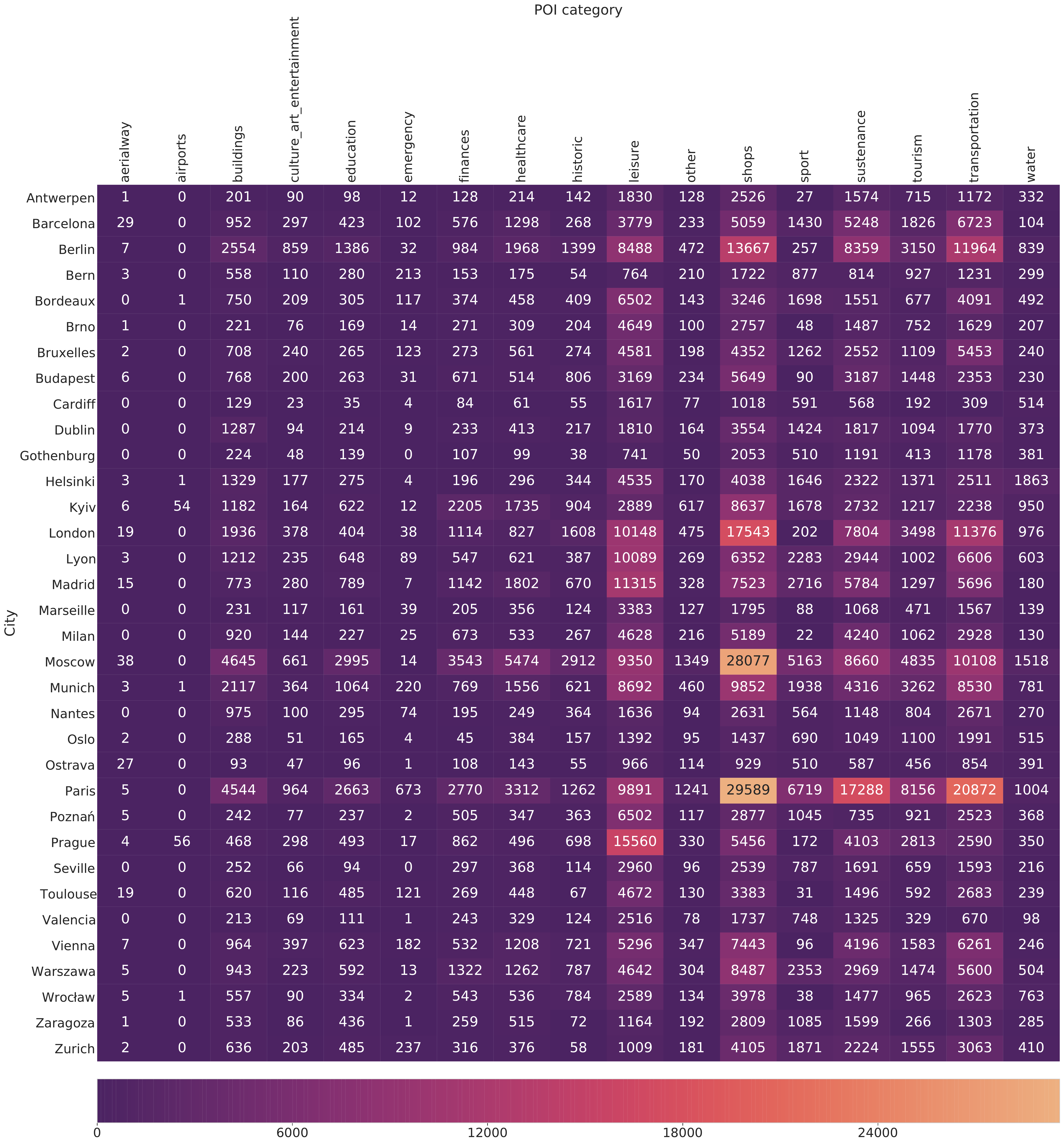}
    \caption[Total number of OSM objects in each category in a researched region]%
    {Total number of OSM objects in each category in a researched region. Roads have been omitted for brevity. \par \small Personal work.}
    \label{chart:osm_objects_total}
\end{figure}

Distributions of POIs categories for all 34 selected cities have been visualised in form of a heat map in Figure \ref{chart:osm_objects_total}. Numbers represent the total number of objects in each of the selected hexes of resolution 9. This can create a situation where one object crosses multiple regions and is counted several times, but the hexagons of resolution 9 are large enough that this is not significant overall.

\begin{figure}[H]
    \centering
    \includegraphics[width=\textwidth]{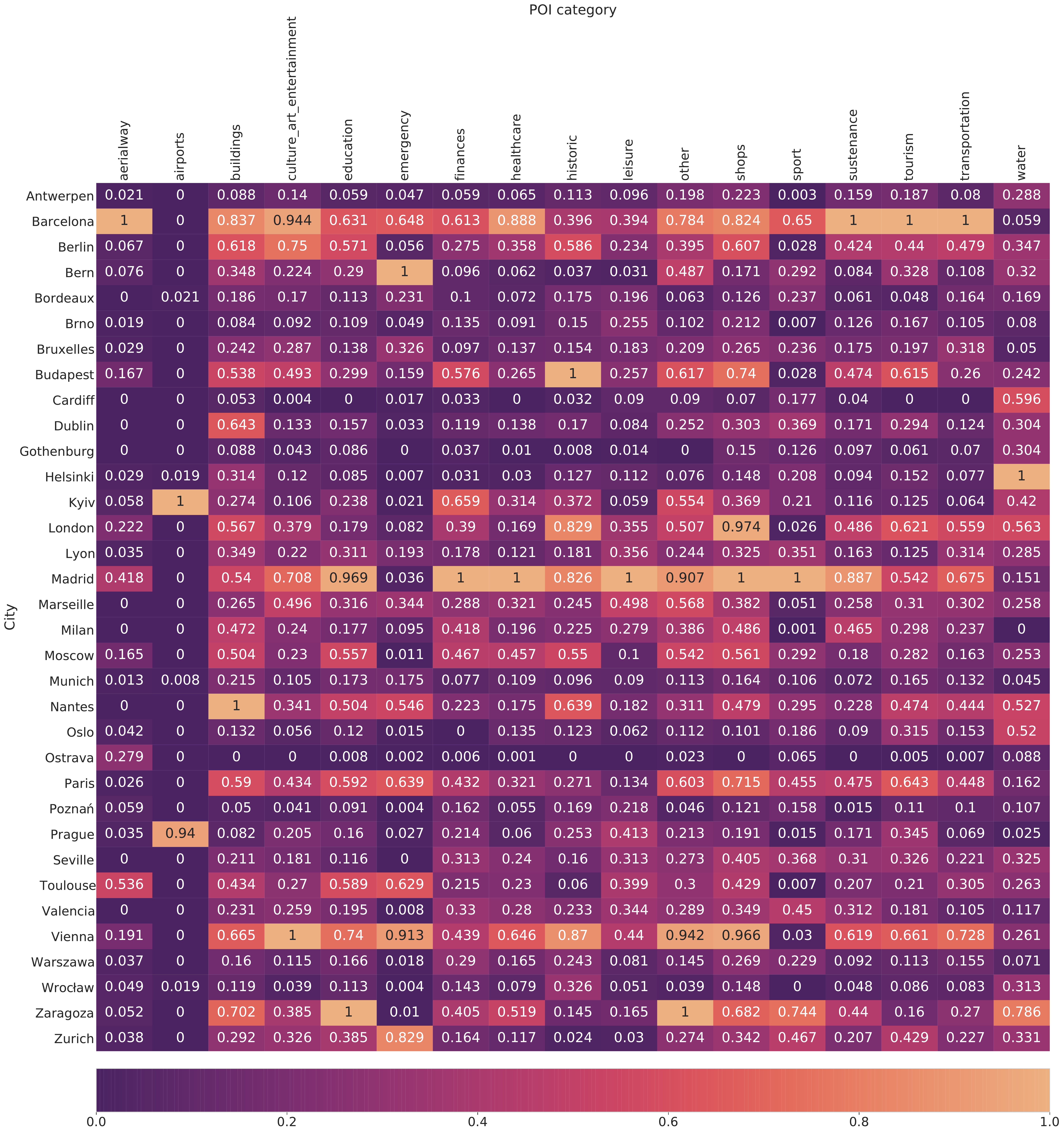}
    \caption[Normalised average number of OSM objects per micro-region in a researched region]%
    {Normalised average number of OSM objects per micro-region in a researched region. Roads have been omitted for brevity. \par \small Personal work.}
    \label{chart:osm_objects_normalised}
\end{figure}

From Figure \ref{chart:osm_objects_total}, it can be seen that the 4 categories are the most numerous: \texttt{shops}, \texttt{transportation}, \texttt{sustenance} and \texttt{leisure}. There are large differences in numbers between some of the categories, which is natural given the categories proposed but makes it difficult to assess how each category is distributed across cities. Therefore, Figure \ref{chart:osm_objects_normalised} shows normalised data. Normalisation was done by dividing the obtained values by the number of resolution hexagons 9 and then normalising min-max per category. 

\newpage

After normalisation, it is immediately apparent which cities are very densely saturated compared to other cities. These are mainly Spanish cities: Barcelona, Madrid, Zaragoza. The Austrian capital Vienna also stands out in this respect. We can also see cities that have a very low density of OSM facilities in comparison with other cities surveyed. Antwerp, Cardiff, Gothenburg, and Ostrava stand out in this respect.

The result of this situation can be both the difference in buildings, which can be dense or sparse, and the quality of the object tagging itself in OpenStreetMap.

\subsection{EA3: Comparison of bicycle-sharing station surroundings}

The comparison of station neighbourhoods will determine whether generalizable and explainable types of neighbourhoods that occur around stations can be determined. The presence of such types may mean that the information contained in the OSM data is sufficient to determine what function the region in which the station is located serves and ultimately to distinguish the region in which the station is located from that in which it is not.

The UMAP dimensionality reduction technique \cite{McInnes2018} will be used to compare station neighbourhoods and the HDBSCAN method \cite{McInnes2017} will be used to determine clusters. All resolution 9 hexagons that contain stations from all 34 cities will be used for clustering. The number of individual objects per category will be used as values, as in the previous step. Before dimensionality reduction, the columns will be normalised using the min-max method.

\begin{figure}[H]
    \centering
    \includegraphics[width=0.7\textwidth]{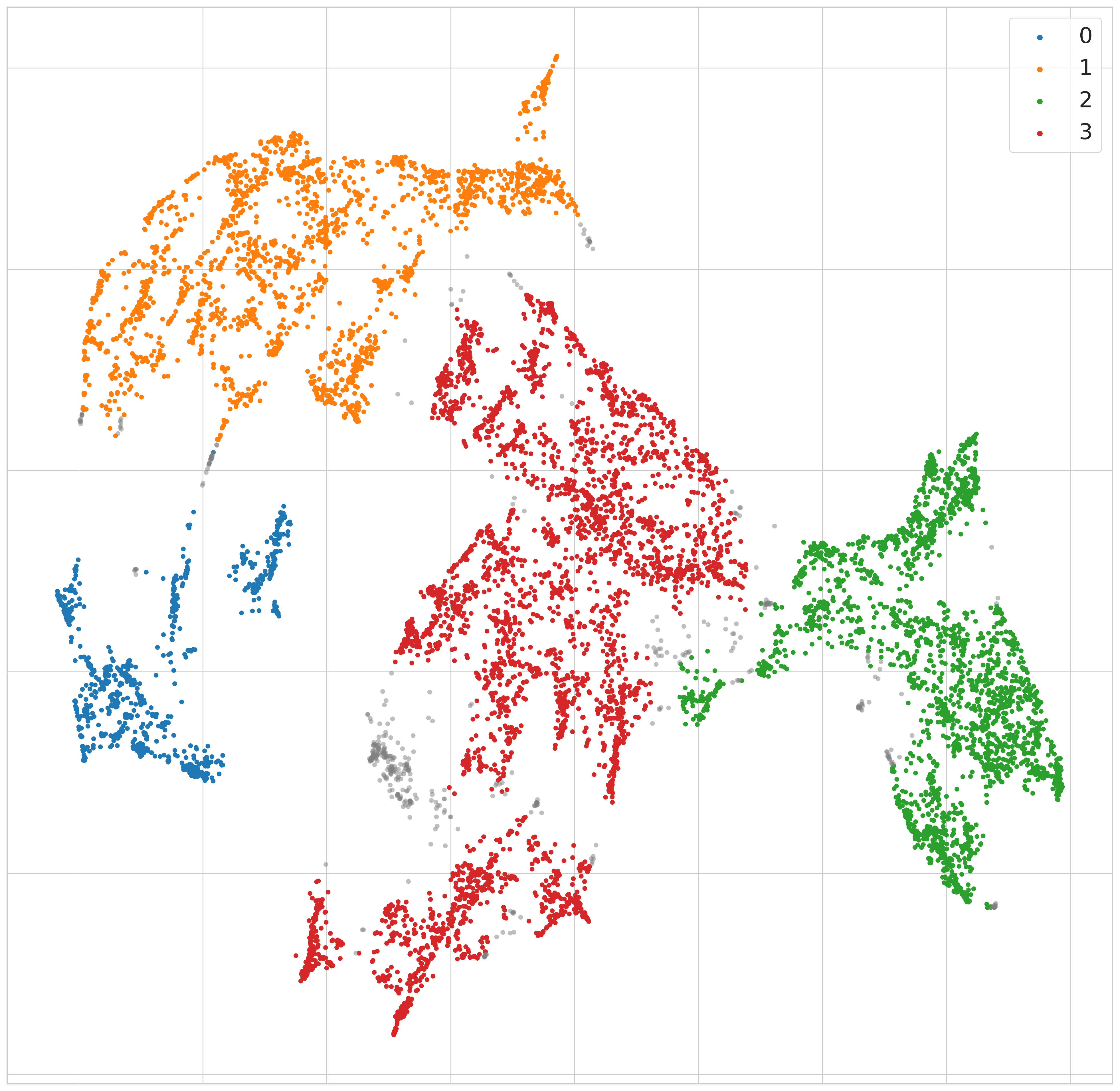}
    \caption[Two-dimensional clustered mapping of all regions with stations using the UMAP and HDBSCAN techniques]%
    {Two-dimensional clustered mapping of all regions with stations (n = 9304) using the UMAP and HDBSCAN techniques. Light-grey dots represent regions that haven't been assigned to any cluster. \par \small Personal work.}
    \label{chart:umap_clusters}
\end{figure}

\begin{figure}[H]
     \centering
    \begin{minipage}{.5\textwidth}
        \centering
        \subfloat{\label{chart:clusters_osm_distribution:1}\includegraphics[width=\linewidth, keepaspectratio]{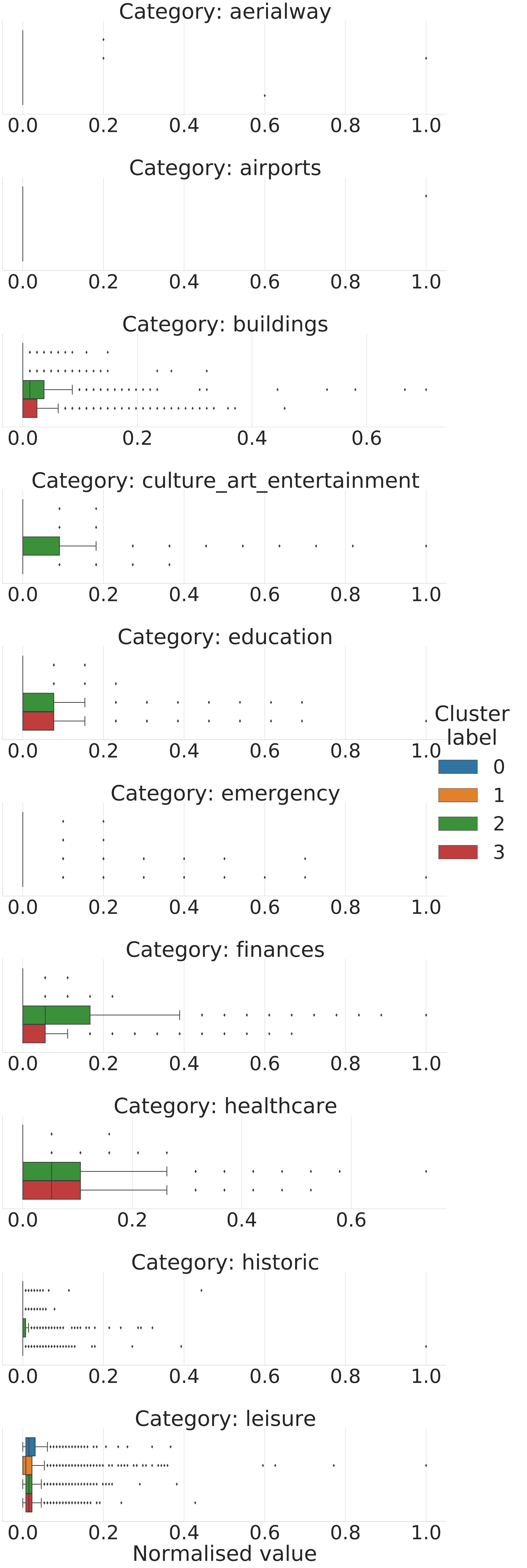}}
    \end{minipage}%
    \begin{minipage}{.5\textwidth}
        \centering
        \subfloat{\label{chart:clusters_osm_distribution:2}\includegraphics[width=\linewidth, keepaspectratio]{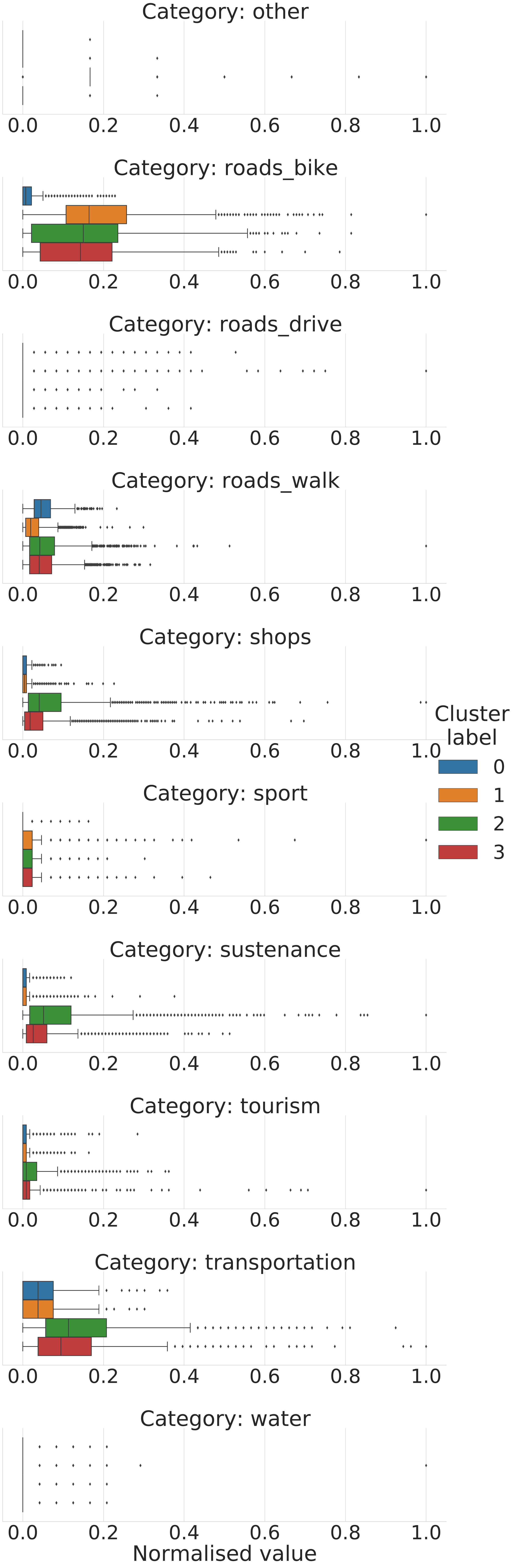}}
    \end{minipage}
    \caption[Normalised distribution of numbers of objects per category per cluster]%
    {Normalised distribution of numbers of objects per category per cluster. \par \small Personal work.}
    \label{chart:clusters_osm_distribution}
\end{figure}

After selecting the appropriate reduction and clustering parameters, 4 clusters were obtained, which can be seen in Figure \ref{chart:umap_clusters}. Although smaller clusters could have been obtained, it was decided to stay with the value 4 for readability. The exact distribution of values per category can be found in Figure \ref{chart:clusters_osm_distribution}.

\newpage

The distribution of values seen in the Figure \ref{chart:clusters_osm_distribution} shows clear differences between the different clusters. Most categories show a clear division into a minimum of two clusters, which means that a deeper division is possible, but this is not the purpose of this thesis. The clear differences (but also similarities) between the regions containing stations indicate firstly that the data available in OSM is sufficient to delineate similar and dissimilar subgroups of regions, and secondly that the machine learning model will be able to learn the relationships derived from these data and be able to recognise differences between regions. 

\begin{figure}[H]
    \centering
    \begin{minipage}{.5\textwidth}
        \centering
        \subfloat[Berlin, Germany]{\label{fig:clusters_cities:b}\includegraphics[width=0.95\linewidth, keepaspectratio]{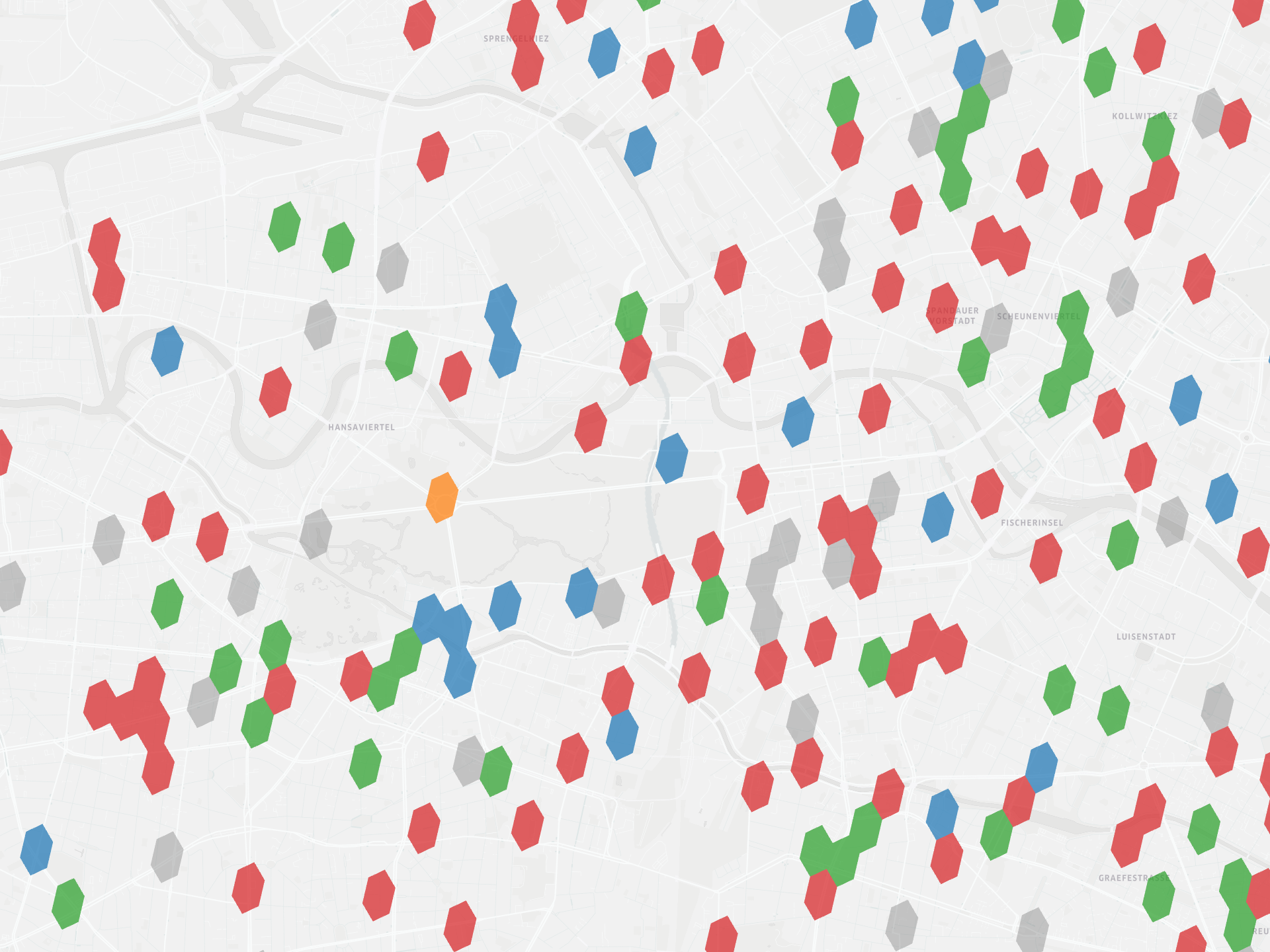}}
    \end{minipage}%
    \begin{minipage}{.5\textwidth}
        \centering
        \subfloat[London, UK]{\label{fig:clusters_cities:l}\includegraphics[width=0.95\linewidth, keepaspectratio]{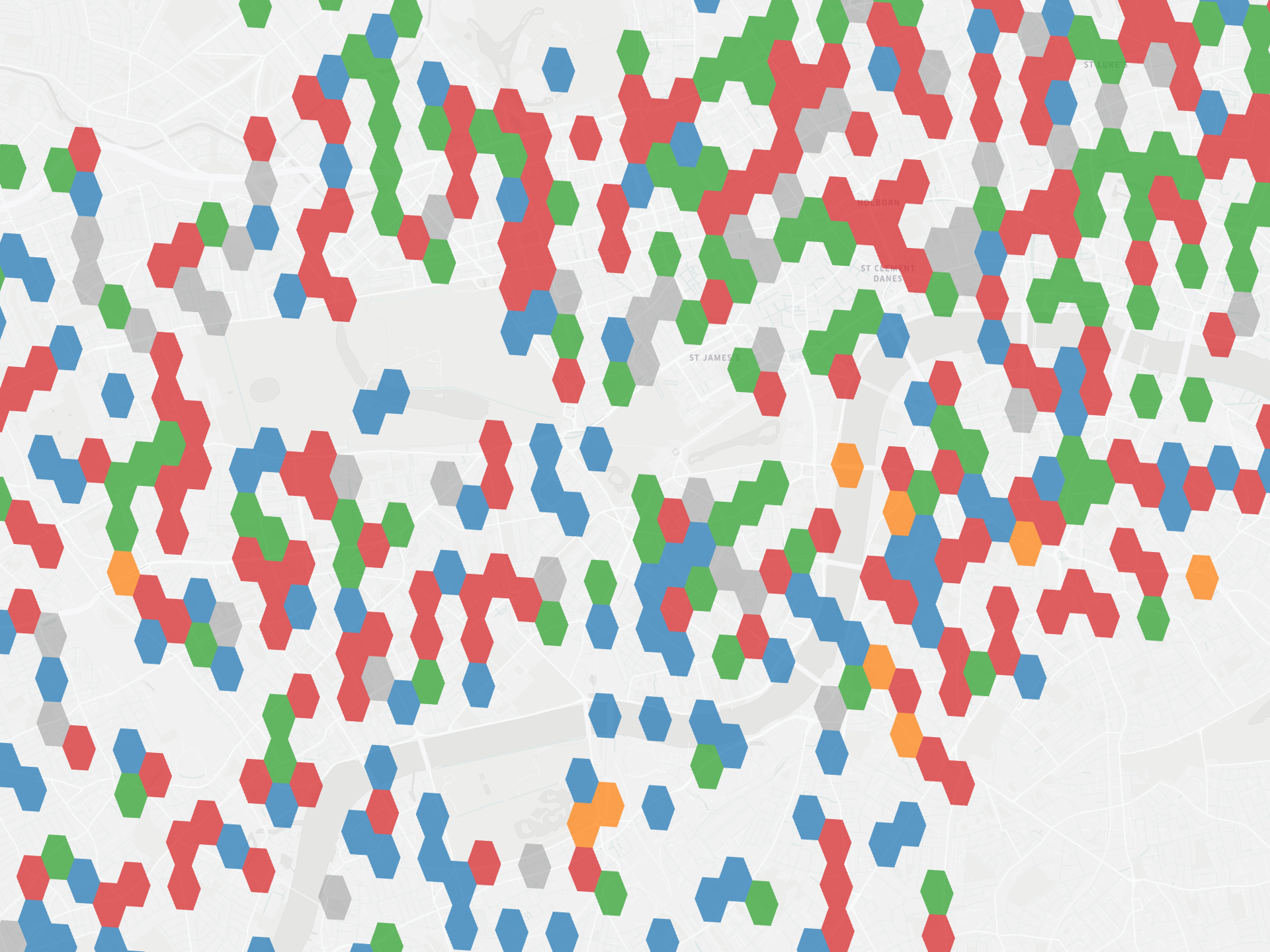}}
    \end{minipage}
    \par
    \begin{minipage}{.5\textwidth}
        \centering
        \subfloat[Paris, France]{\label{fig:clusters_cities:p}\includegraphics[width=0.95\linewidth, keepaspectratio]{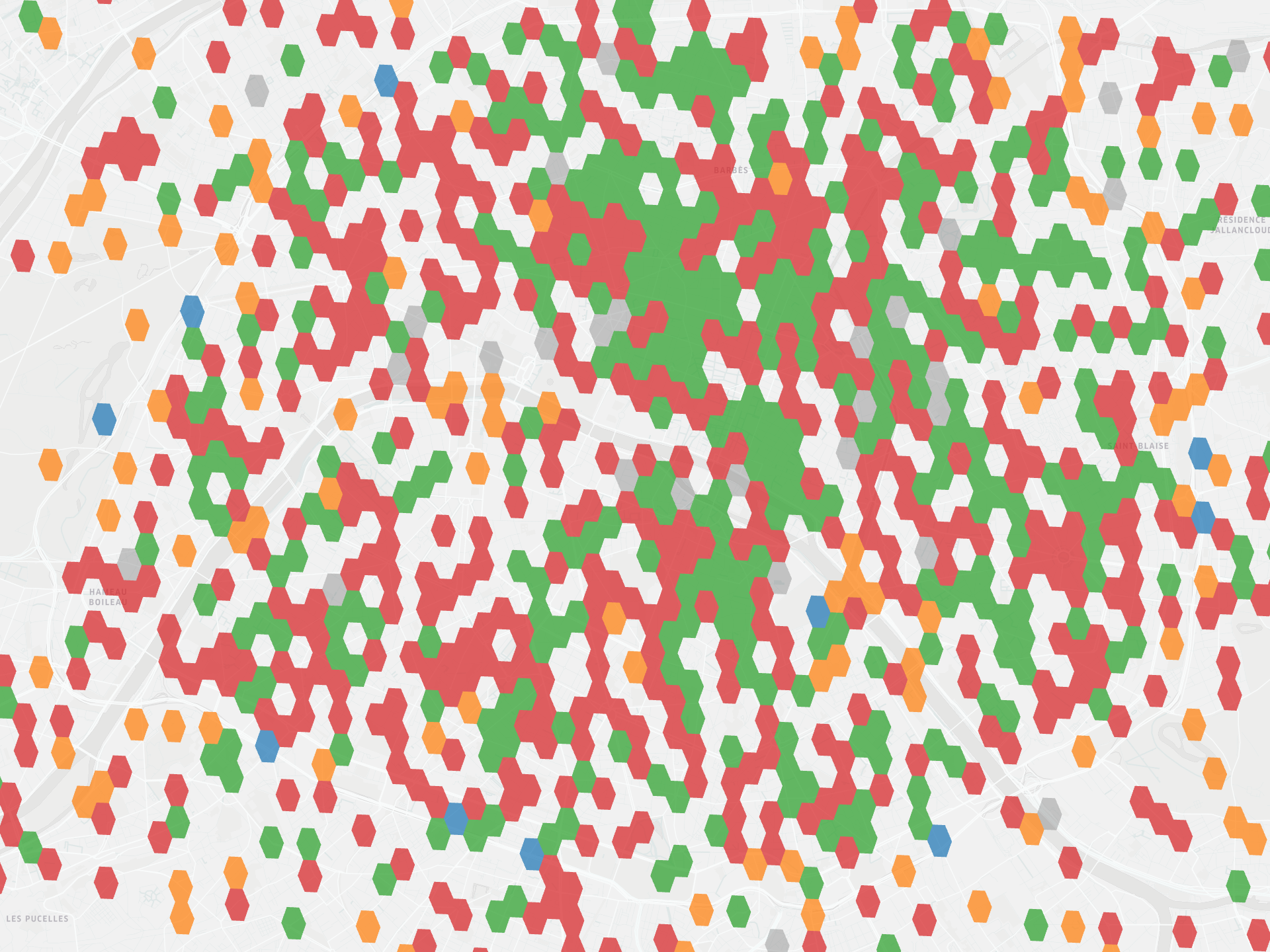}}
    \end{minipage}%
    \begin{minipage}{.5\textwidth}
        \centering
        \subfloat[Wrocław, Poland]{\label{fig:clusters_cities:w}\includegraphics[width=0.95\linewidth, keepaspectratio]{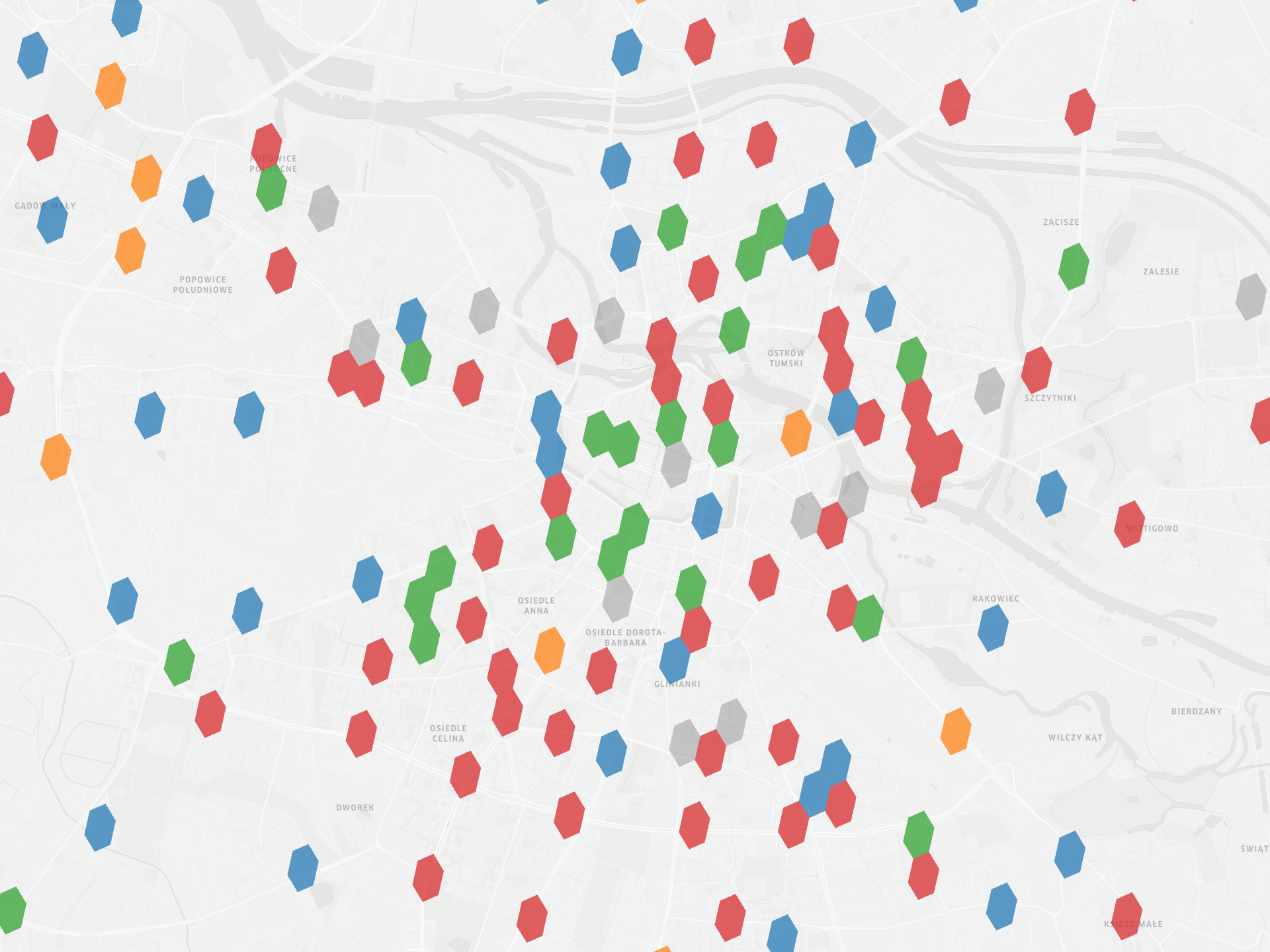}}
    \end{minipage}

    \caption[Example city centres with clustered regions containing stations]%
    {Example city centres with clustered regions containing stations. Light-grey hexagons represent regions that haven't been assigned to any cluster. \par \small Personal work. Rendered using kepler.gl library.}
    \label{fig:clusters_cities}
\end{figure}

The figure \ref{fig:clusters_cities} shows examples of clustered regions for 4 cities.

\newpage

\section{Conclusion of exploratory analysis}
\subsection{EA1: Average density of bicycle-sharing stations per city}

Determining the station density in cities by calculating the average population per station allowed to discover clear differences between some bicycle-sharing systems implemented in Europe. Most of the surveyed cities have less than 2 million inhabitants and their bicycle sharing systems have less than 400 stations. This may indicate that these cities have quite similar systems in terms of size and the method may work better among cities in this subgroup than in other cities that are clearly outliers, which will be investigated in research question 7.

\subsection{EA2: Distribution of POIs categories in different cities}

By determining the average number of OSM objects per studied region, it was possible to know the characteristics of the obtained set. It indicates that the selected cities are clearly different from each other and it cannot be stated unequivocally whether this is due to the construction and urban planning of a given city, or possible errors in tagging objects by users. The set shows the differences between cities that have on average very few objects per category and those which have very many. As the method will be taught mainly on data coming from one city only, there is no risk that differences in distribution will influence the quality of learning. Research question 6 will investigate the quality of the model performance per city for a subgroup of 11 cities and research question 7 will present the total cross-validation between all cities.

\subsection{EA3: Comparison of bicycle-sharing station surroundings}

An attempt to cluster bicycle station neighbourhoods based only on count data of different categories showed that this information is sufficient to make a clear and explainable division into subgroups. This means that the regions where the stations are located fulfil different functions in the urban context. However, this analysis does not imply that the regions without stations are clearly different from the regions that contain stations and that the method will be able to easily distinguish them from each other. However, as bicycle sharing stations are not placed randomly, but at strategic points, these regions are important for urban transport and it is hoped that from the OSM data alone the method will be able to deduce the logic behind the choice of the position of different stations. The impact of the method embedding OSM data, the size of the resolution, and the size of the neighbourhood around the region, on the quality of the method's performance will be investigated in research questions 3 and 6 respectively.

\section{RQ1: How different baseline classifiers perform in the station presence prediction task?}

Before investigating the precise effect of individual parameters on the prediction results, it is important to select a suitable classifier for further study. Four implementations from the popular scikit-learn library were selected for comparison: k-nearest neighbours, support vector machine with radial-basis function kernel, random forest, and AdaBoost metaclassifier. Each classifier adapts differently to the trained domain, so this question will select the best one for further study.

\subsection{Method of RQ1}

Averaged results from 10 iterations for the test set per parameter configuration were used to test the prediction quality of individual classifiers. The results were calculated using data from 3 cities separately: Poznań, Warsaw and Wrocław. All regions containing stations were used for learning and the same number of regions not containing stations were sampled (imbalance ratio = 1). For feature generation, 2 region embedding methods were used: Category counting and Shape analysis per category and 4 methods for embedding neighbourhoods: Concatenation, Averaging, Diminishing Averaging and Squared Diminishing Averaging. No feature processing methods were used on the data. The quality of prediction was tested by the 3 metrics mentioned in the previous section.

\subsection{Results of RQ1}

\begin{figure}[H]
    \centering
    \includegraphics[width=0.9\textwidth]{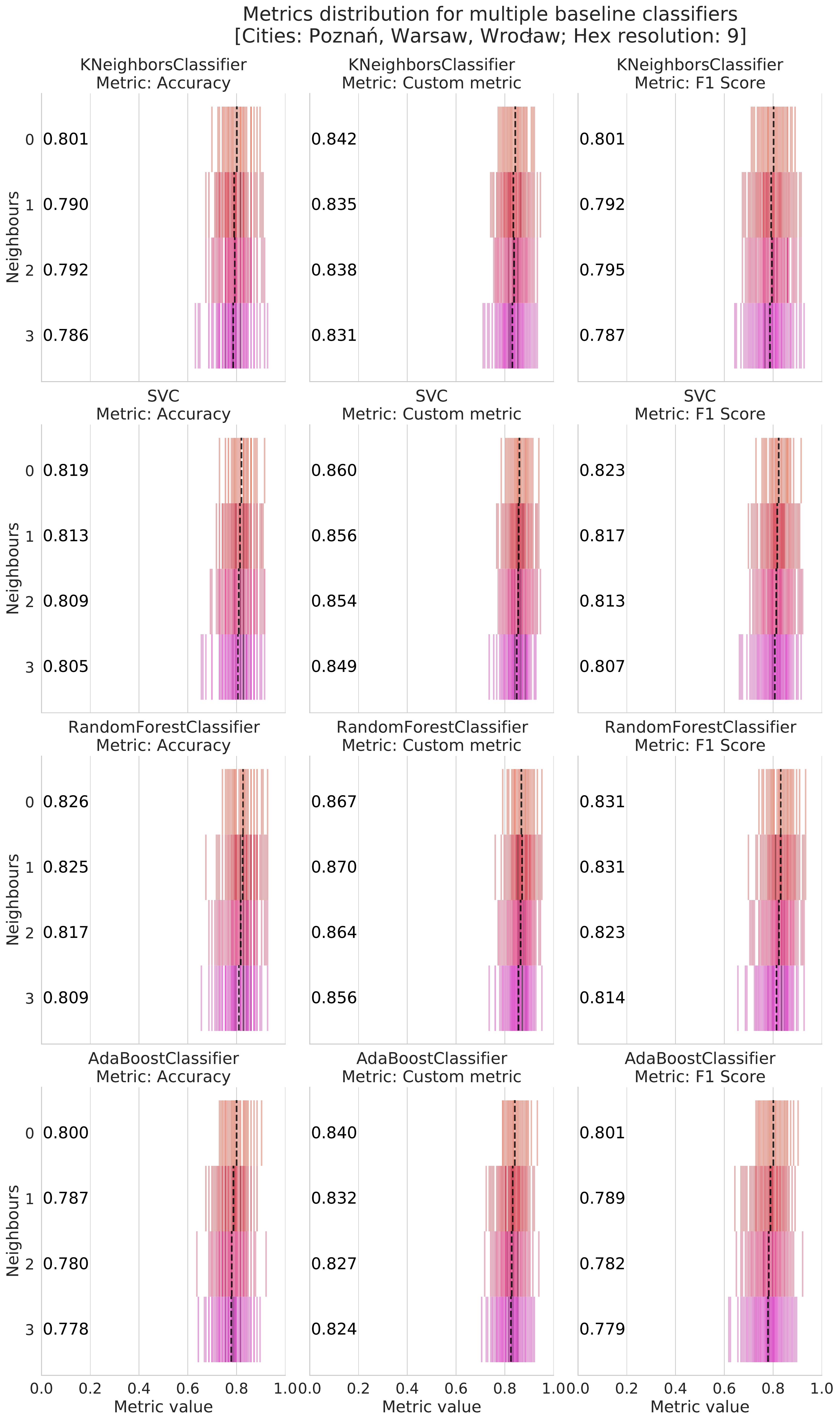}
    \caption[Distribution of prediction results for different baseline classifiers for hex resolution 9]%
    {Distribution of prediction results for different baseline classifiers for hex resolution 9. The vertical dashed lines indicate the average value in a particular chart per size of the neighbourhood. \par \small Personal work.}
    \label{chart:classifiers_9}
\end{figure}

\begin{figure}[H]
    \centering
    \includegraphics[width=0.9\textwidth]{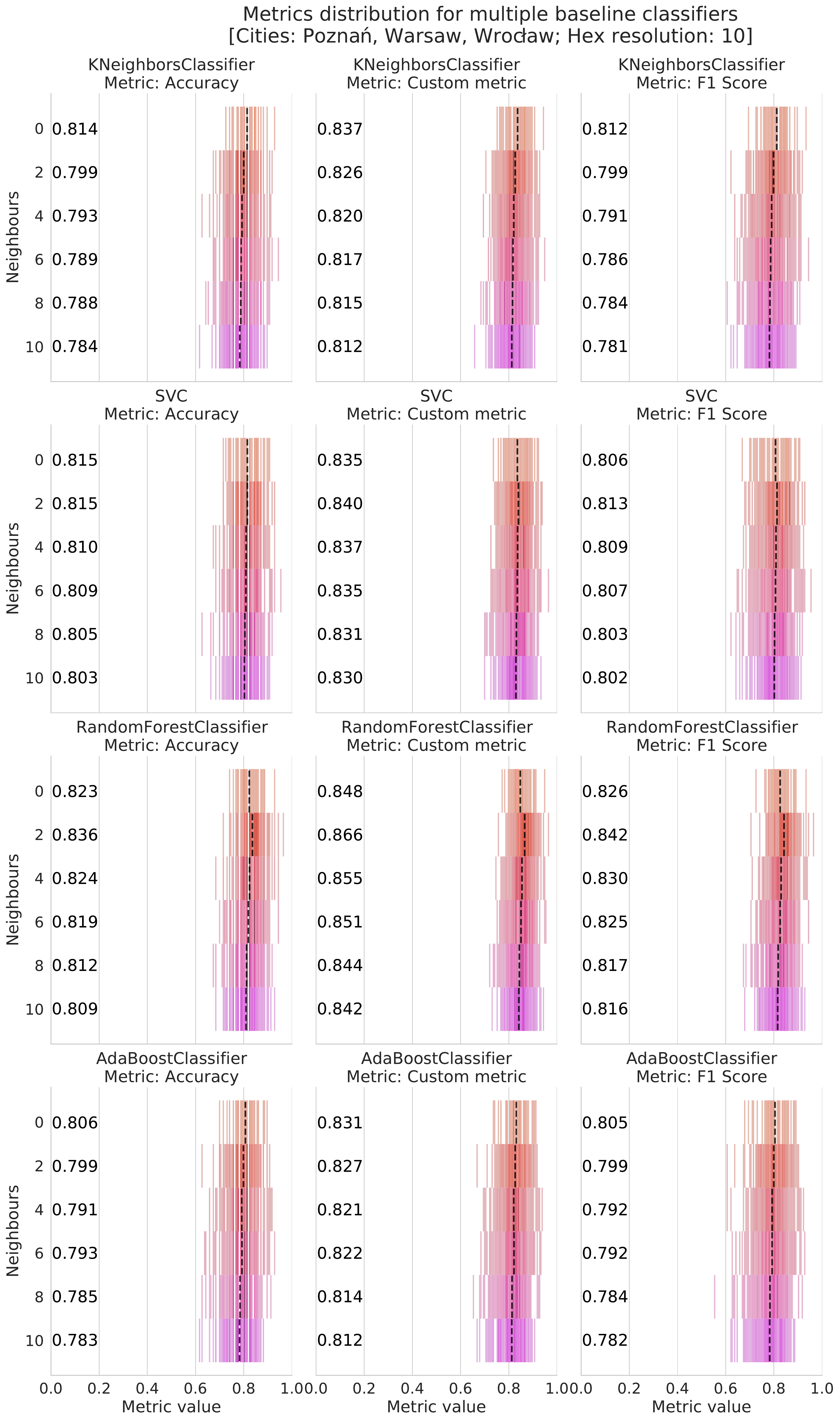}
    \caption[Distribution of prediction results for different baseline classifiers for hex resolution 10]%
    {Distribution of prediction results for different baseline classifiers for hex resolution 10. The vertical dashed lines indicate the average value in a particular chart per size of the neighbourhood. \par \small Personal work.}
    \label{chart:classifiers_10}
\end{figure}

\begin{figure}[H]
    \centering
    \includegraphics[width=0.9\textwidth]{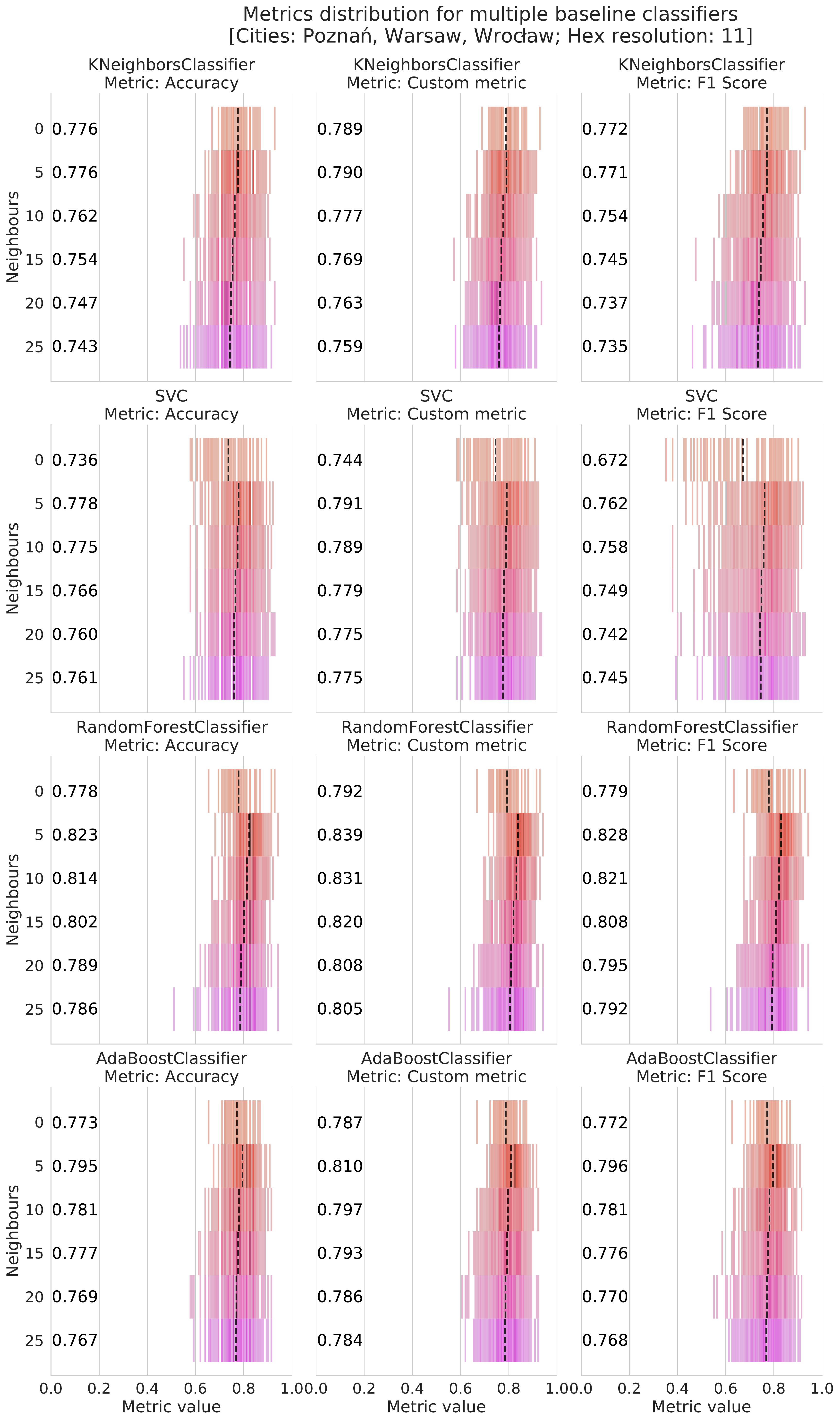}
    \caption[Distribution of prediction results for different baseline classifiers for hex resolution 11]%
    {Distribution of prediction results for different baseline classifiers for hex resolution 11. The vertical dashed lines indicate the average value in a particular chart per size of the neighbourhood. \par \small Personal work.}
    \label{chart:classifiers_11}
\end{figure}

The results were grouped per resolution of the region (9, 10, 11) and within each resolution, the results were divided per size of the neighbourhood examined. All results are plotted in Figures \ref{chart:classifiers_9}, \ref{chart:classifiers_10}, \ref{chart:classifiers_11} and also summarised in the Table \ref{tab:comparing_baseline_classifiers}.

\TopicSetWidth{*}
\TopicSetVPos{t}
\TopicSetContinuationCode{}
\begin{topiclongtable}{@{}FcTcTc|TcTcTcTc@{}}

\toprule
    \begin{tabular}[c]{@{}c@{}}Hex\\ resolution\end{tabular} &
    \begin{tabular}[c]{@{}c@{}}Neigh.\\ size\end{tabular} &
    Metric &
    kNN &
    SVM &
    \begin{tabular}[c]{@{}c@{}}Random\\ Forest\end{tabular} &
    AdaBoost 
\\* \midrule
\endhead
\hline\endfoot
\endlastfoot
\TopicLine \Topic[9] & \Topic[0] & Accuracy & 0.801 & 0.819 & \textbf{0.826} & 0.800 \\
\TopicLine \Topic[9] & \Topic[0] & Custom   & 0.842 & 0.860 & \textbf{0.867} & 0.840 \\
\TopicLine \Topic[9] & \Topic[0] & F1 Score & 0.801 & 0.823 & \textbf{0.831} & 0.801 \\

\TopicLine \Topic[9] & \Topic[1] & Accuracy & 0.790 & 0.813 & \textbf{0.825} & 0.787 \\
\TopicLine \Topic[9] & \Topic[1] & Custom   & 0.835 & 0.856 & \textbf{0.870} & 0.832 \\
\TopicLine \Topic[9] & \Topic[1] & F1 Score & 0.792 & 0.817 & \textbf{0.831} & 0.789 \\

\TopicLine \Topic[9] & \Topic[2] & Accuracy & 0.792 & 0.809 & \textbf{0.817} & 0.780 \\
\TopicLine \Topic[9] & \Topic[2] & Custom   & 0.838 & 0.854 & \textbf{0.864} & 0.827 \\
\TopicLine \Topic[9] & \Topic[2] & F1 Score & 0.795 & 0.813 & \textbf{0.823} & 0.782 \\

\TopicLine \Topic[9] & \Topic[3] & Accuracy & 0.786 & 0.805 & \textbf{0.809} & 0.778 \\
\TopicLine \Topic[9] & \Topic[3] & Custom   & 0.831 & 0.849 & \textbf{0.856} & 0.824 \\
\TopicLine \Topic[9] & \Topic[3] & F1 Score & 0.787 & 0.807 & \textbf{0.814} & 0.779 \\

\TopicLine \Topic[10] & \Topic[0] & Accuracy & 0.814 & 0.815 & \textbf{0.823} & 0.806 \\
\TopicLine \Topic[10] & \Topic[0] & Custom   & 0.837 & 0.835 & \textbf{0.848} & 0.831 \\
\TopicLine \Topic[10] & \Topic[0] & F1 Score & 0.812 & 0.806 & \textbf{0.826} & 0.805 \\

\TopicLine \Topic[10] & \Topic[2] & Accuracy & 0.799 & 0.815 & \textbf{0.836} & 0.799 \\
\TopicLine \Topic[10] & \Topic[2] & Custom   & 0.826 & 0.840 & \textbf{0.866} & 0.827 \\
\TopicLine \Topic[10] & \Topic[2] & F1 Score & 0.799 & 0.813 & \textbf{0.842} & 0.799 \\

\TopicLine \Topic[10] & \Topic[4] & Accuracy & 0.793 & 0.810 & \textbf{0.824} & 0.791 \\
\TopicLine \Topic[10] & \Topic[4] & Custom   & 0.820 & 0.837 & \textbf{0.855} & 0.821 \\
\TopicLine \Topic[10] & \Topic[4] & F1 Score & 0.791 & 0.809 & \textbf{0.830} & 0.792 \\

\TopicLine \Topic[10] & \Topic[6] & Accuracy & 0.789 & 0.809 & \textbf{0.819} & 0.793 \\
\TopicLine \Topic[10] & \Topic[6] & Custom   & 0.817 & 0.835 & \textbf{0.851} & 0.822 \\
\TopicLine \Topic[10] & \Topic[6] & F1 Score & 0.786 & 0.807 & \textbf{0.825} & 0.792 \\

\TopicLine \Topic[10] & \Topic[8] & Accuracy & 0.788 & 0.805 & \textbf{0.812} & 0.785 \\
\TopicLine \Topic[10] & \Topic[8] & Custom   & 0.815 & 0.831 & \textbf{0.844} & 0.814 \\
\TopicLine \Topic[10] & \Topic[8] & F1 Score & 0.784 & 0.802 & \textbf{0.817} & 0.784 \\

\TopicLine \Topic[10] & \Topic[10] & Accuracy & 0.784 & 0.803 & \textbf{0.809} & 0.783 \\
\TopicLine \Topic[10] & \Topic[10] & Custom   & 0.812 & 0.830 & \textbf{0.842} & 0.812 \\
\TopicLine \Topic[10] & \Topic[10] & F1 Score & 0.781 & 0.802 & \textbf{0.816} & 0.782 \\

\TopicLine \Topic[11] & \Topic[0] & Accuracy & 0.776 & 0.736 & \textbf{0.778} & 0.773 \\
\TopicLine \Topic[11] & \Topic[0] & Custom   & 0.789 & 0.744 & \textbf{0.792} & 0.787 \\
\TopicLine \Topic[11] & \Topic[0] & F1 Score & 0.772 & 0.672 & \textbf{0.779} & 0.772 \\

\TopicLine \Topic[11] & \Topic[5] & Accuracy & 0.776 & 0.778 & \textbf{0.823} & 0.795 \\
\TopicLine \Topic[11] & \Topic[5] & Custom   & 0.790 & 0.791 & \textbf{0.839} & 0.810 \\
\TopicLine \Topic[11] & \Topic[5] & F1 Score & 0.771 & 0.762 & \textbf{0.828} & 0.796 \\

\TopicLine \Topic[11] & \Topic[10] & Accuracy & 0.762 & 0.775 & \textbf{0.814} & 0.781 \\
\TopicLine \Topic[11] & \Topic[10] & Custom   & 0.777 & 0.789 & \textbf{0.831} & 0.797 \\
\TopicLine \Topic[11] & \Topic[10] & F1 Score & 0.754 & 0.758 & \textbf{0.821} & 0.781 \\

\TopicLine \Topic[11] & \Topic[15] & Accuracy & 0.754 & 0.766 & \textbf{0.802} & 0.777 \\
\TopicLine \Topic[11] & \Topic[15] & Custom   & 0.769 & 0.779 & \textbf{0.820} & 0.793 \\
\TopicLine \Topic[11] & \Topic[15] & F1 Score & 0.745 & 0.749 & \textbf{0.808} & 0.776 \\

\TopicLine \Topic[11] & \Topic[20] & Accuracy & 0.747 & 0.760 & \textbf{0.789} & 0.769 \\
\TopicLine \Topic[11] & \Topic[20] & Custom   & 0.763 & 0.775 & \textbf{0.808} & 0.786 \\
\TopicLine \Topic[11] & \Topic[20] & F1 Score & 0.737 & 0.742 & \textbf{0.795} & 0.770 \\

\TopicLine \Topic[11] & \Topic[25] & Accuracy & 0.743 & 0.761 & \textbf{0.786} & 0.767 \\
\TopicLine \Topic[11] & \Topic[25] & Custom   & 0.759 & 0.775 & \textbf{0.805} & 0.784 \\
\TopicLine \Topic[11] & \Topic[25] & F1 Score & 0.735 & 0.745 & \textbf{0.792} & 0.768 \\

\bottomrule
\caption[Comparison of different baseline classifiers]%
{Comparison of different baseline classifiers. Results for the cities of Poznań, Warsaw, and Wrocław calculated on the test set. Values represent the average of all grouped results for different parameters. Abbreviations: \textit{Neigh.} - Neighbourhood, \textit{kNN} - k-Nearest Neighbours, \textit{SVM} - Support Vector Machine. The best values in each row were highlighted.}
\label{tab:comparing_baseline_classifiers}\\
\end{topiclongtable}

All 4 classifiers were able to obtain an average F1 score metric value above 0.6, which is not a bad performance. Based on the results, the random forest classifier was selected for further study. It obtained similar but better results than the AdaBoost metaclassifier in every case in all metrics. With increasing resolution (i.e., decreasing regions) a decreasing accuracy of the first two classifiers can be seen and the SVM classifier performed particularly unstably (looking at the scatter of the results).

\newpage

\section{RQ2: How neighbourhood embedding methods affect performance?}

The first hyperparameter investigated is the choice of the neighbourhood embedding method. Its performance directly affects the quality and speed of the classifier. The first method, vector concatenation, does not lose any data obtained from successive vectors of neighbourhoods, but the size of the final vector increases with the number of neighbourhoods selected for the study, which is important for the speed of prediction, especially at a higher resolution of regions. The other three methods preserve the size of the vector, so the speed of the classifier does not change, but the data are combined in different ways and partial data loss or cluttering of the obtained neighbourhood vectors may occur.

\subsection{Method of RQ2}

All results in this step were calculated only using the random forest classifier. As with the previous question, the averaged results from 10 iterations for the test set for each parameter configuration were used to test the effect of embedding methods on prediction quality. The results were calculated on data from 3 cities separately: Poznań, Warsaw and Wrocław. All regions containing stations and an equal number of regions not containing stations were used for learning (imbalance ratio = 1). For feature generation, 2 region embedding methods were used: Category counting and Shape analysis per category. No feature processing methods were applied to the data. Prediction quality was tested using the 3 metrics mentioned in the previous section.

\subsection{Results of RQ2}

\begin{figure}[H]
    \centering
    \includegraphics[width=\textwidth]{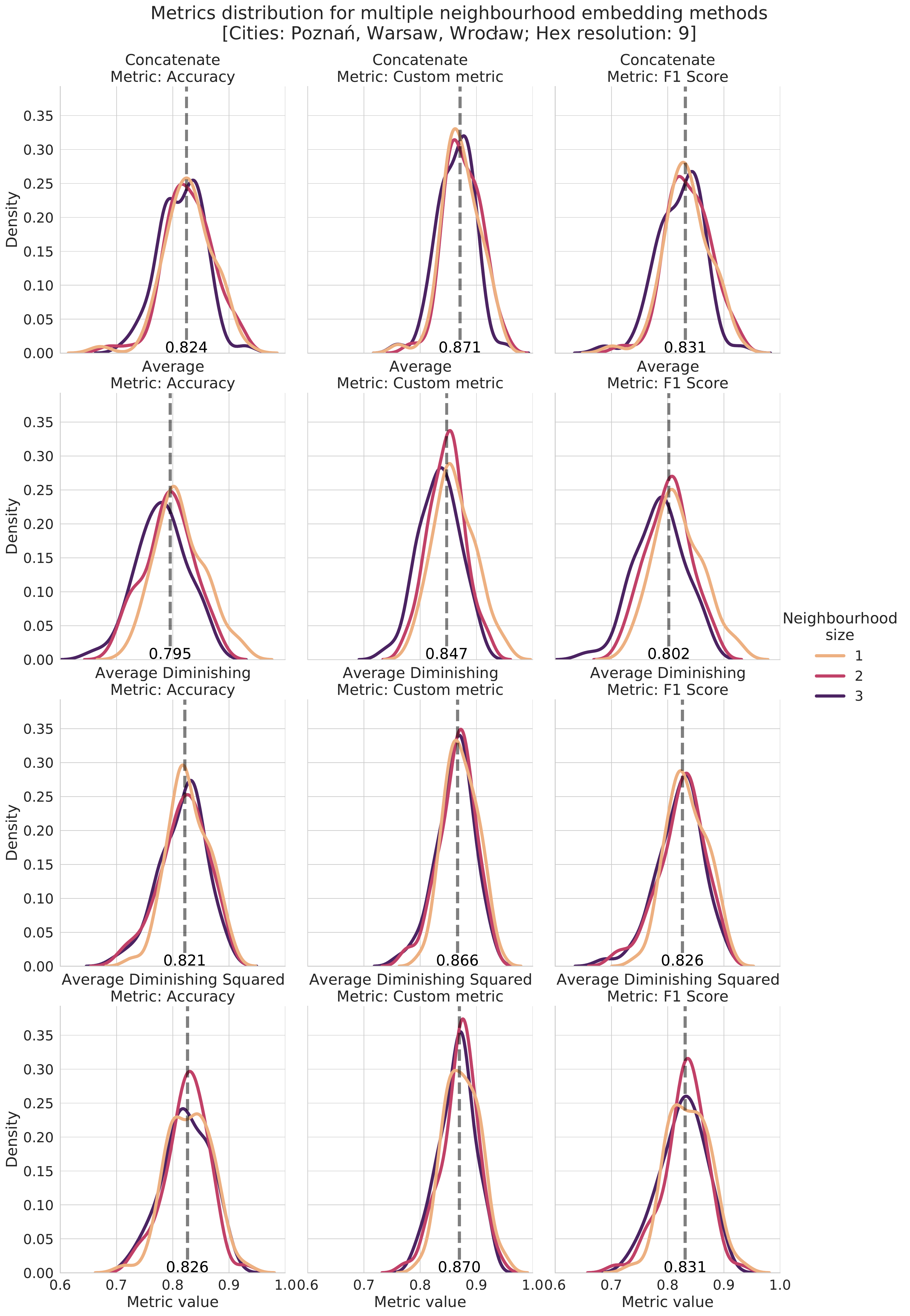}
    \caption[Distribution of prediction results for different neighbourhood embedding methods for hex resolution 9]%
    {Distribution of prediction results for different neighbourhood embedding methods for hex resolution 9. The vertical dashed lines indicate the average value in a particular chart. \par \small Personal work.}
    \label{chart:neighbour_embedding_methods_9}
\end{figure}

\begin{figure}[H]
    \centering
    \includegraphics[width=\textwidth]{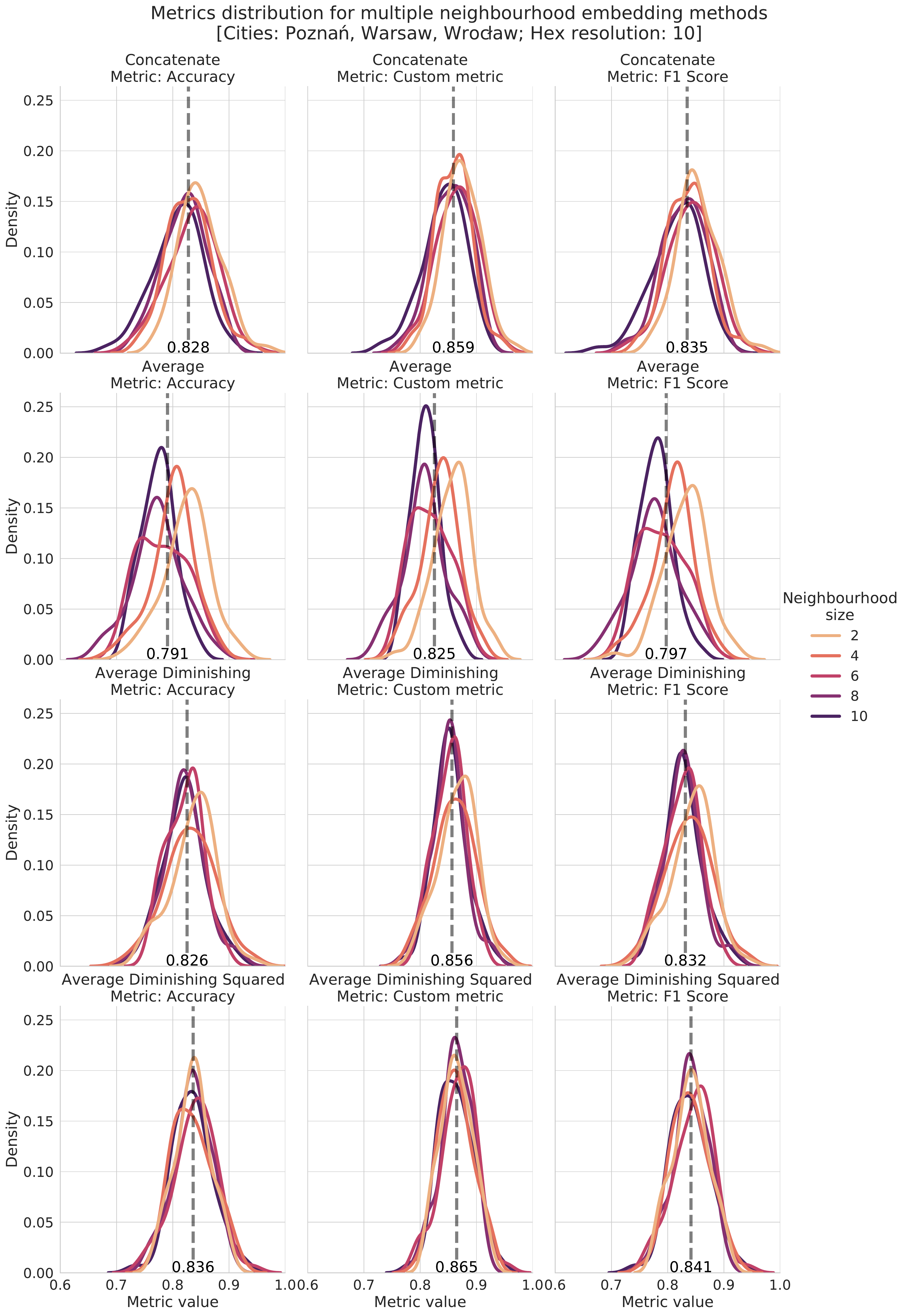}
    \caption[Distribution of prediction results for different neighbourhood embedding methods for hex resolution 10]%
    {Distribution of prediction results for different neighbourhood embedding methods for hex resolution 10. The vertical dashed lines indicate the average value in a particular chart. \par \small Personal work.}
    \label{chart:neighbour_embedding_methods_10}
\end{figure}

\begin{figure}[H]
    \centering
    \includegraphics[width=\textwidth]{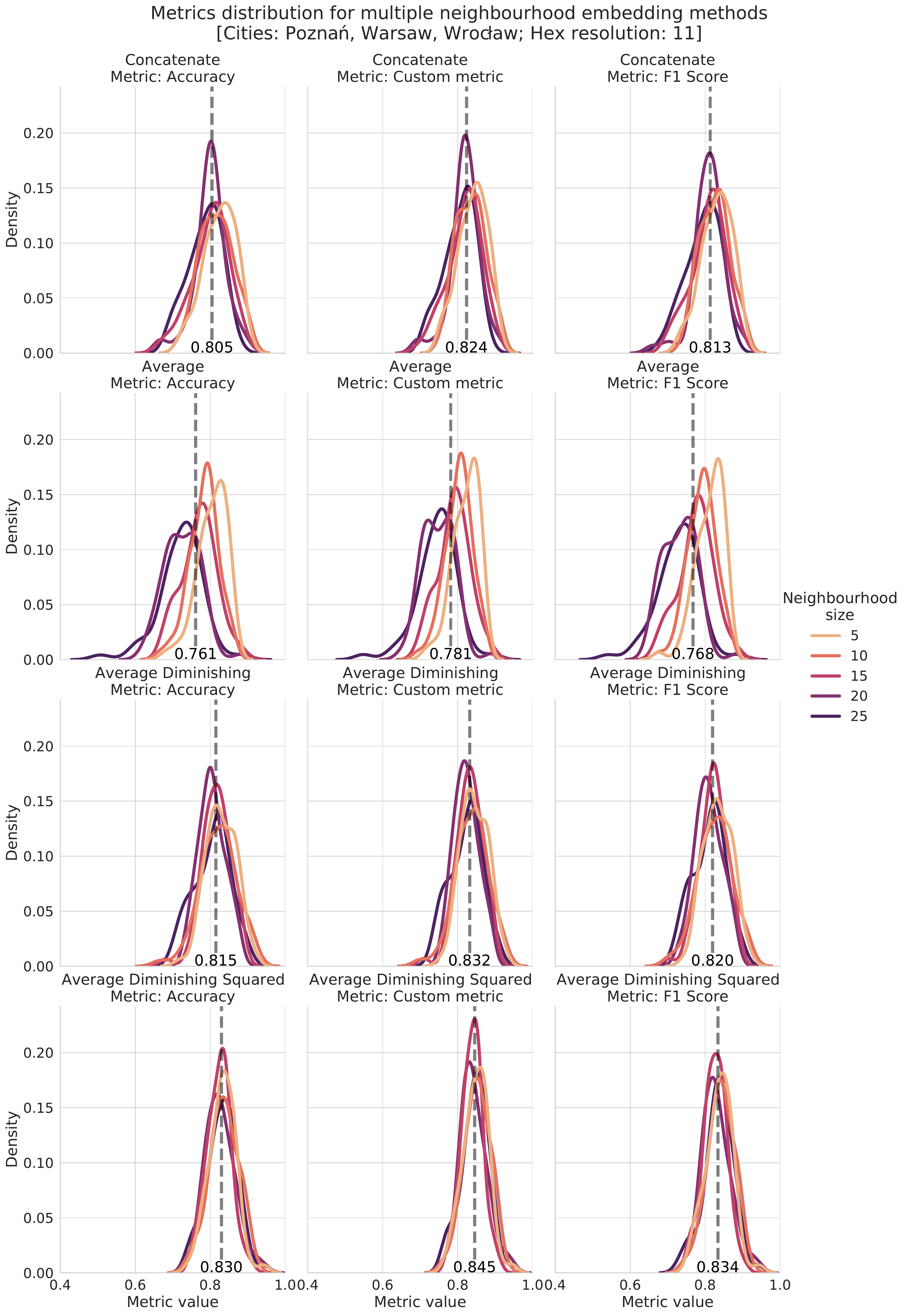}
    \caption[Distribution of prediction results for different neighbourhood embedding methods for hex resolution 11]%
    {Distribution of prediction results for different neighbourhood embedding methods for hex resolution 11. The vertical dashed lines indicate the average value in a particular chart. \par \small Personal work.}
    \label{chart:neighbour_embedding_methods_11}
\end{figure}

The results were grouped per resolution of the region (9, 10, 11). All results are plotted in Figures \ref{chart:neighbour_embedding_methods_9}, \ref{chart:neighbour_embedding_methods_10}, \ref{chart:neighbour_embedding_methods_11} and also summarised in the Table \ref{tab:comparing_neighbourhood_emb}.

\TopicSetWidth{*}
\TopicSetVPos{t}
\TopicSetContinuationCode{}
\begin{topiclongtable}{@{}FcTc|TcTcTcTcTc@{}}

\toprule
    \begin{tabular}[c]{@{}c@{}}Hex\\ resolution\end{tabular} &
    Metric &
    Concatenate &
    Average &
    \begin{tabular}[c]{@{}c@{}}Average\\ Diminishing\end{tabular} &
    \begin{tabular}[c]{@{}c@{}}Average\\ Diminishing\\ Squared\end{tabular}
\\* \midrule
\endhead
\hline\endfoot
\endlastfoot
\TopicLine \Topic[9] & Accuracy & 0.824 & 0.795 & 0.821 & \textbf{0.826} \\
\TopicLine \Topic[9] & Custom   & \textbf{0.871} & 0.847 & 0.866 & 0.870 \\
\TopicLine \Topic[9] & F1 Score & \textbf{0.831} & 0.802 & 0.826 & \textbf{0.831} \\

\TopicLine \Topic[10] & Accuracy & 0.828 & 0.791 & 0.826 & \textbf{0.836} \\
\TopicLine \Topic[10] & Custom   & 0.859 & 0.825 & 0.856 & \textbf{0.865} \\
\TopicLine \Topic[10] & F1 Score & 0.835 & 0.797 & 0.832 & \textbf{0.841} \\

\TopicLine \Topic[11] & Accuracy & 0.805 & 0.761 & 0.815 & \textbf{0.830} \\
\TopicLine \Topic[11] & Custom   & 0.824 & 0.781 & 0.832 & \textbf{0.845} \\
\TopicLine \Topic[11] & F1 Score & 0.813 & 0.768 & 0.820 & \textbf{0.834} \\

\bottomrule
\caption[Comparison of different neighbourhood embedding methods]%
{Comparison of different neighbourhood embedding methods. Results for the cities of Poznań, Warsaw, and Wrocław calculated on the test set. Values represent the average of all grouped results for different parameters. The best values in each row were highlighted.}
\label{tab:comparing_neighbourhood_emb}\\
\end{topiclongtable}

The collected results indicate that the best proposed method is the squared diminishing averaging of the embedding vectors of successive neighbourhoods. For resolution 9, the concatenation of vectors was able to equal the prediction quality, but for higher resolutions, it deteriorated more and more. It can also be seen that the last method produces the most stable results regardless of the size of the neighbourhood under study, where the averaging alone method significantly degraded the prediction quality as the neighbourhood size increased.
An additional argument for choosing the last method is the constant size of the final vector regardless of the size of the neighbourhood, thanks to which the classifier will need more or less the same time for learning and prediction.

\newpage

\section{RQ3: How region embedding method affects the prediction performance?}

Another hyperparameter investigated is the single region embedding method. The methods differ in complexity and generate vectors of different lengths, which may have a direct impact on the results obtained. The first method, Category Counting, is the simplest of the proposed ones and it simply counts the number of objects present inside the region per 20 defined categories. The second method, Shape Analysis, takes into account the appearance aspect of the object by counting the area and length. It is used in 3 modes: analysis per category, analysis per all tags available in OSM, and analysis per selected OSM tags. The first two methods, Category Counting and Shape Analysis per category are tested using both a random forest classifier and a neural network. The other methods were passed through an autoencoder with different final vector lengths and then used in a neural network.

\subsection{Method of RQ3}

In this step, the results were calculated only with a random forest classifier and using the average diminishing squared neighbour embedding method. As in the two previous cases, averaged results from 10 iterations for the test set for each parameter configuration were used to test the effect of embedding methods on prediction quality. The results were calculated on data from 3 cities separately: Poznań, Warsaw and Wrocław. All regions containing stations and an equal number of regions not containing stations were used for learning (imbalance ratio = 1). No feature processing methods were applied to the data. Prediction quality was tested using the 3 metrics mentioned in the previous section.

\subsection{Results of RQ3}

\begin{figure}[H]
    \centering
    \includegraphics[width=\textwidth]{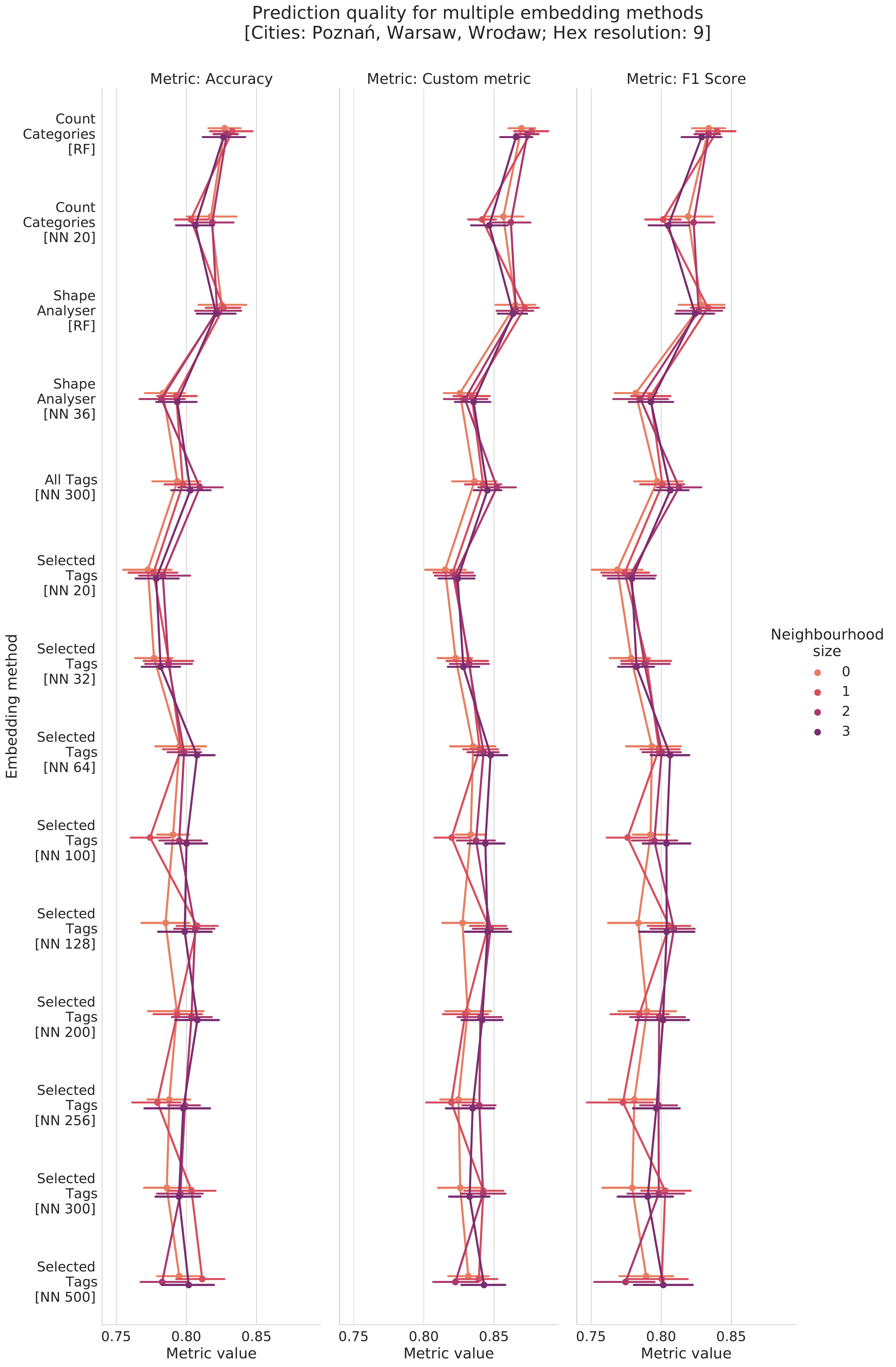}
    \caption[Prediction quality for multiple embedding methods for hex resolution 9]%
    {Prediction quality for multiple embedding methods for hex resolution 9. \par \small Personal work.}
    \label{chart:embedding_methods_9}
\end{figure}

\begin{figure}[H]
    \centering
    \includegraphics[width=\textwidth]{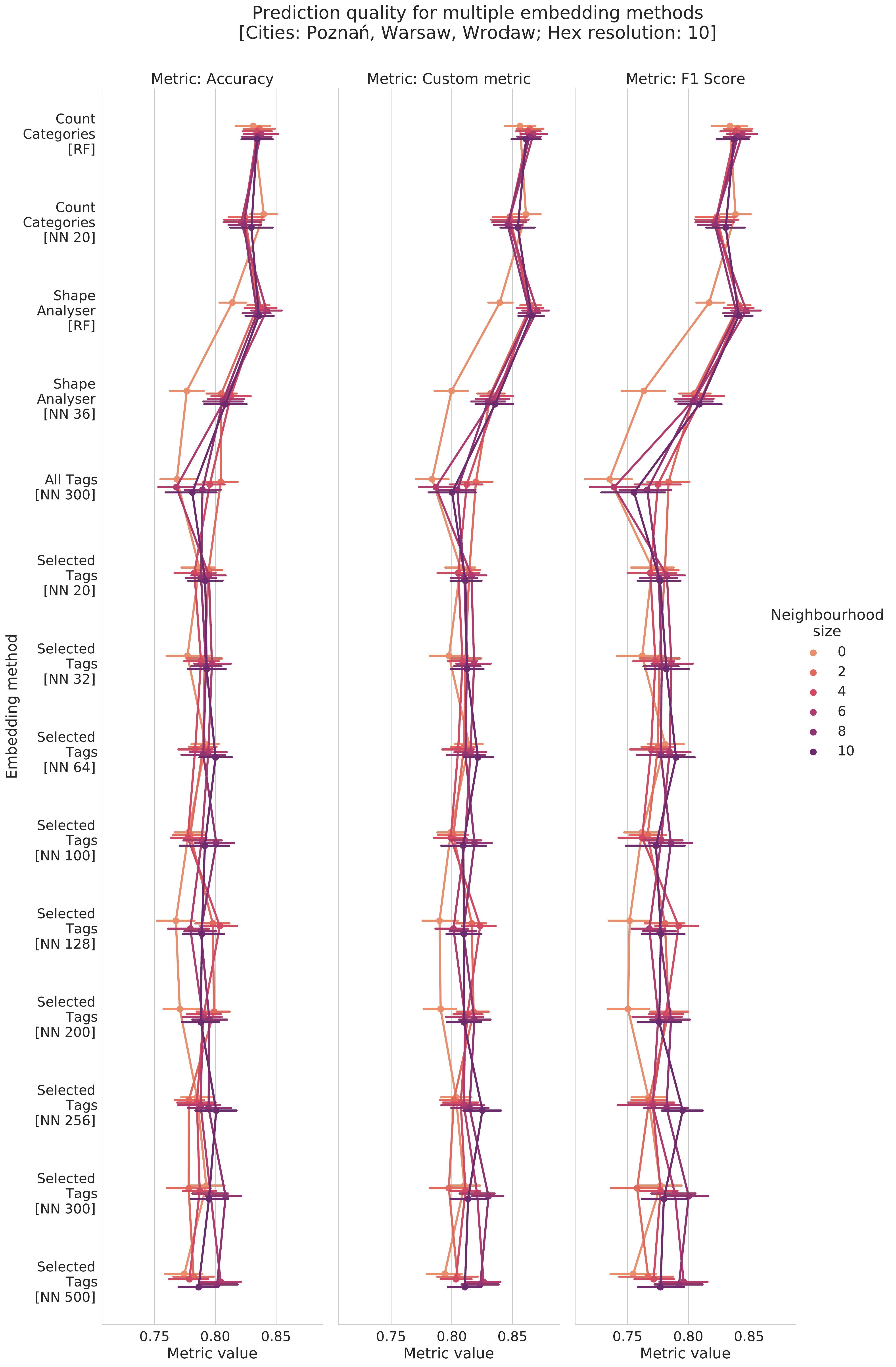}
    \caption[Prediction quality for multiple embedding methods for hex resolution 10]%
    {Prediction quality for multiple embedding methods for hex resolution 10. \par \small Personal work.}
    \label{chart:embedding_methods_10}
\end{figure}

\begin{figure}[H]
    \centering
    \includegraphics[width=\textwidth]{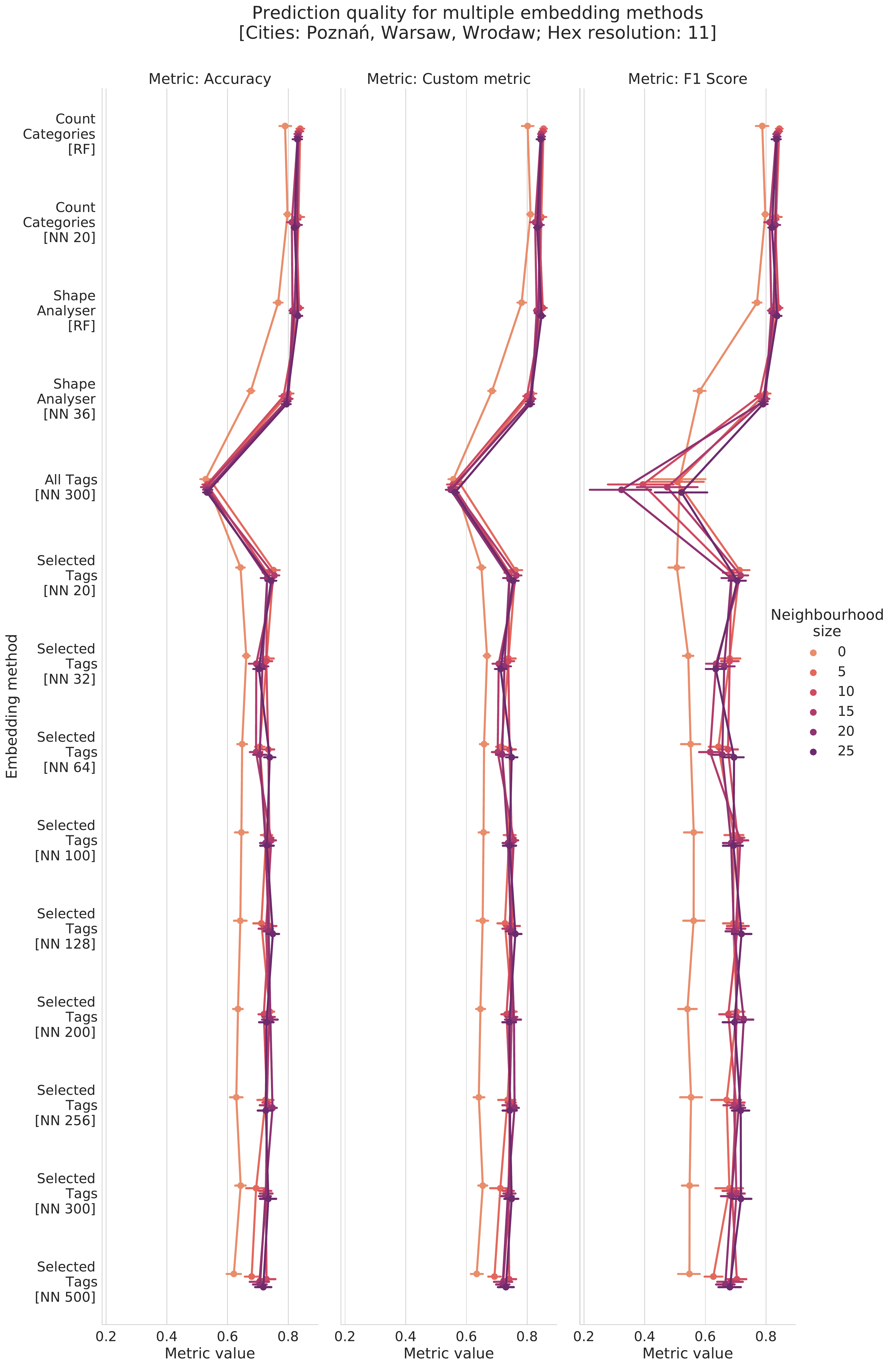}
    \caption[Prediction quality for multiple embedding methods for hex resolution 11]%
    {Prediction quality for multiple embedding methods for hex resolution 11. \par \small Personal work.}
    \label{chart:embedding_methods_11}
\end{figure}

After investigating different neural network architectures, the results for the best one found in the research (input layer equal N, hidden layer equal 20, output layer equal 2) are presented below in comparison with the random forest classifier. 
The results were grouped per resolution of the region (9, 10, 11). All results are plotted in Figures \ref{chart:embedding_methods_9}, \ref{chart:embedding_methods_10}, \ref{chart:embedding_methods_11} and also summarised in the Table \ref{tab:comparing_region_emb}.

\TopicSetWidth{*}
\TopicSetVPos{t}
\TopicSetContinuationCode{}
\begin{topiclongtable}{@{}FcTc|TcTcTcTcTcTcTc@{}}
\toprule
    \begin{tabular}[c]{@{}c@{}}Hex\\ res.\end{tabular} &
    Metric &
    \begin{tabular}[c]{@{}c@{}}CC\\ RF\end{tabular} &
    \begin{tabular}[c]{@{}c@{}}CC\\ NN 20\end{tabular} &
    \begin{tabular}[c]{@{}c@{}}SA\\ RF\end{tabular} &
    \begin{tabular}[c]{@{}c@{}}SA\\ NN 36\end{tabular} &
    \begin{tabular}[c]{@{}c@{}}AT\\ NN 300\end{tabular} &
    \begin{tabular}[c]{@{}c@{}}ST\\ NN 20\end{tabular} &
    \begin{tabular}[c]{@{}c@{}}ST\\ NN 32\end{tabular} 
\\* \midrule
\endhead
\hline\endfoot
\endlastfoot

\TopicLine \Topic[9] & Acc. & 
\begin{tabular}[c]{@{}c@{}} \textbf{0.829} \\ (\textbf{0.829}) \end{tabular} & 
\begin{tabular}[c]{@{}c@{}} 0.811 \\ (0.809) \end{tabular} & 
\begin{tabular}[c]{@{}c@{}} 0.824 \\ (0.823) \end{tabular} & 
\begin{tabular}[c]{@{}c@{}} 0.788 \\ (0.789) \end{tabular} & 
\begin{tabular}[c]{@{}c@{}} 0.801 \\ (0.803) \end{tabular} & 
\begin{tabular}[c]{@{}c@{}} 0.778 \\ (0.779) \end{tabular} & 
\begin{tabular}[c]{@{}c@{}} 0.783 \\ (0.785) \end{tabular} \\
\TopicLine \Topic[9] & Cust. & 
\begin{tabular}[c]{@{}c@{}} \textbf{0.872} \\ (\textbf{0.872}) \end{tabular} & 
\begin{tabular}[c]{@{}c@{}} 0.852 \\ (0.850) \end{tabular} & 
\begin{tabular}[c]{@{}c@{}} 0.867 \\ (0.867) \end{tabular} & 
\begin{tabular}[c]{@{}c@{}} 0.831 \\ (0.833) \end{tabular} & 
\begin{tabular}[c]{@{}c@{}} 0.844 \\ (0.847) \end{tabular} & 
\begin{tabular}[c]{@{}c@{}} 0.821 \\ (0.822) \end{tabular} & 
\begin{tabular}[c]{@{}c@{}} 0.829 \\ (0.831) \end{tabular} \\
\TopicLine \Topic[9] & F1 & 
\begin{tabular}[c]{@{}c@{}} \textbf{0.834} \\ (\textbf{0.834}) \end{tabular} & 
\begin{tabular}[c]{@{}c@{}} 0.812 \\ (0.810) \end{tabular} & 
\begin{tabular}[c]{@{}c@{}} 0.828 \\ (0.828) \end{tabular} & 
\begin{tabular}[c]{@{}c@{}} 0.788 \\ (0.790) \end{tabular} & 
\begin{tabular}[c]{@{}c@{}} 0.804 \\ (0.807) \end{tabular} & 
\begin{tabular}[c]{@{}c@{}} 0.775 \\ (0.777) \end{tabular} & 
\begin{tabular}[c]{@{}c@{}} 0.785 \\ (0.787) \end{tabular} \\

\TopicLine \Topic[10] & Acc. & 
\begin{tabular}[c]{@{}c@{}} \textbf{0.835} \\ (0.835) \end{tabular} & 
\begin{tabular}[c]{@{}c@{}} 0.827 \\ (0.825) \end{tabular} & 
\begin{tabular}[c]{@{}c@{}} 0.833 \\ (\textbf{0.837}) \end{tabular} & 
\begin{tabular}[c]{@{}c@{}} 0.803 \\ (0.808) \end{tabular} & 
\begin{tabular}[c]{@{}c@{}} 0.784 \\ (0.788) \end{tabular} & 
\begin{tabular}[c]{@{}c@{}} 0.790 \\ (0.790) \end{tabular} & 
\begin{tabular}[c]{@{}c@{}} 0.790 \\ (0.792) \end{tabular} \\
\TopicLine \Topic[10] & Cust.   & 
\begin{tabular}[c]{@{}c@{}} \textbf{0.863} \\ (0.864) \end{tabular} & 
\begin{tabular}[c]{@{}c@{}} 0.851 \\ (0.848) \end{tabular} & 
\begin{tabular}[c]{@{}c@{}} 0.861 \\ (\textbf{0.866}) \end{tabular} & 
\begin{tabular}[c]{@{}c@{}} 0.828 \\ (0.834) \end{tabular} & 
\begin{tabular}[c]{@{}c@{}} 0.801 \\ (0.805) \end{tabular} & 
\begin{tabular}[c]{@{}c@{}} 0.811 \\ (0.811) \end{tabular} & 
\begin{tabular}[c]{@{}c@{}} 0.810 \\ (0.813) \end{tabular} \\
\TopicLine \Topic[10] & F1 & 
\begin{tabular}[c]{@{}c@{}} \textbf{0.839} \\ (0.840) \end{tabular} & 
\begin{tabular}[c]{@{}c@{}} 0.827 \\ (0.824) \end{tabular} & 
\begin{tabular}[c]{@{}c@{}} 0.838 \\ (\textbf{0.843}) \end{tabular} & 
\begin{tabular}[c]{@{}c@{}} 0.800 \\ (0.807) \end{tabular} & 
\begin{tabular}[c]{@{}c@{}} 0.759 \\ (0.764) \end{tabular} & 
\begin{tabular}[c]{@{}c@{}} 0.776 \\ (0.777) \end{tabular} & 
\begin{tabular}[c]{@{}c@{}} 0.776 \\ (0.779) \end{tabular} \\

\TopicLine \Topic[11] & Acc. & 
\begin{tabular}[c]{@{}c@{}} \textbf{0.827} \\ (\textbf{0.834}) \end{tabular} & 
\begin{tabular}[c]{@{}c@{}} 0.820 \\ (0.825) \end{tabular} & 
\begin{tabular}[c]{@{}c@{}} 0.816 \\ (0.826) \end{tabular} & 
\begin{tabular}[c]{@{}c@{}} 0.775 \\ (0.795) \end{tabular} & 
\begin{tabular}[c]{@{}c@{}} 0.534 \\ (0.535) \end{tabular} & 
\begin{tabular}[c]{@{}c@{}} 0.726 \\ (0.742) \end{tabular} & 
\begin{tabular}[c]{@{}c@{}} 0.705 \\ (0.714) \end{tabular} \\
\TopicLine \Topic[11] & Cust. & 
\begin{tabular}[c]{@{}c@{}} \textbf{0.841} \\ (\textbf{0.849}) \end{tabular} & 
\begin{tabular}[c]{@{}c@{}} 0.833 \\ (0.837) \end{tabular} & 
\begin{tabular}[c]{@{}c@{}} 0.832 \\ (0.842) \end{tabular} & 
\begin{tabular}[c]{@{}c@{}} 0.789 \\ (0.810) \end{tabular} & 
\begin{tabular}[c]{@{}c@{}} 0.559 \\ (0.559) \end{tabular} & 
\begin{tabular}[c]{@{}c@{}} 0.735 \\ (0.752) \end{tabular} & 
\begin{tabular}[c]{@{}c@{}} 0.715 \\ (0.724) \end{tabular} \\
\TopicLine \Topic[11] & F1 & 
\begin{tabular}[c]{@{}c@{}} \textbf{0.830} \\ (\textbf{0.838}) \end{tabular} & 
\begin{tabular}[c]{@{}c@{}} 0.820 \\ (0.825) \end{tabular} & 
\begin{tabular}[c]{@{}c@{}} 0.820 \\ (0.830) \end{tabular} & 
\begin{tabular}[c]{@{}c@{}} 0.756 \\ (0.791) \end{tabular} & 
\begin{tabular}[c]{@{}c@{}} 0.456 \\ (0.444) \end{tabular} & 
\begin{tabular}[c]{@{}c@{}} 0.669 \\ (0.701) \end{tabular} & 
\begin{tabular}[c]{@{}c@{}} 0.639 \\ (0.658) \end{tabular} \\

\bottomrule
\end{topiclongtable}

\addtocounter{table}{-1}

\begin{topiclongtable}{@{}FcTc|TcTcTcTcTcTcTc@{}}
    \toprule
        \begin{tabular}[c]{@{}c@{}}Hex\\ res.\end{tabular} &
        Metric &
        \begin{tabular}[c]{@{}c@{}}ST\\ NN 64\end{tabular} &
        \begin{tabular}[c]{@{}c@{}}ST\\ NN 100\end{tabular} &
        \begin{tabular}[c]{@{}c@{}}ST\\ NN 128\end{tabular} &
        \begin{tabular}[c]{@{}c@{}}ST\\ NN 200\end{tabular} &
        \begin{tabular}[c]{@{}c@{}}ST\\ NN 256\end{tabular} &
        \begin{tabular}[c]{@{}c@{}}ST\\ NN 300\end{tabular} &
        \begin{tabular}[c]{@{}c@{}}ST\\ NN 500\end{tabular}
    \\* \midrule
\endhead
\hline\endfoot
\endlastfoot

\TopicLine \Topic[9] & Acc. & 
\begin{tabular}[c]{@{}c@{}} 0.799 \\ (0.801) \end{tabular} &
\begin{tabular}[c]{@{}c@{}} 0.790 \\ (0.790) \end{tabular} &
\begin{tabular}[c]{@{}c@{}} 0.799 \\ (0.804) \end{tabular} &
\begin{tabular}[c]{@{}c@{}} 0.799 \\ (0.802) \end{tabular} & 
\begin{tabular}[c]{@{}c@{}} 0.791 \\ (0.792) \end{tabular} & 
\begin{tabular}[c]{@{}c@{}} 0.795 \\ (0.798) \end{tabular} & 
\begin{tabular}[c]{@{}c@{}} 0.798 \\ (0.799) \end{tabular} \\
\TopicLine \Topic[9] & Cust. & 
\begin{tabular}[c]{@{}c@{}} 0.841 \\ (0.843) \end{tabular} & 
\begin{tabular}[c]{@{}c@{}} 0.834 \\ (0.834) \end{tabular} & 
\begin{tabular}[c]{@{}c@{}} 0.842 \\ (0.846) \end{tabular} & 
\begin{tabular}[c]{@{}c@{}} 0.836 \\ (0.837) \end{tabular} & 
\begin{tabular}[c]{@{}c@{}} 0.830 \\ (0.831) \end{tabular} & 
\begin{tabular}[c]{@{}c@{}} 0.836 \\ (0.839) \end{tabular} & 
\begin{tabular}[c]{@{}c@{}} 0.834 \\ (0.835) \end{tabular} \\
\TopicLine \Topic[9] & F1 & 
\begin{tabular}[c]{@{}c@{}} 0.800 \\ (0.802) \end{tabular} & 
\begin{tabular}[c]{@{}c@{}} 0.792 \\ (0.792) \end{tabular} & 
\begin{tabular}[c]{@{}c@{}} 0.801 \\ (0.806) \end{tabular} & 
\begin{tabular}[c]{@{}c@{}} 0.794 \\ (0.795) \end{tabular} & 
\begin{tabular}[c]{@{}c@{}} 0.787 \\ (0.789) \end{tabular} & 
\begin{tabular}[c]{@{}c@{}} 0.793 \\ (0.797) \end{tabular} & 
\begin{tabular}[c]{@{}c@{}} 0.791 \\ (0.792) \end{tabular} \\

\TopicLine \Topic[10] & Acc. & 
\begin{tabular}[c]{@{}c@{}} 0.792 \\ (0.792) \end{tabular} & 
\begin{tabular}[c]{@{}c@{}} 0.786 \\ (0.788) \end{tabular} & 
\begin{tabular}[c]{@{}c@{}} 0.788 \\ (0.792) \end{tabular} & 
\begin{tabular}[c]{@{}c@{}} 0.789 \\ (0.792) \end{tabular} & 
\begin{tabular}[c]{@{}c@{}} 0.789 \\ (0.789) \end{tabular} & 
\begin{tabular}[c]{@{}c@{}} 0.793 \\ (0.793) \end{tabular} & 
\begin{tabular}[c]{@{}c@{}} 0.788 \\ (0.791) \end{tabular} \\
\TopicLine \Topic[10] & Cust. & 
\begin{tabular}[c]{@{}c@{}} 0.813 \\ (0.813) \end{tabular} & 
\begin{tabular}[c]{@{}c@{}} 0.806 \\ (0.808) \end{tabular} & 
\begin{tabular}[c]{@{}c@{}} 0.808 \\ (0.812) \end{tabular} & 
\begin{tabular}[c]{@{}c@{}} 0.810 \\ (0.814) \end{tabular} & 
\begin{tabular}[c]{@{}c@{}} 0.810 \\ (0.812) \end{tabular} & 
\begin{tabular}[c]{@{}c@{}} 0.814 \\ (0.815) \end{tabular} & 
\begin{tabular}[c]{@{}c@{}} 0.810 \\ (0.814) \end{tabular} \\
\TopicLine \Topic[10] & F1 & 
\begin{tabular}[c]{@{}c@{}} 0.779 \\ (0.779) \end{tabular} & 
\begin{tabular}[c]{@{}c@{}} 0.771 \\ (0.773) \end{tabular} & 
\begin{tabular}[c]{@{}c@{}} 0.774 \\ (0.779) \end{tabular} & 
\begin{tabular}[c]{@{}c@{}} 0.775 \\ (0.780) \end{tabular} & 
\begin{tabular}[c]{@{}c@{}} 0.776 \\ (0.777) \end{tabular} & 
\begin{tabular}[c]{@{}c@{}} 0.780 \\ (0.781) \end{tabular} & 
\begin{tabular}[c]{@{}c@{}} 0.776 \\ (0.781) \end{tabular} \\

\TopicLine \Topic[11] & Acc. & 
\begin{tabular}[c]{@{}c@{}} 0.705 \\ (0.716) \end{tabular} & 
\begin{tabular}[c]{@{}c@{}} 0.718 \\ (0.733) \end{tabular} & 
\begin{tabular}[c]{@{}c@{}} 0.717 \\ (0.732) \end{tabular} & 
\begin{tabular}[c]{@{}c@{}} 0.717 \\ (0.733) \end{tabular} & 
\begin{tabular}[c]{@{}c@{}} 0.715 \\ (0.732) \end{tabular} & 
\begin{tabular}[c]{@{}c@{}} 0.708 \\ (0.721) \end{tabular} & 
\begin{tabular}[c]{@{}c@{}} 0.693 \\ (0.708) \end{tabular} \\
\TopicLine \Topic[11] & Cust. & 
\begin{tabular}[c]{@{}c@{}} 0.714 \\ (0.725) \end{tabular} & 
\begin{tabular}[c]{@{}c@{}} 0.730 \\ (0.745) \end{tabular} & 
\begin{tabular}[c]{@{}c@{}} 0.730 \\ (0.745) \end{tabular} & 
\begin{tabular}[c]{@{}c@{}} 0.729 \\ (0.746) \end{tabular} & 
\begin{tabular}[c]{@{}c@{}} 0.727 \\ (0.745) \end{tabular} & 
\begin{tabular}[c]{@{}c@{}} 0.722 \\ (0.735) \end{tabular} & 
\begin{tabular}[c]{@{}c@{}} 0.706 \\ (0.721) \end{tabular} \\
\TopicLine \Topic[11] & F1 & 
\begin{tabular}[c]{@{}c@{}} 0.639 \\ (0.657) \end{tabular} & 
\begin{tabular}[c]{@{}c@{}} 0.676 \\ (0.698) \end{tabular} & 
\begin{tabular}[c]{@{}c@{}} 0.679 \\ (0.703) \end{tabular} & 
\begin{tabular}[c]{@{}c@{}} 0.674 \\ (0.701) \end{tabular} & 
\begin{tabular}[c]{@{}c@{}} 0.674 \\ (0.698) \end{tabular} & 
\begin{tabular}[c]{@{}c@{}} 0.670 \\ (0.694) \end{tabular} & 
\begin{tabular}[c]{@{}c@{}} 0.651 \\ (0.672) \end{tabular} \\

\bottomrule

\caption[Comparison of different region embedding methods]%
{Comparison of different region embedding methods. Results for the cities of Poznań, Warsaw, and Wrocław calculated on the test set. Values represent the average of all grouped results for different parameters. Values in brackets represent the average score excluding the neighbourhood size of 0. Abbreviations: \textit{Acc.} - Accuracy, \textit{Cust.} - Custom metric, \textit{F1} - F1 Score, \textit{CC} - Category counting, \textit{SA} - Shape analysis per category, \textit{AT} - Shape analysis per all tags, \textit{ST} - Shape analysis per selected tags, \textit{RF} - Random forest classifier, \textit{NN} - Neural network classifier. The numbers next to \textit{NN} represent input embedding vector size. The best values in each row were highlighted.}
\label{tab:comparing_region_emb}\\
\end{topiclongtable}

The best results were obtained with the simplest method (Category counting) in combination with the random forest classifier. Due to the limited resources in this research, no further attempt was made to find a better neural network architecture that would produce better results. 

\section{RQ4: How vector preprocessing affects performance?}

Feature scaling is a commonly used process in processing variables before running a machine learning model. They allow to improve the performance of the model and in some cases, it is a required step to be able to run some machine learning methods at all. The impact of normalisation (min-max) and also standardisation on the results obtained by the method will be investigated.

\subsection{Method of RQ4}

Having already selected several hyperparameters of the method, additional cities for learning were added to increase the set for averaging the results. In this step, the results were calculated only with a random forest classifier, using the category counting embedding method and average diminishing squared neighbour embedding method. As in the two previous cases, averaged results from 10 iterations for the test set for each parameter configuration were used to test the effect of embedding methods on prediction quality. The results were calculated on data from 7 cities separately: Barcelona, Berlin, Budapest, Poznań, Prague, Warsaw and Wrocław. All regions containing stations and an equal number of regions not containing stations were used for learning (imbalance ratio = 1). Prediction quality was tested using the 3 metrics mentioned in the previous section.

\subsection{Results of RQ4}

\begin{figure}[H]
    \centering
    \includegraphics[width=\textwidth]{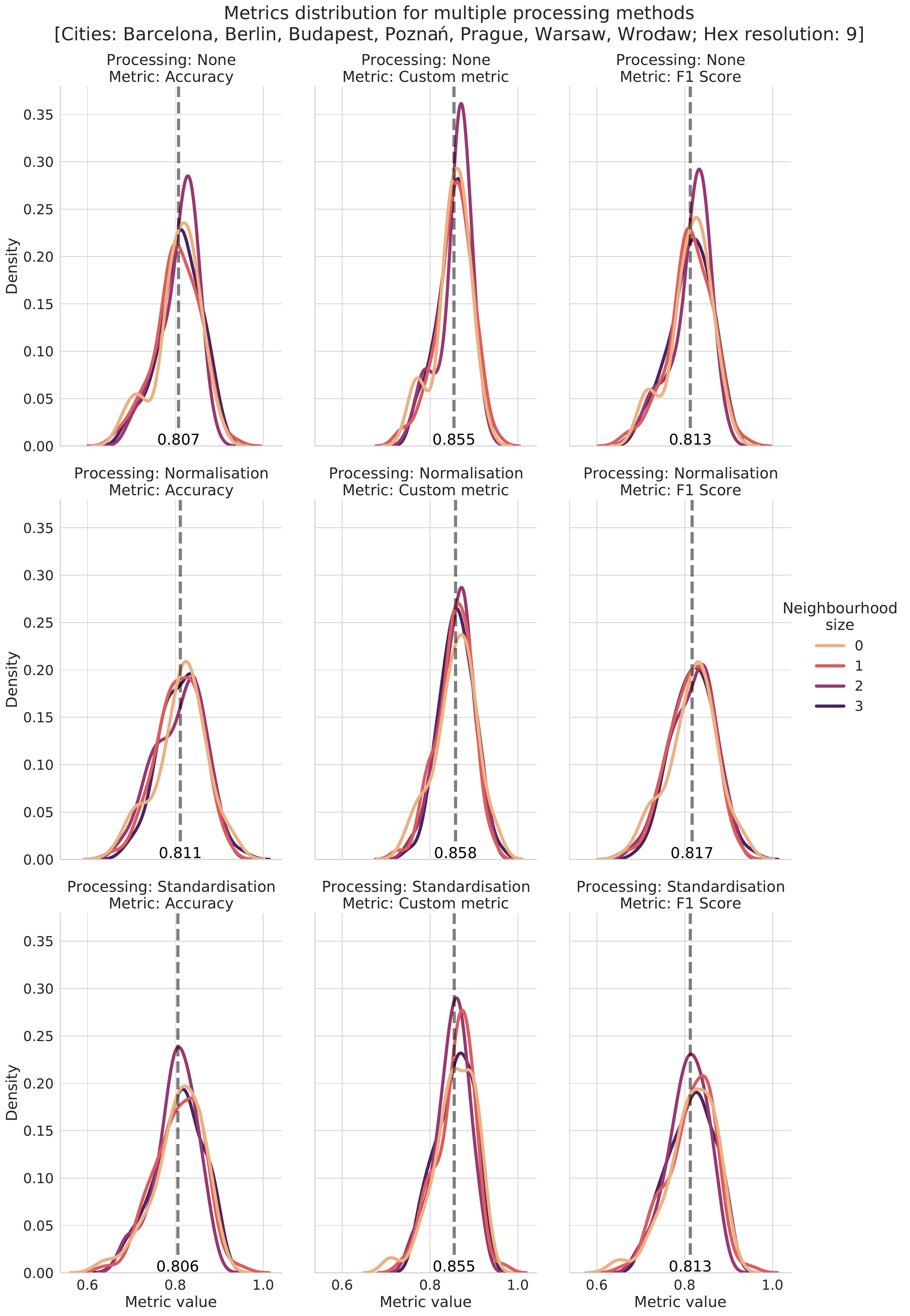}
    \caption[Distribution of prediction results for different features processing methods for hex resolution 9]%
    {Distribution of prediction results for different features processing methods for hex resolution 9. \par \small Personal work.}
    \label{chart:processing_methods_9}
\end{figure}

\begin{figure}[H]
    \centering
    \includegraphics[width=\textwidth]{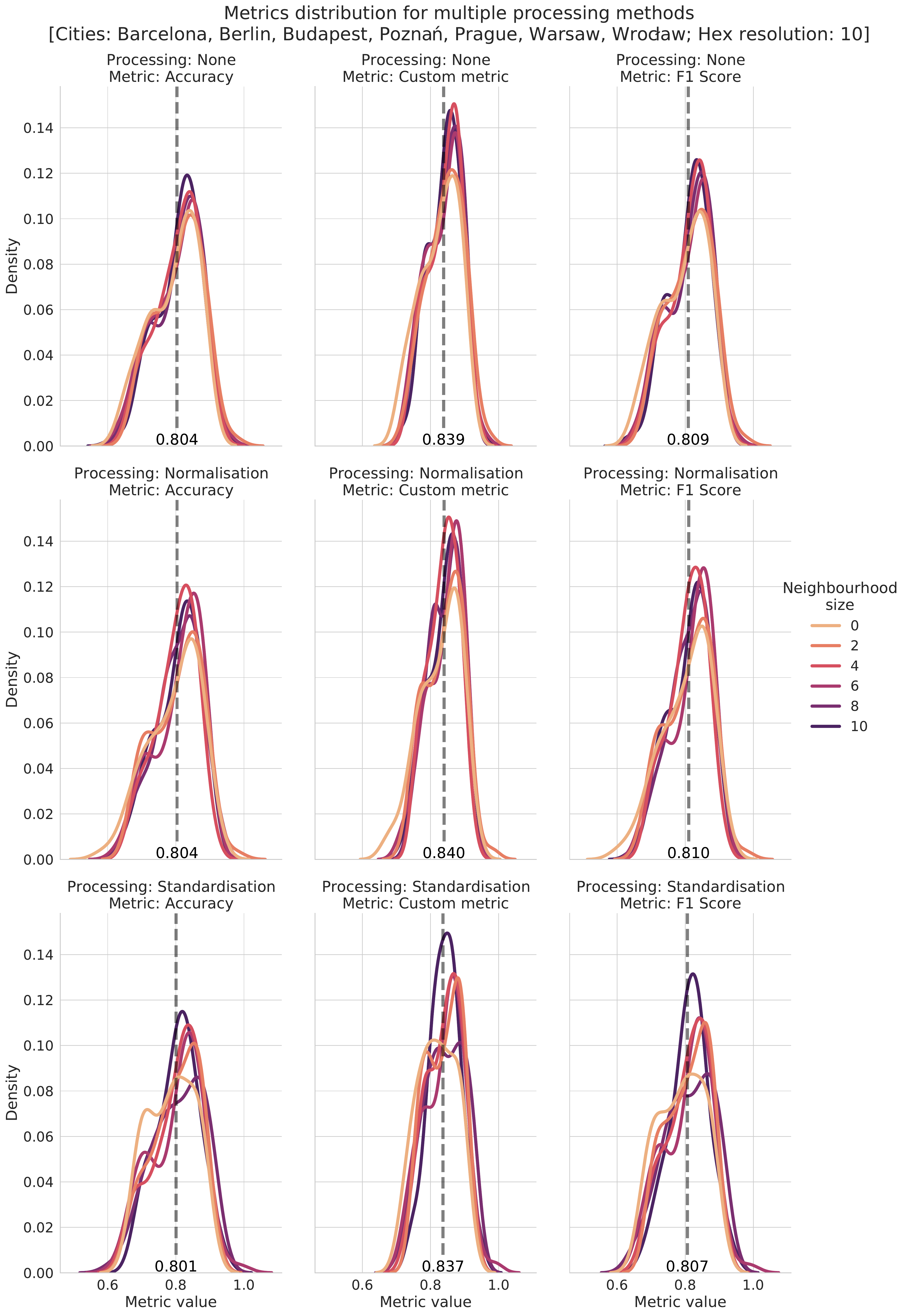}
    \caption[Distribution of prediction results for different features processing methods for hex resolution 10]%
    {Distribution of prediction results for different features processing methods for hex resolution 10. \par \small Personal work.}
    \label{chart:processing_methods_10}
\end{figure}

\begin{figure}[H]
    \centering
    \includegraphics[width=\textwidth]{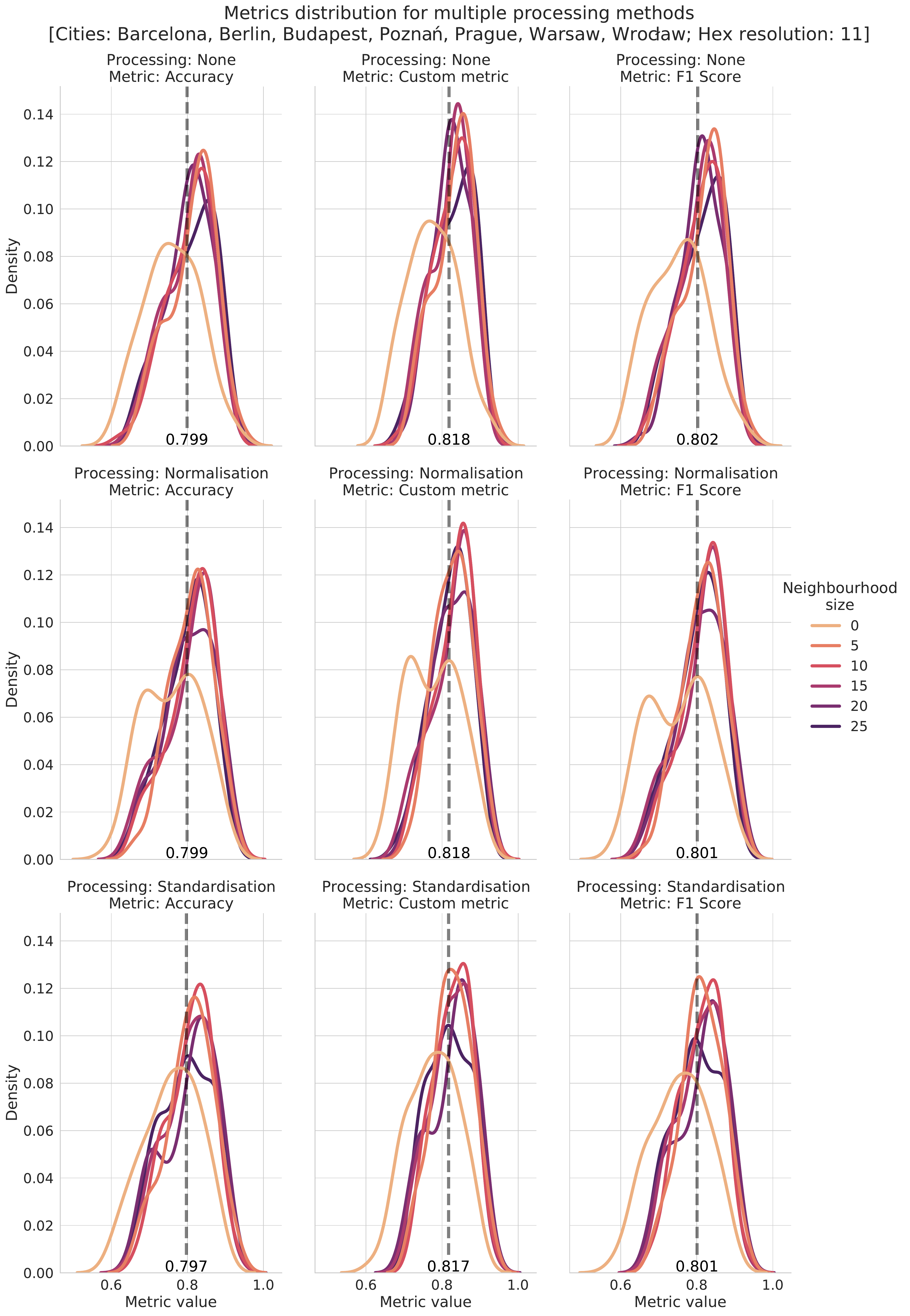}
    \caption[Distribution of prediction results for different features processing methods for hex resolution 11]%
    {Distribution of prediction results for different features processing methods for hex resolution 11. \par \small Personal work.}
    \label{chart:processing_methods_11}
\end{figure}

The results were grouped per resolution of the region (9, 10, 11). All results are plotted in Figures \ref{chart:processing_methods_9}, \ref{chart:processing_methods_10}, \ref{chart:processing_methods_11} and also summarised in the Table \ref{tab:comparing_processing}.

\TopicSetWidth{*}
\TopicSetVPos{t}
\TopicSetContinuationCode{}
\begin{topiclongtable}{@{}FcTc|TcTcTcTcTc@{}}

\toprule
    \begin{tabular}[c]{@{}c@{}}Hex\\ resolution\end{tabular} &
    Metric &
    None &
    Normalisation &
    Standardisation
\\* \midrule
\endhead
\hline\endfoot
\endlastfoot
\TopicLine \Topic[9] & Accuracy & 0.807 & \textbf{0.811} & 0.806 \\
\TopicLine \Topic[9] & Custom   & 0.855 & \textbf{0.858} & 0.855 \\
\TopicLine \Topic[9] & F1 Score & 0.813 & \textbf{0.817} & 0.813 \\

\TopicLine \Topic[10] & Accuracy & \textbf{0.804} & \textbf{0.804} & 0.801 \\
\TopicLine \Topic[10] & Custom   & 0.839 & \textbf{0.840} & 0.837 \\
\TopicLine \Topic[10] & F1 Score & 0.809 & \textbf{0.810} & 0.807 \\

\TopicLine \Topic[11] & Accuracy & \textbf{0.799} & \textbf{0.799} & 0.797 \\
\TopicLine \Topic[11] & Custom   & \textbf{0.818} & \textbf{0.818} & 0.817 \\
\TopicLine \Topic[11] & F1 Score & \textbf{0.802} & 0.801 & 0.801 \\

\bottomrule
\caption[Comparison of different feature processing methods]%
{Comparison of different feature processing methods. Results for the cities of Barcelona, Berlin, Budapest, Poznań, Prague, Warsaw and Wrocław calculated on the test set. Values represent the average of all grouped results for different parameters. The best values in each row were highlighted.}
\label{tab:comparing_processing}\\
\end{topiclongtable}

Although the results are very close, it is clear that normalisation allows the results to improve in each case, so this method of feature scaling will be used in further research.

\newpage

\section{RQ5: How the imbalance ratio affects performance?}

Until now, to maintain a balance in learning between regions with and without stations, all micro-regions with stations were used in the learning set and an equal number of microregions without stations were sampled. The ratio between the classes was 1:1. Unfortunately, in this approach only a small subset of the regions from the entire city coverage is selected during the draw, thus a large part of the city area is missed. The effect is exacerbated at higher resolutions as the size of the regions decreases. Therefore, this step will investigate the effect of the imbalance ratio on the obtained results. The goal is to use the highest possible ratio to cover as much city size as possible in the learning, without significant loss of F1 Score measure prediction quality. In addition to the baseline random forest classifier, two values of the class balance parameter will be investigated: balanced and balanced per sample, to see if these parameters will produce better results than the baseline classifier.

\subsection{Method of RQ5}

To investigate how the classifier behaves as the class imbalance ratio increases, measures of precision, recall and balanced accuracy were additionally recorded. As in the previous steps, the results were averaged over 10 iterations on the test set. Experiments were conducted on data from 7 cities (Barcelona, Berlin, Budapest, Poznań, Prague, Warsaw, and Wrocław), with random forest classifier, using the category counting embedding method, average diminishing squared neighbour embedding method and feature normalisation. Additionally, the random forest has been tested in 3 modes: normal, balanced and balanced per sample using \texttt{class\_balance} parameter from the scikit-learn library.

\subsection{Results of RQ5}

\begin{figure}[H]
    \centering
    \includegraphics[width=\textwidth]{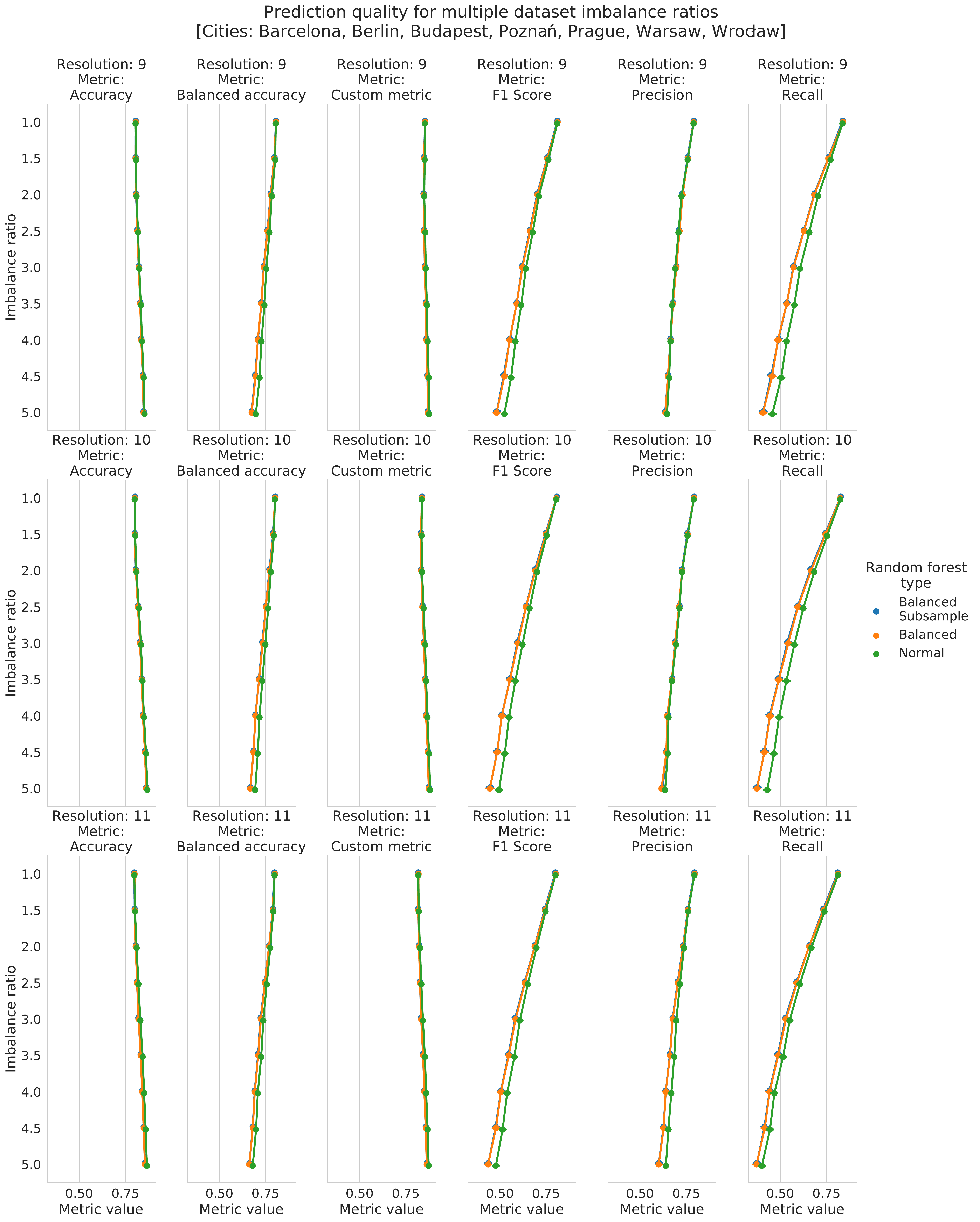}
    \caption[Prediction quality for multiple different imbalance ratios]%
    {Prediction quality for multiple different imbalance ratios. \par \small Personal work.}
    \label{chart:imbalance_ratios}
\end{figure}

The results were grouped per resolution of the region (9, 10, 11). All results are plotted in Figure \ref{chart:imbalance_ratios}. Additionally, the results for the random forest classifier without any class balancing are summarised in the Table \ref{tab:comparing_imbalance}.

\TopicSetWidth{*}
\TopicSetVPos{t}
\TopicSetContinuationCode{}
\begin{topiclongtable}{@{}FcTc|TcTcTcTcTcTcTcTcTc@{}}

\toprule
    \begin{tabular}[c]{@{}c@{}}Hex\\ res.\end{tabular} &
    Metric &
    1.0 & 1.5 & 2.0 & 2.5 & 3.0 & 3.5 & 4.0 & 4.5 & 5.0
\\* \midrule
\endhead
\hline\endfoot
\endlastfoot
\TopicLine \Topic[9] & Acc.  & 0.805 & 0.808 & 0.810 & 0.818 & 0.825 & 0.833 & 0.840 & 0.849 & 0.853 \\
\TopicLine \Topic[9] & \begin{tabular}[c]{@{}c@{}}
Bal.\\ acc. \end{tabular}    & 0.805 & 0.802 & 0.783 & 0.771 & 0.753 & 0.743 & 0.727 & 0.717 & 0.697 \\
\TopicLine \Topic[9] & Cust. & 0.853 & 0.852 & 0.850 & 0.854 & 0.858 & 0.864 & 0.868 & 0.874 & 0.876 \\
\TopicLine \Topic[9] & F1    & 0.810 & 0.762 & 0.711 & 0.677 & 0.640 & 0.616 & 0.584 & 0.561 & 0.524 \\
\TopicLine \Topic[9] & Prec. & 0.789 & 0.757 & 0.724 & 0.708 & 0.690 & 0.674 & 0.664 & 0.657 & 0.646 \\
\TopicLine \Topic[9] & Rec.  & 0.836 & 0.772 & 0.704 & 0.656 & 0.606 & 0.575 & 0.534 & 0.504 & 0.457 \\

\TopicLine \Topic[10] & Acc.  & 0.800 & 0.803 & 0.809 & 0.823 & 0.834 & 0.842 & 0.850 & 0.861 & 0.868 \\
\TopicLine \Topic[10] & \begin{tabular}[c]{@{}c@{}}
Bal.\\ acc. \end{tabular}     & 0.800 & 0.795 & 0.778 & 0.763 & 0.748 & 0.732 & 0.717 & 0.708 & 0.693 \\
\TopicLine \Topic[10] & Cust. & 0.836 & 0.835 & 0.837 & 0.846 & 0.854 & 0.860 & 0.866 & 0.875 & 0.881 \\
\TopicLine \Topic[10] & F1    & 0.805 & 0.753 & 0.702 & 0.661 & 0.622 & 0.584 & 0.549 & 0.527 & 0.495 \\
\TopicLine \Topic[10] & Prec. & 0.791 & 0.757 & 0.727 & 0.713 & 0.694 & 0.672 & 0.654 & 0.650 & 0.635 \\
\TopicLine \Topic[10] & Rec.  & 0.824 & 0.753 & 0.683 & 0.624 & 0.576 & 0.533 & 0.493 & 0.465 & 0.429 \\

\TopicLine \Topic[11] & Acc.  & 0.798 & 0.802 & 0.811 & 0.820 & 0.831 & 0.843 & 0.851 & 0.859 & 0.866 \\
\TopicLine \Topic[11] & \begin{tabular}[c]{@{}c@{}}
Bal.\\ acc. \end{tabular}     & 0.798 & 0.791 & 0.775 & 0.756 & 0.737 & 0.726 & 0.707 & 0.698 & 0.680 \\
\TopicLine \Topic[11] & Cust. & 0.817 & 0.820 & 0.826 & 0.834 & 0.842 & 0.853 & 0.860 & 0.868 & 0.873 \\
\TopicLine \Topic[11] & F1    & 0.801 & 0.747 & 0.698 & 0.651 & 0.608 & 0.578 & 0.539 & 0.515 & 0.478 \\
\TopicLine \Topic[11] & Prec. & 0.794 & 0.760 & 0.738 & 0.715 & 0.696 & 0.685 & 0.669 & 0.653 & 0.640 \\
\TopicLine \Topic[11] & Rec.  & 0.812 & 0.739 & 0.668 & 0.605 & 0.551 & 0.515 & 0.468 & 0.444 & 0.401 \\

\bottomrule
\caption[Comparison of different imbalance ratios]%
{Comparison of different imbalance ratios. Results for the cities of Barcelona, Berlin, Budapest, Poznań, Prague, Warsaw and Wrocław calculated on the test set. Values represent the average of all grouped results for different parameters for Random Forest without any balance adjustments.}
\label{tab:comparing_imbalance}\\
\end{topiclongtable}

Intuitively, as the class imbalance increases, the accuracy increases and the F1 score decreases. The quite high precision value indicates that a high percentage of all regions predicted to contain stations actually contain them. In contrast, the rapidly decreasing recall value indicates that the model has difficulty predicting all regions containing stations and marks some of them as not containing stations. This may be due to the situation when the direct neighbourhood of the region that contains stations is used in the learning - then there is a probability that 6 neighbouring regions have very similar embedding vector as the central region and there is an overrepresentation of similar regions as not containing stations in the training dataset.

\newpage

\begin{figure}[H]
    \centering
    \begin{minipage}{.5\textwidth}
        \centering
        \subfloat[Imbalance ratio: 1.0]{\label{fig:wro_avg_imb:1}\includegraphics[width=0.95\linewidth, keepaspectratio]{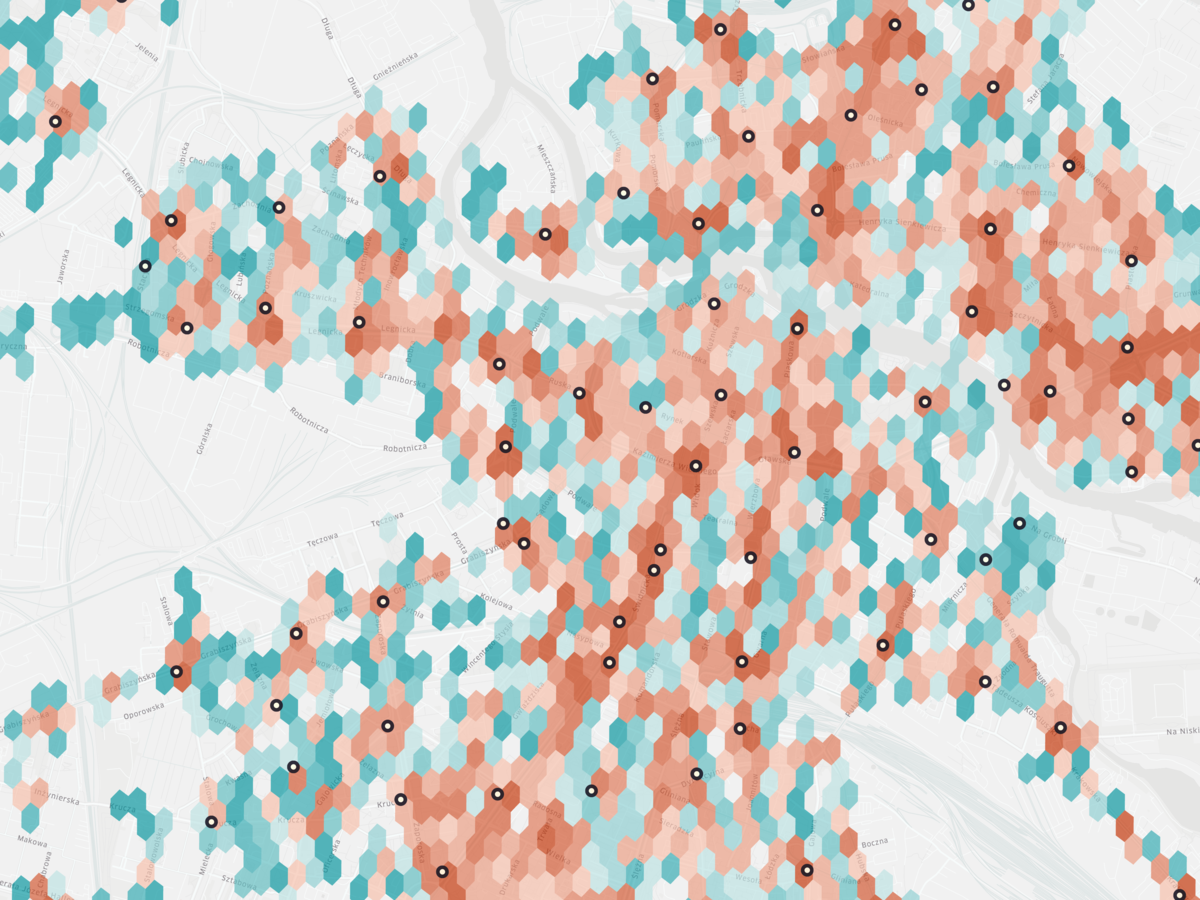}}
    \end{minipage}%
    \begin{minipage}{.5\textwidth}
        \centering
        \subfloat[Imbalance ratio: 2.0]{\label{fig:wro_avg_imb:2}\includegraphics[width=0.95\linewidth, keepaspectratio]{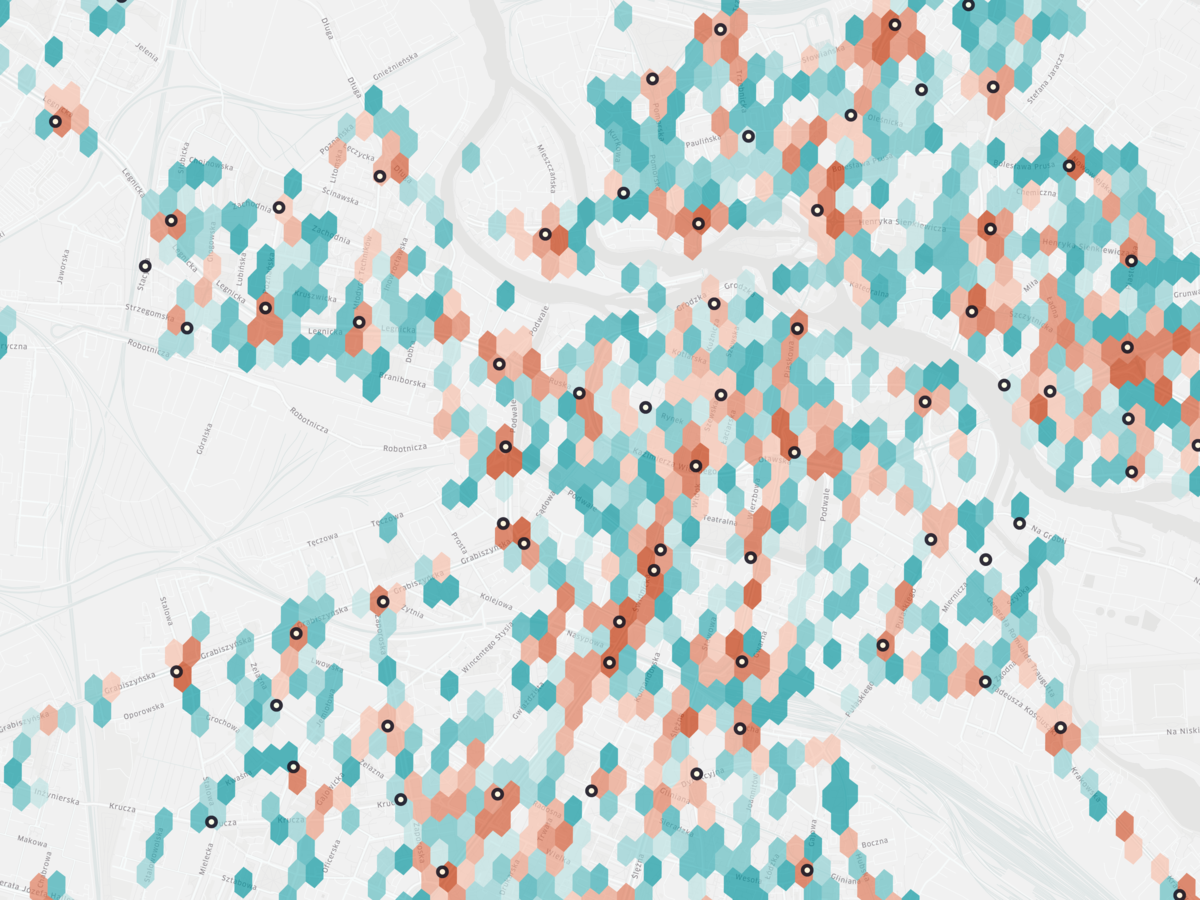}}
    \end{minipage}
    \par\medskip
    \begin{minipage}{.5\textwidth}
        \centering
        \subfloat[Imbalance ratio: 3.0]{\label{fig:wro_avg_imb:3}\includegraphics[width=0.95\linewidth, keepaspectratio]{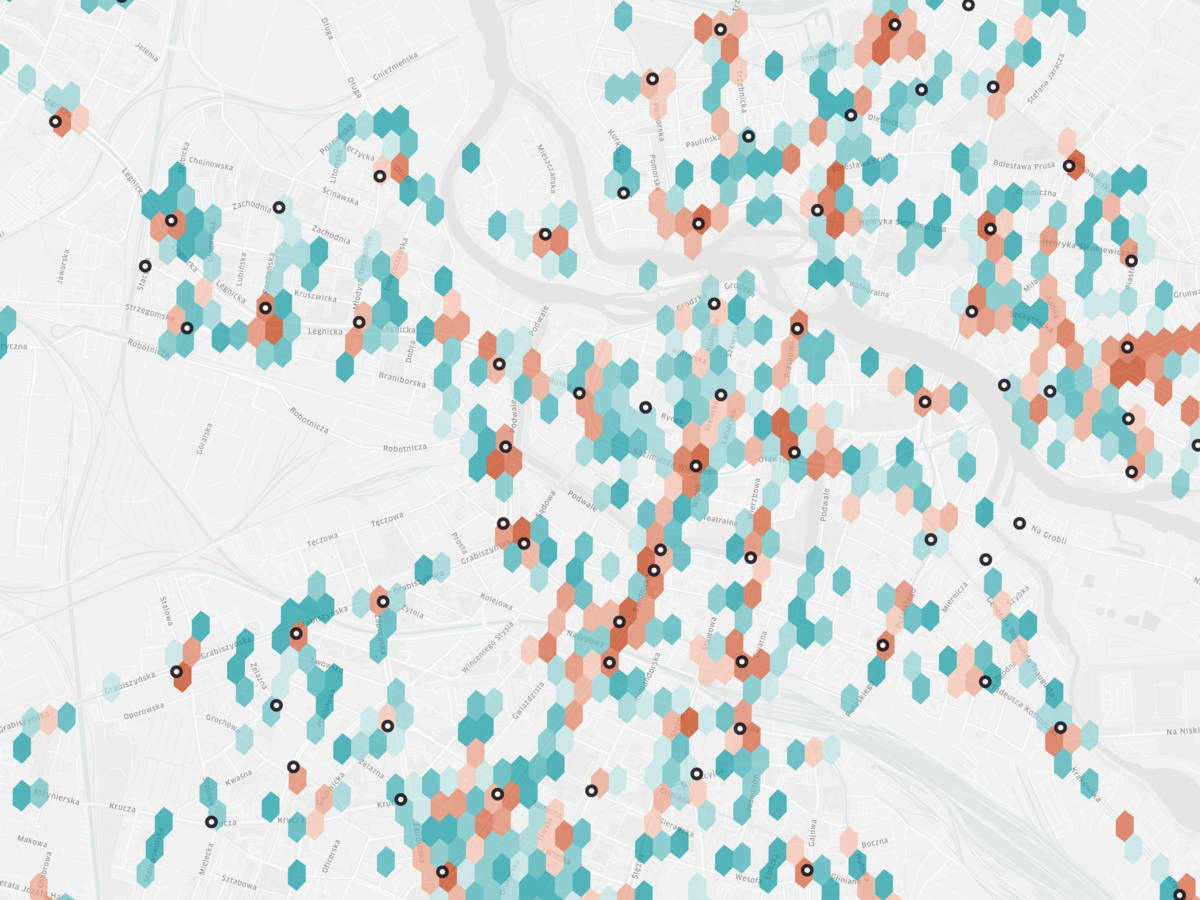}}
    \end{minipage}%
    \begin{minipage}{.5\textwidth}
        \centering
        \subfloat[Imbalance ratio: 4.0]{\label{fig:wro_avg_imb:4}\includegraphics[width=0.95\linewidth, keepaspectratio]{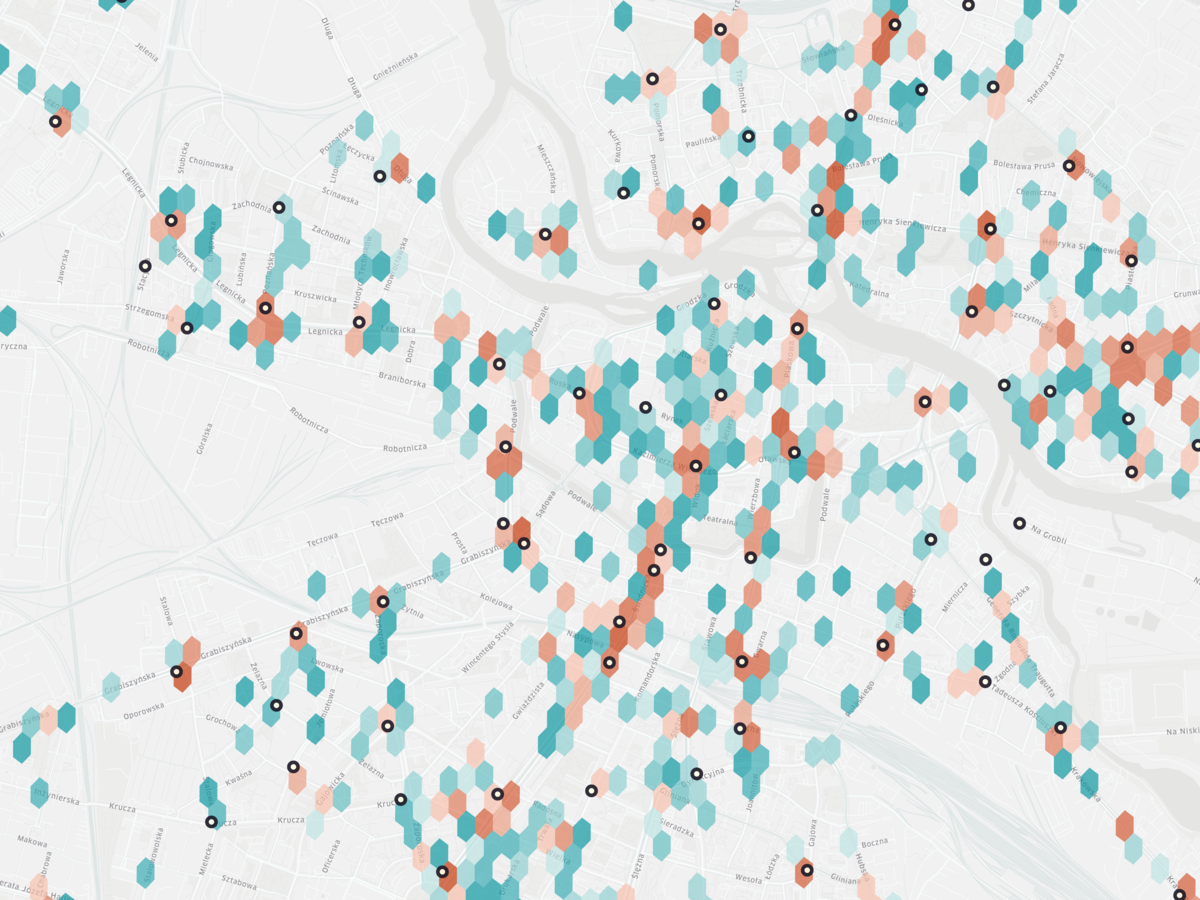}}
    \end{minipage}
    \par\medskip
    \begin{minipage}{\textwidth}
        \centering
        \subfloat[Imbalance ratio: 5.0]{\label{fig:wro_avg_imb:5}\includegraphics[width=0.475\linewidth, keepaspectratio]{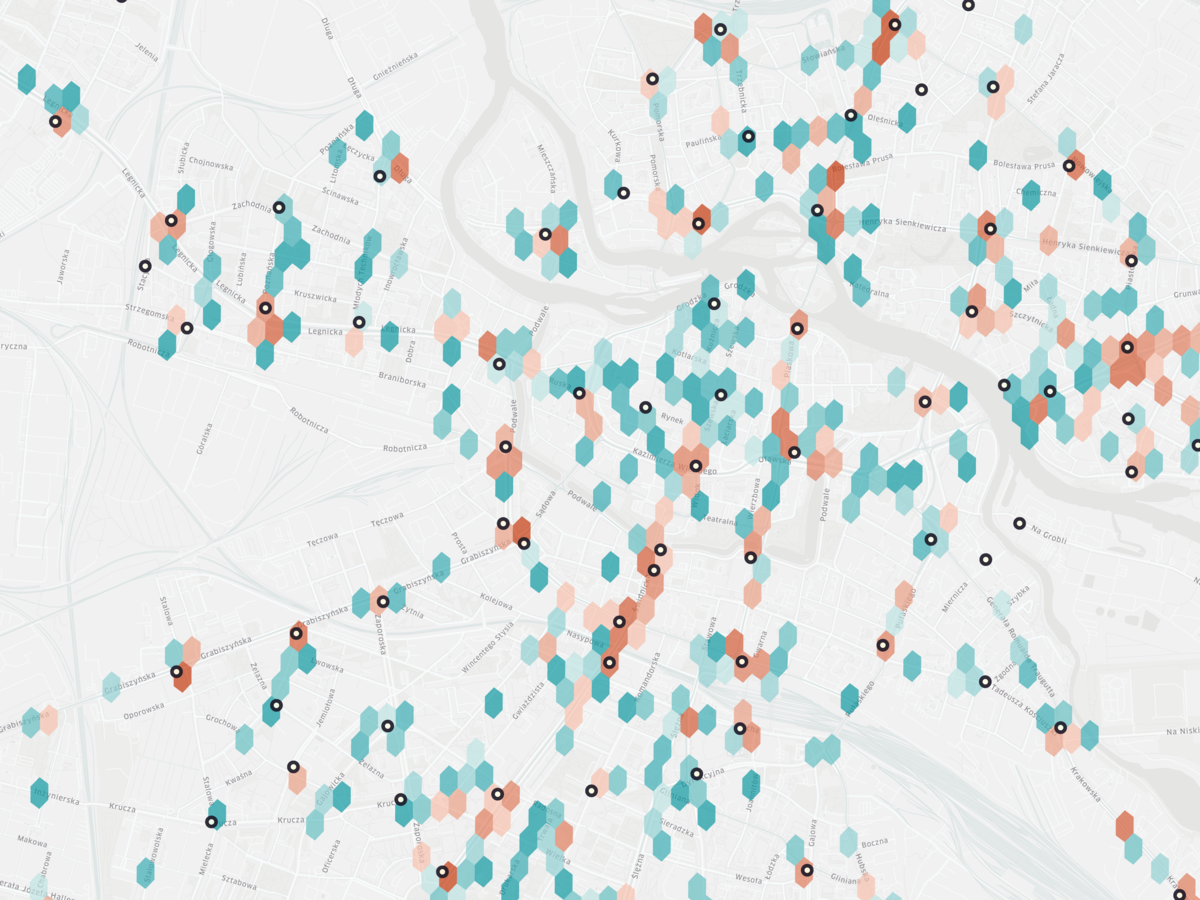}}
    \end{minipage}
    
    \caption[Comparison of different imbalance ratios on the example of Wrocław]%
    {Comparison of different imbalance ratios on the example of Wrocław. \par
    Maps show the probability of station occurrence as coloured hexes. Results have been generated for resolution 10 with a neighbourhood size of 10 and averaged from 100 iterations. Results below probability 0.5 have been filtered. The dark teal colour represents the probability of 50\% and the dark red colour represents the probability of 100\%. White dots with black borders represent existing stations.
    \par \small Personal work. Rendered using kepler.gl library.}
    \label{fig:wro_avg_imb}
\end{figure}

To better understand the influence of the hyperparameter on the performance quality, simulations for values 1, 2, 3, 4 and 5 were performed on the example of Wroclaw and visualised as heat maps. The simulations were run on resolution regions of 10 with a neighbourhood size of 10 and repeated 100 times and then averaged. The maps can be seen in Figure \ref{fig:wro_avg_imb}.

\newpage

\begin{figure}[H]
    \centering
    \includegraphics[width=\textwidth]{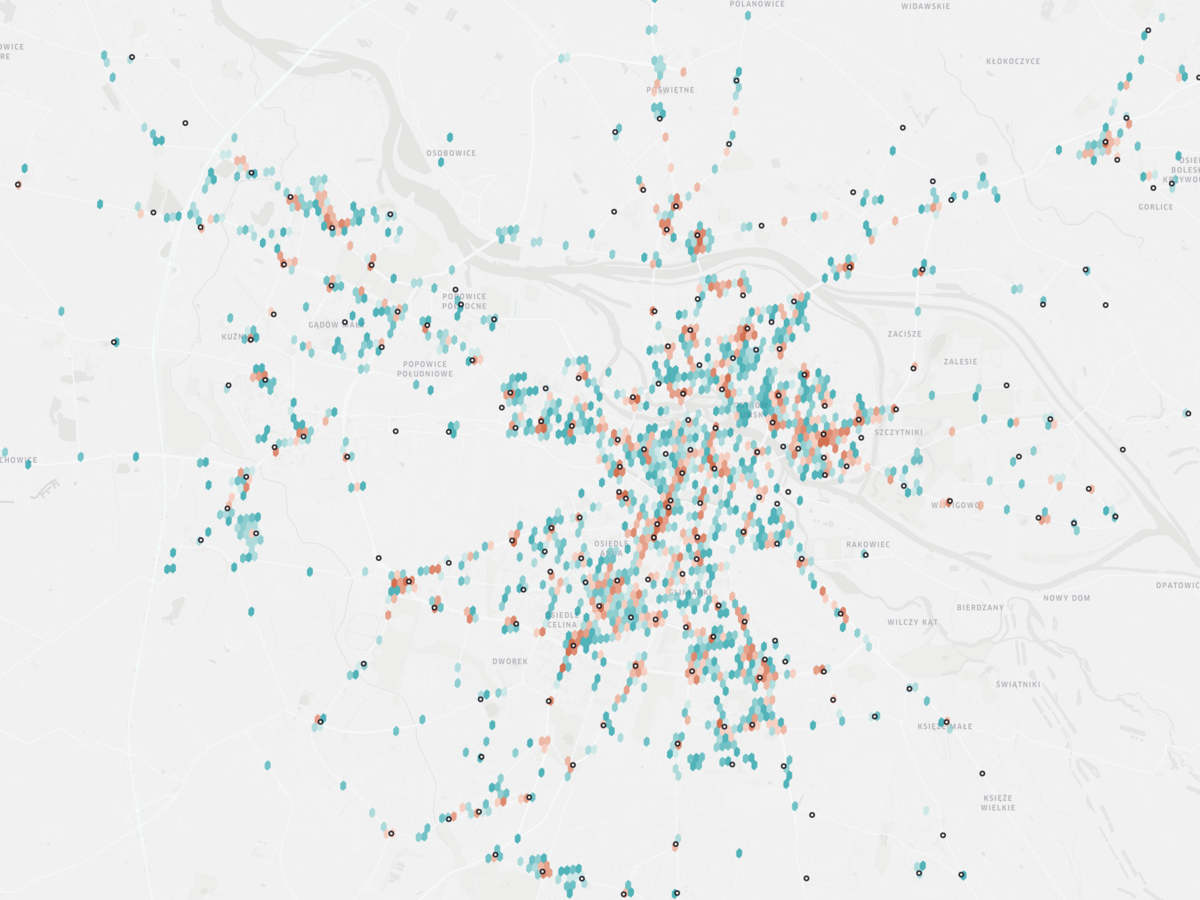}
    \caption[Performance of the method for the city of Wroclaw for class imbalance ratio of 2.5]%
    {Performance of the method for the city of Wroclaw for a resolution of 10 with a neighbourhood size of 10 and a class imbalance ratio of 2.5 averaged from 100 iterations. Results below probability 0.5 have been filtered. The dark teal colour represents the probability of 50\% and the dark red colour represents the probability of 100\%. White dots with black borders represent existing stations. \par \small Personal work. Rendered using kepler.gl library.}
    \label{fig:wro_avg_imb_2.5}
\end{figure}

Based on the visualisation and precision and recall results, it was decided to choose a value for the class imbalance parameter equal to 2.5. It can be seen from the maps that for this value, not too many regions are marked as containing stations, which is a good sign because it is useful that the method does not mark the whole city as potential positions for stations. The figure \ref{fig:wro_avg_imb_2.5} shows a distant view of the city with probability predictions for a class imbalance ratio equal to 2.5.

\newpage

\section{RQ6: How the resolution of regions and the size of region neighbourhood affect the prediction performance?}

From the point of view of the urban planner, the proposed method will be more useful, if it could more accurately indicate the region of the city in which the station should be located. Based on the previous results, it can be seen that as the resolution increases, the quality of the prediction decreases slightly. However, these were results averaged per resolution for many neighbourhood sizes. This question aims to find a suitable neighbourhood size that will produce good results but at the same time low enough not to unnecessarily increase the computational complexity of the method.

\subsection{Method of RQ6}

As in the previous steps, the results were averaged over 10 repetitions for a particular set of parameters: imbalance ratio - 2.5, region embedding method - Category counting, neighbourhood embedding method - Averaging diminishing squared, classifier - Random forest. Additionally, to enlarge the set of results, the set of cities studied was increased to 11: Barcelona, Berlin, Brussels, Budapest, Dublin, Helsinki, Lyon, Poznań, Prague, Warsaw, Wrocław.

\subsection{Results of RQ6}

\begin{figure}[H]
    \centering
    \includegraphics[width=\textwidth]{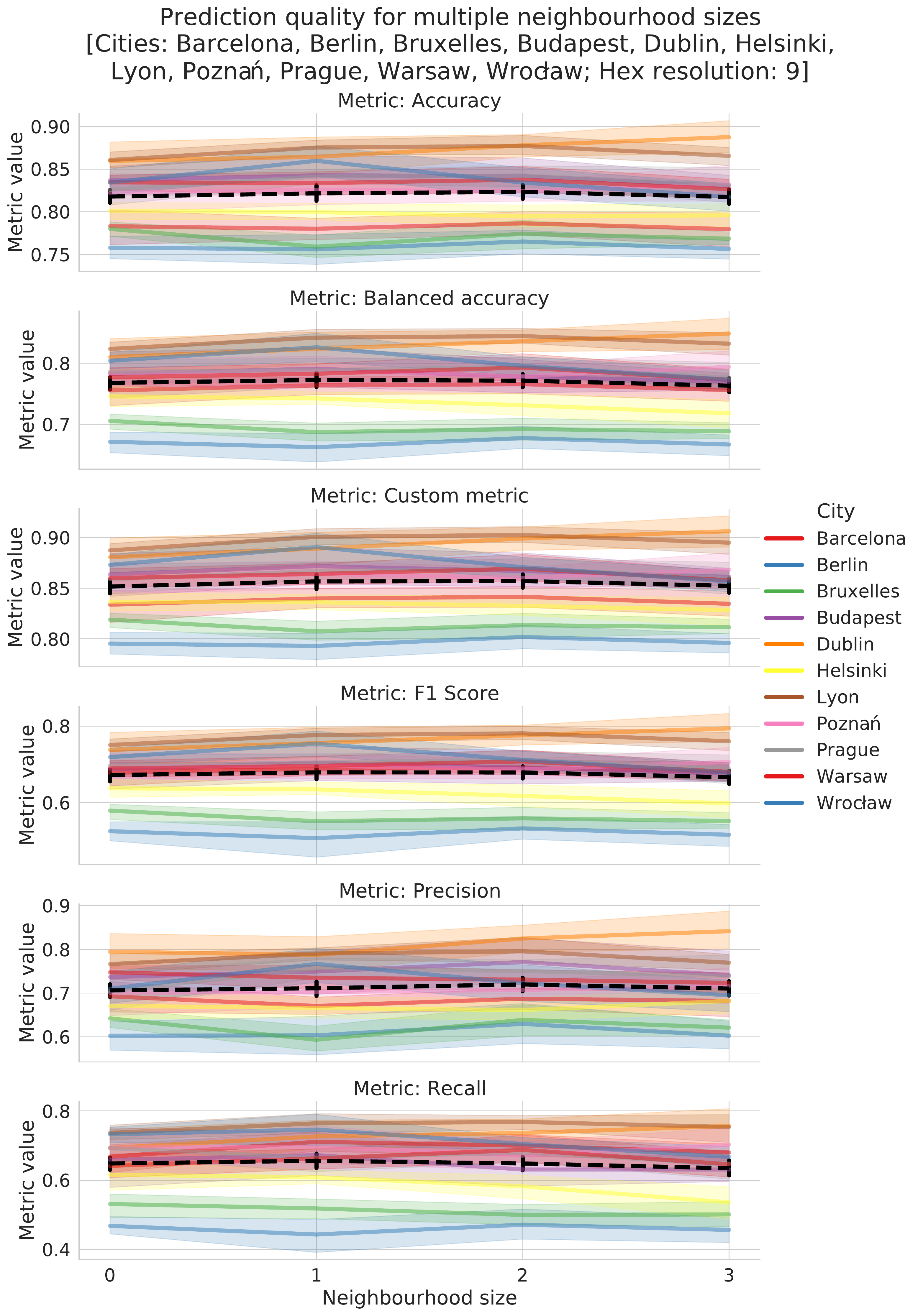}
    \caption[Prediction quality for multiple neighbourhood sizes for hex resolution 9]%
    {Prediction quality for multiple neighbourhood sizes for hex resolution 9. The dashed black line represents the average value from all cities. \par \small Personal work.}
    \label{chart:neighbourhood_size_9}
\end{figure}

\begin{figure}[H]
    \centering
    \includegraphics[width=\textwidth]{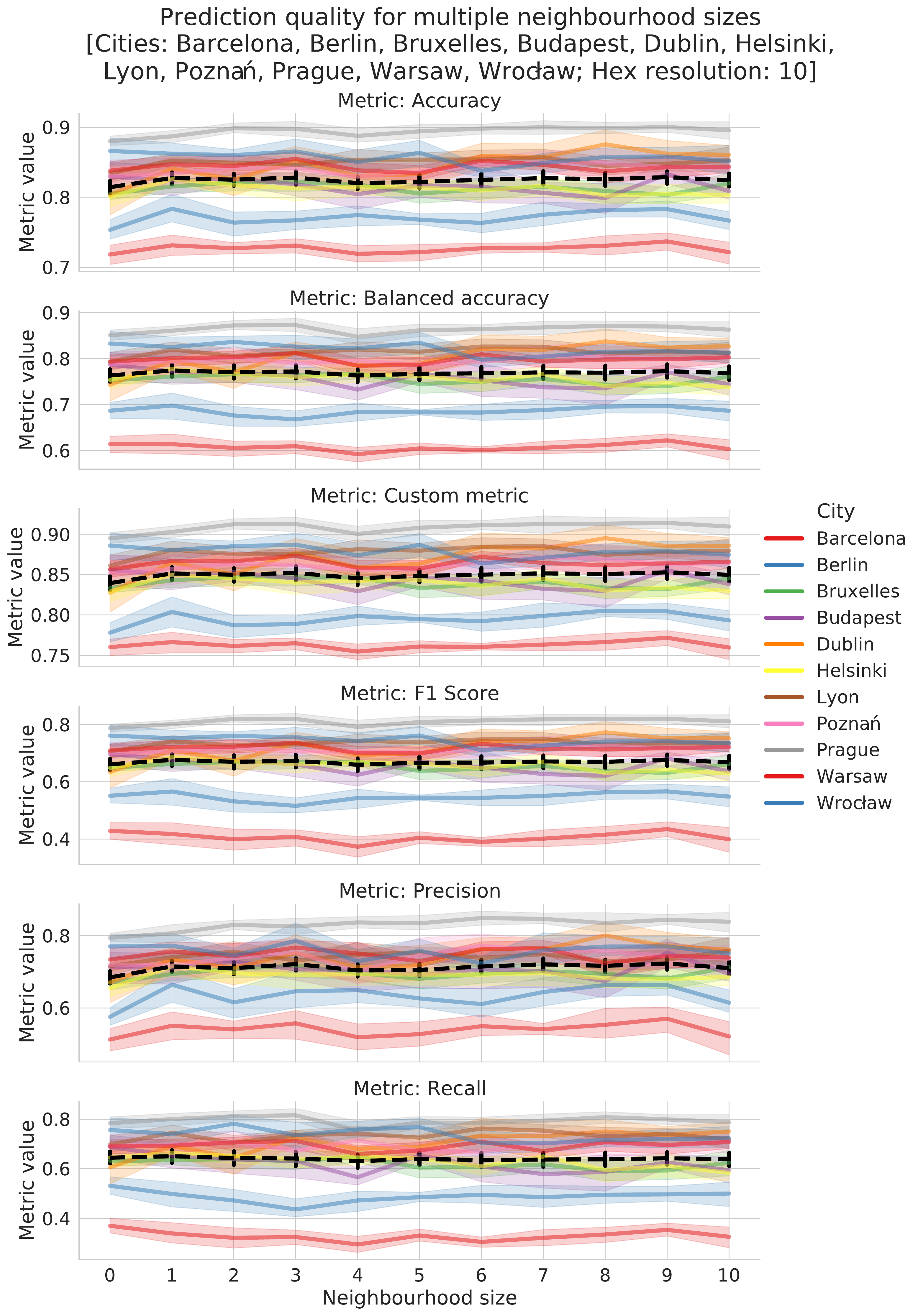}
    \caption[Prediction quality for multiple neighbourhood sizes for hex resolution 10]%
    {Prediction quality for multiple neighbourhood sizes for hex resolution 10. The dashed black line represents the average value from all cities. \par \small Personal work.}
    \label{chart:neighbourhood_size_10}
\end{figure}

\begin{figure}[H]
    \centering
    \includegraphics[width=\textwidth]{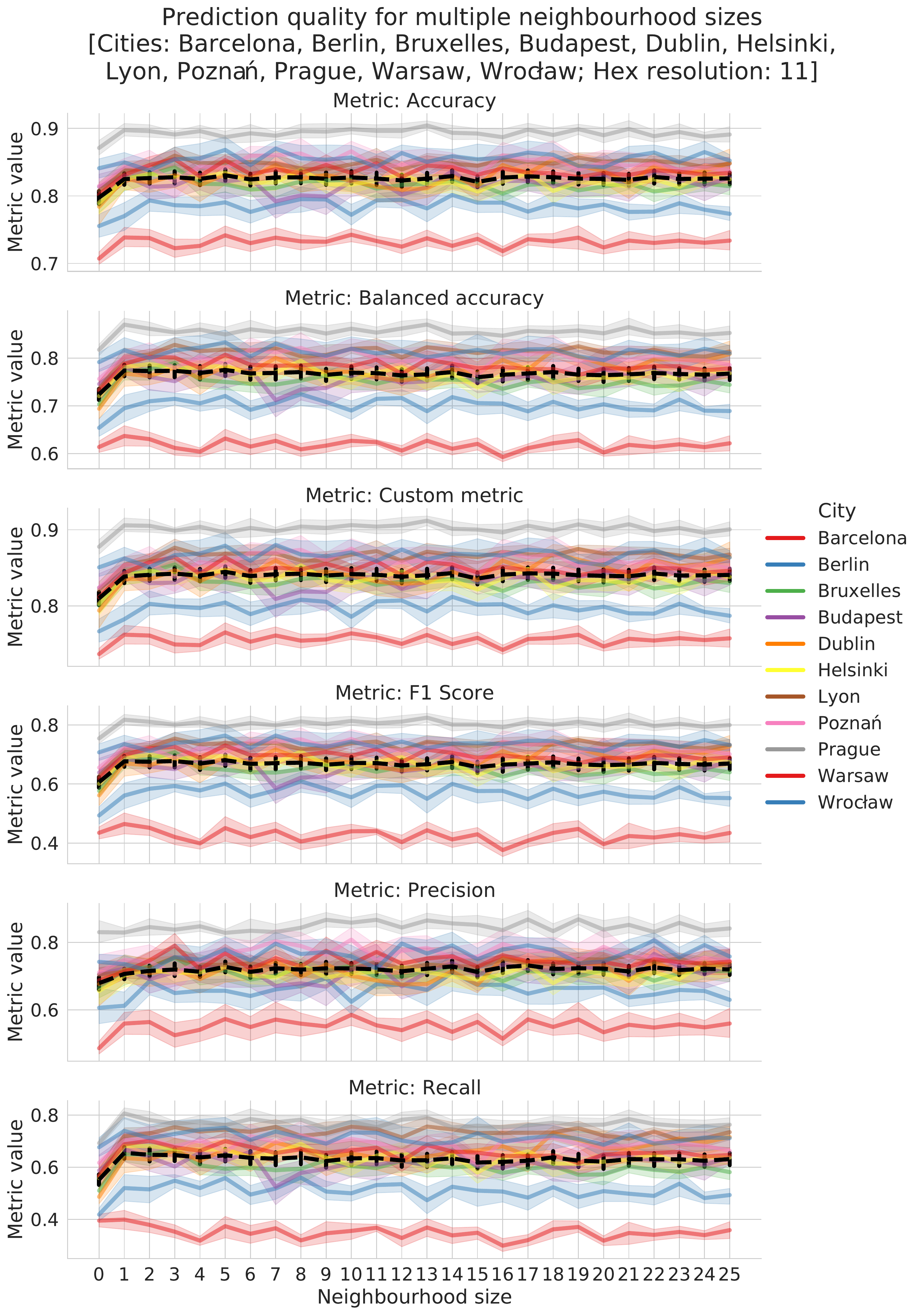}
    \caption[Prediction quality for multiple neighbourhood sizes for hex resolution 11]%
    {Prediction quality for multiple neighbourhood sizes for hex resolution 11. The dashed black line represents the average value from all cities. \par \small Personal work.}
    \label{chart:neighbourhood_size_11}
\end{figure}

All results are plotted in Figures \ref{chart:neighbourhood_size_9}, \ref{chart:neighbourhood_size_10}, \ref{chart:neighbourhood_size_11},  and also summarised in the Table \ref{tab:comparing_neighbourhood}.

\TopicSetWidth{*}
\TopicSetVPos{t}
\TopicSetContinuationCode{}
\begin{topiclongtable}{@{}FcTc|TcTcTcTcTcTc@{}}

\toprule
    \begin{tabular}[c]{@{}c@{}}Hex\\ res.\end{tabular} &
    \begin{tabular}[c]{@{}c@{}}Neigh.\\ size\end{tabular} &
    Accuracy &
    \begin{tabular}[c]{@{}c@{}}Balanced\\ accuracy\end{tabular} &
    \begin{tabular}[c]{@{}c@{}}Custom\\ metric\end{tabular} &
    F1 Score &
    Precision &
    Recall
\\* \midrule
\endhead
\hline\endfoot
\endlastfoot
\TopicLine \Topic[9] & 0 & 0.818 & 0.768 & 0.852 & 0.673 & 0.706 & 0.649 \\
\TopicLine \Topic[9] & 1 & 0.822 & 0.772 & 0.857 & 0.679 & 0.711 & 0.656 \\
\TopicLine \Topic[9] & \textbf{2} & \textbf{0.823} & \textbf{0.771} & \textbf{0.857} & \textbf{0.679} & \textbf{0.720} & \textbf{0.649} \\
\TopicLine \Topic[9] & 3 & 0.817 & 0.763 & 0.852 & 0.667 & 0.710 & 0.635 \\

\TopicLine \Topic[10] & 0  & 0.814 & 0.763 & 0.840 & 0.661 & 0.685 & 0.644 \\
\TopicLine \Topic[10] & \textbf{1} & \textbf{0.828} & \textbf{0.775} & \textbf{0.851} & \textbf{0.677} & \textbf{0.715} & \textbf{0.650} \\
\TopicLine \Topic[10] & 2  & 0.825 & 0.771 & 0.850 & 0.670 & 0.710 & 0.643 \\
\TopicLine \Topic[10] & 3  & 0.828 & 0.772 & 0.852 & 0.673 & 0.721 & 0.641 \\
\TopicLine \Topic[10] & 4  & 0.820 & 0.763 & 0.846 & 0.660 & 0.704 & 0.631 \\
\TopicLine \Topic[10] & 5  & 0.822 & 0.767 & 0.848 & 0.667 & 0.705 & 0.639 \\
\TopicLine \Topic[10] & 6  & 0.825 & 0.768 & 0.850 & 0.667 & 0.715 & 0.635 \\
\TopicLine \Topic[10] & 7  & 0.827 & 0.770 & 0.852 & 0.671 & 0.721 & 0.637 \\
\TopicLine \Topic[10] & 8  & 0.826 & 0.770 & 0.851 & 0.670 & 0.717 & 0.638 \\
\TopicLine \Topic[10] & 9  & 0.829 & 0.773 & 0.853 & 0.676 & 0.723 & 0.642 \\
\TopicLine \Topic[10] & 10 & 0.824 & 0.769 & 0.850 & 0.669 & 0.710 & 0.639 \\

\TopicLine \Topic[11] & 0 & 0.798 & 0.724 & 0.810 & 0.606 & 0.679 & 0.552 \\
\TopicLine \Topic[11] & 1 & 0.825 & 0.774 & 0.839 & 0.677 & 0.706 & 0.655 \\
\TopicLine \Topic[11] & 2 & 0.826 & 0.773 & 0.841 & 0.675 & 0.715 & 0.648 \\
\TopicLine \Topic[11] & 3 & 0.828 & 0.773 & 0.842 & 0.678 & 0.720 & 0.647 \\
\TopicLine \Topic[11] & 4 & 0.825 & 0.769 & 0.840 & 0.669 & 0.713 & 0.638 \\
\TopicLine \Topic[11] & \textbf{5} & \textbf{0.831} & \textbf{0.775} & \textbf{0.845} & \textbf{0.681} & \textbf{0.728} & \textbf{0.646} \\
\TopicLine \Topic[11] & 6 & 0.824 & 0.768 & 0.839 & 0.668 & 0.712 & 0.636 \\
\TopicLine \Topic[11] & 7 & 0.827 & 0.769 & 0.842 & 0.671 & 0.723 & 0.633 \\
\TopicLine \Topic[11] & 8 & 0.828 & 0.771 & 0.842 & 0.672 & 0.719 & 0.638 \\
\TopicLine \Topic[11] & 9 & 0.825 & 0.765 & 0.840 & 0.665 & 0.723 & 0.624 \\
\TopicLine \Topic[11] & 10 & 0.827 & 0.769 & 0.842 & 0.671 & 0.723 & 0.634 \\
\TopicLine \Topic[11] & 11 & 0.826 & 0.768 & 0.841 & 0.671 & 0.720 & 0.635 \\
\TopicLine \Topic[11] & 12 & 0.823 & 0.764 & 0.838 & 0.663 & 0.714 & 0.626 \\
\TopicLine \Topic[11] & 13 & 0.826 & 0.766 & 0.840 & 0.667 & 0.722 & 0.627 \\
\TopicLine \Topic[11] & 14 & 0.829 & 0.771 & 0.843 & 0.674 & 0.729 & 0.634 \\
\TopicLine \Topic[11] & 15 & 0.821 & 0.760 & 0.836 & 0.657 & 0.712 & 0.618 \\
\TopicLine \Topic[11] & 16 & 0.827 & 0.765 & 0.841 & 0.665 & 0.726 & 0.621 \\
\TopicLine \Topic[11] & 17 & 0.829 & 0.768 & 0.843 & 0.669 & 0.732 & 0.625 \\
\TopicLine \Topic[11] & 18 & 0.828 & 0.771 & 0.842 & 0.673 & 0.721 & 0.638 \\
\TopicLine \Topic[11] & 19 & 0.826 & 0.765 & 0.840 & 0.667 & 0.723 & 0.625 \\
\TopicLine \Topic[11] & 20 & 0.825 & 0.764 & 0.840 & 0.664 & 0.722 & 0.622 \\
\TopicLine \Topic[11] & 21 & 0.824 & 0.766 & 0.839 & 0.666 & 0.714 & 0.631 \\
\TopicLine \Topic[11] & 22 & 0.828 & 0.769 & 0.843 & 0.671 & 0.726 & 0.630 \\
\TopicLine \Topic[11] & 23 & 0.825 & 0.767 & 0.840 & 0.668 & 0.718 & 0.630 \\
\TopicLine \Topic[11] & 24 & 0.826 & 0.765 & 0.840 & 0.666 & 0.722 & 0.625 \\
\TopicLine \Topic[11] & 25 & 0.826 & 0.767 & 0.841 & 0.669 & 0.720 & 0.631 \\

\bottomrule
\caption[Comparison of different neighbourhood sizes]%
{Comparison of different neighbourhood sizes. Results for the cities of Barcelona, Berlin, Brussels, Budapest, Dublin, Helsinki, Lyon, Poznań, Prague, Warsaw and Wrocław calculated on the test set. Values represent the average of all grouped results for parameters selected in previous steps. One best candidate neighbour size for each resolution has been highlighted.}
\label{tab:comparing_neighbourhood}\\
\end{topiclongtable}

\begin{figure}[H]
    \centering
    \includegraphics[width=\textwidth]{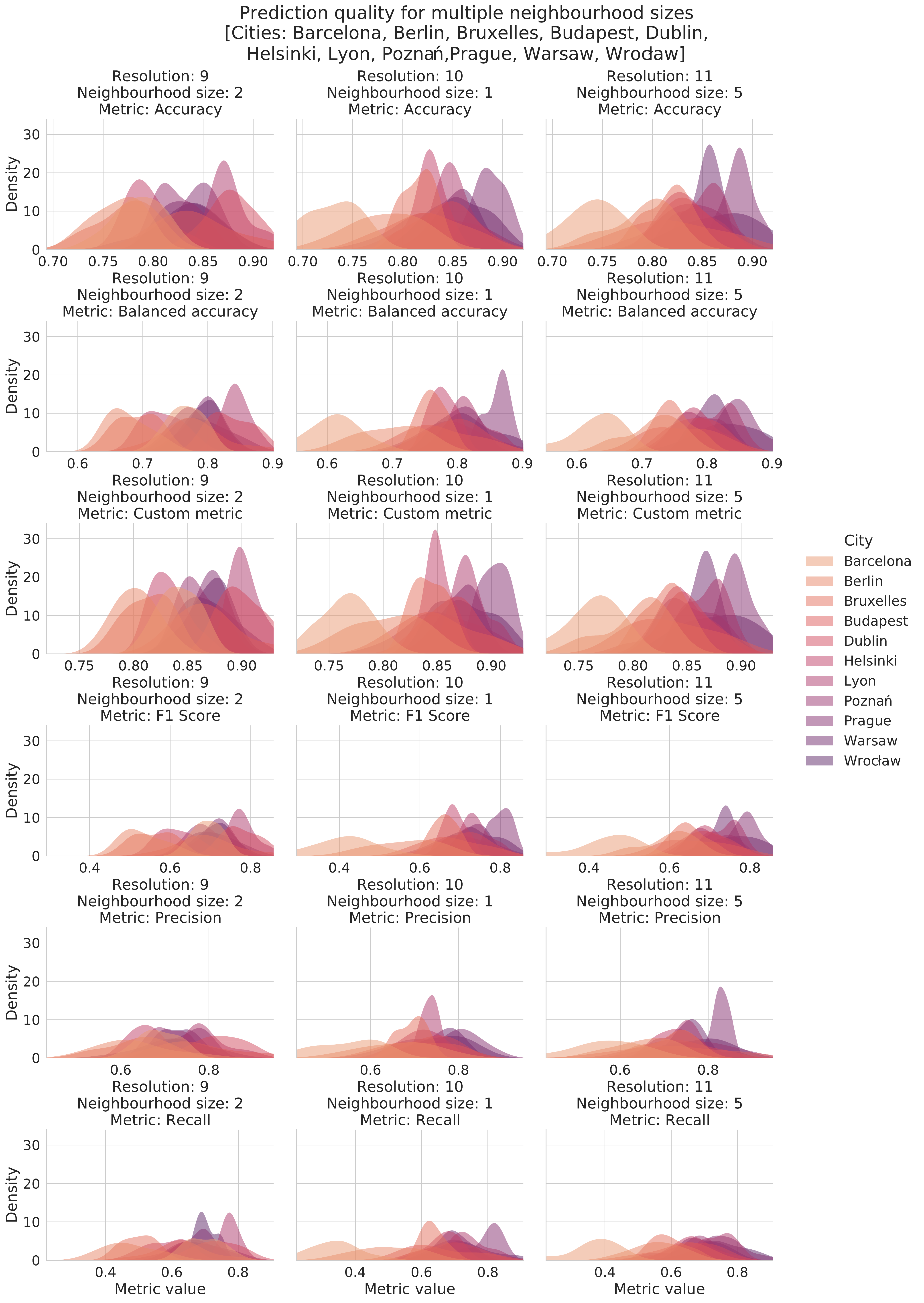}
    \caption[Prediction quality for multiple neighbourhood sizes]%
    {Prediction quality for multiple neighbourhood sizes. Results for 3 best candidates (one from each resolution). \par \small Personal work.}
    \label{chart:comparing_neighbourhood_best}
\end{figure}

The results for the top 3 candidates 
are further detailed in Figure \ref{chart:comparing_neighbourhood_best}.

\newpage

\begin{figure}[H]
    \centering
    \includegraphics[width=\textwidth]{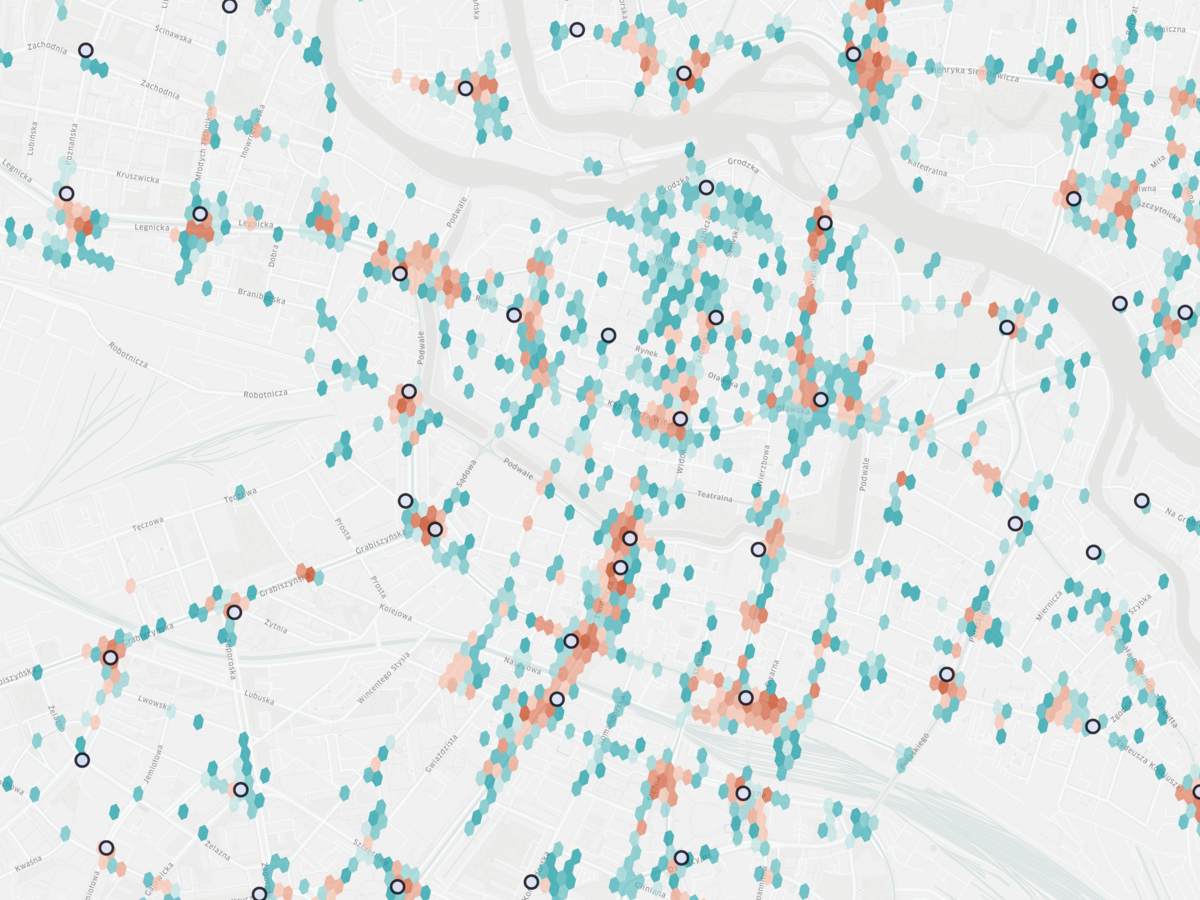}
    \caption[Performance of the method for the city of Wroclaw for resolution 11 with neighbourhood size of 5]%
    {Performance of the method for the city of Wroclaw for resolution 11 with neighbourhood size of 5 and a class imbalance ratio of 2.5 averaged from 100 iterations. Results below probability 0.5 have been filtered. The dark teal colour represents the probability of 50\% and the dark red colour represents the probability of 100\%. White dots with black borders represent existing stations. \par \small Personal work. Rendered using kepler.gl library.}
    \label{fig:wro_avg_res_11_neigh_5}
\end{figure}

As the results between the best candidates are similar, according to the decision to choose the highest resolution, value 11 was chosen. An example of how the method works for the city of Wroclaw for the final combination of parameters can be seen in Figure \ref{fig:wro_avg_res_11_neigh_5}.

The method predicts much more regions than only those that contain stations, which makes the method hard to evaluate using standard metrics for classification. However, it can be seen that the method manages to determine regions that contain stations with more certainty than their neighbourhood. As stations are not very densely distributed in cities, there may be situations where the model considers that a neighbourhood that does not actually contain a station should have one, based on the similarity of the urban significance of the neighbourhood regions.

\section{RQ7: How does the model perform in predicting stations between cities?}
\label{ch:rq7}
 
Once all the hyperparameters have been selected, the next step is to test the usability of the model in predicting stations in another city to be able to use the model learned on a city (or cities) that contains an existing bicycle sharing system and to propose a station layout on a city that does not have such a station layout. This will be tested by learning the model on a dataset from one city and trying to predict stations on a dataset from another city. However, a possible low quality of the prediction does not necessarily imply a weakness of the proposed method but could indicate that the city planners made different decisions about station layout and the station neighbourhoods are functionally completely different.

\subsection{Method of RQ7}

Using the final combination of hyperparameters, a cross-prediction of all 34 cities against each other individually was made. All hexes with stations were selected from the learning city and 2.5 times as many hexes without stations were drawn, which were then used to learn the model. The results were validated on the whole set of the second city. The experiments were repeated 100 times and averaged.

\subsection{Results of RQ7}

\begin{figure}[H]
    \centering
    \includegraphics[width=\textwidth]{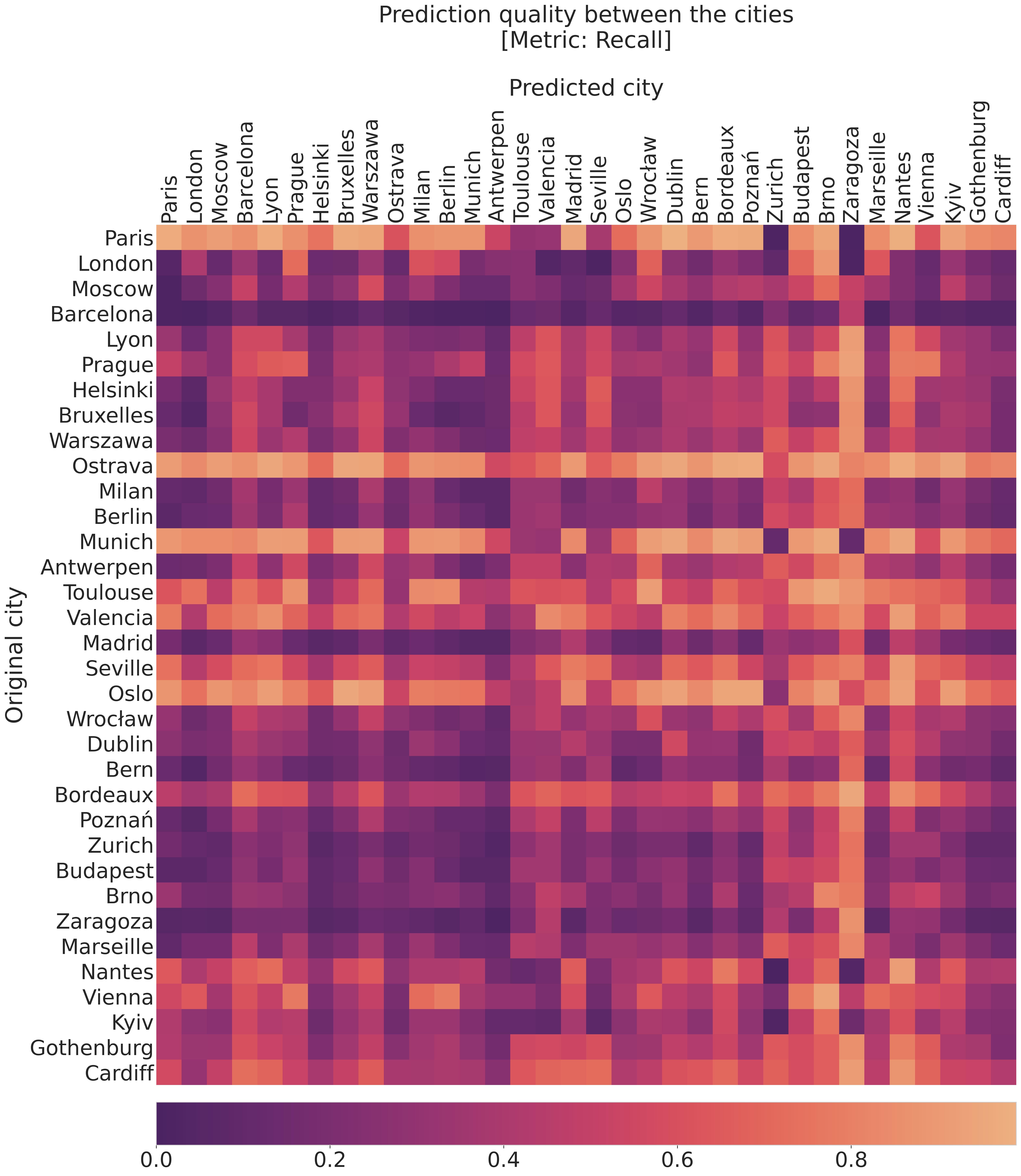}
    \caption[Recall metric in cross-city prediction]%
    {Recall metric in cross-city prediction. Cities were sorted in descending order of the number of stations in the city. \par \small Personal work.}
    \label{chart:transfer_rec}
\end{figure}

\begin{figure}[H]
    \centering
    \includegraphics[width=\textwidth]{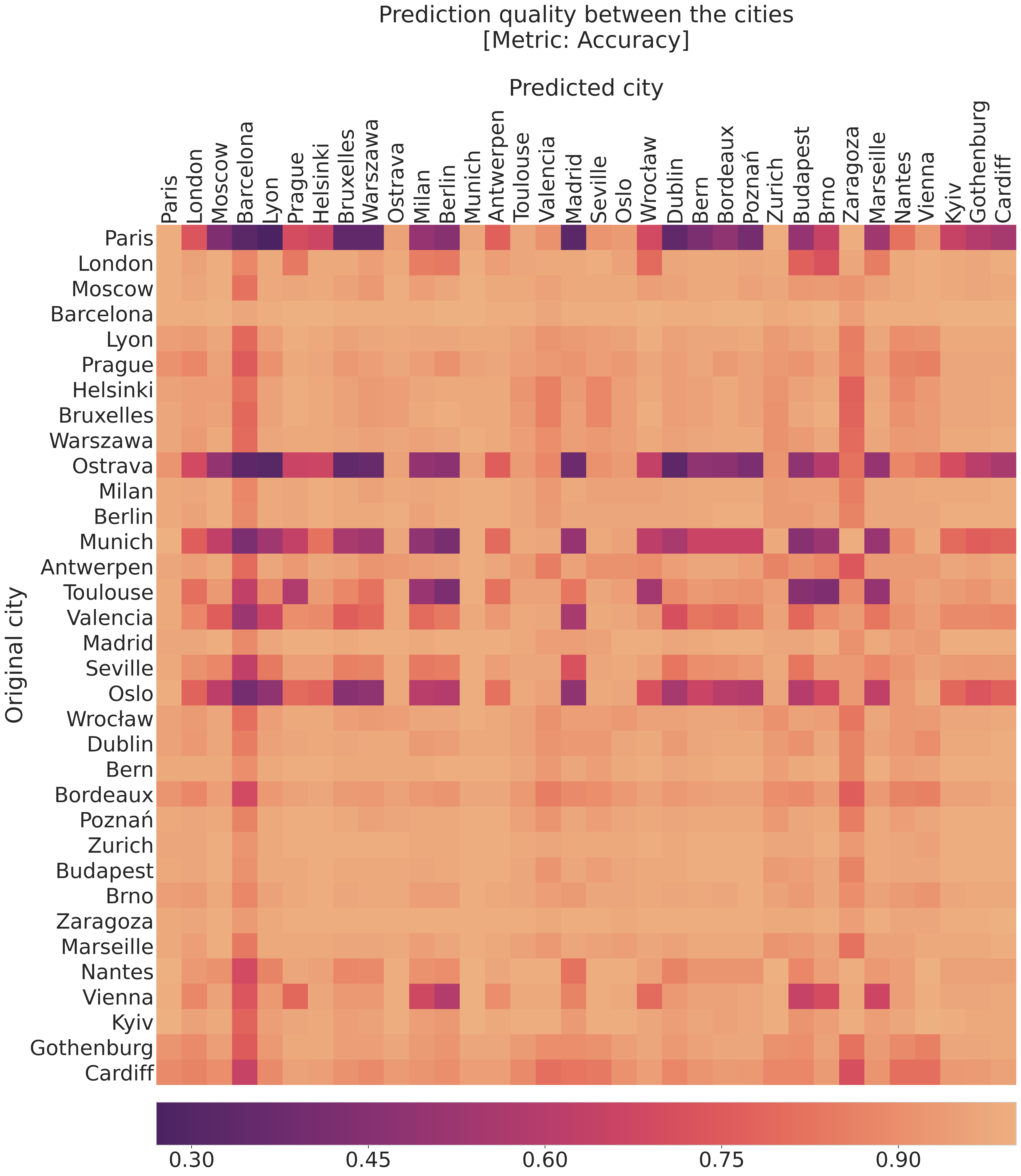}
    \caption[Accuracy metric in cross-city prediction]%
    {Accuracy metric in cross-city prediction. Cities were sorted in descending order of the number of stations in the city. \par \small Personal work.}
    \label{chart:transfer_acc}
\end{figure}

Within the results, it was decided to present only the recall and accuracy metric, as it is the only one that brings some information about the behaviour of the model. Since the validation set included all hexes in the entire city, the ratio between hexes containing stations to those that did not contain stations was often 100 to 50-100 thousand. Therefore, the value of the accuracy metric was mostly above 98\% with single cases where the value was much lower. The precision in most cases was very close to zero because the model predicted a lot of hexes that do not have stations as hexes that should contain a station (often proximity to stations). Along with the low precision, the F1 score was equally low. In the case of the recall metric, more correlations can be seen. The value of this measure varies between 0 and 0.95, and the heat map shows that some cities perform very well as base cities for station occurrence prediction (Munich, Oslo, Ostrava, Paris) and some cities, it is very difficult for the model to pick out all existing stations (Antwerp, Helsinki, Ostrava). Results are presented in the Figures \ref{chart:transfer_rec} and \ref{chart:transfer_acc}.

The high value of the recall measure for some cities indicates that for those 4 best performing cities the model can predict well the occurrence of all stations within a city, and a low measure of accuracy (and precision) indicates that the model predicts many more positions as proposed than there are actually stations present, which is quite an expected result since cities often have multiple regions (especially at such high resolution) with similar structures that the model indicates slightly larger regions than a single hexagon. 

\begin{figure}[H]
    \centering
    \includegraphics[width=\textwidth]{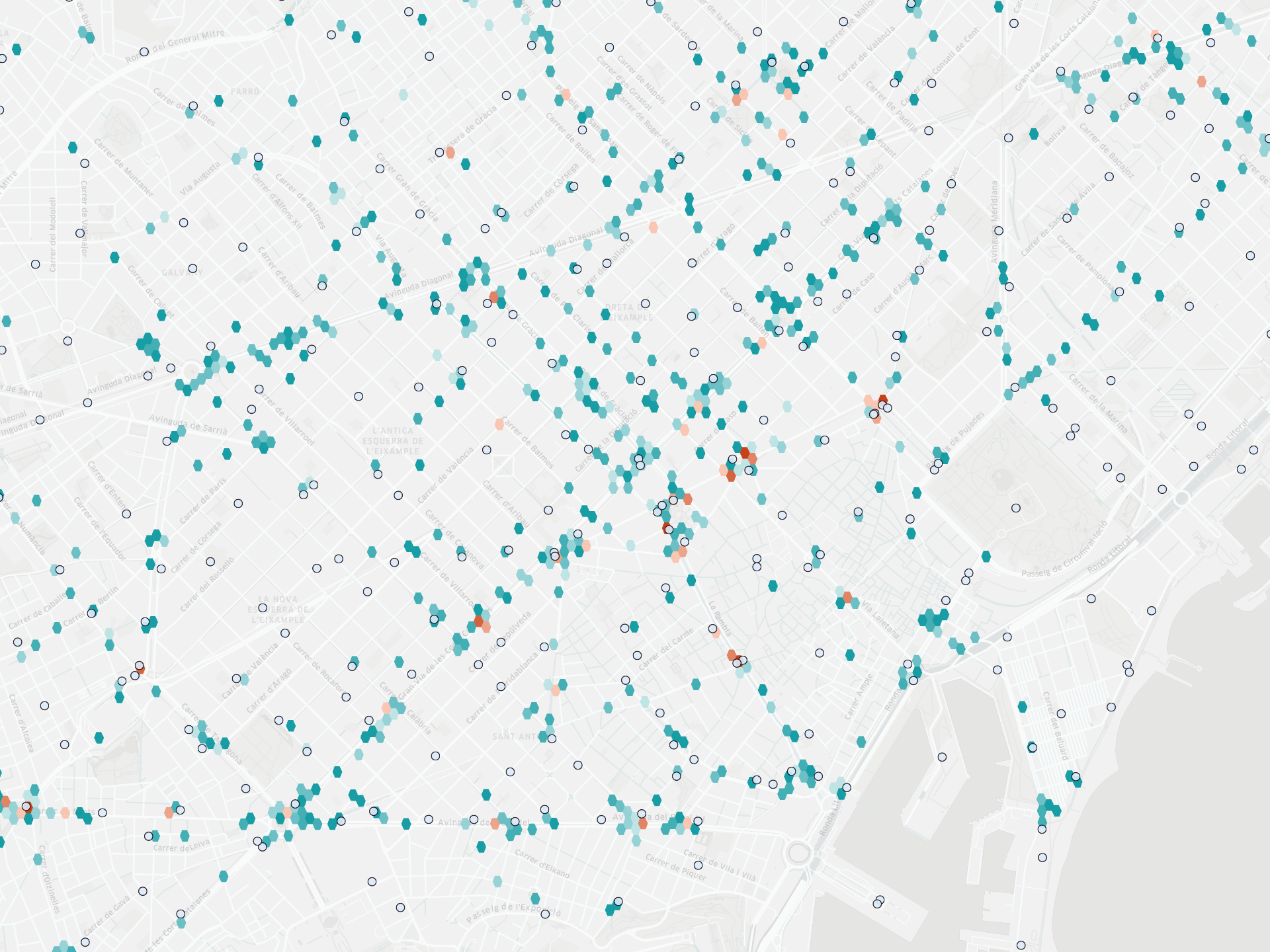}
    \caption[Example of Barcelona city predictions for method trained on a dataset from Barcelona]%
    {Example of Barcelona city predictions for method trained on a dataset from Barcelona. Results have been averaged from 100 iterations and filtered below probability 0.5. The dark teal colour represents the probability of 50\% and the dark red colour represents the probability of around 75\%. White dots with black borders represent existing stations. \par \small Personal work. Rendered using kepler.gl library.}
    \label{fig:barcelona_ex}
\end{figure}

Finally, it was decided to investigate how the predictions were distributed for the city of Barcelona, as the results indicated a very low recall measure for this city. The figure \ref{fig:barcelona_ex} shows a close-up map of the city centre with the predictions marked. It can be clearly seen that the method is not able to determine all station positions when the confidence threshold is equal to 0.5, which is the default value in the classifier. However, it cannot be said that the model generates completely random predictions and can reflect partial station layout infrastructure or predicts the occurrence of a station right next to where it occurs and because of the binary approach, the classifier is penalised for this result and the quality metric drops. This may be due to the density of the city, which in combination with the good quality tagging in OpenStreetMap causes many regions to generate embedding vectors that are close to each other in space. Additionally, it can be seen that the model performs better in the inner city than on the eastern coast of the city, for example.

\newpage

\section{Example analysis for cities without bicycle sharing systems}

To demonstrate the usefulness of the method, the results are presented below in the form of heat maps distributed over four European cities of different sizes: Naples (Italy) with around 1 million inhabitants, Florence (Italy) with over 350,000 inhabitants, Salzburg (Austria) with over 150,000 inhabitants, and Świdnica (Poland) with over 50,000 inhabitants.

The 4 cities that had the best recall metric in research question 7 (Section \ref{ch:rq7}) were used to teach the model: Munich, Oslo, Ostrava, Paris. The results are presented as 4 separate examples to show the differences in the performance of the model depending on the city chosen, which may allow other types of predictions to be made, which may depend on the layout of the city, the quality of its OpenStreetMap tagging and the layout of the bicycle sharing stations themselves. Under the examples for each city, a close-up map of the city centre with predictions averaged over the 4 cities is placed. The examples are left without any judgement.

Results were averaged from 100 iterations. Each map represents only a selected subset of hexes above a selected threshold that varies from example to example because sometimes the model predicted (i.e., has a probability higher than 0.5) less than 50 hexes in the whole city. The threshold value varies between 0.2 and 0.5. Red colour represents low and yellow colour represents a high value of the predicted probability of occurrence of a station in a given microregion.

\newpage

\subsection{Naples, Italy}

\begin{figure}[H]
    \centering
    \begin{minipage}{.5\textwidth}
        \centering
        \subfloat[Munich, Germany]{\label{fig:naples:38}\includegraphics[width=0.95\linewidth, keepaspectratio]{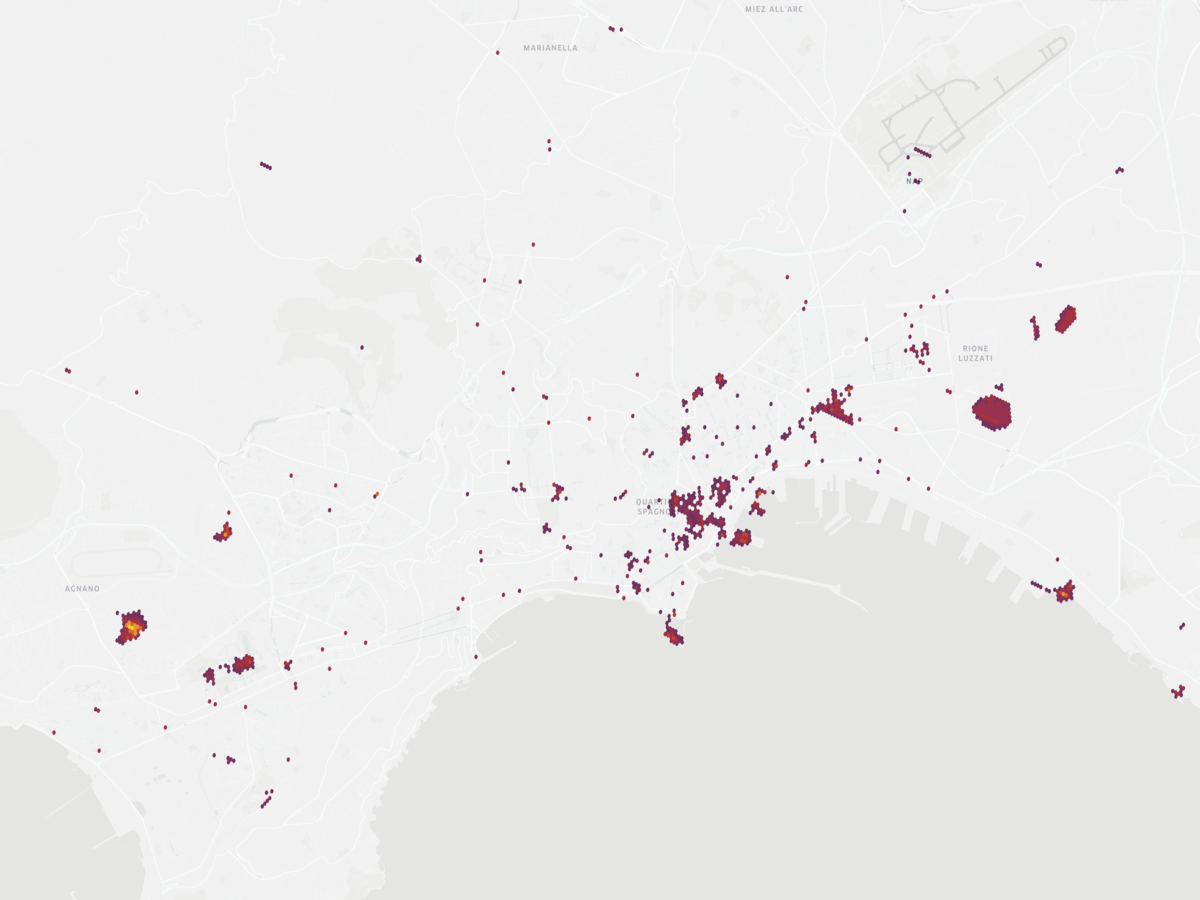}}
    \end{minipage}%
    \begin{minipage}{.5\textwidth}
        \centering
        \subfloat[Oslo, Norway]{\label{fig:naples:41}\includegraphics[width=0.95\linewidth, keepaspectratio]{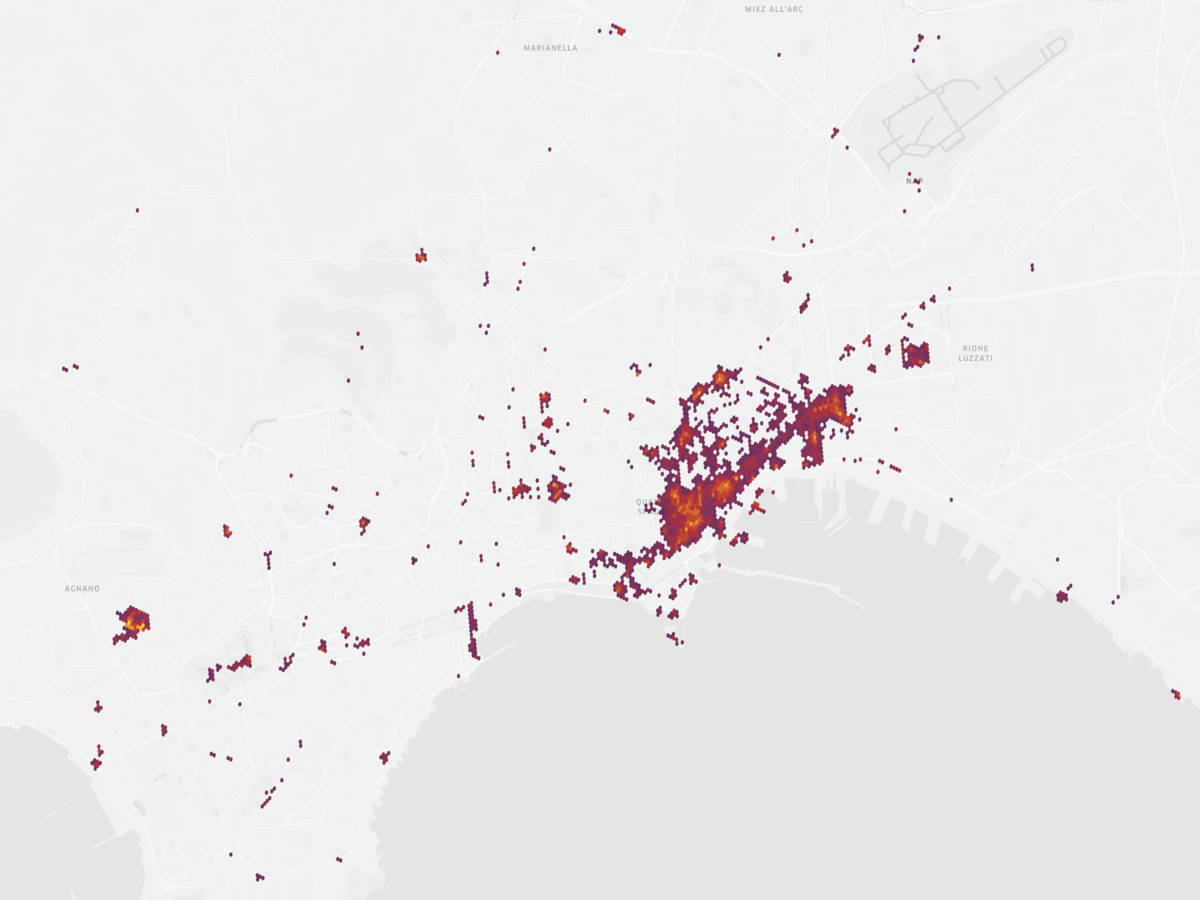}}
    \end{minipage}
    \par
    \begin{minipage}{.5\textwidth}
        \centering
        \subfloat[Ostrava, Czech Republic]{\label{fig:naples:42}\includegraphics[width=0.95\linewidth, keepaspectratio]{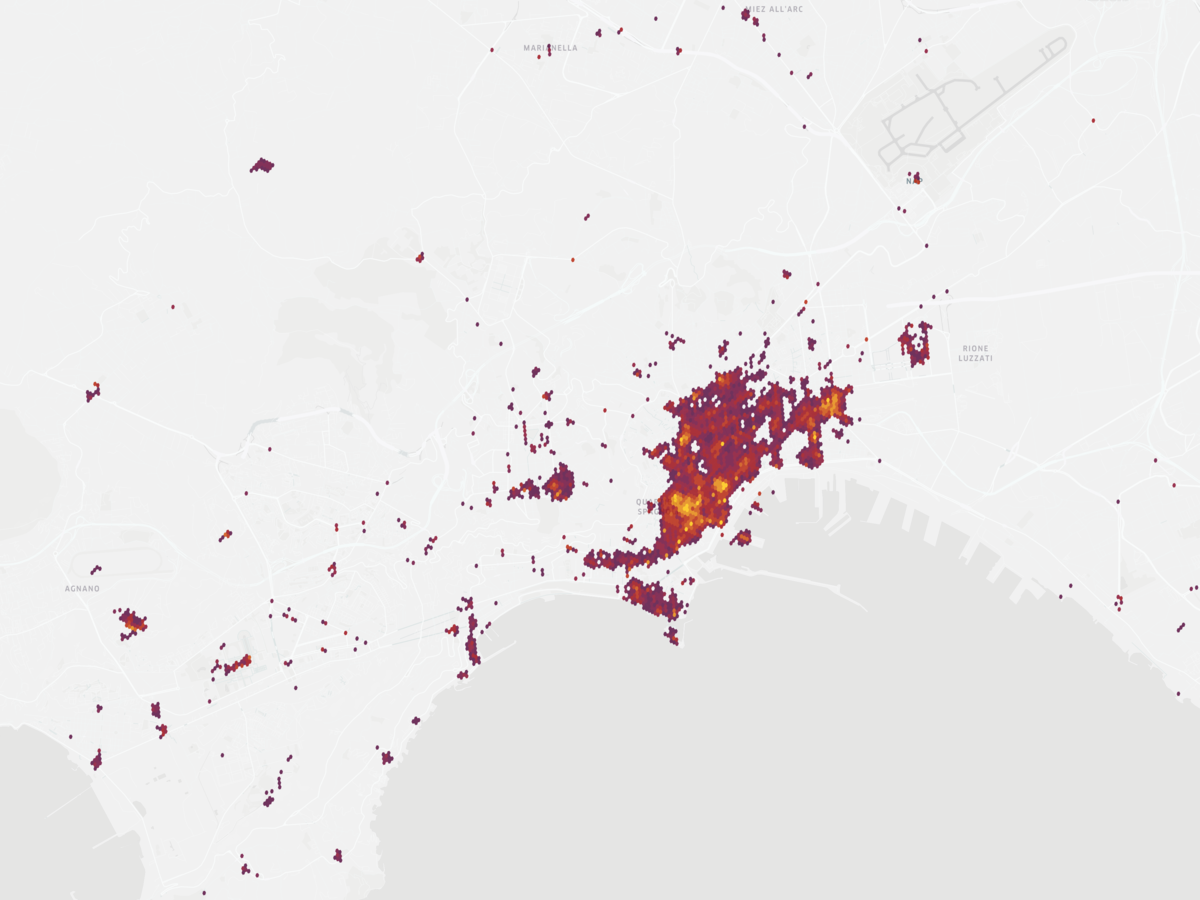}}
    \end{minipage}%
    \begin{minipage}{.5\textwidth}
        \centering
        \subfloat[Paris, France]{\label{fig:naples:45}\includegraphics[width=0.95\linewidth, keepaspectratio]{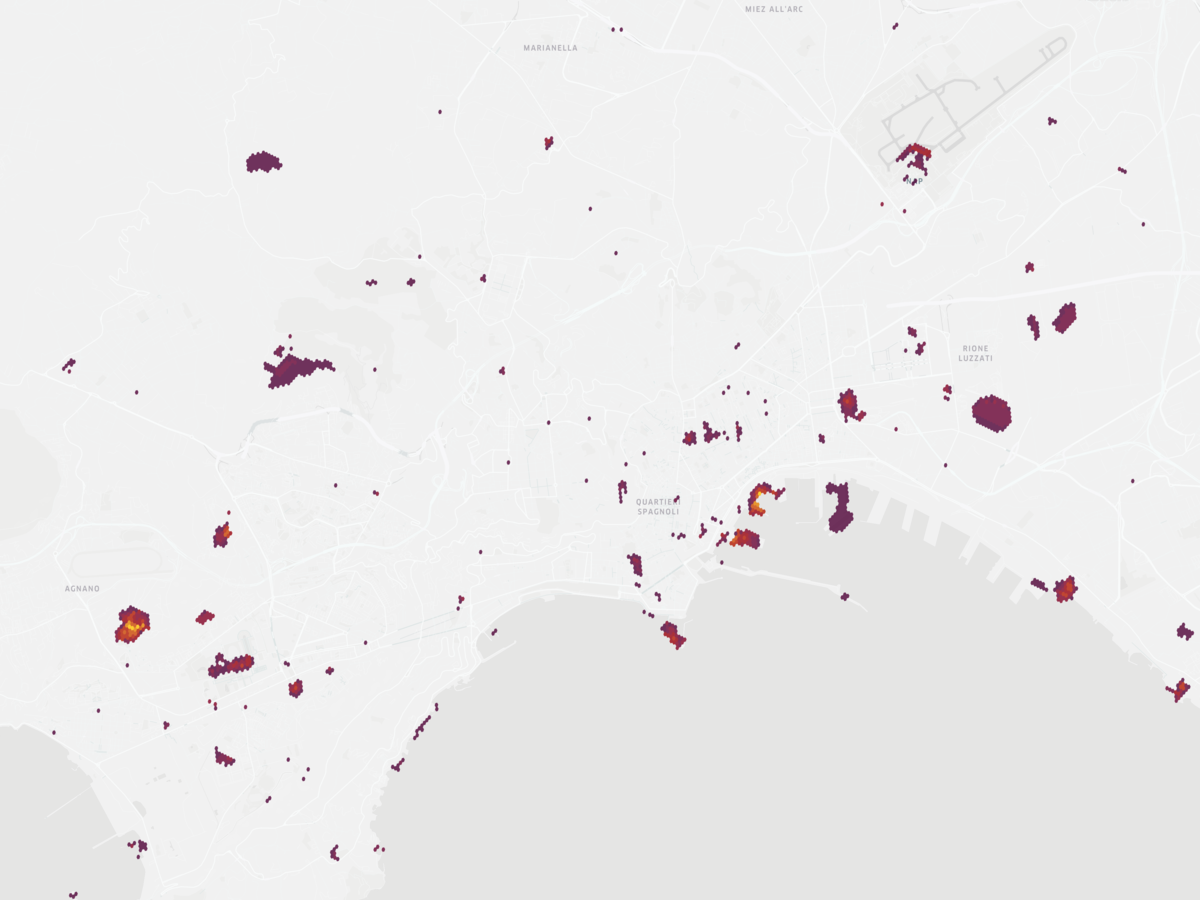}}
    \end{minipage}
    
    \begin{minipage}{\textwidth}
        \centering
        \subfloat[All predictions averaged]{\label{fig:naples:avg}\includegraphics[width=0.975\linewidth, keepaspectratio]{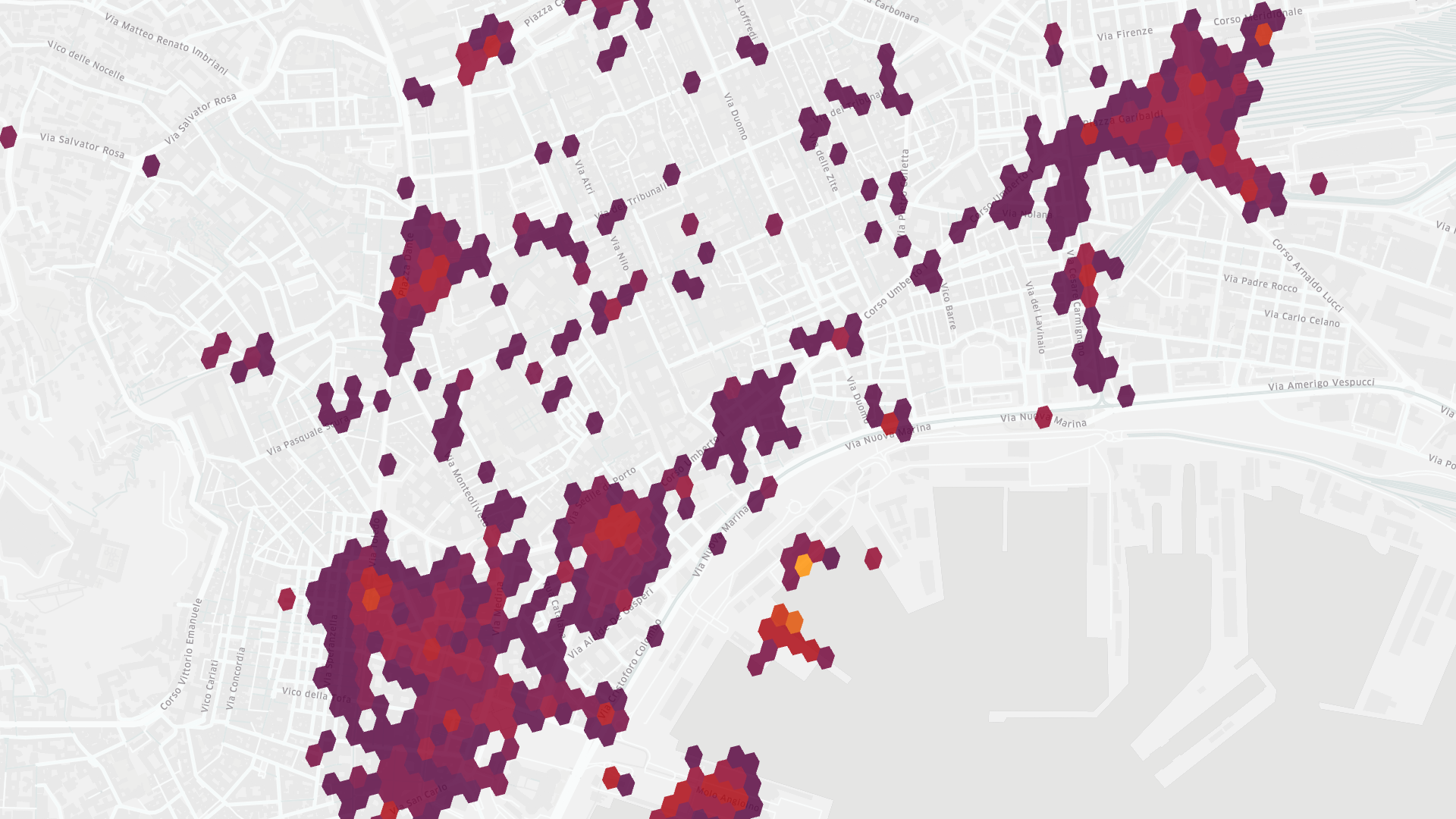}}
    \end{minipage}
    
    \caption[Example predictions for the city of Naples, Italy]%
    {Example predictions for the city of Naples, Italy. \par \small Personal work. Rendered using kepler.gl library.}
    \label{fig:naples}
\end{figure}

\subsection{Florence, Italy}

\begin{figure}[H]
    \centering
    \begin{minipage}{.5\textwidth}
        \centering
        \subfloat[Munich, Germany]{\label{fig:florence:38}\includegraphics[width=0.95\linewidth, keepaspectratio]{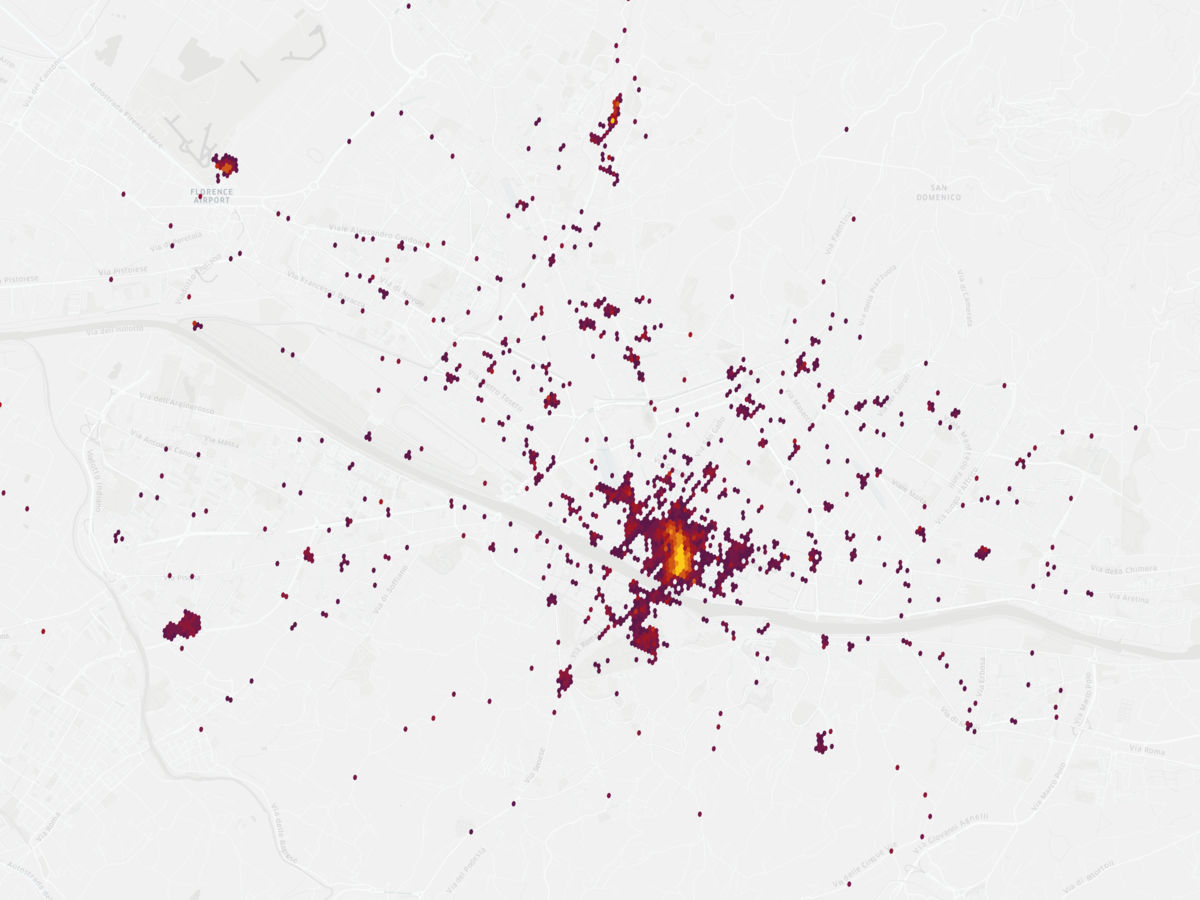}}
    \end{minipage}%
    \begin{minipage}{.5\textwidth}
        \centering
        \subfloat[Oslo, Norway]{\label{fig:florence:41}\includegraphics[width=0.95\linewidth, keepaspectratio]{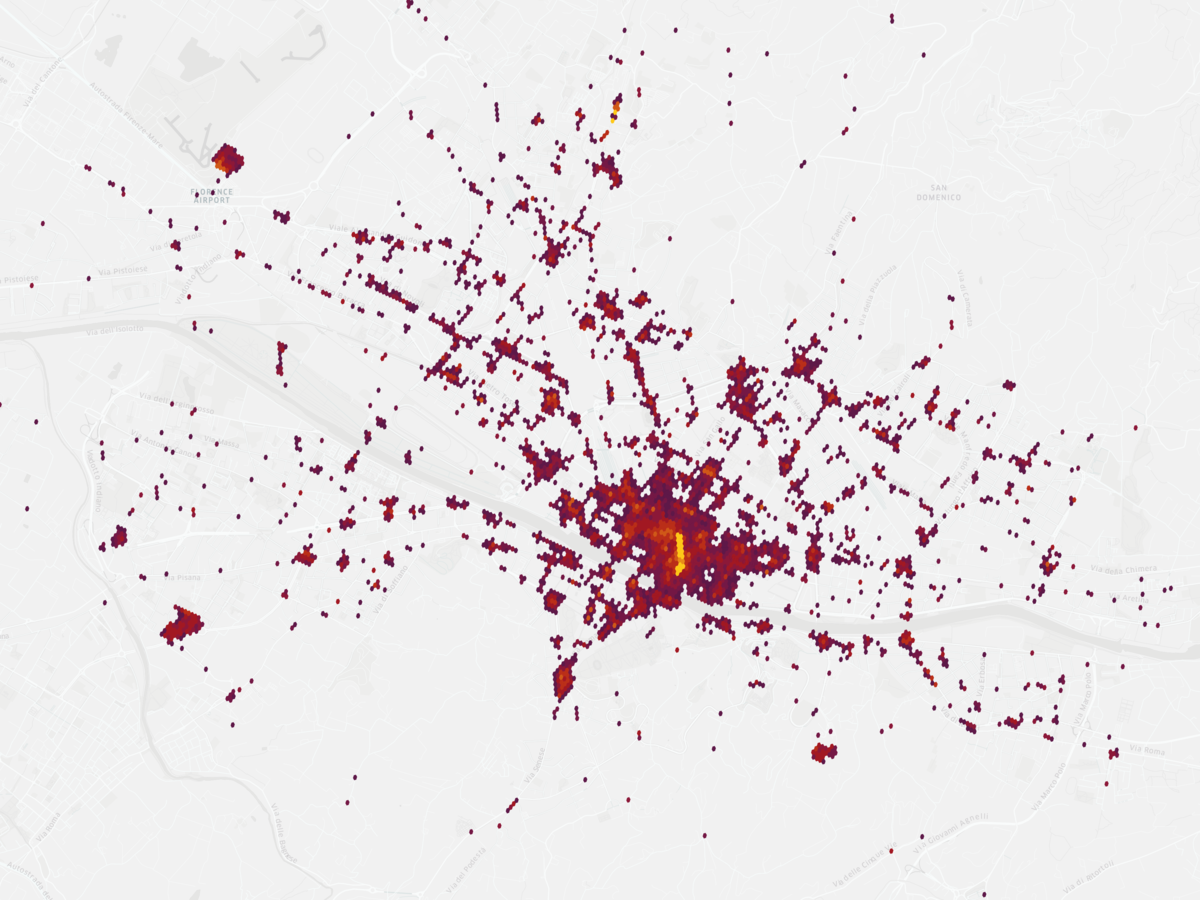}}
    \end{minipage}
    \par
    \begin{minipage}{.5\textwidth}
        \centering
        \subfloat[Ostrava, Czech Republic]{\label{fig:florence:42}\includegraphics[width=0.95\linewidth, keepaspectratio]{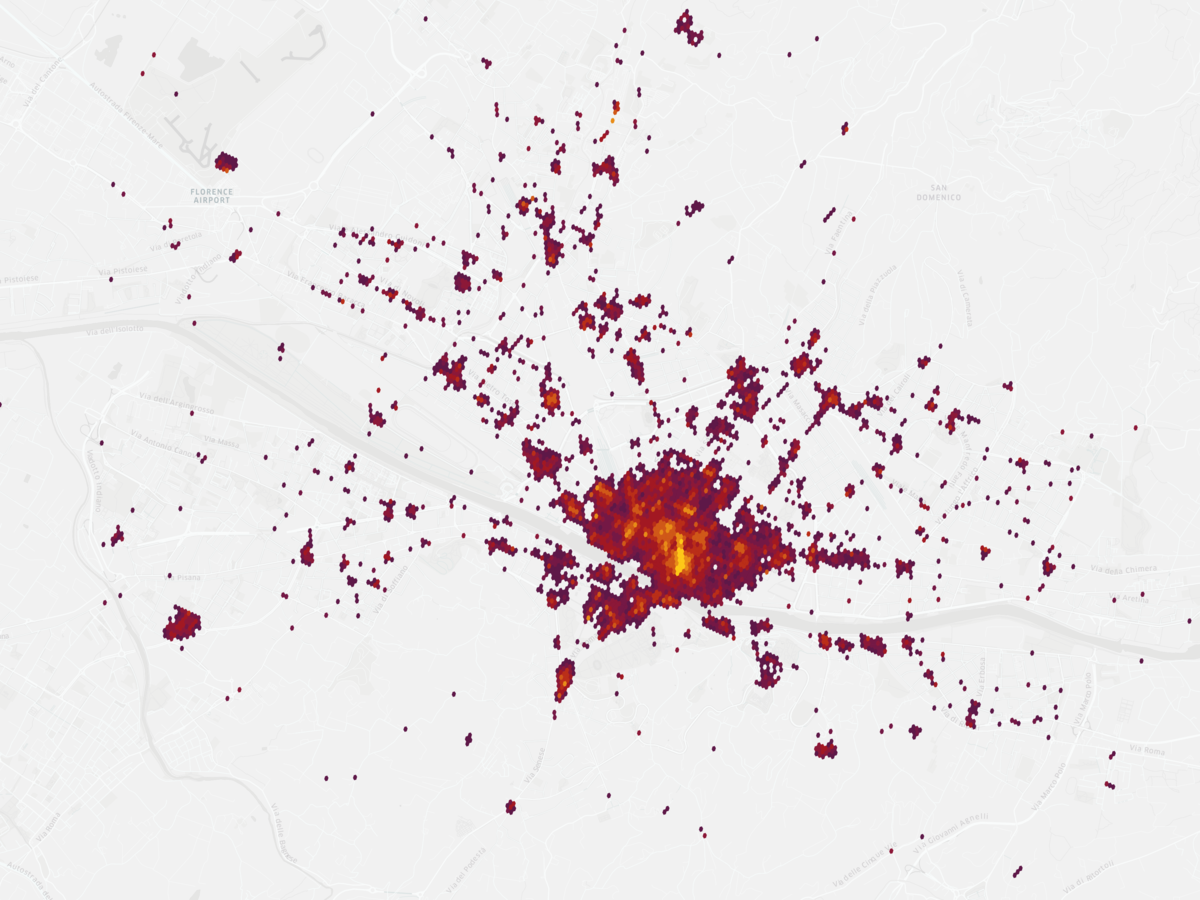}}
    \end{minipage}%
    \begin{minipage}{.5\textwidth}
        \centering
        \subfloat[Paris, France]{\label{fig:florence:45}\includegraphics[width=0.95\linewidth, keepaspectratio]{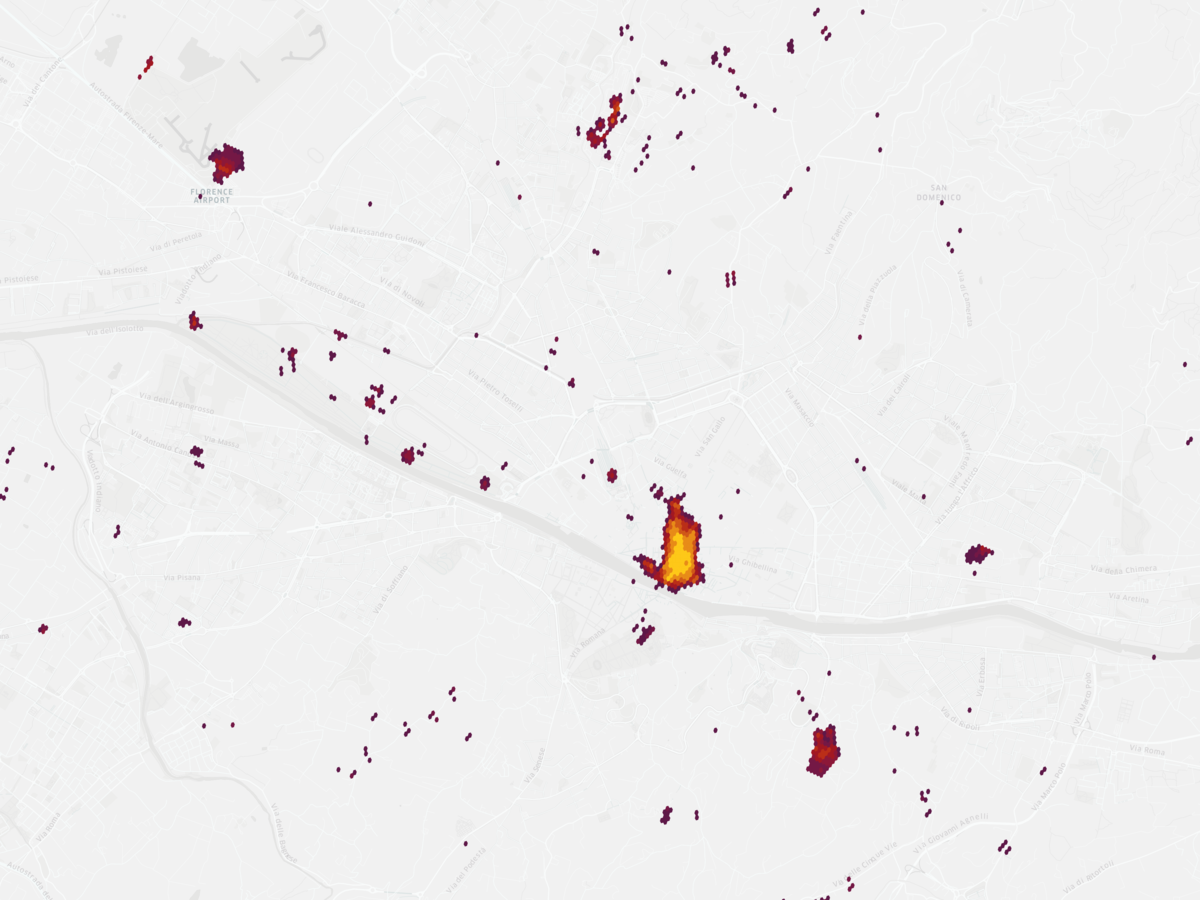}}
    \end{minipage}
    
    \begin{minipage}{\textwidth}
        \centering
        \subfloat[All predictions averaged]{\label{fig:florence:avg}\includegraphics[width=0.975\linewidth, keepaspectratio]{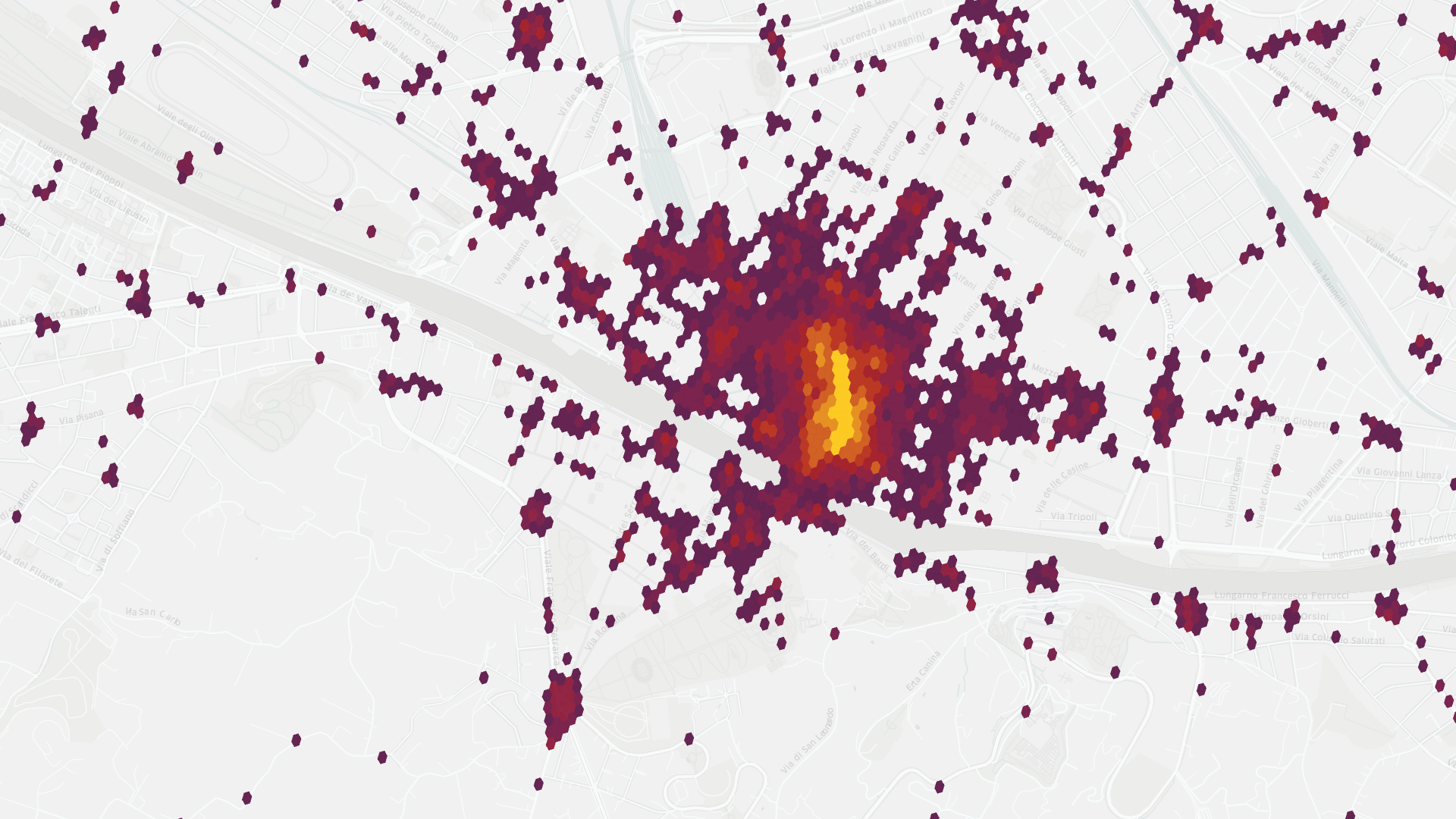}}
    \end{minipage}
    
    \caption[Example predictions for the city of Florence, Italy]%
    {Example predictions for the city of Florence, Italy. \par \small Personal work. Rendered using kepler.gl library.}
    \label{fig:florence}
\end{figure}

\subsection{Salzburg, Austria}

\begin{figure}[H]
    \centering
    \begin{minipage}{.5\textwidth}
        \centering
        \subfloat[Munich, Germany]{\label{fig:salzburg:38}\includegraphics[width=0.95\linewidth, keepaspectratio]{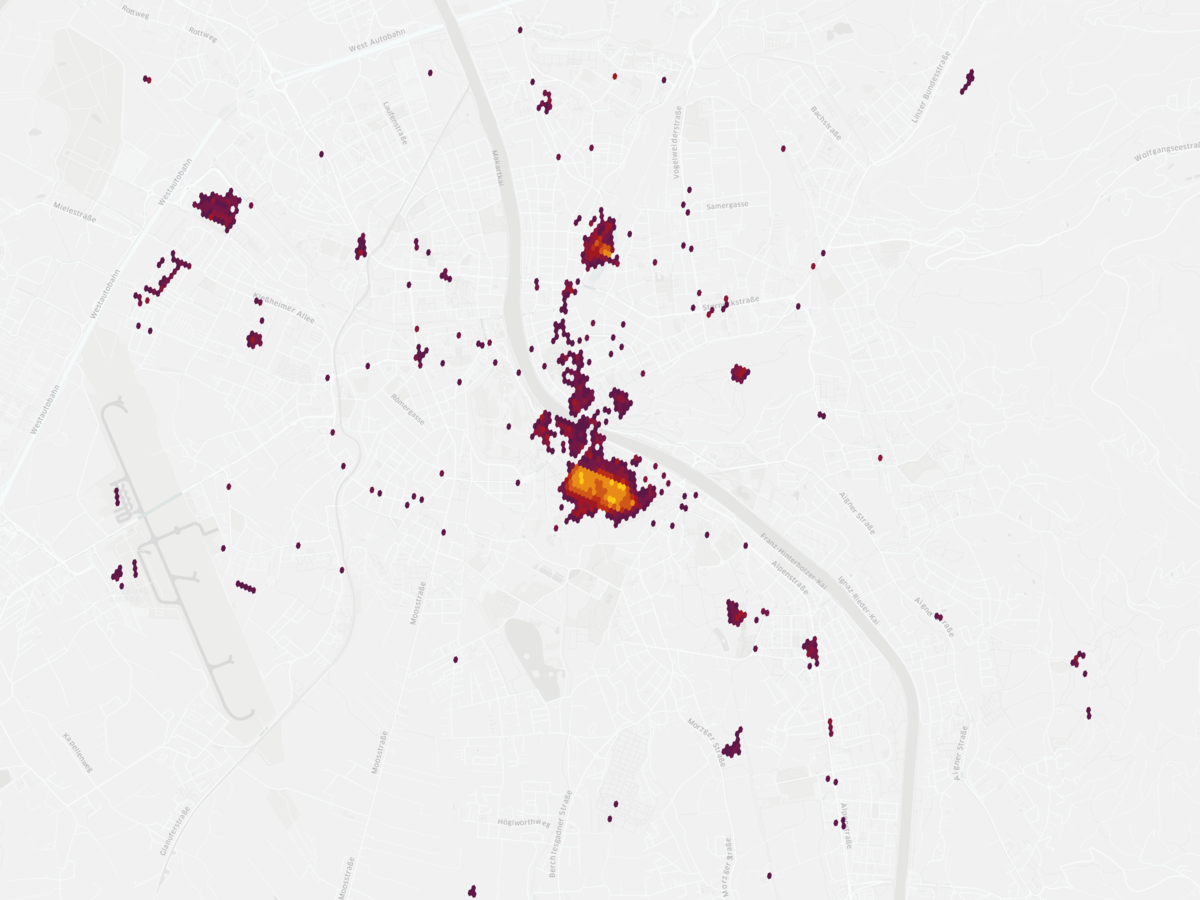}}
    \end{minipage}%
    \begin{minipage}{.5\textwidth}
        \centering
        \subfloat[Oslo, Norway]{\label{fig:salzburg:41}\includegraphics[width=0.95\linewidth, keepaspectratio]{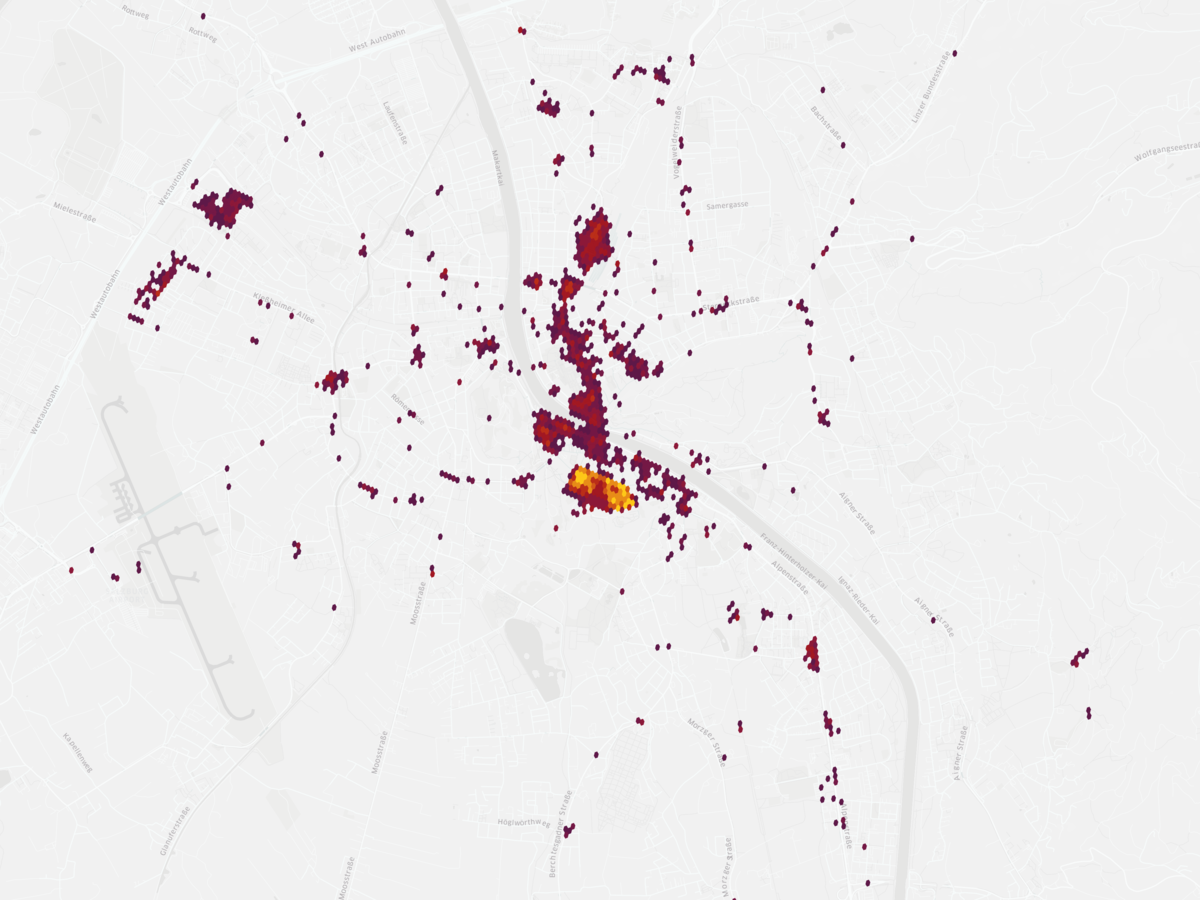}}
    \end{minipage}
    \par
    \begin{minipage}{.5\textwidth}
        \centering
        \subfloat[Ostrava, Czech Republic]{\label{fig:salzburg:42}\includegraphics[width=0.95\linewidth, keepaspectratio]{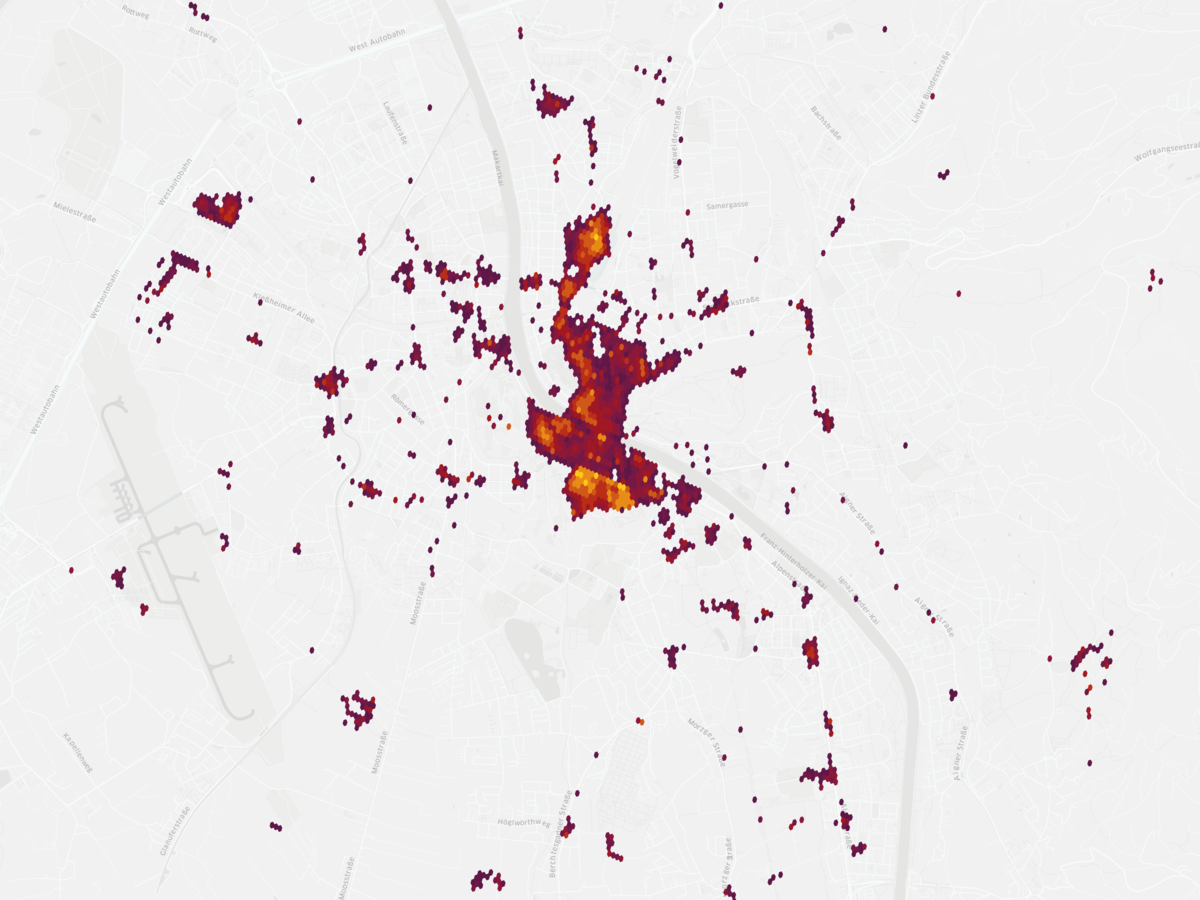}}
    \end{minipage}%
    \begin{minipage}{.5\textwidth}
        \centering
        \subfloat[Paris, France]{\label{fig:salzburg:45}\includegraphics[width=0.95\linewidth, keepaspectratio]{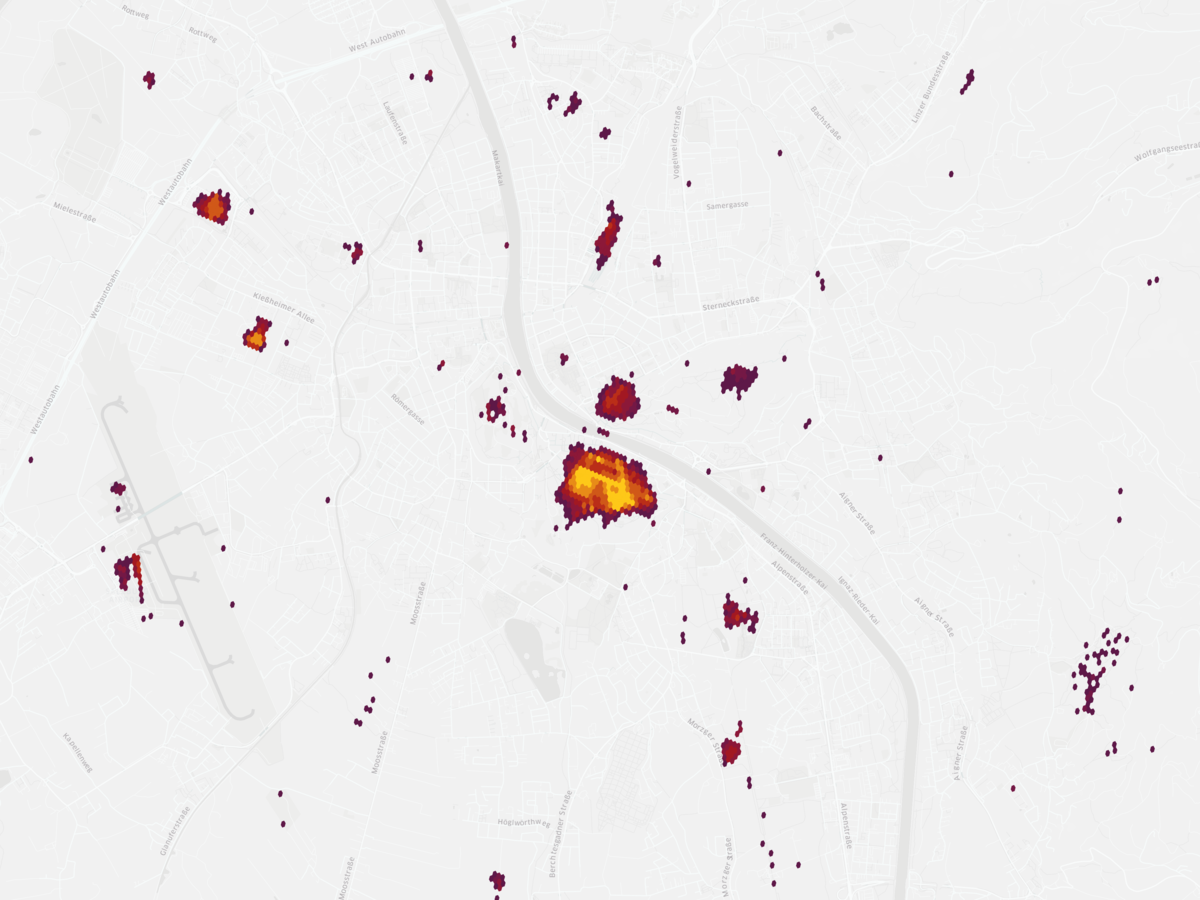}}
    \end{minipage}
    
    \begin{minipage}{\textwidth}
        \centering
        \subfloat[All predictions averaged]{\label{fig:salzburg:avg}\includegraphics[width=0.975\linewidth, keepaspectratio]{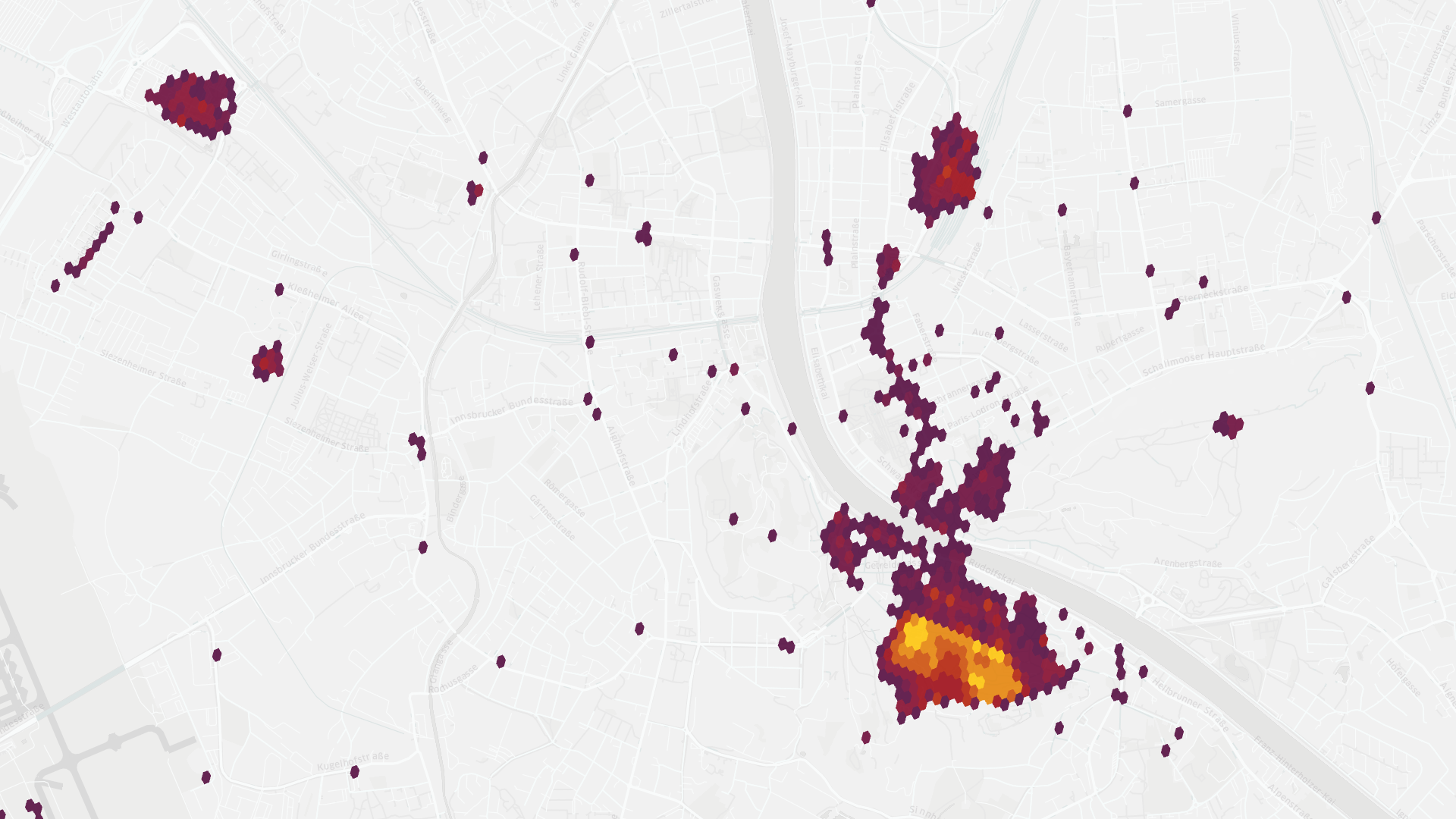}}
    \end{minipage}
    
    \caption[Example predictions for the city of Salzburg, Austria]%
    {Example predictions for the city of Salzburg, Austria. \par \small Personal work. Rendered using kepler.gl library.}
    \label{fig:salzburg}
\end{figure}

\subsection{Świdnica, Poland}

\begin{figure}[H]
    \centering
    \begin{minipage}{.5\textwidth}
        \centering
        \subfloat[Munich, Germany]{\label{fig:swidnica:38}\includegraphics[width=0.95\linewidth, keepaspectratio]{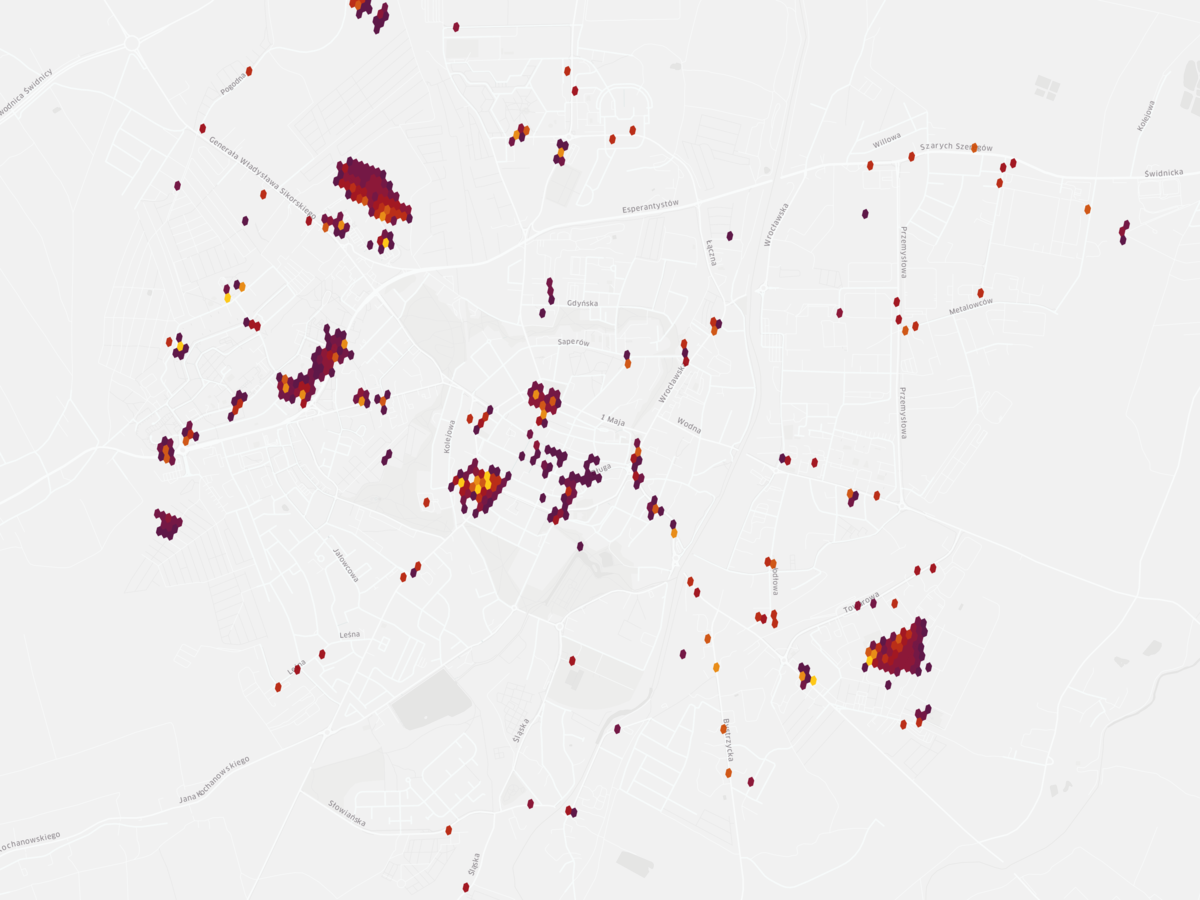}}
    \end{minipage}%
    \begin{minipage}{.5\textwidth}
        \centering
        \subfloat[Oslo, Norway]{\label{fig:swidnica:41}\includegraphics[width=0.95\linewidth, keepaspectratio]{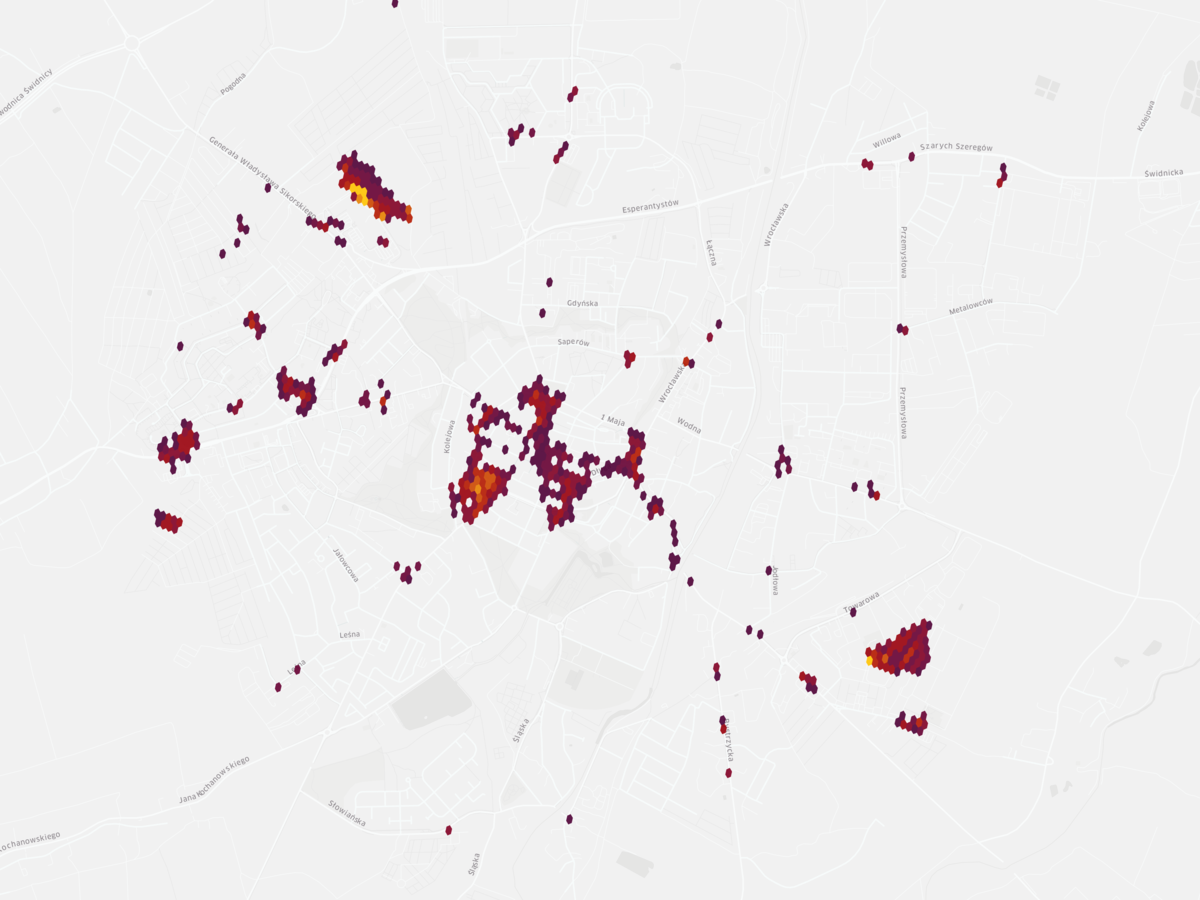}}
    \end{minipage}
    \par
    \begin{minipage}{.5\textwidth}
        \centering
        \subfloat[Ostrava, Czech Republic]{\label{fig:swidnica:42}\includegraphics[width=0.95\linewidth, keepaspectratio]{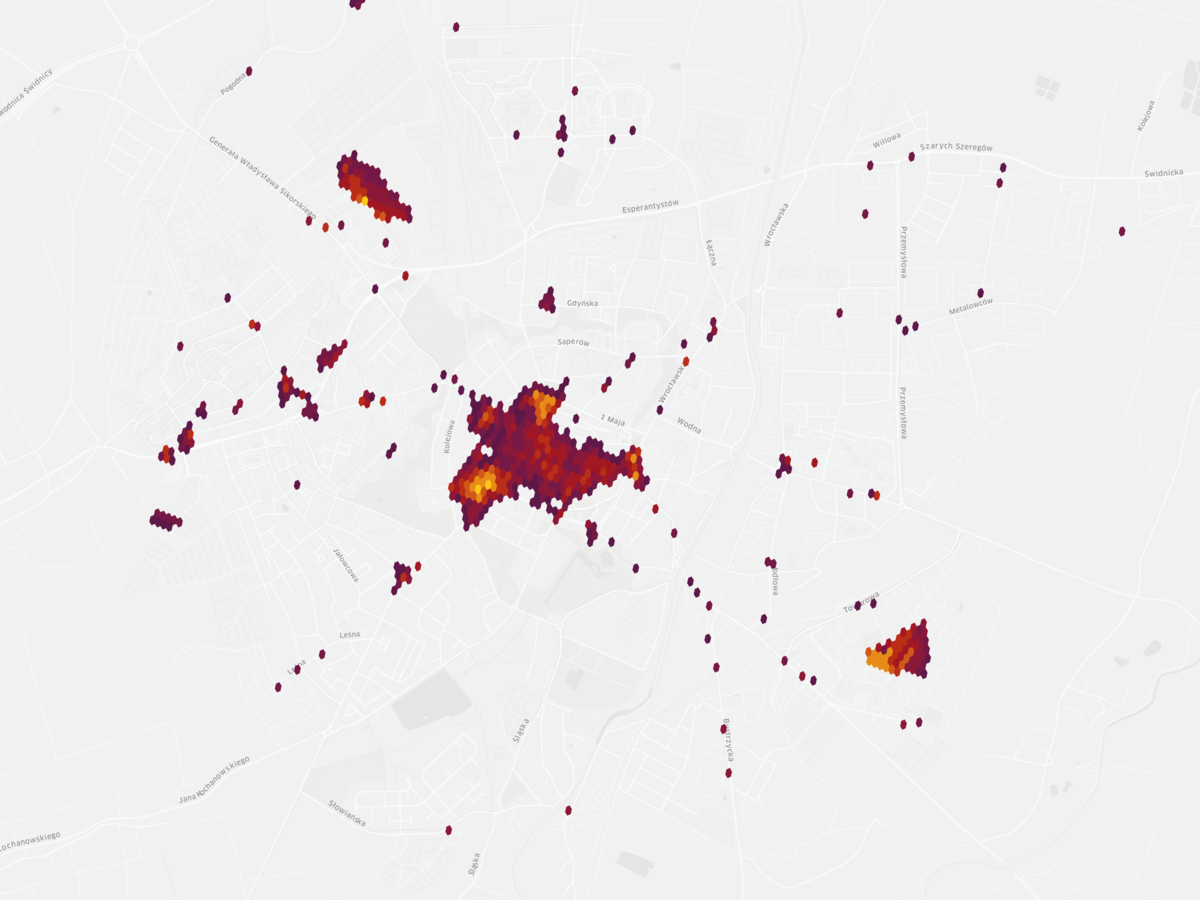}}
    \end{minipage}%
    \begin{minipage}{.5\textwidth}
        \centering
        \subfloat[Paris, France]{\label{fig:swidnica:45}\includegraphics[width=0.95\linewidth, keepaspectratio]{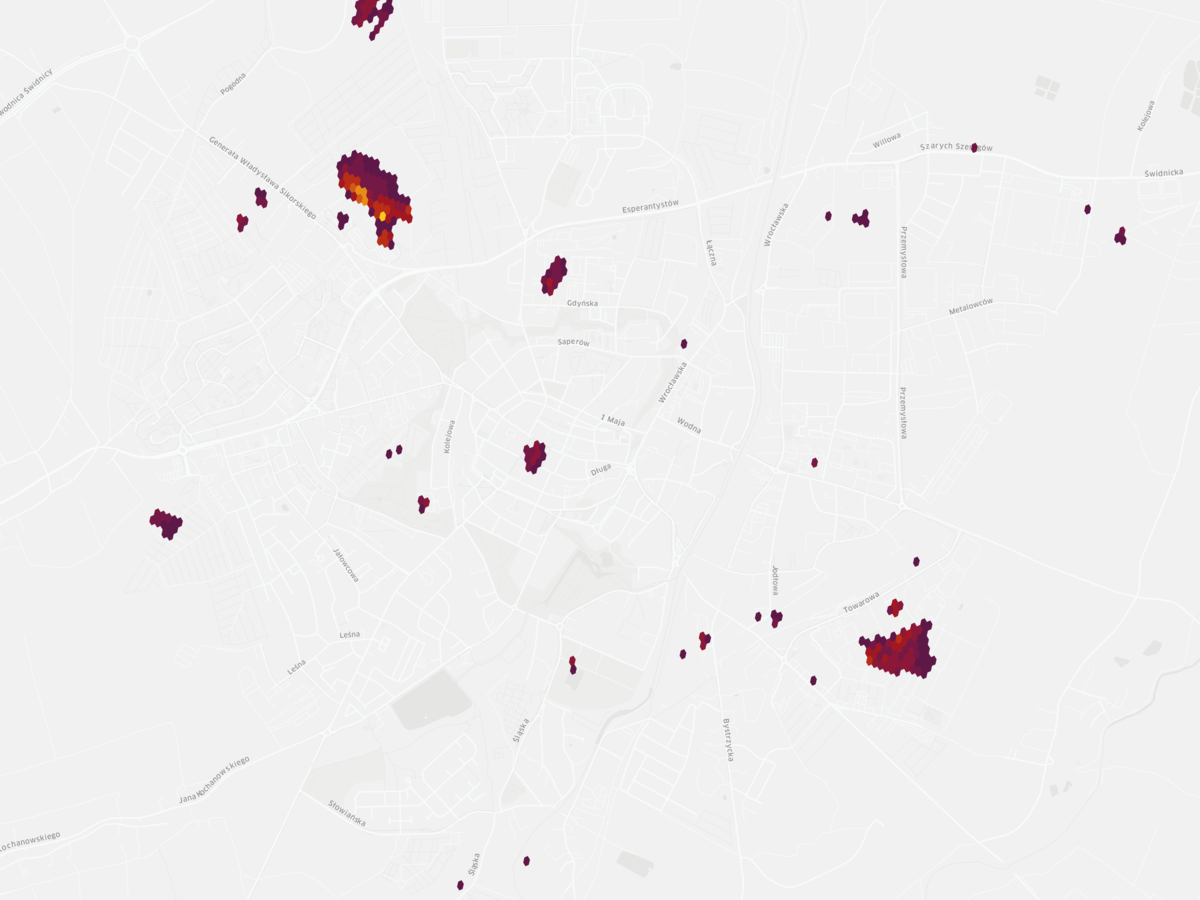}}
    \end{minipage}
    
    \begin{minipage}{\textwidth}
        \centering
        \subfloat[All predictions averaged]{\label{fig:swidnica:avg}\includegraphics[width=0.975\linewidth, keepaspectratio]{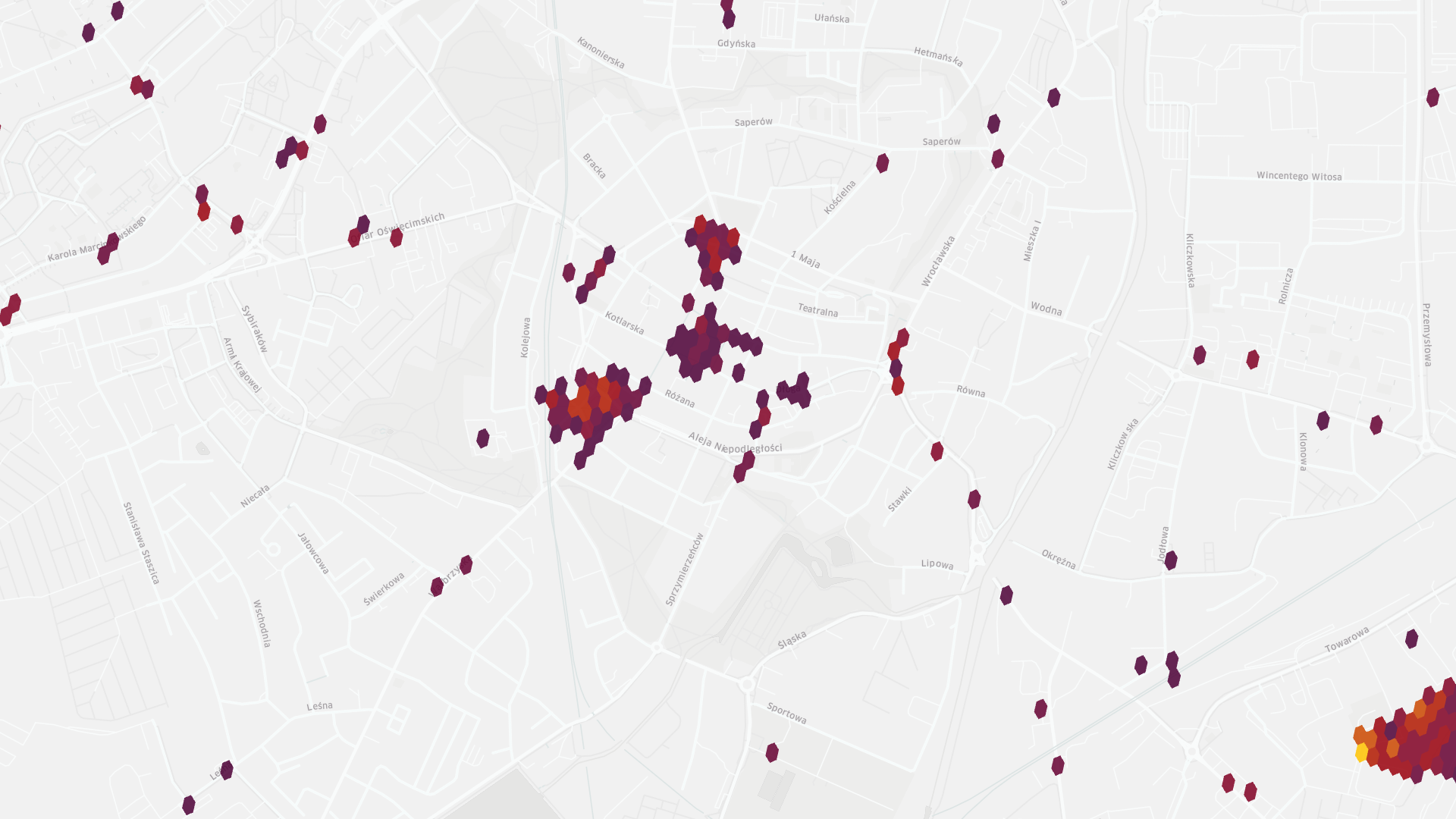}}
    \end{minipage}
    
    \caption[Example predictions for the city of Świdnica, Poland]%
    {Example predictions for the city of Świdnica, Poland. \par \small Personal work. Rendered using kepler.gl library.}
    \label{fig:swidnica}
\end{figure}

\chapter{Conclusions}
\label{ch:conclusions}

To develop a method that meets the objectives of the problem statement, 7 research questions were formulated. The information obtained in the exploratory analysis allowed some assumptions to be made about the performance of the method and confirmed that the data obtained could be used in machine learning for inference. The results obtained confirm that the method is capable of making valuable predictions, but the quality of the method's performance varies depending on the city whose data was used for learning. In this section, there is a detailed discussion of the 7 research questions, the answer to the problem statement, and also directions for future research.

\section{Discussion of research questions}

The first 6 research questions aimed to discover the influence of different hyperparameters and the last one tested how the method performs in the task of predicting the occurrence of stations between cities.

\paragraph*{RQ1: How different baseline classifiers perform in the station presence prediction task?} \ \\*
The analysis of the different classifiers showed that each classifier performed quite well in the station occurrence prediction task when the number of regions containing stations and not containing stations was in balance in the training and test sets. All classifiers were able to score above 0.7 for the F1-Score measure. Finally, it was decided to choose the random forest classifier because it obtained better results than all other classifiers and showed a low variance in the results obtained.

\paragraph*{RQ2: How neighbourhood embedding methods affect performance?} \ \\*

Analysis of methods for combining neighbourhood vectors showed that a simple concatenation method performs quite well regardless of the neighbourhood size. Unfortunately, this method of concatenating vectors negatively affects the model learning performance as the dimensionality of the features increases significantly. The proposed averaging method, which is quite popular in natural language processing where the task vector is built by averaging vectors of single words, appeared to work well for low neighbourhood values, but with each successive ring of neighbours, the quality of prediction decreased. In addition to simple averaging, 2 weighted averaging methods were investigated: diminishing and squared diminishing. The results show that the last method behaves the most stable and obtains results similar to concatenation, so it was chosen as the method for neighbourhood embedding.

\paragraph*{RQ3: How region embedding method affects the prediction performance?} \ \\*

A comparison of the different proposed methods for embedding single regions yielded unexpected results. It turned out that the basic method based only on the basic 20 categories without paying attention to the shapes of the studied objects works best. The method analyzing the shapes and taking into account the lengths of the roads in the regions gave similar results to the baseline method, but not higher values, so it was decided to stay with the first method. The more elaborate methods focusing on a higher number of tags and dimensionality reduction using an autoencoder were found to perform worse, however, this could be due to both too much granularity and sparseness of objects and differences in quality of tagging between cities, poor autoencoder infrastructure or poor architecture of the final NN classifier.

\paragraph*{RQ4: How vector preprocessing affects performance?} \ \\*

Often, machine learning uses data scaling methods to improve the performance of the classifier. As the underlying method returned values from a set of natural numbers, the maximum value was not limited in advance. In the case of large disparities in distributions between features, a situation may arise where one feature becomes more important than others in the inference process. The application of normalisation slightly improved the performance of the model and standardisation worsened its quality, which may result from the fact that most of the values in the vectors were equal to 0.

\paragraph*{RQ5: How the imbalance ratio affects performance?} \ \\*

Examination of the class imbalance ratio showed that as it increased, the quality of both the F1 score and precision and recall measures decreased, which was a worrying sign. This could be because previously the model was taught on a subset but also tested on a subset rather than the whole city, and as the elements in the test set increased, the quality of the method performance decreased. Therefore, it was decided to visualise the results obtained by the method on the example of one of the cities. Based on the author's own assessment supported by the obtained heat maps and to maintain the average value of the F1 score measured higher than 0.6, it was decided to choose the ratio equal to 2.5. Thanks to it, the model was able to predict most of the existing stations, but on the other hand, it predicted a very large number of regions that do not have stations. However, showing a probability value of an occurrence rather than a binary score allows some regions to be filtered out by the user.

\paragraph*{RQ6: How the resolution of regions and the size of region neighbourhood affect the prediction performance?} \ \\*

The final important element was the choice of the appropriate resolution and size of the neighbourhood considered for prediction. The analysis showed that all 3 resolutions with appropriate neighbourhood size were able to obtain similar results. As the aim of the prediction is to indicate the appropriate region as accurately as possible, it was decided to choose the highest resolution. It can be noticed that the differences between successive values of neighbourhoods are almost imperceptible, which may result from the chosen method of neighbourhood embedding. The difference between the vectors embedding the region with neighbourhood sizes 24 and 25 will be at most 1/125 or 0.008, which may be a small difference overall.

\paragraph*{RQ7: How does the model perform in predicting stations between cities?} \ \\*

The previous research steps aimed to find the best set of hyperparameters allowing the method to reflect the station structure for the same city for which the classifier was learned. In this analysis, predictions were made between all pairs of cities to obtain information on whether the method can make valuable predictions between different cities.  The results indicate that the method obtains different results depending on the city, both the one used for learning and the one predicted. Going back to the exploratory analysis, it is difficult to establish any relationship between the calculated average population per station and the quality of prediction between cities that are close to each other in this space. However, one can see a certain correlation related to the average number of POIs per region - the cities of Bordeaux, Cardiff, Gothenburg, Munich, Oslo, Ostrava, Poznań, Seville, and Valencia, which have relatively few OSMs within their regions, allowing the method to obtain recall values above 0.8. On the other hand, Barcelona and Madrid, which are quite dense compared to other cities, caused the method to perform poorly in the presented binary station prediction task.
There are also a few city pairs that perform exceptionally poorly compared to other pairs in a row or column, namely Paris-Zurich, Paris-Zaragoza, Munich-Zurich, Munich-Zaragoza, Nantes-Zurich, Nantes-Zaragoza, Kyiv-Zurich, Kyiv-Zaragoza. Looking at the differences between these cities, it can be seen that for each of the pairs there are significant differences between the different characteristics. In some cities there is a significant difference in educational buildings, others in historical or leisure facilities, which are quite rare in Zaragoza and Zurich. However, it is difficult to judge whether this is the reason for such differences or whether it is due to another reason.

Due to the very poor recall that occurred for the city of Barcelona, it was decided to test how the method performs in predicting regions for this city. It turned out that the method does indeed omit existing stations or predicts the presence of a station right next to the place where the station actually occurs, which means that the method can learn from the significant features of the regions and what causes the station to be placed where it is, but it is possible that due to the dense city centre, the method averages the positions and the method is not able to perfectly reflect the existing station layout. 

\paragraph*{Summary of research questions} \ \\*

Based on the experiments performed, the following list of hyperparameter values was selected:
\begin{itemize}
    \item \lbrack \textbf{RQ1}\rbrack \ Baseline classifier - Random forest
    \item \lbrack \textbf{RQ2}\rbrack \ Neighbourhood embedding method - Diminishing averaging squared
    \item \lbrack \textbf{RQ3}\rbrack \ Region embedding method - Category counting
    \item \lbrack \textbf{RQ4}\rbrack \ Features preprocessing - Normalisation (Min-Max)
    \item \lbrack \textbf{RQ5}\rbrack \ Imbalance ratio - 2.5
    \item \lbrack \textbf{RQ6}\rbrack \ Resolution - 11
    \item \lbrack \textbf{RQ6}\rbrack \ Neighbourhood size - 5
\end{itemize}

It is possible that for some cities it is worth using a different set of hyperparameters and the one chosen in the previous steps is not necessarily universal. It is possible that better results would be obtained in a linear regression task where each region would be assigned a value corresponding to the distance to the station (1 for the region with the station and decreasing for subsequent neighbourhoods). Another unexplored hyperparameter is the threshold level and its effect on prediction quality. Examining the prediction and recall curve would allow a more accurate diagnosis of the method's performance.

\section{Answer to the problem statement}

\paragraph*{Problem statement: Define a function that returns the probability of occurrence of a bicycle-sharing station in a specific city region based on OpenStreetMap data.} \ \\*

According to the assumption from the third step of the exploratory analysis, sensing the sufficiency of the obtained data to distinguish between different regions in cities, it can be stated that this is true. The method can make inferences based only on OpenStreetMap data and return valuable information regarding the importance of different regions in the bicycle-sharing station layouts. The prediction itself for the whole city without filtering by different thresholds provides an easy to read analysis that can help planners to quickly identify areas that can be taken into account when designing station layouts. Both the function generating embedding vectors for the regions and the function converting these vectors into station probabilities have been thoroughly tested with different hyperparameters and allow valuable results to be obtained.

The method treats each region separately and returns station occurrence probabilities without paying attention to whether the region next to it has already been proposed as a station so that the method is not able to return a very accurate layout at this point but only a heat map that can assist decision-makers. However, this can be improved by adding another function that will select a user-specified number of stations together with a minimum distance between stations.

An important value of this method is that it is easy to use - no nonpublic data was used for the analysis and it does not require any complex mobility analysis of the city.

This thesis has focused entirely on the layout of bicycle stations, but nothing prevents it from being used for other tasks such as planning the layout of charging stations for electric vehicles.

Using only public and readily available data from OpenStreetMap opens the way to use spatial data in machine learning for more applications, not necessarily only for those related to administrative entities that use nonpublic data. However, focusing only on the static data that is available in OSM may not be sufficient for many spatial data tasks.

\section{Future research}

A natural development for further research is to extend the method to other continents and see how it will work on data from North America or Asia. For this thesis, it was decided to rely only on data from Europe to start the research with a small set.

Although the method performs quite well for some of the cities, it may be worth exploring the impact of individual hyperparameters on performance by taking each city separately and adjusting its performance as appropriate to the data. Additionally, it is possible that more reliable results would be obtained if the method was validated on the whole set of regions from a single city and not only on its subset. It would then be possible to immediately detect when the model tends to predict too many stations in cities.

In addition, as mentioned in the discussion of the research questions, it would be worthwhile to investigate the effect of the confidence threshold, which in the experiments was set to a standard value of 0.5. This, together with an examination of the precision and recall curve, would give a better understanding of the behaviour of the method.

In order not to rely only on category-based basis vectors, it might be good to use another embedding method in the form of an autoencoder or to train a neural network using a triplet loss function.

Another direction for the development of the method is to investigate what effect changing the algorithm from classification to regression would have. At first glance, the task seemed ideal to be solved by the binary classification method, however, the use of very high-resolution regions causes neighbouring regions to be close to each other in the embedding space and the method has problems drawing a definite line between the region where the station should be and where it should not be. Using regression and assigning different confidence values depending on, for example, the Euclidean distance or the distance calculated from the city's road layout, the method would be set to predict confidence from 0 to 1 from the start, instead of teaching the classification model and using the confidence returned by the classifier thanks to the implementation in the scikit-learn library.

Although the results can be easily interpreted by a person living in the city that the analysis concerns or by an expert dealing with urban planning topics, the added value would be to accurately determine the importance of the features on the decisions to be made to develop an explanatory model that can be better understood by the user.

To increase the accuracy of the model, an algorithm could be added that takes into account how many stations should be distributed on the city grid and the minimum distances between stations to propose the exact regions that the stations should have, instead of returning a heat map of the city to the end-user.

\listoffigures

\listoftables

\bibliography{bibliography}
\bibliographystyle{abbrvnat}

\appendix

\chapter{List of filtered OpenStreetMap tags}
\label{osm_tags}

\TopicSetVPos{t}
\TopicSetWidth{=}
\TopicSetContinuationCode{\ (cont.)}
\begin{topiclongtable}{@{}Fp{0.25\linewidth}Tp{0.25\linewidth}Tp{0.5\linewidth}@{}}

\toprule
Category & Tag & Value \\* \midrule
\endhead
\hline\endfoot
\TopicLine \Topic[Aerialway] & \Topic[\texttt{aerialway}] & \texttt{cable\_car} \\
\TopicLine \Topic[Aerialway] & \Topic[\texttt{aerialway}] & \texttt{chair\_lift} \\
\TopicLine \Topic[Aerialway] & \Topic[\texttt{aerialway}] & \texttt{drag\_lift} \\
\TopicLine \Topic[Aerialway] & \Topic[\texttt{aerialway}] & \texttt{gondola} \\
\TopicLine \Topic[Aerialway] & \Topic[\texttt{aerialway}] & \texttt{goods} \\
\TopicLine \Topic[Aerialway] & \Topic[\texttt{aerialway}] & \texttt{j-bar} \\
\TopicLine \Topic[Aerialway] & \Topic[\texttt{aerialway}] & \texttt{magic\_carpet} \\
\TopicLine \Topic[Aerialway] & \Topic[\texttt{aerialway}] & \texttt{mixed\_lift} \\
\TopicLine \Topic[Aerialway] & \Topic[\texttt{aerialway}] & \texttt{platter} \\
\TopicLine \Topic[Aerialway] & \Topic[\texttt{aerialway}] & \texttt{pylon} \\
\TopicLine \Topic[Aerialway] & \Topic[\texttt{aerialway}] & \texttt{rope\_tow} \\
\TopicLine \Topic[Aerialway] & \Topic[\texttt{aerialway}] & \texttt{station} \\
\TopicLine \Topic[Aerialway] & \Topic[\texttt{aerialway}] & \texttt{t-bar} \\
\TopicLine \Topic[Aerialway] & \Topic[\texttt{aerialway}] & \texttt{zip\_line} \\

\TopicLine \Topic[Airports] & \Topic[\texttt{aeroway}] & \texttt{aerodrome} \\
\TopicLine \Topic[Airports] & \Topic[\texttt{aeroway}] & \texttt{heliport} \\

\TopicLine \Topic[Buildings] & \Topic[\texttt{building}] & - \\

\TopicLine \Topic[Culture, Art \& \allowbreak Entertainment] & \Topic[\texttt{amenity}] & \texttt{arts\_centre} \\
\TopicLine \Topic[Culture, Art \& \allowbreak Entertainment] & \Topic[\texttt{amenity}] & \texttt{brothel} \\
\TopicLine \Topic[Culture, Art \& \allowbreak Entertainment] & \Topic[\texttt{amenity}] & \texttt{casino} \\
\TopicLine \Topic[Culture, Art \& \allowbreak Entertainment] & \Topic[\texttt{amenity}] & \texttt{cinema} \\
\TopicLine \Topic[Culture, Art \& \allowbreak Entertainment] & \Topic[\texttt{amenity}] & \texttt{community\_centre} \\
\TopicLine \Topic[Culture, Art \& \allowbreak Entertainment] & \Topic[\texttt{amenity}] & \texttt{gambling} \\
\TopicLine \Topic[Culture, Art \& \allowbreak Entertainment] & \Topic[\texttt{amenity}] & \texttt{nightclub} \\
\TopicLine \Topic[Culture, Art \& \allowbreak Entertainment] & \Topic[\texttt{amenity}] & \texttt{planetarium} \\
\TopicLine \Topic[Culture, Art \& \allowbreak Entertainment] & \Topic[\texttt{amenity}] & \texttt{public\_bookcase} \\
\TopicLine \Topic[Culture, Art \& \allowbreak Entertainment] & \Topic[\texttt{amenity}] & \texttt{social\_centre} \\
\TopicLine \Topic[Culture, Art \& \allowbreak Entertainment] & \Topic[\texttt{amenity}] & \texttt{stripclub} \\
\TopicLine \Topic[Culture, Art \& \allowbreak Entertainment] & \Topic[\texttt{amenity}] & \texttt{studio} \\
\TopicLine \Topic[Culture, Art \& \allowbreak Entertainment] & \Topic[\texttt{amenity}] & \texttt{theatre} \\

\TopicLine \Topic[Education] & \Topic[\texttt{amenity}] & \texttt{college} \\
\TopicLine \Topic[Education] & \Topic[\texttt{amenity}] & \texttt{driving\_school} \\
\TopicLine \Topic[Education] & \Topic[\texttt{amenity}] & \texttt{kindergarten} \\
\TopicLine \Topic[Education] & \Topic[\texttt{amenity}] & \texttt{language\_school} \\
\TopicLine \Topic[Education] & \Topic[\texttt{amenity}] & \texttt{library} \\
\TopicLine \Topic[Education] & \Topic[\texttt{amenity}] & \texttt{music\_school} \\
\TopicLine \Topic[Education] & \Topic[\texttt{amenity}] & \texttt{school} \\
\TopicLine \Topic[Education] & \Topic[\texttt{amenity}] & \texttt{toy\_library} \\
\TopicLine \Topic[Education] & \Topic[\texttt{amenity}] & \texttt{university} \\

\TopicLine \Topic[Emergency] & \Topic[\texttt{emergency}] & - \\

\TopicLine \Topic[Finances] & \Topic[\texttt{amenity}] & \texttt{atm} \\
\TopicLine \Topic[Finances] & \Topic[\texttt{amenity}] & \texttt{bank} \\
\TopicLine \Topic[Finances] & \Topic[\texttt{amenity}] & \texttt{bureau\_de\_change} \\

\TopicLine \Topic[Healthcare] & \Topic[\texttt{amenity}] & \texttt{baby\_hatch} \\
\TopicLine \Topic[Healthcare] & \Topic[\texttt{amenity}] & \texttt{clinic} \\
\TopicLine \Topic[Healthcare] & \Topic[\texttt{amenity}] & \texttt{dentist} \\
\TopicLine \Topic[Healthcare] & \Topic[\texttt{amenity}] & \texttt{doctors} \\
\TopicLine \Topic[Healthcare] & \Topic[\texttt{amenity}] & \texttt{hospital} \\
\TopicLine \Topic[Healthcare] & \Topic[\texttt{amenity}] & \texttt{nursing\_home} \\
\TopicLine \Topic[Healthcare] & \Topic[\texttt{amenity}] & \texttt{pharmacy} \\
\TopicLine \Topic[Healthcare] & \Topic[\texttt{amenity}] & \texttt{social\_facility} \\
\TopicLine \Topic[Healthcare] & \Topic[\texttt{amenity}] & \texttt{veterinary} \\

\TopicLine \Topic[Historic] & \Topic[\texttt{historic}] & \texttt{aqueduct} \\
\TopicLine \Topic[Historic] & \Topic[\texttt{historic}] & \texttt{battlefield} \\
\TopicLine \Topic[Historic] & \Topic[\texttt{historic}] & \texttt{building} \\
\TopicLine \Topic[Historic] & \Topic[\texttt{historic}] & \texttt{castle} \\
\TopicLine \Topic[Historic] & \Topic[\texttt{historic}] & \texttt{church} \\
\TopicLine \Topic[Historic] & \Topic[\texttt{historic}] & \texttt{citywalls} \\
\TopicLine \Topic[Historic] & \Topic[\texttt{historic}] & \texttt{fort} \\
\TopicLine \Topic[Historic] & \Topic[\texttt{historic}] & \texttt{memorial} \\
\TopicLine \Topic[Historic] & \Topic[\texttt{historic}] & \texttt{monastery} \\
\TopicLine \Topic[Historic] & \Topic[\texttt{historic}] & \texttt{monument} \\
\TopicLine \Topic[Historic] & \Topic[\texttt{historic}] & \texttt{ruins} \\
\TopicLine \Topic[Historic] & \Topic[\texttt{historic}] & \texttt{tower} \\

\TopicLine \Topic[Leisure] & \Topic[\texttt{amenity}] & \texttt{dive\_centre} \\
\TopicLine \Topic[Leisure] & \Topic[\texttt{amenity}] & \texttt{public\_bath} \\

\TopicLine \Topic[Leisure] & \Topic[\texttt{leisure}] & \texttt{adult\_gaming\_centre} \\
\TopicLine \Topic[Leisure] & \Topic[\texttt{leisure}] & \texttt{amusement\_arcade} \\
\TopicLine \Topic[Leisure] & \Topic[\texttt{leisure}] & \texttt{beach\_resort} \\
\TopicLine \Topic[Leisure] & \Topic[\texttt{leisure}] & \texttt{common} \\
\TopicLine \Topic[Leisure] & \Topic[\texttt{leisure}] & \texttt{dance} \\
\TopicLine \Topic[Leisure] & \Topic[\texttt{leisure}] & \texttt{dog\_park} \\
\TopicLine \Topic[Leisure] & \Topic[\texttt{leisure}] & \texttt{escape\_game} \\
\TopicLine \Topic[Leisure] & \Topic[\texttt{leisure}] & \texttt{fitness\_centre} \\
\TopicLine \Topic[Leisure] & \Topic[\texttt{leisure}] & \texttt{fitness\_station} \\
\TopicLine \Topic[Leisure] & \Topic[\texttt{leisure}] & \texttt{garden} \\
\TopicLine \Topic[Leisure] & \Topic[\texttt{leisure}] & \texttt{hackerspace} \\
\TopicLine \Topic[Leisure] & \Topic[\texttt{leisure}] & \texttt{horse\_riding} \\
\TopicLine \Topic[Leisure] & \Topic[\texttt{leisure}] & \texttt{ice\_rink} \\
\TopicLine \Topic[Leisure] & \Topic[\texttt{leisure}] & \texttt{marina} \\
\TopicLine \Topic[Leisure] & \Topic[\texttt{leisure}] & \texttt{miniature\_golf} \\
\TopicLine \Topic[Leisure] & \Topic[\texttt{leisure}] & \texttt{nature\_reserve} \\
\TopicLine \Topic[Leisure] & \Topic[\texttt{leisure}] & \texttt{park} \\
\TopicLine \Topic[Leisure] & \Topic[\texttt{leisure}] & \texttt{pitch} \\
\TopicLine \Topic[Leisure] & \Topic[\texttt{leisure}] & \texttt{sauna} \\
\TopicLine \Topic[Leisure] & \Topic[\texttt{leisure}] & \texttt{slipway} \\
\TopicLine \Topic[Leisure] & \Topic[\texttt{leisure}] & \texttt{sports\_centre} \\
\TopicLine \Topic[Leisure] & \Topic[\texttt{leisure}] & \texttt{stadium} \\
\TopicLine \Topic[Leisure] & \Topic[\texttt{leisure}] & \texttt{summer\_camp} \\
\TopicLine \Topic[Leisure] & \Topic[\texttt{leisure}] & \texttt{swimming\_area} \\
\TopicLine \Topic[Leisure] & \Topic[\texttt{leisure}] & \texttt{swimming\_pool} \\
\TopicLine \Topic[Leisure] & \Topic[\texttt{leisure}] & \texttt{track} \\
\TopicLine \Topic[Leisure] & \Topic[\texttt{leisure}] & \texttt{water\_park} \\

\TopicLine \Topic[Other] & \Topic[\texttt{amenity}] & \texttt{animal\_boarding} \\
\TopicLine \Topic[Other] & \Topic[\texttt{amenity}] & \texttt{animal\_shelter} \\
\TopicLine \Topic[Other] & \Topic[\texttt{amenity}] & \texttt{childcare} \\
\TopicLine \Topic[Other] & \Topic[\texttt{amenity}] & \texttt{conference\_centre} \\
\TopicLine \Topic[Other] & \Topic[\texttt{amenity}] & \texttt{courthouse} \\
\TopicLine \Topic[Other] & \Topic[\texttt{amenity}] & \texttt{crematorium} \\
\TopicLine \Topic[Other] & \Topic[\texttt{amenity}] & \texttt{embassy} \\
\TopicLine \Topic[Other] & \Topic[\texttt{amenity}] & \texttt{fire\_station} \\
\TopicLine \Topic[Other] & \Topic[\texttt{amenity}] & \texttt{grave\_yard} \\
\TopicLine \Topic[Other] & \Topic[\texttt{amenity}] & \texttt{internet\_cafe} \\
\TopicLine \Topic[Other] & \Topic[\texttt{amenity}] & \texttt{marketplace} \\
\TopicLine \Topic[Other] & \Topic[\texttt{amenity}] & \texttt{monastery} \\
\TopicLine \Topic[Other] & \Topic[\texttt{amenity}] & \texttt{place\_of\_worship} \\
\TopicLine \Topic[Other] & \Topic[\texttt{amenity}] & \texttt{police} \\
\TopicLine \Topic[Other] & \Topic[\texttt{amenity}] & \texttt{post\_office} \\
\TopicLine \Topic[Other] & \Topic[\texttt{amenity}] & \texttt{prison} \\
\TopicLine \Topic[Other] & \Topic[\texttt{amenity}] & \texttt{ranger\_station} \\
\TopicLine \Topic[Other] & \Topic[\texttt{amenity}] & \texttt{townhall} \\

\TopicLine \Topic[Shops] & \Topic[\texttt{shop}] & \texttt{agrarian} \\
\TopicLine \Topic[Shops] & \Topic[\texttt{shop}] & \texttt{alcohol} \\
\TopicLine \Topic[Shops] & \Topic[\texttt{shop}] & \texttt{anime} \\
\TopicLine \Topic[Shops] & \Topic[\texttt{shop}] & \texttt{antiques} \\
\TopicLine \Topic[Shops] & \Topic[\texttt{shop}] & \texttt{appliance} \\
\TopicLine \Topic[Shops] & \Topic[\texttt{shop}] & \texttt{art} \\
\TopicLine \Topic[Shops] & \Topic[\texttt{shop}] & \texttt{atv} \\
\TopicLine \Topic[Shops] & \Topic[\texttt{shop}] & \texttt{baby\_goods} \\
\TopicLine \Topic[Shops] & \Topic[\texttt{shop}] & \texttt{bag} \\
\TopicLine \Topic[Shops] & \Topic[\texttt{shop}] & \texttt{bakery} \\
\TopicLine \Topic[Shops] & \Topic[\texttt{shop}] & \texttt{bathroom\_furnishing} \\
\TopicLine \Topic[Shops] & \Topic[\texttt{shop}] & \texttt{beauty} \\
\TopicLine \Topic[Shops] & \Topic[\texttt{shop}] & \texttt{bed} \\
\TopicLine \Topic[Shops] & \Topic[\texttt{shop}] & \texttt{beverages} \\
\TopicLine \Topic[Shops] & \Topic[\texttt{shop}] & \texttt{bicycle} \\
\TopicLine \Topic[Shops] & \Topic[\texttt{shop}] & \texttt{boat} \\
\TopicLine \Topic[Shops] & \Topic[\texttt{shop}] & \texttt{bookmaker} \\
\TopicLine \Topic[Shops] & \Topic[\texttt{shop}] & \texttt{books} \\
\TopicLine \Topic[Shops] & \Topic[\texttt{shop}] & \texttt{boutique} \\
\TopicLine \Topic[Shops] & \Topic[\texttt{shop}] & \texttt{brewing\_supplies} \\
\TopicLine \Topic[Shops] & \Topic[\texttt{shop}] & \texttt{butcher} \\
\TopicLine \Topic[Shops] & \Topic[\texttt{shop}] & \texttt{camera} \\
\TopicLine \Topic[Shops] & \Topic[\texttt{shop}] & \texttt{candles} \\
\TopicLine \Topic[Shops] & \Topic[\texttt{shop}] & \texttt{cannabis} \\
\TopicLine \Topic[Shops] & \Topic[\texttt{shop}] & \texttt{car} \\
\TopicLine \Topic[Shops] & \Topic[\texttt{shop}] & \texttt{car\_parts} \\
\TopicLine \Topic[Shops] & \Topic[\texttt{shop}] & \texttt{car\_repair} \\
\TopicLine \Topic[Shops] & \Topic[\texttt{shop}] & \texttt{caravan} \\
\TopicLine \Topic[Shops] & \Topic[\texttt{shop}] & \texttt{carpet} \\
\TopicLine \Topic[Shops] & \Topic[\texttt{shop}] & \texttt{charity} \\
\TopicLine \Topic[Shops] & \Topic[\texttt{shop}] & \texttt{cheese} \\
\TopicLine \Topic[Shops] & \Topic[\texttt{shop}] & \texttt{chemist} \\
\TopicLine \Topic[Shops] & \Topic[\texttt{shop}] & \texttt{chocolate} \\
\TopicLine \Topic[Shops] & \Topic[\texttt{shop}] & \texttt{clothes} \\
\TopicLine \Topic[Shops] & \Topic[\texttt{shop}] & \texttt{coffee} \\
\TopicLine \Topic[Shops] & \Topic[\texttt{shop}] & \texttt{collector} \\
\TopicLine \Topic[Shops] & \Topic[\texttt{shop}] & \texttt{computer} \\
\TopicLine \Topic[Shops] & \Topic[\texttt{shop}] & \texttt{confectionery} \\
\TopicLine \Topic[Shops] & \Topic[\texttt{shop}] & \texttt{convenience} \\
\TopicLine \Topic[Shops] & \Topic[\texttt{shop}] & \texttt{copyshop} \\
\TopicLine \Topic[Shops] & \Topic[\texttt{shop}] & \texttt{cosmetics} \\
\TopicLine \Topic[Shops] & \Topic[\texttt{shop}] & \texttt{craft} \\
\TopicLine \Topic[Shops] & \Topic[\texttt{shop}] & \texttt{curtain} \\
\TopicLine \Topic[Shops] & \Topic[\texttt{shop}] & \texttt{dairy} \\
\TopicLine \Topic[Shops] & \Topic[\texttt{shop}] & \texttt{deli} \\
\TopicLine \Topic[Shops] & \Topic[\texttt{shop}] & \texttt{department\_store} \\
\TopicLine \Topic[Shops] & \Topic[\texttt{shop}] & \texttt{doityourself} \\
\TopicLine \Topic[Shops] & \Topic[\texttt{shop}] & \texttt{doors} \\
\TopicLine \Topic[Shops] & \Topic[\texttt{shop}] & \texttt{dry\_cleaning} \\
\TopicLine \Topic[Shops] & \Topic[\texttt{shop}] & \texttt{e-cigarette} \\
\TopicLine \Topic[Shops] & \Topic[\texttt{shop}] & \texttt{electrical} \\
\TopicLine \Topic[Shops] & \Topic[\texttt{shop}] & \texttt{electronics} \\
\TopicLine \Topic[Shops] & \Topic[\texttt{shop}] & \texttt{energy} \\
\TopicLine \Topic[Shops] & \Topic[\texttt{shop}] & \texttt{erotic} \\
\TopicLine \Topic[Shops] & \Topic[\texttt{shop}] & \texttt{fabric} \\
\TopicLine \Topic[Shops] & \Topic[\texttt{shop}] & \texttt{farm} \\
\TopicLine \Topic[Shops] & \Topic[\texttt{shop}] & \texttt{fashion\_accessories} \\
\TopicLine \Topic[Shops] & \Topic[\texttt{shop}] & \texttt{fireplace} \\
\TopicLine \Topic[Shops] & \Topic[\texttt{shop}] & \texttt{fishing} \\
\TopicLine \Topic[Shops] & \Topic[\texttt{shop}] & \texttt{flooring} \\
\TopicLine \Topic[Shops] & \Topic[\texttt{shop}] & \texttt{florist} \\
\TopicLine \Topic[Shops] & \Topic[\texttt{shop}] & \texttt{frame} \\
\TopicLine \Topic[Shops] & \Topic[\texttt{shop}] & \texttt{frozen\_food} \\
\TopicLine \Topic[Shops] & \Topic[\texttt{shop}] & \texttt{fuel} \\
\TopicLine \Topic[Shops] & \Topic[\texttt{shop}] & \texttt{funeral\_directors} \\
\TopicLine \Topic[Shops] & \Topic[\texttt{shop}] & \texttt{furniture} \\
\TopicLine \Topic[Shops] & \Topic[\texttt{shop}] & \texttt{games} \\
\TopicLine \Topic[Shops] & \Topic[\texttt{shop}] & \texttt{garden\_centre} \\
\TopicLine \Topic[Shops] & \Topic[\texttt{shop}] & \texttt{garden\_furniture} \\
\TopicLine \Topic[Shops] & \Topic[\texttt{shop}] & \texttt{gas} \\
\TopicLine \Topic[Shops] & \Topic[\texttt{shop}] & \texttt{general} \\
\TopicLine \Topic[Shops] & \Topic[\texttt{shop}] & \texttt{gift} \\
\TopicLine \Topic[Shops] & \Topic[\texttt{shop}] & \texttt{glaziery} \\
\TopicLine \Topic[Shops] & \Topic[\texttt{shop}] & \texttt{golf} \\
\TopicLine \Topic[Shops] & \Topic[\texttt{shop}] & \texttt{greengrocer} \\
\TopicLine \Topic[Shops] & \Topic[\texttt{shop}] & \texttt{groundskeeping} \\
\TopicLine \Topic[Shops] & \Topic[\texttt{shop}] & \texttt{hairdresser} \\
\TopicLine \Topic[Shops] & \Topic[\texttt{shop}] & \texttt{hairdresser\_supply} \\
\TopicLine \Topic[Shops] & \Topic[\texttt{shop}] & \texttt{hardware} \\
\TopicLine \Topic[Shops] & \Topic[\texttt{shop}] & \texttt{health\_food} \\
\TopicLine \Topic[Shops] & \Topic[\texttt{shop}] & \texttt{hearing\_aids} \\
\TopicLine \Topic[Shops] & \Topic[\texttt{shop}] & \texttt{herbalist} \\
\TopicLine \Topic[Shops] & \Topic[\texttt{shop}] & \texttt{hifi} \\
\TopicLine \Topic[Shops] & \Topic[\texttt{shop}] & \texttt{household\_linen} \\
\TopicLine \Topic[Shops] & \Topic[\texttt{shop}] & \texttt{houseware} \\
\TopicLine \Topic[Shops] & \Topic[\texttt{shop}] & \texttt{hunting} \\
\TopicLine \Topic[Shops] & \Topic[\texttt{shop}] & \texttt{ice\_cream} \\
\TopicLine \Topic[Shops] & \Topic[\texttt{shop}] & \texttt{interior\_decoration} \\
\TopicLine \Topic[Shops] & \Topic[\texttt{shop}] & \texttt{jetski} \\
\TopicLine \Topic[Shops] & \Topic[\texttt{shop}] & \texttt{jewelry} \\
\TopicLine \Topic[Shops] & \Topic[\texttt{shop}] & \texttt{kiosk} \\
\TopicLine \Topic[Shops] & \Topic[\texttt{shop}] & \texttt{kitchen} \\
\TopicLine \Topic[Shops] & \Topic[\texttt{shop}] & \texttt{lamps} \\
\TopicLine \Topic[Shops] & \Topic[\texttt{shop}] & \texttt{laundry} \\
\TopicLine \Topic[Shops] & \Topic[\texttt{shop}] & \texttt{leather} \\
\TopicLine \Topic[Shops] & \Topic[\texttt{shop}] & \texttt{lighting} \\
\TopicLine \Topic[Shops] & \Topic[\texttt{shop}] & \texttt{locksmith} \\
\TopicLine \Topic[Shops] & \Topic[\texttt{shop}] & \texttt{lottery} \\
\TopicLine \Topic[Shops] & \Topic[\texttt{shop}] & \texttt{mall} \\
\TopicLine \Topic[Shops] & \Topic[\texttt{shop}] & \texttt{massage} \\
\TopicLine \Topic[Shops] & \Topic[\texttt{shop}] & \texttt{medical\_supply} \\
\TopicLine \Topic[Shops] & \Topic[\texttt{shop}] & \texttt{military\_surplus} \\
\TopicLine \Topic[Shops] & \Topic[\texttt{shop}] & \texttt{mobile\_phone} \\
\TopicLine \Topic[Shops] & \Topic[\texttt{shop}] & \texttt{model} \\
\TopicLine \Topic[Shops] & \Topic[\texttt{shop}] & \texttt{money\_lender} \\
\TopicLine \Topic[Shops] & \Topic[\texttt{shop}] & \texttt{motorcycle} \\
\TopicLine \Topic[Shops] & \Topic[\texttt{shop}] & \texttt{music} \\
\TopicLine \Topic[Shops] & \Topic[\texttt{shop}] & \texttt{musical\_instrument} \\
\TopicLine \Topic[Shops] & \Topic[\texttt{shop}] & \texttt{newsagent} \\
\TopicLine \Topic[Shops] & \Topic[\texttt{shop}] & \texttt{nutrition\_supplements} \\
\TopicLine \Topic[Shops] & \Topic[\texttt{shop}] & \texttt{optician} \\
\TopicLine \Topic[Shops] & \Topic[\texttt{shop}] & \texttt{organic} \\
\TopicLine \Topic[Shops] & \Topic[\texttt{shop}] & \texttt{outdoor} \\
\TopicLine \Topic[Shops] & \Topic[\texttt{shop}] & \texttt{outpost} \\
\TopicLine \Topic[Shops] & \Topic[\texttt{shop}] & \texttt{paint} \\
\TopicLine \Topic[Shops] & \Topic[\texttt{shop}] & \texttt{party} \\
\TopicLine \Topic[Shops] & \Topic[\texttt{shop}] & \texttt{pasta} \\
\TopicLine \Topic[Shops] & \Topic[\texttt{shop}] & \texttt{pastry} \\
\TopicLine \Topic[Shops] & \Topic[\texttt{shop}] & \texttt{pawnbroker} \\
\TopicLine \Topic[Shops] & \Topic[\texttt{shop}] & \texttt{perfumery} \\
\TopicLine \Topic[Shops] & \Topic[\texttt{shop}] & \texttt{pest\_control} \\
\TopicLine \Topic[Shops] & \Topic[\texttt{shop}] & \texttt{pet} \\
\TopicLine \Topic[Shops] & \Topic[\texttt{shop}] & \texttt{pet\_grooming} \\
\TopicLine \Topic[Shops] & \Topic[\texttt{shop}] & \texttt{photo} \\
\TopicLine \Topic[Shops] & \Topic[\texttt{shop}] & \texttt{pyrotechnics} \\
\TopicLine \Topic[Shops] & \Topic[\texttt{shop}] & \texttt{radiotechnics} \\
\TopicLine \Topic[Shops] & \Topic[\texttt{shop}] & \texttt{religion} \\
\TopicLine \Topic[Shops] & \Topic[\texttt{shop}] & \texttt{scuba\_diving} \\
\TopicLine \Topic[Shops] & \Topic[\texttt{shop}] & \texttt{seafood} \\
\TopicLine \Topic[Shops] & \Topic[\texttt{shop}] & \texttt{second\_hand} \\
\TopicLine \Topic[Shops] & \Topic[\texttt{shop}] & \texttt{security} \\
\TopicLine \Topic[Shops] & \Topic[\texttt{shop}] & \texttt{sewing} \\
\TopicLine \Topic[Shops] & \Topic[\texttt{shop}] & \texttt{shoes} \\
\TopicLine \Topic[Shops] & \Topic[\texttt{shop}] & \texttt{ski} \\
\TopicLine \Topic[Shops] & \Topic[\texttt{shop}] & \texttt{snowmobile} \\
\TopicLine \Topic[Shops] & \Topic[\texttt{shop}] & \texttt{spices} \\
\TopicLine \Topic[Shops] & \Topic[\texttt{shop}] & \texttt{sports} \\
\TopicLine \Topic[Shops] & \Topic[\texttt{shop}] & \texttt{stationery} \\
\TopicLine \Topic[Shops] & \Topic[\texttt{shop}] & \texttt{storage\_rental} \\
\TopicLine \Topic[Shops] & \Topic[\texttt{shop}] & \texttt{supermarket} \\
\TopicLine \Topic[Shops] & \Topic[\texttt{shop}] & \texttt{swimming\_pool} \\
\TopicLine \Topic[Shops] & \Topic[\texttt{shop}] & \texttt{tailor} \\
\TopicLine \Topic[Shops] & \Topic[\texttt{shop}] & \texttt{tattoo} \\
\TopicLine \Topic[Shops] & \Topic[\texttt{shop}] & \texttt{tea} \\
\TopicLine \Topic[Shops] & \Topic[\texttt{shop}] & \texttt{ticket} \\
\TopicLine \Topic[Shops] & \Topic[\texttt{shop}] & \texttt{tiles} \\
\TopicLine \Topic[Shops] & \Topic[\texttt{shop}] & \texttt{tobacco} \\
\TopicLine \Topic[Shops] & \Topic[\texttt{shop}] & \texttt{toys} \\
\TopicLine \Topic[Shops] & \Topic[\texttt{shop}] & \texttt{trade} \\
\TopicLine \Topic[Shops] & \Topic[\texttt{shop}] & \texttt{trailer} \\
\TopicLine \Topic[Shops] & \Topic[\texttt{shop}] & \texttt{travel\_agency} \\
\TopicLine \Topic[Shops] & \Topic[\texttt{shop}] & \texttt{trophy} \\
\TopicLine \Topic[Shops] & \Topic[\texttt{shop}] & \texttt{tyres} \\
\TopicLine \Topic[Shops] & \Topic[\texttt{shop}] & \texttt{vacant} \\
\TopicLine \Topic[Shops] & \Topic[\texttt{shop}] & \texttt{vacuum\_cleaner} \\
\TopicLine \Topic[Shops] & \Topic[\texttt{shop}] & \texttt{variety\_store} \\
\TopicLine \Topic[Shops] & \Topic[\texttt{shop}] & \texttt{video} \\
\TopicLine \Topic[Shops] & \Topic[\texttt{shop}] & \texttt{video\_games} \\
\TopicLine \Topic[Shops] & \Topic[\texttt{shop}] & \texttt{watches} \\
\TopicLine \Topic[Shops] & \Topic[\texttt{shop}] & \texttt{water} \\
\TopicLine \Topic[Shops] & \Topic[\texttt{shop}] & \texttt{weapons} \\
\TopicLine \Topic[Shops] & \Topic[\texttt{shop}] & \texttt{wholesale} \\
\TopicLine \Topic[Shops] & \Topic[\texttt{shop}] & \texttt{window\_blind} \\
\TopicLine \Topic[Shops] & \Topic[\texttt{shop}] & \texttt{wool} \\

\TopicLine \Topic[Sport] & \Topic[\texttt{sport}] & \texttt{10pin} \\
\TopicLine \Topic[Sport] & \Topic[\texttt{sport}] & \texttt{9pin} \\
\TopicLine \Topic[Sport] & \Topic[\texttt{sport}] & \texttt{aikido} \\
\TopicLine \Topic[Sport] & \Topic[\texttt{sport}] & \texttt{american\_football} \\
\TopicLine \Topic[Sport] & \Topic[\texttt{sport}] & \texttt{archery} \\
\TopicLine \Topic[Sport] & \Topic[\texttt{sport}] & \texttt{athletics} \\
\TopicLine \Topic[Sport] & \Topic[\texttt{sport}] & \texttt{australian\_football} \\
\TopicLine \Topic[Sport] & \Topic[\texttt{sport}] & \texttt{badminton} \\
\TopicLine \Topic[Sport] & \Topic[\texttt{sport}] & \texttt{bandy} \\
\TopicLine \Topic[Sport] & \Topic[\texttt{sport}] & \texttt{baseball} \\
\TopicLine \Topic[Sport] & \Topic[\texttt{sport}] & \texttt{basketball} \\
\TopicLine \Topic[Sport] & \Topic[\texttt{sport}] & \texttt{beachvolleyball} \\
\TopicLine \Topic[Sport] & \Topic[\texttt{sport}] & \texttt{biathlon} \\
\TopicLine \Topic[Sport] & \Topic[\texttt{sport}] & \texttt{billiards} \\
\TopicLine \Topic[Sport] & \Topic[\texttt{sport}] & \texttt{bmx} \\
\TopicLine \Topic[Sport] & \Topic[\texttt{sport}] & \texttt{bobsleigh} \\
\TopicLine \Topic[Sport] & \Topic[\texttt{sport}] & \texttt{boules} \\
\TopicLine \Topic[Sport] & \Topic[\texttt{sport}] & \texttt{bowls} \\
\TopicLine \Topic[Sport] & \Topic[\texttt{sport}] & \texttt{boxing} \\
\TopicLine \Topic[Sport] & \Topic[\texttt{sport}] & \texttt{bullfighting} \\
\TopicLine \Topic[Sport] & \Topic[\texttt{sport}] & \texttt{canadian\_football} \\
\TopicLine \Topic[Sport] & \Topic[\texttt{sport}] & \texttt{canoe} \\
\TopicLine \Topic[Sport] & \Topic[\texttt{sport}] & \texttt{chess} \\
\TopicLine \Topic[Sport] & \Topic[\texttt{sport}] & \texttt{cliff\_diving} \\
\TopicLine \Topic[Sport] & \Topic[\texttt{sport}] & \texttt{climbing} \\
\TopicLine \Topic[Sport] & \Topic[\texttt{sport}] & \texttt{climbing\_adventure} \\
\TopicLine \Topic[Sport] & \Topic[\texttt{sport}] & \texttt{cockfighting} \\
\TopicLine \Topic[Sport] & \Topic[\texttt{sport}] & \texttt{cricket} \\
\TopicLine \Topic[Sport] & \Topic[\texttt{sport}] & \texttt{croquet} \\
\TopicLine \Topic[Sport] & \Topic[\texttt{sport}] & \texttt{crossfit} \\
\TopicLine \Topic[Sport] & \Topic[\texttt{sport}] & \texttt{curling} \\
\TopicLine \Topic[Sport] & \Topic[\texttt{sport}] & \texttt{cycle\_polo} \\
\TopicLine \Topic[Sport] & \Topic[\texttt{sport}] & \texttt{cycling} \\
\TopicLine \Topic[Sport] & \Topic[\texttt{sport}] & \texttt{darts} \\
\TopicLine \Topic[Sport] & \Topic[\texttt{sport}] & \texttt{dog\_agility} \\
\TopicLine \Topic[Sport] & \Topic[\texttt{sport}] & \texttt{dog\_racing} \\
\TopicLine \Topic[Sport] & \Topic[\texttt{sport}] & \texttt{equestrian} \\
\TopicLine \Topic[Sport] & \Topic[\texttt{sport}] & \texttt{fencing} \\
\TopicLine \Topic[Sport] & \Topic[\texttt{sport}] & \texttt{field\_hockey} \\
\TopicLine \Topic[Sport] & \Topic[\texttt{sport}] & \texttt{fitness} \\
\TopicLine \Topic[Sport] & \Topic[\texttt{sport}] & \texttt{five-a-side} \\
\TopicLine \Topic[Sport] & \Topic[\texttt{sport}] & \texttt{floorball} \\
\TopicLine \Topic[Sport] & \Topic[\texttt{sport}] & \texttt{free\_flying} \\
\TopicLine \Topic[Sport] & \Topic[\texttt{sport}] & \texttt{futsal} \\
\TopicLine \Topic[Sport] & \Topic[\texttt{sport}] & \texttt{gaelic\_games} \\
\TopicLine \Topic[Sport] & \Topic[\texttt{sport}] & \texttt{golf} \\
\TopicLine \Topic[Sport] & \Topic[\texttt{sport}] & \texttt{gymnastics} \\
\TopicLine \Topic[Sport] & \Topic[\texttt{sport}] & \texttt{handball} \\
\TopicLine \Topic[Sport] & \Topic[\texttt{sport}] & \texttt{hapkido} \\
\TopicLine \Topic[Sport] & \Topic[\texttt{sport}] & \texttt{horse\_racing} \\
\TopicLine \Topic[Sport] & \Topic[\texttt{sport}] & \texttt{horseshoes} \\
\TopicLine \Topic[Sport] & \Topic[\texttt{sport}] & \texttt{ice\_hockey} \\
\TopicLine \Topic[Sport] & \Topic[\texttt{sport}] & \texttt{ice\_skating} \\
\TopicLine \Topic[Sport] & \Topic[\texttt{sport}] & \texttt{ice\_stock} \\
\TopicLine \Topic[Sport] & \Topic[\texttt{sport}] & \texttt{jiu-jitsu} \\
\TopicLine \Topic[Sport] & \Topic[\texttt{sport}] & \texttt{judo} \\
\TopicLine \Topic[Sport] & \Topic[\texttt{sport}] & \texttt{karate} \\
\TopicLine \Topic[Sport] & \Topic[\texttt{sport}] & \texttt{karting} \\
\TopicLine \Topic[Sport] & \Topic[\texttt{sport}] & \texttt{kickboxing} \\
\TopicLine \Topic[Sport] & \Topic[\texttt{sport}] & \texttt{kitesurfing} \\
\TopicLine \Topic[Sport] & \Topic[\texttt{sport}] & \texttt{korfball} \\
\TopicLine \Topic[Sport] & \Topic[\texttt{sport}] & \texttt{krachtbal} \\
\TopicLine \Topic[Sport] & \Topic[\texttt{sport}] & \texttt{lacrosse} \\
\TopicLine \Topic[Sport] & \Topic[\texttt{sport}] & \texttt{martial\_arts} \\
\TopicLine \Topic[Sport] & \Topic[\texttt{sport}] & \texttt{miniature\_golf} \\
\TopicLine \Topic[Sport] & \Topic[\texttt{sport}] & \texttt{model\_aerodrome} \\
\TopicLine \Topic[Sport] & \Topic[\texttt{sport}] & \texttt{motocross} \\
\TopicLine \Topic[Sport] & \Topic[\texttt{sport}] & \texttt{motor} \\
\TopicLine \Topic[Sport] & \Topic[\texttt{sport}] & \texttt{multi} \\
\TopicLine \Topic[Sport] & \Topic[\texttt{sport}] & \texttt{netball} \\
\TopicLine \Topic[Sport] & \Topic[\texttt{sport}] & \texttt{obstacle\_course} \\
\TopicLine \Topic[Sport] & \Topic[\texttt{sport}] & \texttt{orienteering} \\
\TopicLine \Topic[Sport] & \Topic[\texttt{sport}] & \texttt{paddle\_tennis} \\
\TopicLine \Topic[Sport] & \Topic[\texttt{sport}] & \texttt{padel} \\
\TopicLine \Topic[Sport] & \Topic[\texttt{sport}] & \texttt{parachuting} \\
\TopicLine \Topic[Sport] & \Topic[\texttt{sport}] & \texttt{parkour} \\
\TopicLine \Topic[Sport] & \Topic[\texttt{sport}] & \texttt{pedal\_car\_racing} \\
\TopicLine \Topic[Sport] & \Topic[\texttt{sport}] & \texttt{pelota} \\
\TopicLine \Topic[Sport] & \Topic[\texttt{sport}] & \texttt{pesäpallo} \\
\TopicLine \Topic[Sport] & \Topic[\texttt{sport}] & \texttt{pickleball} \\
\TopicLine \Topic[Sport] & \Topic[\texttt{sport}] & \texttt{pilates} \\
\TopicLine \Topic[Sport] & \Topic[\texttt{sport}] & \texttt{pole\_dance} \\
\TopicLine \Topic[Sport] & \Topic[\texttt{sport}] & \texttt{racquet} \\
\TopicLine \Topic[Sport] & \Topic[\texttt{sport}] & \texttt{rc\_car} \\
\TopicLine \Topic[Sport] & \Topic[\texttt{sport}] & \texttt{roller\_skating} \\
\TopicLine \Topic[Sport] & \Topic[\texttt{sport}] & \texttt{rowing} \\
\TopicLine \Topic[Sport] & \Topic[\texttt{sport}] & \texttt{rugby\_league} \\
\TopicLine \Topic[Sport] & \Topic[\texttt{sport}] & \texttt{rugby\_union} \\
\TopicLine \Topic[Sport] & \Topic[\texttt{sport}] & \texttt{running} \\
\TopicLine \Topic[Sport] & \Topic[\texttt{sport}] & \texttt{sailing} \\
\TopicLine \Topic[Sport] & \Topic[\texttt{sport}] & \texttt{scuba\_diving} \\
\TopicLine \Topic[Sport] & \Topic[\texttt{sport}] & \texttt{shooting} \\
\TopicLine \Topic[Sport] & \Topic[\texttt{sport}] & \texttt{shot-put} \\
\TopicLine \Topic[Sport] & \Topic[\texttt{sport}] & \texttt{skateboard} \\
\TopicLine \Topic[Sport] & \Topic[\texttt{sport}] & \texttt{ski\_jumping} \\
\TopicLine \Topic[Sport] & \Topic[\texttt{sport}] & \texttt{skiing} \\
\TopicLine \Topic[Sport] & \Topic[\texttt{sport}] & \texttt{snooker} \\
\TopicLine \Topic[Sport] & \Topic[\texttt{sport}] & \texttt{soccer} \\
\TopicLine \Topic[Sport] & \Topic[\texttt{sport}] & \texttt{speedway} \\
\TopicLine \Topic[Sport] & \Topic[\texttt{sport}] & \texttt{squash} \\
\TopicLine \Topic[Sport] & \Topic[\texttt{sport}] & \texttt{sumo} \\
\TopicLine \Topic[Sport] & \Topic[\texttt{sport}] & \texttt{surfing} \\
\TopicLine \Topic[Sport] & \Topic[\texttt{sport}] & \texttt{swimming} \\
\TopicLine \Topic[Sport] & \Topic[\texttt{sport}] & \texttt{table\_soccer} \\
\TopicLine \Topic[Sport] & \Topic[\texttt{sport}] & \texttt{table\_tennis} \\
\TopicLine \Topic[Sport] & \Topic[\texttt{sport}] & \texttt{taekwondo} \\
\TopicLine \Topic[Sport] & \Topic[\texttt{sport}] & \texttt{tennis} \\
\TopicLine \Topic[Sport] & \Topic[\texttt{sport}] & \texttt{toboggan} \\
\TopicLine \Topic[Sport] & \Topic[\texttt{sport}] & \texttt{ultimate} \\
\TopicLine \Topic[Sport] & \Topic[\texttt{sport}] & \texttt{volleyball} \\
\TopicLine \Topic[Sport] & \Topic[\texttt{sport}] & \texttt{wakeboarding} \\
\TopicLine \Topic[Sport] & \Topic[\texttt{sport}] & \texttt{water\_polo} \\
\TopicLine \Topic[Sport] & \Topic[\texttt{sport}] & \texttt{water\_ski} \\
\TopicLine \Topic[Sport] & \Topic[\texttt{sport}] & \texttt{weightlifting} \\
\TopicLine \Topic[Sport] & \Topic[\texttt{sport}] & \texttt{wrestling} \\
\TopicLine \Topic[Sport] & \Topic[\texttt{sport}] & \texttt{yoga} \\
\TopicLine \Topic[Sport] & \Topic[\texttt{sport}] & \texttt{zurkhaneh\_sport} \\

\TopicLine \Topic[Sustenance] & \Topic[\texttt{amenity}] & \texttt{bar} \\
\TopicLine \Topic[Sustenance] & \Topic[\texttt{amenity}] & \texttt{bbq} \\
\TopicLine \Topic[Sustenance] & \Topic[\texttt{amenity}] & \texttt{biergarten} \\
\TopicLine \Topic[Sustenance] & \Topic[\texttt{amenity}] & \texttt{cafe} \\
\TopicLine \Topic[Sustenance] & \Topic[\texttt{amenity}] & \texttt{fast\_food} \\
\TopicLine \Topic[Sustenance] & \Topic[\texttt{amenity}] & \texttt{food\_court} \\
\TopicLine \Topic[Sustenance] & \Topic[\texttt{amenity}] & \texttt{ice\_cream} \\
\TopicLine \Topic[Sustenance] & \Topic[\texttt{amenity}] & \texttt{pub} \\
\TopicLine \Topic[Sustenance] & \Topic[\texttt{amenity}] & \texttt{restaurant} \\

\TopicLine \Topic[Tourism] & \Topic[\texttt{tourism}] & \texttt{alpine\_hut} \\
\TopicLine \Topic[Tourism] & \Topic[\texttt{tourism}] & \texttt{apartment} \\
\TopicLine \Topic[Tourism] & \Topic[\texttt{tourism}] & \texttt{aquarium} \\
\TopicLine \Topic[Tourism] & \Topic[\texttt{tourism}] & \texttt{artwork} \\
\TopicLine \Topic[Tourism] & \Topic[\texttt{tourism}] & \texttt{attraction} \\
\TopicLine \Topic[Tourism] & \Topic[\texttt{tourism}] & \texttt{camp\_pitch} \\
\TopicLine \Topic[Tourism] & \Topic[\texttt{tourism}] & \texttt{camp\_site} \\
\TopicLine \Topic[Tourism] & \Topic[\texttt{tourism}] & \texttt{caravan\_site} \\
\TopicLine \Topic[Tourism] & \Topic[\texttt{tourism}] & \texttt{chalet} \\
\TopicLine \Topic[Tourism] & \Topic[\texttt{tourism}] & \texttt{gallery} \\
\TopicLine \Topic[Tourism] & \Topic[\texttt{tourism}] & \texttt{guest\_house} \\
\TopicLine \Topic[Tourism] & \Topic[\texttt{tourism}] & \texttt{hostel} \\
\TopicLine \Topic[Tourism] & \Topic[\texttt{tourism}] & \texttt{hotel} \\
\TopicLine \Topic[Tourism] & \Topic[\texttt{tourism}] & \texttt{information} \\
\TopicLine \Topic[Tourism] & \Topic[\texttt{tourism}] & \texttt{motel} \\
\TopicLine \Topic[Tourism] & \Topic[\texttt{tourism}] & \texttt{museum} \\
\TopicLine \Topic[Tourism] & \Topic[\texttt{tourism}] & \texttt{picnic\_site} \\
\TopicLine \Topic[Tourism] & \Topic[\texttt{tourism}] & \texttt{theme\_park} \\
\TopicLine \Topic[Tourism] & \Topic[\texttt{tourism}] & \texttt{viewpoint} \\
\TopicLine \Topic[Tourism] & \Topic[\texttt{tourism}] & \texttt{wilderness\_hut} \\
\TopicLine \Topic[Tourism] & \Topic[\texttt{tourism}] & \texttt{zoo} \\

\TopicLine \Topic[Transportation] & \Topic[\texttt{amenity}] & \texttt{bicycle\_parking} \\
\TopicLine \Topic[Transportation] & \Topic[\texttt{amenity}] & \texttt{bicycle\_rental} \\
\TopicLine \Topic[Transportation] & \Topic[\texttt{amenity}] & \texttt{bicycle\_repair\_station} \\
\TopicLine \Topic[Transportation] & \Topic[\texttt{amenity}] & \texttt{boat\_rental} \\
\TopicLine \Topic[Transportation] & \Topic[\texttt{amenity}] & \texttt{boat\_sharing} \\
\TopicLine \Topic[Transportation] & \Topic[\texttt{amenity}] & \texttt{bus\_station} \\
\TopicLine \Topic[Transportation] & \Topic[\texttt{amenity}] & \texttt{car\_rental} \\
\TopicLine \Topic[Transportation] & \Topic[\texttt{amenity}] & \texttt{car\_sharing} \\
\TopicLine \Topic[Transportation] & \Topic[\texttt{amenity}] & \texttt{car\_wash} \\
\TopicLine \Topic[Transportation] & \Topic[\texttt{amenity}] & \texttt{charging\_station} \\
\TopicLine \Topic[Transportation] & \Topic[\texttt{amenity}] & \texttt{clock} \\
\TopicLine \Topic[Transportation] & \Topic[\texttt{amenity}] & \texttt{ferry\_terminal} \\
\TopicLine \Topic[Transportation] & \Topic[\texttt{amenity}] & \texttt{motorcycle\_parking} \\
\TopicLine \Topic[Transportation] & \Topic[\texttt{amenity}] & \texttt{parking} \\
\TopicLine \Topic[Transportation] & \Topic[\texttt{amenity}] & \texttt{shelter} \\
\TopicLine \Topic[Transportation] & \Topic[\texttt{amenity}] & \texttt{taxi} \\

\TopicLine \Topic[Transportation] & \Topic[\texttt{public\_transport}] & \texttt{platform} \\
\TopicLine \Topic[Transportation] & \Topic[\texttt{public\_transport}] & \texttt{station} \\
\TopicLine \Topic[Transportation] & \Topic[\texttt{public\_transport}] & \texttt{stop\_area} \\
\TopicLine \Topic[Transportation] & \Topic[\texttt{public\_transport}] & \texttt{stop\_position} \\

\TopicLine \Topic[Water] & \Topic[-] & - \\

\TopicLine \Topic[Roads: \\ bicycle paths] & \Topic[-] & - \\
           \Topic[Roads: \\ bicycle paths] & \Topic[-] &   \\

\TopicLine \Topic[Roads: motorways] & \Topic[-] & - \\
           \Topic[Roads: motorways] & \Topic[-] &   \\

\TopicLine \Topic[Roads: pavements] & \Topic[-] & - \\
           \Topic[Roads: pavements] & \Topic[-] &   \\
\bottomrule

\caption[List of all OpenStreetMap tags used for filtering relations]{List of all OpenStreetMap tags used for filtering relations. These tags allowed the generation of a fixed-length feature vector ($n = 888$) that was later embedded using an autoencoder into lower dimensionality vectors. Entries without subcategories were used collectively without division.}
\label{tab:app-osm-tags}\\
\end{topiclongtable}

\end{document}